\documentclass{article}

\usepackage{microtype}
\usepackage{graphicx}
\usepackage{subfigure}
\usepackage{amsfonts}
\usepackage{bm}
\usepackage{amsmath}
\usepackage{enumitem}
\usepackage{booktabs} % for professional tables
\usepackage{setspace} % for supplementary
\usepackage{color}
\definecolor{bgcolor}{RGB}{255,209,220} % for bg of image recon error
\usepackage{hyperref}
\usepackage[accepted]{icml2018}

\icmltitlerunning{Neural Inverse Rendering for General Reflectance Photometric Stereo (To appear in ICML 2018)}

\usepackage{xspace}
\newcommand*{\eg}{\emph{e.g.}\@\xspace}
\newcommand*{\ie}{\emph{i.e.}\@\xspace}
\newcommand{\nvec}[1]{\bar{\bm{#1}}}

\usepackage{color}
\usepackage{CJKutf8}
\usepackage{ifthen}
\newcommand{\COMM}[2]{{
\begin{CJK}{UTF8}{ipxm}
\ifthenelse{\equal{#1}{TT}}{\color{blue}}{
\ifthenelse{\equal{#1}{TM}}{\color{red}}{
\ifthenelse{\equal{#1}{AA}}{\color{cyan}}{
\ifthenelse{\equal{#1}{BB}}{\color{magenta}}}}}
[#1: #2]
\end{CJK}
}}

\makeatletter
\def\tabcaption{\def\@captype{table}\caption}
\def\figcaption{\def\@captype{figure}\caption}
\makeatother
\begin{document}

\twocolumn[
%\icmltitle{Neural Photometric Stereo Reconstruction for General Reflectance Surfaces}
\icmltitle{Neural Inverse Rendering for General Reflectance Photometric Stereo}



\icmlsetsymbol{equal}{*}

\begin{icmlauthorlist}
\icmlauthor{Tatsunori Taniai}{to}
\icmlauthor{Takanori Maehara}{to}
\end{icmlauthorlist}

\icmlaffiliation{to}{RIKEN Center for Advanced Intelligence Project (RIKEN AIP), Nihonbashi, Tokyo, Japan}
\icmlcorrespondingauthor{Tatsunori Taniai}{tatsunori.taniai@riken.jp}
%\icmlcorrespondingauthor{Takanori Maehara}{takanori.maehara@riken.jp}

\icmlkeywords{photometric stereo, physics-based vision, surface normal estimation, reflectance analysis, BRDF}

\vskip 0.3in
]



\printAffiliationsAndNotice{}  % leave blank if no need to mention equal contribution

\begin{abstract}
We present a novel convolutional neural network architecture for photometric stereo~\cite{Woodham80}, a problem of recovering 3D object surface normals from multiple images observed under varying illuminations. Despite its long history in computer vision, the problem still shows fundamental challenges for surfaces with unknown general reflectance properties (BRDFs). 
Leveraging deep neural networks to learn complicated reflectance models is promising, but studies in this direction are very limited due to difficulties in  acquiring accurate ground truth for training and also in designing networks invariant to permutation of input images. In order to address these challenges, we propose a physics based unsupervised learning framework where surface normals and BRDFs are predicted by the network and fed into the rendering equation to synthesize observed images. The network weights are optimized during testing by minimizing reconstruction loss between observed and synthesized images. Thus, our learning process does not require ground truth normals or even pre-training on external images. Our method is shown to achieve the state-of-the-art performance on a challenging real-world scene benchmark.
\end{abstract}

\begin{figure}[t]
\centering
\includegraphics[width=\linewidth]{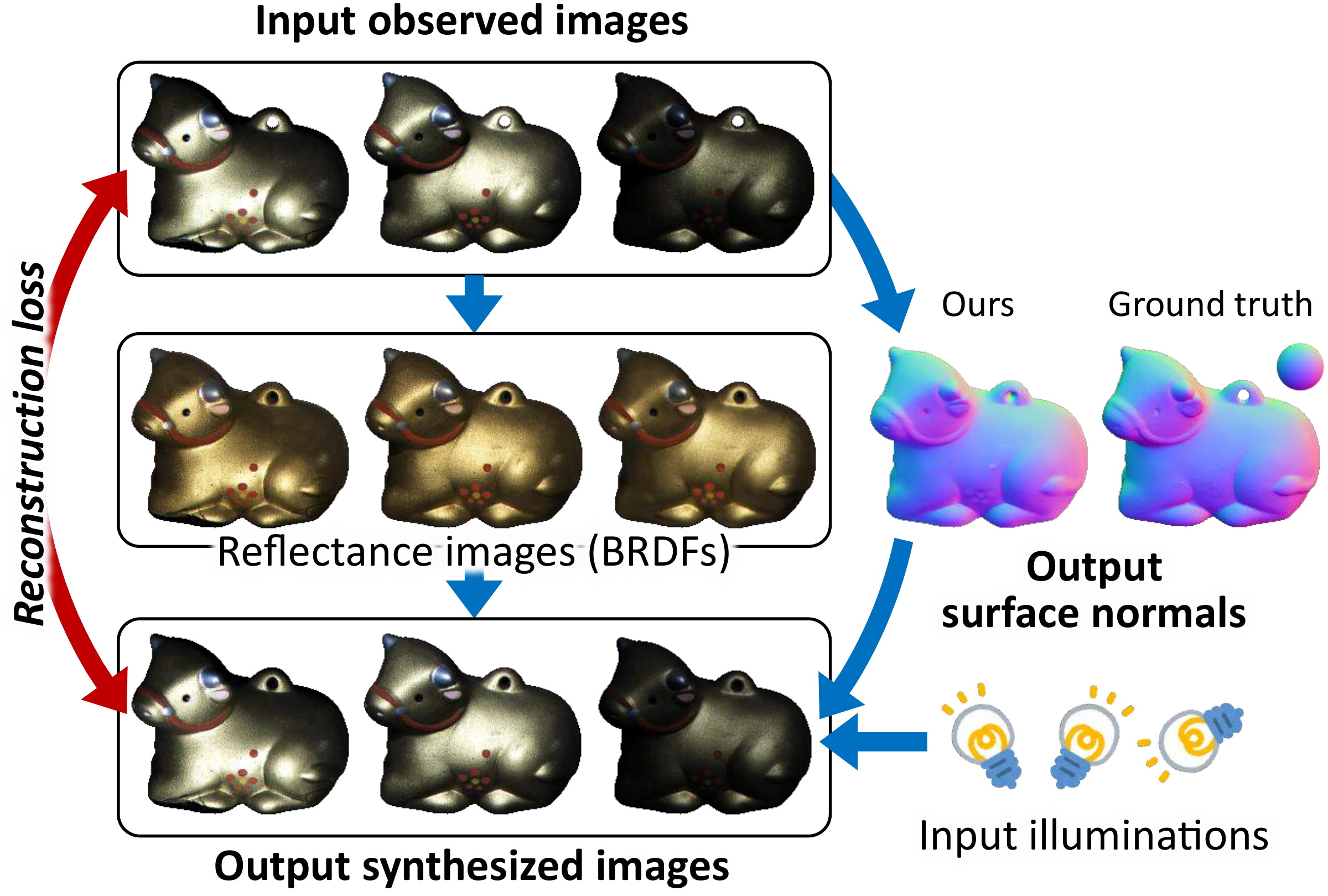}\\
\caption{\textbf{Reconstruction based photometric stereo.}
Given multiple images observed under varying illuminations, our inverse-rendering network estimates a surface normal map and reflectance images. We then reconstruct (or render) the observed images using these estimates and input illuminations. The synthesized images are used to define reconstruction loss for unsupervised learning.}
\label{fig:teaser}
\end{figure}

\section{Introduction}
\label{sec:introduction}
3D shape recovery from images is a central problem in computer vision.
While geometric approaches such as binocular~\cite{Kendall17,Taniai18} and multi-view stereo~\cite{Furukawa10} use images from different viewpoints to triangulate 3D points, photometric stereo~\cite{Woodham80} uses varying shading cues of multiple images to recover 3D surface normals. It is well known that photometric methods prevail in recovering fine details of surfaces, and play an essential role for highly accurate 3D shape recovery in combined approaches~\cite{Nehab05,Esteban08,Park17}. 
Although there exists a closed-form least squares solution to the simplest Lambertian surfaces, such ideally diffuse materials rarely exist in the real word. Photometric stereo for surfaces with unknown general reflectance properties (\ie, bidirectional reflectance distribution functions or BRDFs) still remains as a fundamental challenge~\cite{Shi18}.

%For non-Lambertian materials, some methods use robust statistical techniques to reject non-diffuse components as sparse outliers to the Lambertian model~[Ikehata12,Wu,Bo17]. However, they have limited performances to broad and soft specular reflectance~[Shi18]. Other methods use more realistic reflectance models or 
Meanwhile, deep learning technologies have drastically pushed the envelope of state-of-the-art in many computer vision tasks such as image recognition~\cite{Alex12,He15,He16}, segmentation~\cite{He17} and stereo vision~\cite{Kendall17}. As for photometric stereo, it is promising to replace hand-crafted reflectance models with deep neural networks to learn complicated BRDFs. However, studies in this direction so far are surprisingly limited~\cite{Santo17,Geoffroy18}. This is possibly due to difficulties of making a large amount of training data with ground truth. Accurately measuring surface normals of real objects is very difficult, because we need highly accurate 3D shapes to reliably compute surface gradients. In fact, a real-world scene benchmark of photometric stereo with ground truth has only recently been introduced by precisely registering laser-scanned 3D meshes onto 2D images~\cite{Shi18}. Using synthetic training data is possible~\cite{Santo17}, but we need photo-realistic rendering that should ideally account for various realistic BRDFs and object shapes, spatially-varying BRDFs and materials, presence of cast shadows and interreflections,~etc.
This is more demanding than training-data synthesis for stereo and optical flow~\cite{Mayer16} where rendering by the simplest Lambertian reflectance often suffices.
Also, measuring BRDFs of real materials requires efforts and an existing BRDF database~\cite{Matusik03} provides only a limited number of materials.

As another difficulty of applying deep learning to photometric stereo, when networks are pre-trained, they need to be invariant to permutation of inputs, \ie, permuting input images (and corresponding illuminations) should not change the resulting  surface normals. Existing neural network methods~\cite{Santo17} avoid this problem by assuming the same illumination patterns throughout training and testing phases, which limits application scenarios of methods.

In this paper, we propose a novel convolutional neural network (CNN) architecture for general BRDF photometric stereo. Given observed images and corresponding lighting directions, our network \emph{inverse renders} surface normals and spatially-varying BRDFs from the images, which are further fed into the reflectance (or rendering) equation to synthesize observed images (see Fig.~\ref{fig:teaser}). The network weights are optimized by minimizing reconstruction loss between observed and synthesized images, enabling unsupervised learning that does not use ground truth normals. Furthermore, learning is performed directly on a test scene during the testing phase without any pre-training. Therefore, the permutation invariance problem does not matter in our framework.
Our method is evaluated on a challenging real-world scene benchmark~\cite{Shi18} and is shown to outperform state-of-the-art learning-based~\cite{Santo17} and other classical unsupervised methods~\cite{Shi14,Shi12,Ikehata14,Ikehata12,Wu10,Goldman10,Higo10,Alldrin08}.
We summarize the advantages of our method as follows.
\setlist[itemize]{topsep=-3pt}
\begin{itemize}
\setlength{\itemsep}{0pt}    % ブロック間の余白
\setlength{\parskip}{0pt}    % 段落間余白．
\item Existing neural network methods require pre-training using synthetic data, whenever illumination conditions of test scenes change from the trained ones. In contrast, our physics-based approach can directly fit network weights for a test scene in an unsupervised fashion. 
\item Compared to classical physics-based approaches, we leverage deep neural networks to learn complicated reflectance models, rather than manually analyzing and inventing reflectance properties and models.
\item Yet, our physics-based network architecture allows us to exploit prior knowledge about reflectance properties that have been broadly studied in the literature. 
\end{itemize}

\section{Preliminaries}
\label{sec:preliminaries}
\begin{figure}[t]
\centering
\includegraphics[width=0.85\linewidth]{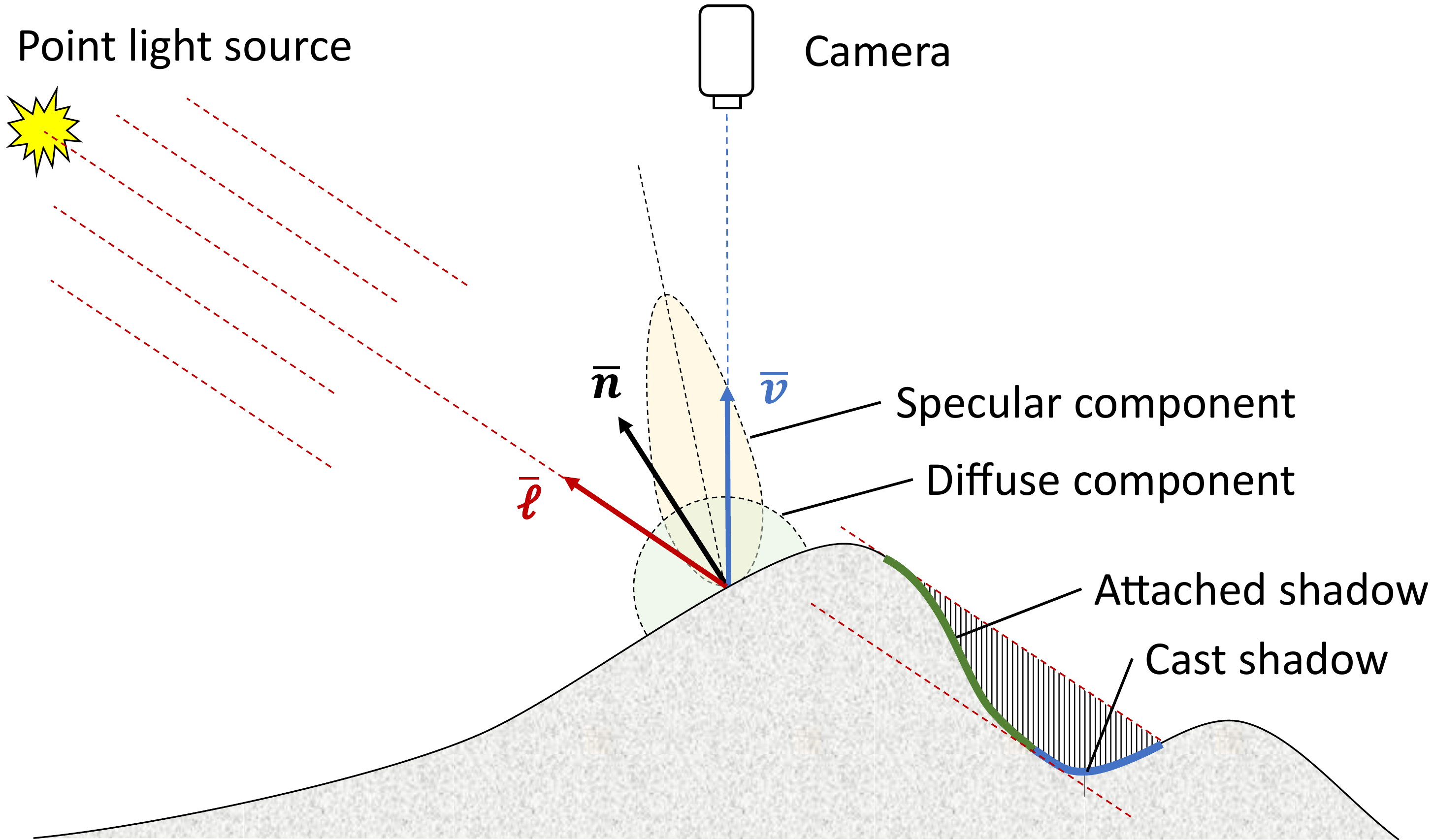}\\
\caption{\textbf{Surface reflectance and BRDFs.} We illustrate a situation where an object surface point with a normal vector $\nvec{n}$ is illuminated by an infinitely distant point light source in a direction $\nvec{\ell}$, and is observed by a camera in a view direction $\nvec{v}$. Unknown BRDFs have major components of diffuse and specular reflections. Shadows occur at surfaces where $\nvec{\ell}^T \nvec{n} \le 0$ (attached shadow) or the light ray is occluded by objects (cast shadow).}
\label{fig:reflectance}
\end{figure}

Before presenting our method, we recap basic settings and approaches in photometric stereo.
Suppose a reflective surface with a unit normal vector $\nvec{n} \in \mathbb{R}^3$ is illuminated by a point light source $\bm{\ell}\in \mathbb{R}^3$ (where $\bm{\ell} = \ell \cdot \nvec{\ell}$ has an intensity $\ell > 0$ and a unit direction $\nvec{\ell}$), without interreflection and ambient lighting.  When this surface is observed by a linear-response camera in a view direction $\nvec{v} \in \mathbb{R}^3$, its pixel intensity $I \in \mathbb{R}_+$ is determined as follows.
\begin{equation}
I = s\rho(\nvec{n}, \nvec{\ell}, \nvec{v}) \max(\bm{\ell}^T \nvec{n}, 0) \label{eq:reflectance}
\end{equation}
Here, $s\in\{0, 1\}$ is a binary function for the presence of a cast shadow, $\rho(\nvec{n}, \nvec{\ell}, \nvec{v})$ is a BRDF, and $\max(\cdot, 0)$ represents an attached shadow. Figure~\ref{fig:reflectance} illustrates this situation.

The goal of photometric stereo is to recover the surface normal $\nvec{n}$ from intensities $I$, when changing illuminations~$\bm{\ell}$. Here, we usually assume a camera with a fixed viewpoint and an orthogonal projection model, in which case the view direction $\nvec{v}$ is constant and typically $\nvec{v}=(0,0,1)^T$.
Also, light sources are assume to be infinitely distant so that $\bm{\ell}$ is uniform over the entire object surfaces.

\subsection{Lambertian model and least squares method}
\label{sec:lambert}
When the BRDF $\rho(\nvec{n}, \nvec{\ell}, \nvec{v})$ is constant as $\rho_0$, the surface is purely diffuse. Such a model is called the Lambertian reflectance and the value $\rho_0$ is called {albedo}. In this case,  estimation of $\nvec{n}$ is relatively easy, because for bright pixels ($I > 0$) the reflectance equation of Eq.~(\ref{eq:reflectance}) becomes a linear equation: $I =\bm{\ell}^T \bm{n}$ 
%\begin{equation}
%I =\bm{\ell}^T \bm{n},
%\label{eq:lambert}
%\end{equation}
where $\bm{n} = \rho_0 \nvec{n}$. 
Therefore, if we know at least three intensity measurements $\bm{I}\in \mathbb{R}^M_+$ ($M \ge 3$) and their lighting conditions $\bm{L}=[\bm{\ell}_1, \bm{\ell}_2, \ldots, \bm{\ell}_M]^T \in \mathbb{R}^{M\times 3}$, then we obtain a linear system
% \begin{equation}
% \underbrace{
% \left[
%     \begin{array}{c}
%       I_1 \\
%       I_2 \\
%       \vdots \\
%       I_M
%     \end{array}
%   \right]
% }_{\bm{I}} = 
% \underbrace{\left[
%     \begin{array}{c}
%       \bm{\ell}_1^T \\
%       \bm{\ell}_2^T \\
%       \vdots \\
%       \bm{\ell}_M^T
%     \end{array}
%   \right]
% }_{\bm{L}}
% \bm{n},
% \label{eq:linear}
% \end{equation}
\begin{equation}
\bm{I} = \bm{L}\bm{n},
\label{eq:linear}
\end{equation}
which is solved by least squares as
\begin{equation}
\bm{n} = \bm{L}^\dagger \bm{I}. \label{eq:ls}
\end{equation}
Here, $\bm{L}^\dagger$ is the pseudo inverse of $\bm{L}$, and the resulting vector $\bm{n} = \rho_0 \nvec{n}$ is then L2-normalized to obtain the unit normal~$\nvec{n}$.

In practice, images are contaminated as $\bm{I} + \bm{e}$ due to sensor noises, interreflections, etc. Therefore, we often set a threshold $\tau$ for selecting inlier observation pixels $I_i > \tau$.

When the lighting conditions $\bm{L}$ are unknown, the problem is called \emph{uncalibrated photometric stereo}. It is known that the problem has the so-called \emph{bas-relief ambiguity}~\cite{Belhumeur99}, and is difficult even for the Lambertian surfaces. In this paper, we focus on the \emph{calibrated photometric stereo} settings that assume known lighting conditions.

\subsection{Photometric stereo for general BRDF surfaces}
When the BRDF $\rho(\nvec{n}, \nvec{\ell}, \nvec{v})$ has unknown non-Lambertian properties, photometric stereo  becomes very challenging, because we essentially need to know the form of the BRDF $\rho(\nvec{n}, \nvec{\ell}, \nvec{v})$ by assuming some reflectance model to it or by directly estimating $\rho(\nvec{n}, \nvec{\ell}, \nvec{v})$ along with the surface normal~$\nvec{n}$. Below we briefly review existing such approaches and their limitations. For more comprehensive reviews, please refer to a recent excellent survey by \citet{Shi18}.

\paragraph{Ourlier rejection based methods.}
A group of methods treat non-Lambertian reflectance components including specular highlights and shadows as outliers to the Lambertian model. Thus, Eq.~(\ref{eq:linear}) is rewritten to
\begin{equation}
\bm{I} = \bm{L} \bm{n} + \bm{e}, \label{eq:outliers}
\end{equation}
where non-Gaussian outliers $\bm{e}$ are assume to be sparse. Recent methods solve this sparse regression problem by using robust statistical techniques~\cite{Wu10,Ikehata12} or using learnable optimization networks~\cite{XinW0WW16,HeXIW17}.
However, this approach cannot handle broad and soft specularity due to the collapse of the sparse outlier assumption~\cite{Shi18}.

\paragraph{Analytic BRDF models.} Another type of methods use more realistic BRDF models than the Lambertian model matured in the computer graphics literature, \eg,  the Torrance-Sparrow model~\cite{Georghiades03}, the Ward model~\cite{Chung08}, or a Ward mixture model~\cite{Goldman10}. These models explicitly consider specularity rather than treating it as outliers, and often take a form of the sum of diffuse and specular components as follows.
\begin{equation}
\rho(\nvec{n}, \nvec{\ell}, \nvec{v})  = \rho_\text{diff} + \rho_\text{spec}(\nvec{n}, \nvec{\ell}, \nvec{v}) \label{eq:analytic}
\end{equation}
However, these methods rely on hand-crafted models that can only handle narrow classes of materials.

\paragraph{General isotropic BRDF properties.} 
More advanced methods directly estimate the unknown BRDF $\rho(\nvec{n}, \nvec{\ell}, \nvec{v})$ by exploiting some general BRDF properties.
For example, many materials have an isotopic BRDF that only depends on relative angles between $\nvec{n}$, $\nvec{\ell}$ and $\nvec{v}$. Given the isotropy, \citet{Ikehata14} further assume the following bi-variate BRDF function
\begin{equation}
\rho(\nvec{n}, \nvec{\ell}, \nvec{v})  = \rho(\nvec{n}^T \nvec{\ell}, \nvec{\ell}^T \nvec{v}) \label{eq:ikehata14}
\end{equation}
with monotonicity and non-negativity constraints. 
Similarly, \citet{Shi14} exploit a low-frequency prior of BRDFs and propose a bi-polynomial BRDF:
\begin{equation}
\rho(\nvec{n}, \nvec{\ell}, \nvec{v})  = \sum_{i=0}^k \sum_{j=0}^k C_{ij} x^i y^j, \label{eq:shi14}
\end{equation}
where $x=\nvec{n}^T \nvec{h}$, $y=\nvec{\ell}^T \nvec{h}$, and $\nvec{h} = (\nvec{\ell} + \nvec{v})/\| \nvec{\ell} + \nvec{v}\|_2$.

Our method is close to the last approach in that we learn broad classes of a BRDF from observations without restricting it to a particular reflectance model.
However, unlike those methods that fully rely on careful human analysis of BRDF properties, we leverage the powerful expressibility of deep neural networks to learn general complicated BRDFs.
Yet, our network architecture also explicitly uses the physical reflectance equation of Eq.~(\ref{eq:reflectance}) internally, which allows us to incorporate abundant wisdom about reflectance developed in the literature, into neural network based approaches.
%Furthermore, 

\begin{figure*}[t]
\centering
\includegraphics[width=0.9\linewidth]{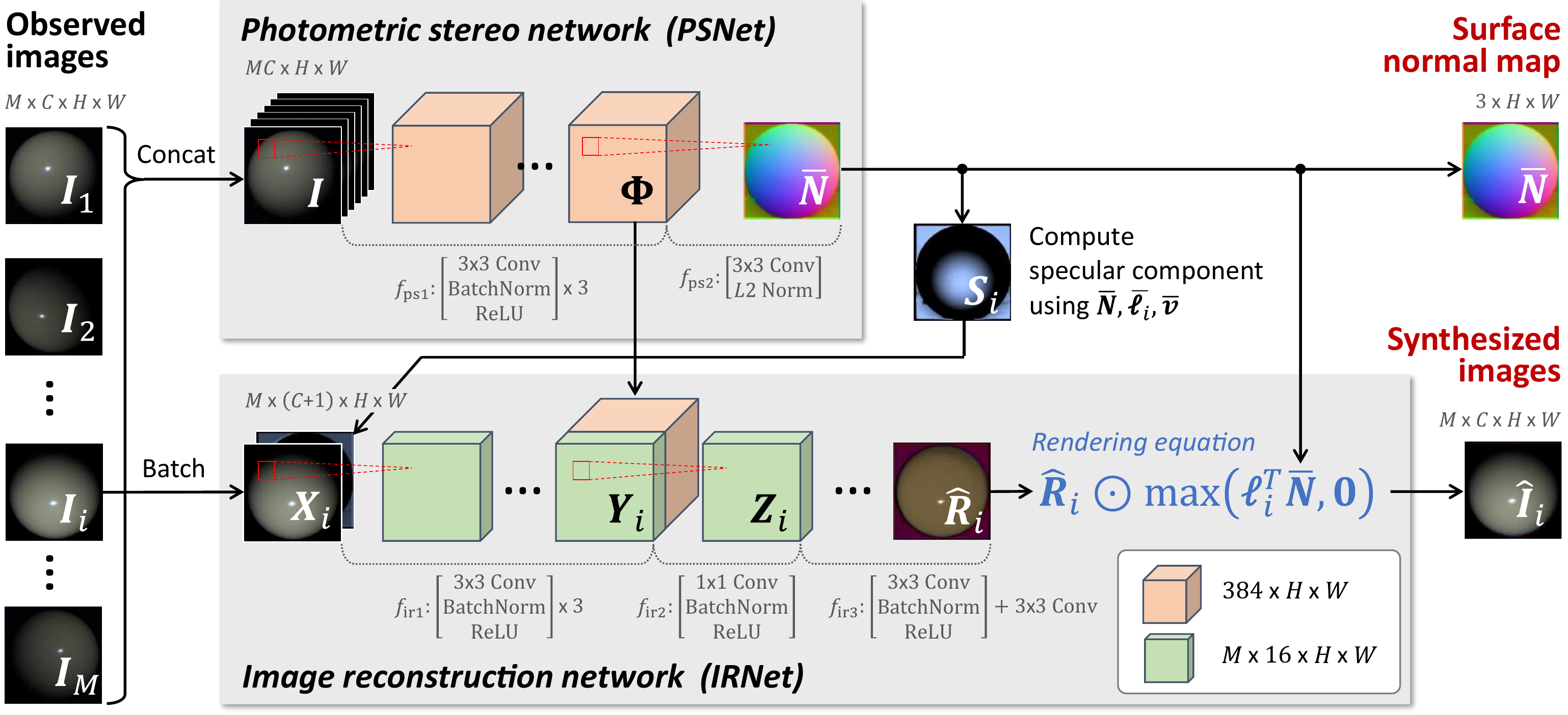}\\
\caption{\textbf{Proposed network architecture for photometric stereo.} 
We use two subnetworks, both are fully convolutional. [\textsc{Top}]~The photometric stereo network (PSNet) outputs a surface normal map $\nvec{N}$ as the desired solution, given an image tensor $\bm{I}$ that concatenates all observed images $\{\bm{I}_1, \bm{I}_2, \ldots, \bm{I}_M\}$ of a test scene. [\textsc{Bottom}]~The image reconstruction network (IRNet) synthesizes each observed image $\bm{I}_i$ using the rendering equation. IRNet is used to define reconstruction loss between the observed and synthesized images for unsupervised learning.
Note that, as calibrated photometric stereo, the lighting directions $\{\bm{\ell}_1, \bm{\ell}_2, \ldots, \bm{\ell}_M\}$  are also provided as inputs, and used for the computations of the rendering equation and a specular component input $\bm{S}_i$. 
Also, dimensionality $B \times D\times H \times W$ of tensors denotes a minibatch size $B$, channel number $D$, and spatial resolution $H \times W$, respectively, where $B=1$ is omitted in PSNet.
}
\label{fig:network}
\end{figure*}

\section{Proposed method}
In this section, we present our novel inverse-rendering based neural network architecture for photometric stereo, and explain its learning procedures with a technique of early-stage weak supervision.
Here, as standard settings of calibrated photometric stereo, we assume $M$ patterns of light source directions $\bm{\ell}_i$ ($i\in\{1,2,\ldots,M\}$) and corresponding image observations $\bm{I}_i$ as inputs. We also assume that the mask $\bm{O}$ of target object regions is provided. Our goal is to estimate the surface normal map $\nvec{N}$ of the target object regions.
%Here, both images and lights can have color channels $C$ (\ie, $\bm{I}_i \in \mathbb{R}_+^{C\times H \times W}$ and $\bm{\ell}_i \in \mathbb{R}^{3\times C}$), but we first assume single gray-scale channels ($C=1$) to simplify discussions. The treatments of color channels are detailed later in this section.

\paragraph{Notations.} We use bold capital letters for tensors and matrices, and bold small letters for vectors. We use tensors of  dimensionality $D\times H\times W$ to represent images, and normal and other feature maps, where $D$ is some channel number and $H\times W$ is the spatial resolution. Thus, $\bm{I}_i \in \mathbb{R}^{C\times H\times W}$ and $\nvec{N} \in \mathbb{R}^{3\times H\times W}$, where $C$ is  the number of color channels of images. We use the subscript $p$ to denote a pixel location of such tensors, \eg, $\nvec{N}_p \in \mathbb{R}^3$ is a normal vector at $p$. The light vectors $\bm{\ell}_i$ can also have color channels, in which case $\bm{\ell}_i \in \mathbb{R}^{3\times C}$ are matrices but we use a small letter for intuitiveness. The index $i$ is always used to denote the observation index $i \in \{1, 2, \ldots, M\}$.
When we use tensors of  dimensionality $B\times D\times H\times W$, the first dimension $B$ denotes a minibatch size processed in one SGD iteration.

\subsection{Network architecture}
We illustrate our network architecture in Fig.~\ref{fig:network}.
Our method uses two subnetworks, which we name the photometric stereo network (PSNet) and image reconstruction network (IRNet), respectively. 
PSNet predicts a surface normal map as the desired output, given the input images. On the other hand,  IRNet synthesizes observed images using the rendering equation of Eq.~(\ref{eq:reflectance}). The synthesized images are used to define reconstruction loss with the observed images,  which produces gradients flowing into both networks and enables learning without ground truth supervision.
We now explain these two networks in more details below.

\subsubsection{Photometric stereo network}
Given a tensor $\bm{I} \in \mathbb{R}^{MC\times H\times W}$ that concatenates all $M$ input images along the channel axis, PSNet first converts it to an abstract feature map $\bm{\Phi} \in \mathbb{R}^{D_\text{ps} \times H\times W}$ as 
\begin{equation}
\bm{\Phi} = f_\text{ps1}(\bm{I}; \bm{\theta}_\text{ps1}), \label{eq:psnet}
\end{equation}
and then outputs a surface normal map $\nvec{N}$ given $\bm{\Phi}$ as 
\begin{equation}
\nvec{N} = f_\text{ps2}(\bm{\Phi}; \bm{\theta}_\text{ps2}).
\end{equation}
Here, $f_\text{ps1}$ is a feed-forward CNN of three  layers with learnable parameters $\bm{\theta}_\text{ps1}$, where each layer applies $3$x$3$ Conv of $D_\text{ps}$ channels, BatchNorm~\cite{Ioffe15}, and ReLU. We use channels of $D_\text{ps} = 384$, and use no skip-connections or pooling. Similarly, $f_\text{ps2}$ applies $3$x$3$ Conv and L2 normalization that makes each $\nvec{N}_p$ a unit vector.

%Then, at the end of PSNet, a surface normal map $\nvec{N}$ is produced given $\bm{\Phi}$ as follows.
%\begin{equation}
%\nvec{N} = f_\text{ps2}(\bm{\Phi}; \bm{\theta}_\text{ps2}).
%\end{equation}
%Here, $f_\text{ps2}$ applies $3$x$3$ Conv and L2 normalization that guarantees that each normal $\nvec{N}_p$ is a unit vector.%, and 

\subsubsection{Image reconstruction network}
\label{sec:irnet}
IRNet synthesizes each observed image $\bm{I}_i$ as $\hat{\bm{I}}_i$ based on the rendering equation of Eq.~(\ref{eq:reflectance}). Specifically, 
IRNet first predicts $R = s \rho(\nvec{n}, \nvec{\ell}, \nvec{v})$, the multiplication of a cast shadow and a BRDF, under a particular illumination~$\bm{\ell}_i$ as 
\begin{equation}
\hat{\bm{R}}_i = f_\text{ir}(\bm{I}_i, \nvec{N}, \nvec{\ell}_i, \nvec{v}, \bm{\Phi}; \bm{\theta}_\text{ir}). \label{eq:brdf_reco}
\end{equation}
Here, we call $\hat{\bm{R}}_i \in \mathbb{R}^{C\times H \times W}$ a reflectance image, which is produced by a CNN 
$f_\text{ir}$ as explained later. Then, IRNet synthesizes each image $\hat{\bm{I}}_i$ by the rendering equation below.
\begin{equation}
\hat{\bm{I}}_i = \hat{\bm{R}}_i \odot \max(\bm{\ell}^T_i \nvec{N}, \bm{0}) \label{eq:reconstruct}
\end{equation}
Here, the inner products between light $\bm{\ell}_i$ and normal vectors $\nvec{N}$ are computed at each pixel $p$ by $\bm{\ell}^T_i \nvec{N}_p$. Note that when $\bm{\ell}_i$ has color channels, we multiply a matrix $\bm{\ell}_i^T \in \mathbb{R}^{C \times 3}$ to $\nvec{N}_p$. Consequently, $\bm{\ell}^T_i \nvec{N}$ and  $\hat{\bm{R}}_i$ have the same dimensions with $\bm{I}$. The $\max(\cdot, \bm{0})$ is done elementwise and is implemented by ReLU, and $\odot$ is elementwise multiplication. 
We now explain details of $f_\text{ir}$ by dividing it  into three parts. 

\paragraph{Individual observation transform.}
The first part transforms each observed image $\bm{I}_i$ (which we denote as $\bm{X}_i$) into a feature map $\bm{Y}_i \in \mathbb{R}^{D_\text{ir}\times H \times W}$ as follows.
\begin{equation}
\bm{Y}_i = f_\text{ir1}(\bm{X}_i; \bm{\theta}_\text{ir1}) \label{eq:ir1}
\end{equation}
The network architecture of $f_\text{ir1}$ is the same with $f_\text{ps1}$ in Eq.~(\ref{eq:psnet}), except that we use channels of $D_\text{ir} = 16$ for $f_\text{ir1}$.
To more effectively learn BRDFs, %we follow ideas of classical analytic approaches by Eq.~(\ref{eq:analytic}) and 
we use an additional specularity channel $\bm{S}_i$ for the input $\bm{X}_i$ as
\begin{equation}
\bm{X}_i = \text{Concat}(\bm{I}_i, \bm{S}_i), \label{eq:xi}
\end{equation}
where $\bm{S}_i \in \mathbb{R}^{1\times H \times W}$ is  computed at each pixel $p$ as 
\begin{equation}
%\bm{D}_{ip} &=& \nvec{\ell}^T_i \nvec{N}_p, \label{eq:di}\\
\bm{S}_{ip} = \nvec{v}^T \nvec{s}_{ip} = \nvec{v}^T \left[ 2(\nvec{\ell}^T_i\nvec{N}_p) \nvec{N}_p - \nvec{\ell}_i \right]. \label{eq:spec}
\end{equation}
Here, $\nvec{s}_{ip}$ is the direction of the specular reflection (dashed line between $\nvec{n}$ and $\nvec{v}$ in Fig.~\ref{fig:reflectance}).
It is well known by past studies that $\bm{S}_{ip}$ is highly correlated with the actual  specular component of a BRDF. Therefore, directly giving it as a hint to the network will promote learning of complex BRDFs. %Note that a diffuse component is constant in $\rho(\nvec{n},\nvec{\ell},\nvec{v})$ and is thus omitted from the addition to $\bm{X}_i$. 

\paragraph{Global observation blending.}
Because $\bm{Y}_i$ has limited observation information under a particular illumination~$\bm{\ell}_i$, we enrich it by $\bm{\Phi}$ in Eq.~(\ref{eq:psnet}) that has more comprehensive information of the scene.
We do this similarly to global and local feature blending in~\cite{Charles17,Iizuka16} as
\begin{equation}
\bm{Z}_i = f_\text{ir2}(\text{Concat}(\bm{Y}_i, \bm{\Phi}); \bm{\theta}_\text{ir2}), \label{eq:ir2}
\end{equation}
where $f_\text{ir2}$ applies $1$x$1$ Conv, BatchNorm, and ReLU.
Note that applying Conv to $\text{Concat}(\bm{Y}_i, \bm{\Phi})$ is efficiently done as $\bm{W}_1 \bm{Y}_i + \bm{W}_2 \bm{\Phi} + \bm{b}$ where Conv of $\bm{W}_2 \bm{\Phi} + \bm{b}$ is computed only once and reused for all observations $i$.

\paragraph{Output.}
After the blending, we finally output $\hat{\bm{R}}_i$ by 
\begin{equation}
\hat{\bm{R}}_i = f_\text{ir3}(\bm{Z}_i; \bm{\theta}_\text{ir3}), \label{eq:ir3}
\end{equation}
where $f_\text{ir3}$ is 3x3 Conv, BatchNorm, ReLU, and 3x3 Conv. As explained in Eq.~(\ref{eq:reconstruct}), the resulting $\hat{\bm{R}}_i$ is used to reconstruct each image as $\hat{\bm{I}}_i$, which is the final output of IRNet.

Note that the internal channels of IRNet are all the same as $D_\text{ir}$. %, except for the last output Conv layer that has $C$ channels.
Also, IRNet simultaneously reconstructs all images during SGD iterations, by treating them as a minibatch: $\hat{\bm{I}} = \text{Batch}(\hat{\bm{I}}_1, \hat{\bm{I}}_2, \ldots, \hat{\bm{I}}_M) \in \mathbb{R}^{M \times C \times H \times W}$. This learning procedure is more explained in the next section.

\subsection{Learning procedures (optimization)}
We optimize the network parameters $\bm{\theta}$ by minimizing the following loss function using SGD.
\begin{equation}
L = L_\text{rec}(\hat{\bm{I}}, \bm{I}) + \lambda_t L_\text{prior}(\nvec{N}, \nvec{N}')
\label{eq:loss}
\end{equation}
The first term defines reconstruction loss between the synthesized $\hat{\bm{I}}$ and observed images $\bm{I}$, which is explained in Sec.~\ref{sec:rec_loss}.
The second term defines weak supervision loss between the predicted  $\nvec{N}$ and some prior normal map $\nvec{N}'$. This term is only activated in early iterations of SGD (\ie, $\lambda_{t} = 0$ when $t>T$) in order to warm up randomly initialized networks and stabilize the learning. This is more explained in Sec.~\ref{sec:weak_sup}.
Other implementation details and hyper-parameter settings are described in Sec.~\ref{sec:details}.

Most importantly, the network is directly fit for a particular test scene without any pre-training on other data, by updating the network parameters $\bm{\theta}$ over SGD iterations. Final results are obtained at convergence.

\subsubsection{Reconstruction loss}
\label{sec:rec_loss}
The reconstruction loss is defined as mean absolute errors between $\hat{\bm{I}}$ and $\bm{I}$ over target object regions $\bm{O}$ as
\begin{equation}
L_\text{rec}(\hat{\bm{I}}, \bm{I}) = \frac{1}{MCO} \sum_{i,c,p}  \bm{O}_{p} |\hat{\bm{I}}_{icp} - {\bm{I}_{icp}}|. \label{eq:rec_loss}
\end{equation}
Here, $\bm{O}$ in $\{1, 0\}^{1 \times H \times W}$ is the binary object mask, and $O =  \sum_{p} \bm{O}_{p}$ is its object area size. Using absolute errors increases the robustness to high-intensity specular highlights.

\subsubsection{Early-stage weak supervision}
\label{sec:weak_sup}
If the target scene has relatively simple reflectance properties, the reconstruction loss alone can often lead to a good solution, even starting with randomly initialized networks. However, for complex scenes, we need to warm up the network by adding the following weak supervision.
\begin{equation}
\lambda_t L_\text{prior}(\nvec{N}, \nvec{N}') = \lambda_t\frac{1}{O} \sum_{p}  \bm{O}_{p} \|\nvec{N}_{p} - {\nvec{N}'_{p}}\|_2^2 \label{eq:ls_loss}
\end{equation}
Here, the prior normal map $\nvec{N}'$ is obtained by the simplest least squares method described in Sec.~\ref{sec:lambert} using all observed pixels without any thresholding. Due to the presence of shadows and non-Lambertian specularity, this least squares solution can be very inaccurate. However, even such priors work well in our method, because we only use them to guide the optimization in its early stage.
For this, we  set $\lambda_t$ to $0.1c$ for  initial 50 iterations, and then set it to zero afterwards. The coefficient $c$ is to adaptively balance weights between $L_\text{rec}$ and $L_\text{prior}$, and is computed as the mean intensities of $\bm{I}$ over target object regions, \ie, $c = L_\text{rec}(\bm{0}, \bm{I})$.

\subsubsection{Implementation details}
\label{sec:details}
We use Adam~\cite{Kingma14} as the optimizer. For each test scene, we iterate SGD updates for $1000$ steps. Adam's hyper-parameter $\alpha$ is set to  $\alpha_0 = 8\times 10^{-4}$ for first 900 iterations, and then decreased to $\alpha_0 /10$ for last 100 iterations for fine-tuning. We use the default values for the other hyper-parameters.
The convolution weights are randomly initialized by He initialization~\cite{He15}.

In each iteration, PSNet predicts a surface normal map $\nvec{N}$, and then IRNet reconstructs all observed images $\hat{\bm{I}}$  as samples of a minibatch. Given $\nvec{N}$ and $\hat{\bm{I}}$, we compute the loss $L$ and update the parameters $\bm{\theta}$ of both networks.

When computing the reconstruction loss $L_\text{rec}$ in Eq.~(\ref{eq:rec_loss}), we randomly dropout 90\% of its elements and rescale $L_\text{rec}$ by a factor of 10 instead. This treatment is to compensate for the well known issue of poor local convergence of SGD by the use of a large minibatch~\cite{Keskar17}.

Because we learn network parameters during testing, we always run BatchNorm by the training mode using statistics of given data (\ie, we never use moving-average statistics).

Before being fed into the network, the input images $\bm{I}$ are cropped by a loose bounding box of the target object regions for reducing redundant computations. Then, the  images are normalized by global scaling as
\begin{equation}
\bm{I}' = \bm{I} / (2 \sigma),
\end{equation}
where $\sigma$ is the square-root of mean squared intensities of $\bm{I}$ over target regions. 
%From our initial experiments, this global normalization worked better than channel-wise or image-wise normalization to $\bm{I}$.
For PSNet, the normalized image tensor $\bm{I}'$ is further concatenated with the binary mask~$\bm{O}$ as input.

\begin{table*}[p]
\centering
\caption{\textbf{Comparisons on ten real-world scenes of the DiLiGenT photometric stereo benchmark~\cite{Shi18}.}
We compare our proposed method with ten existing calibrated photometric stereo methods. Here, we show mean angular errors in degrees (\ie, the mean of $\arccos(\nvec{N}^T_p \nvec{N}^\star_p)$ over the object regions using  ground truth normals $\nvec{N}^\star_p$) for ten scenes, and average scores.
Our method achieves best accuracies for all except two scenes of \textsc{goblet} and \textsc{harvest}.
The second best method~\cite{Santo17} also uses a deep neural network, but it requires supervised pre-training on synthetic data and  outperforms the other existing methods only for \textsc{harvest}.
The results of the baseline least squares method are used in our method as prior normals for weak supervision. Since the priors are used only for an early-stage of learning, their low accuracies are not critical to the performance of our method.
Note that, due to a non-deterministic property of our method, its accuracy for each scene is evaluated as the median score of 11 rounds run.
}
\vskip 2mm
\includegraphics[width=\linewidth]{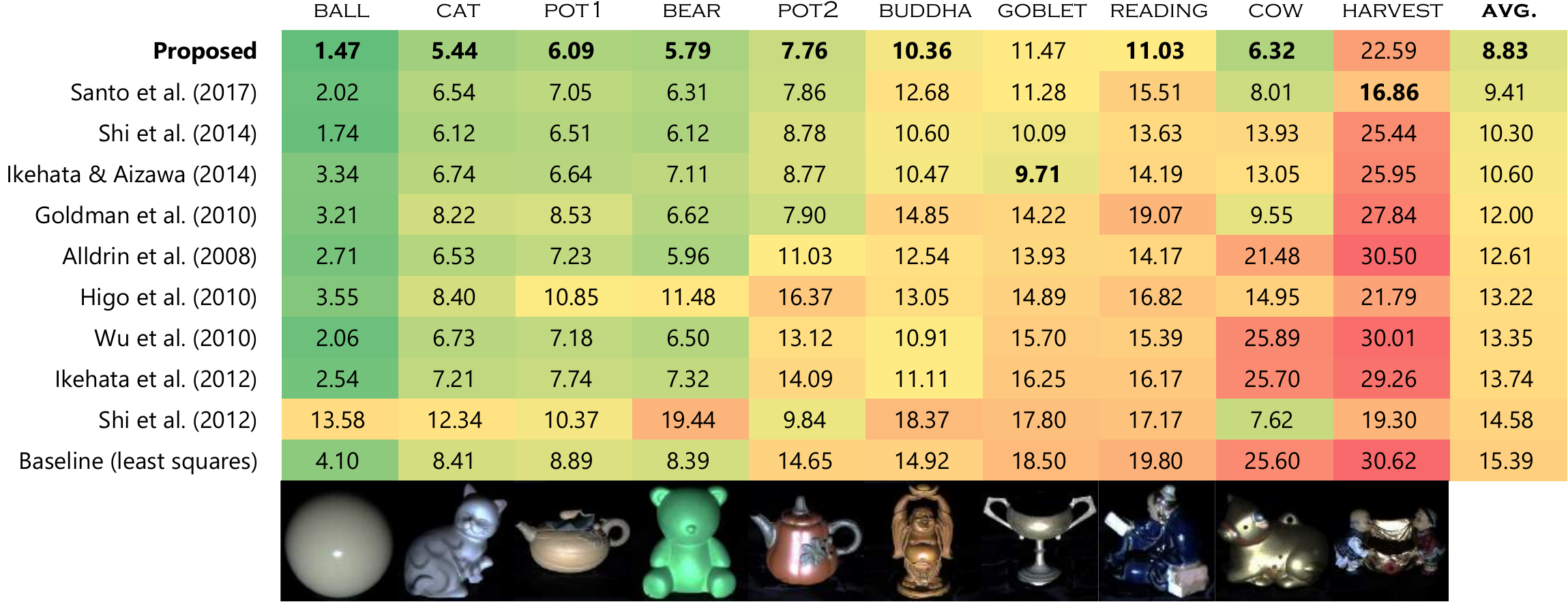}
\label{tab:benchmark}
%\end{table*}
%\begin{figure*}[t]
%\centering
\small
\centering
\begin{minipage}{0.16\linewidth}\centering\small Observed \&\\ synthesized images\end{minipage}\hfil
\begin{minipage}{0.16\linewidth}\centering\small Ground truth /\\ Reconstruction errors\end{minipage}\hfil
\begin{minipage}{0.16\linewidth}\centering\small Ours\end{minipage}\hfil
\begin{minipage}{0.16\linewidth}\centering\small \citet{Santo17}\end{minipage}\hfil
\begin{minipage}{0.16\linewidth}\centering\small \citet{Shi14}\end{minipage}\hfil
\begin{minipage}{0.16\linewidth}\centering\small Baseline\\(least squares)\end{minipage}\\
\vskip 0.5mm
\includegraphics[width=0.15\linewidth]{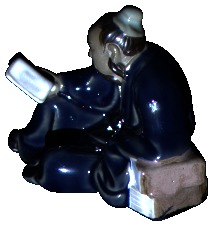}\hfil
\includegraphics[width=0.15\linewidth]{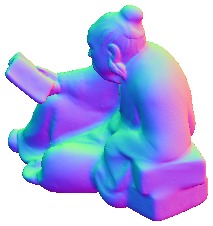}\hfil
\includegraphics[width=0.15\linewidth]{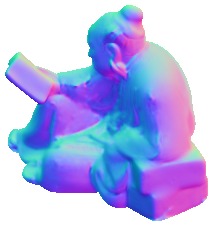}\hfil
\includegraphics[width=0.15\linewidth]{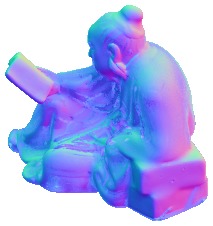}\hfil
\includegraphics[width=0.15\linewidth]{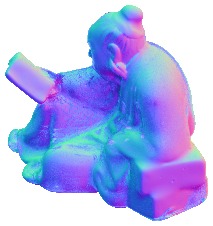}\hfil
\includegraphics[width=0.15\linewidth]{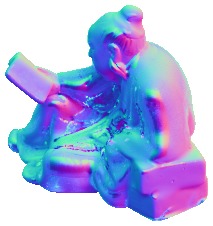}%
\begin{minipage}{0mm}{\vskip -45mm}\hspace*{0mm}\includegraphics[width=5mm]{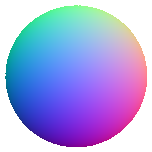}\end{minipage}\\
\includegraphics[width=0.15\linewidth]{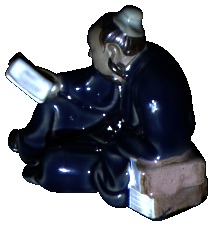}\hfil
\colorbox{bgcolor}{\includegraphics[width=0.15\linewidth]{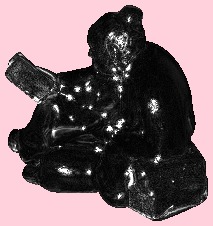}}\hfil
\includegraphics[width=0.15\linewidth]{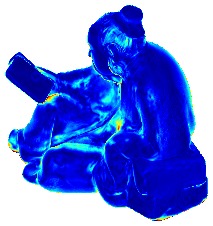}\hfil
\includegraphics[width=0.15\linewidth]{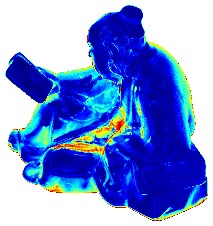}\hfil
\includegraphics[width=0.15\linewidth]{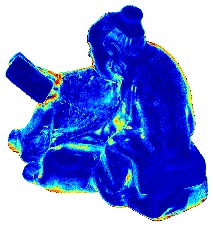}\hfil
\includegraphics[width=0.15\linewidth]{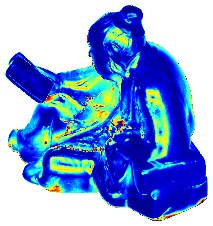}%
\begin{minipage}{0mm}{\vskip -25.7mm}\hspace*{1mm}\includegraphics[width=7.35mm]{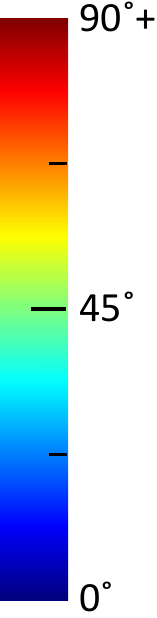}\end{minipage}\\
\begin{minipage}{0.15\linewidth}\centering \textsc{reading}\end{minipage}\hfil
\begin{minipage}{0.15\linewidth}\centering (intensity $\times 10$)\end{minipage}\hfil
\begin{minipage}{0.15\linewidth}\centering \textbf{11.03}\end{minipage}\hfil
\begin{minipage}{0.15\linewidth}\centering 15.51\end{minipage}\hfil
\begin{minipage}{0.15\linewidth}\centering 13.63\end{minipage}\hfil
\begin{minipage}{0.15\linewidth}\centering 19.80\end{minipage}\\
\vskip 3mm
%
%
% Cow
%\includegraphics[width=0.15\linewidth]{figures/results/gt/cowPNG/image}\hfil
%\includegraphics[width=0.15\linewidth]{figures/results/gt/cowPNG/normal}\hfil
%\includegraphics[width=0.15\linewidth]{figures/results/ours/cowPNG/normal}\hfil
%\includegraphics[width=0.15\linewidth]{figures/results/ICCVW17Santo/cowPNG/normal}\hfil
%\includegraphics[width=0.15\linewidth]{figures/results/CVPR12Shi/cowPNG/normal}\hfil
%\includegraphics[width=0.15\linewidth]{figures/results/l2/cowPNG/normal}\\
%%
%\includegraphics[width=0.15\linewidth]{figures/results/ours/cowPNG/image}\hfil
%\includegraphics[width=0.15\linewidth]{figures/results/ours/cowPNG/error_gray}\hfil
%\includegraphics[width=0.15\linewidth]{figures/results/ours/cowPNG/error_jet}\hfil
%\includegraphics[width=0.15\linewidth]{figures/results/ICCVW17Santo/cowPNG/error_jet}\hfil
%\includegraphics[width=0.15\linewidth]{figures/results/CVPR12Shi/cowPNG/error_jet}\hfil
%\includegraphics[width=0.15\linewidth]{figures/results/l2/cowPNG/error_jet}\\
%\begin{minipage}{0.15\linewidth}\centering \textsc{cow}\end{minipage}\hfil
%\begin{minipage}{0.15\linewidth}\centering 0.1 (intensity)\end{minipage}\hfil
%\begin{minipage}{0.15\linewidth}\centering 8.08\end{minipage}\hfil
%\begin{minipage}{0.15\linewidth}\centering 8.01\end{minipage}\hfil
%\begin{minipage}{0.15\linewidth}\centering 13.93\end{minipage}\hfil
%\begin{minipage}{0.15\linewidth}\centering 25.60\end{minipage}
%\vskip 2mm
%
%
% Harvest
\includegraphics[width=0.15\linewidth]{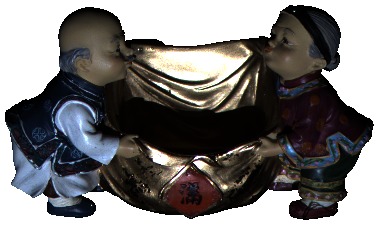}\hfil
\includegraphics[width=0.15\linewidth]{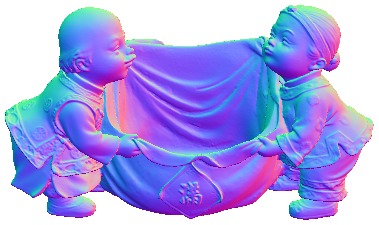}\hfil
\includegraphics[width=0.15\linewidth]{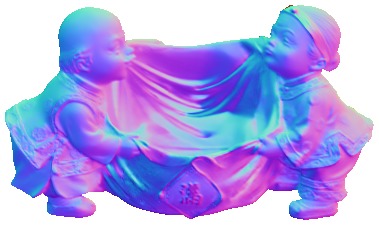}\hfil
\includegraphics[width=0.15\linewidth]{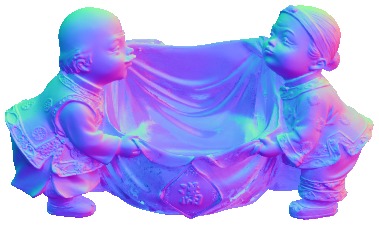}\hfil
\includegraphics[width=0.15\linewidth]{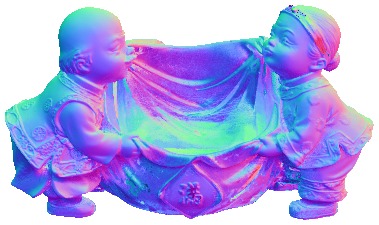}\hfil
\includegraphics[width=0.15\linewidth]{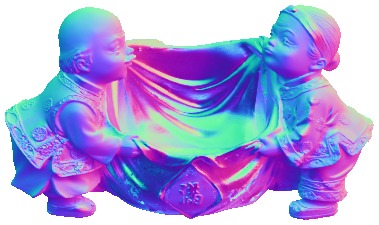}\\
\includegraphics[width=0.15\linewidth]{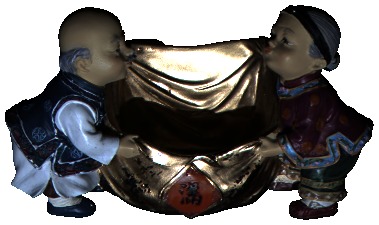}\hfil
\colorbox{bgcolor}{\includegraphics[width=0.15\linewidth]{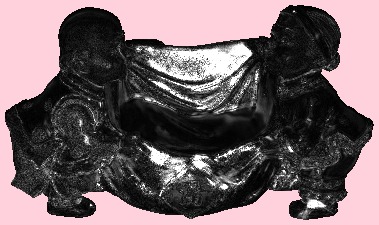}}\hfil
\includegraphics[width=0.15\linewidth]{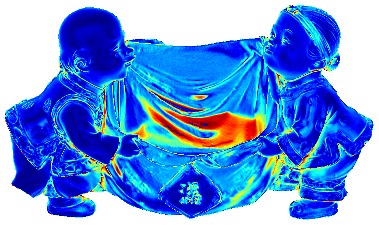}\hfil
\includegraphics[width=0.15\linewidth]{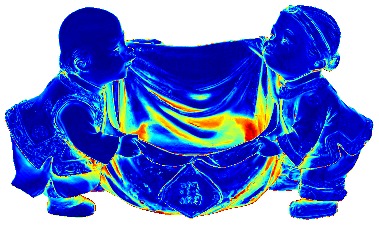}\hfil
\includegraphics[width=0.15\linewidth]{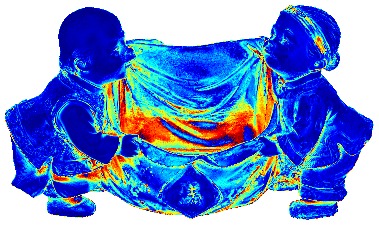}\hfil
\includegraphics[width=0.15\linewidth]{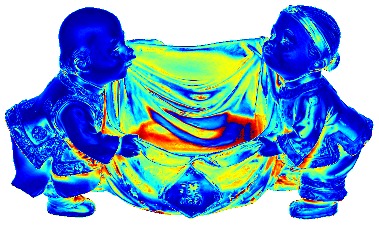}\hfil
\begin{minipage}{0.15\linewidth}\centering \textsc{harvest}\end{minipage}\hfil
\begin{minipage}{0.15\linewidth}\centering (intensity $\times 10$)\end{minipage}\hfil
\begin{minipage}{0.15\linewidth}\centering 22.59\end{minipage}\hfil
\begin{minipage}{0.15\linewidth}\centering \textbf{16.86}\end{minipage}\hfil
\begin{minipage}{0.15\linewidth}\centering 25.44\end{minipage}\hfil
\begin{minipage}{0.15\linewidth}\centering 30.62\end{minipage}\\
\vskip -3mm
\figcaption{\textbf{Visual comparisons for \textsc{reading} and \textsc{harvest} scenes.} From left to right columns in each scene, we show 1) observed and  our synthesized images, 2) ground truth normal and image reconstruction error maps, and 3--6) estimated surface normal and angular error maps by four methods. Numbers under angular error maps show their mean errors. See the supplementary material for more comparisons.
%\COMM{TT}{Our results will be updated later.}
\label{fig:results}
}
%\end{figure*}
\end{table*}

\begin{table*}[t]
	\centering
	\caption{\textbf{Evaluations of the proposed network architecture and weak supervision.}
		For each item we show median and mean scores (in left and right) by 11 rounds run.
		Here, \textbf{S}, \textbf{G} and \textbf{WS} denote the specularity input, global observation blending, and weak supervision by a prior normal map, respectively. Cell colors of red/blue indicate worse/better relative accuracy compared to the proposed settings.
	}
	\vskip 1mm
	\includegraphics[width=0.985\linewidth]{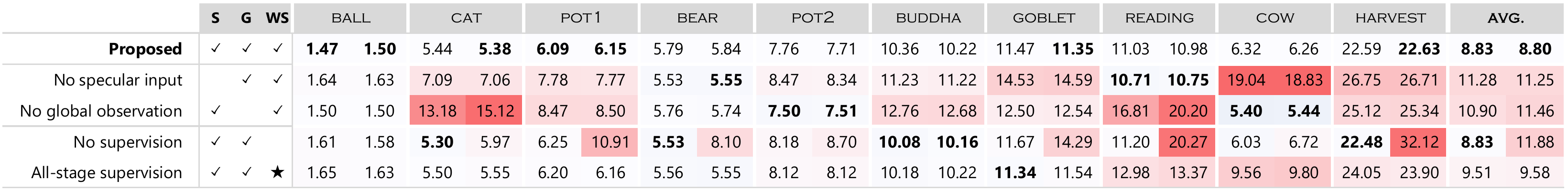}\hfill
	\begin{minipage}{0mm}
		{\vskip -15.8mm}\hspace*{ -2.5mm}\includegraphics[width=6.4mm]{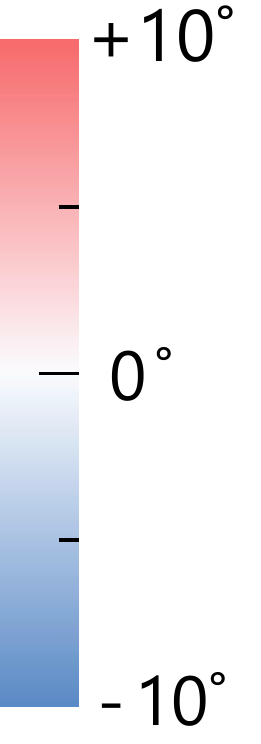}\end{minipage}\\
	\vskip 2mm
    \label{tab:analysis}
\end{table*}

\begin{figure*}[t]
\centering
\small
\def\figw{0.225\linewidth}
% 0:ball, 1:bear, 2:buddha, 3:cat, 4:cow, 5:goblet, 6:harbest, 7:pot1, 8:pot2, 9:reading
%\includegraphics[width=\figw]{figures/plots/m_accuracy_0n}%
%\includegraphics[width=\figw]{figures/plots/m_accuracy_0c}\hfill
\includegraphics[width=\figw]{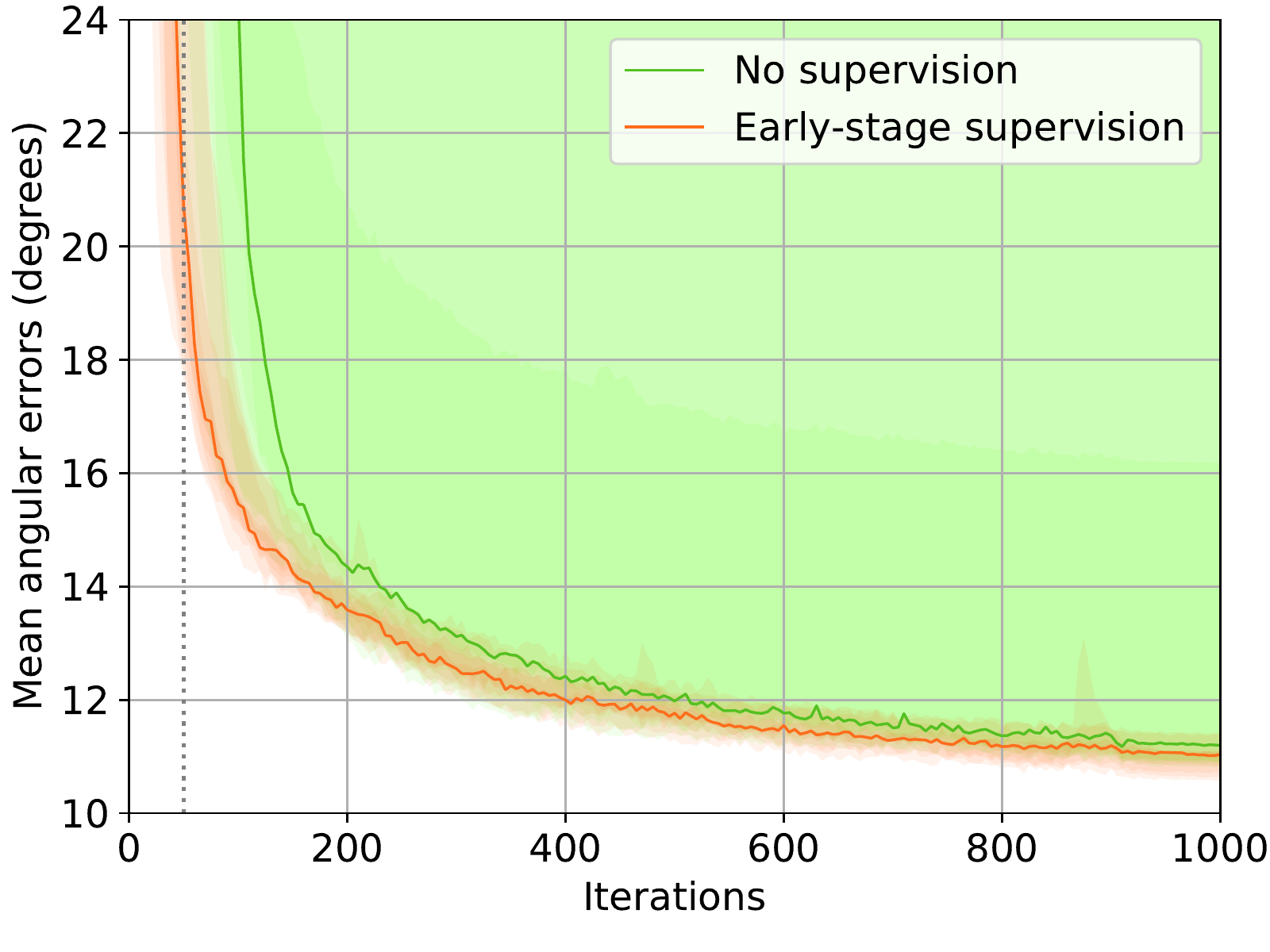}
\includegraphics[width=\figw]{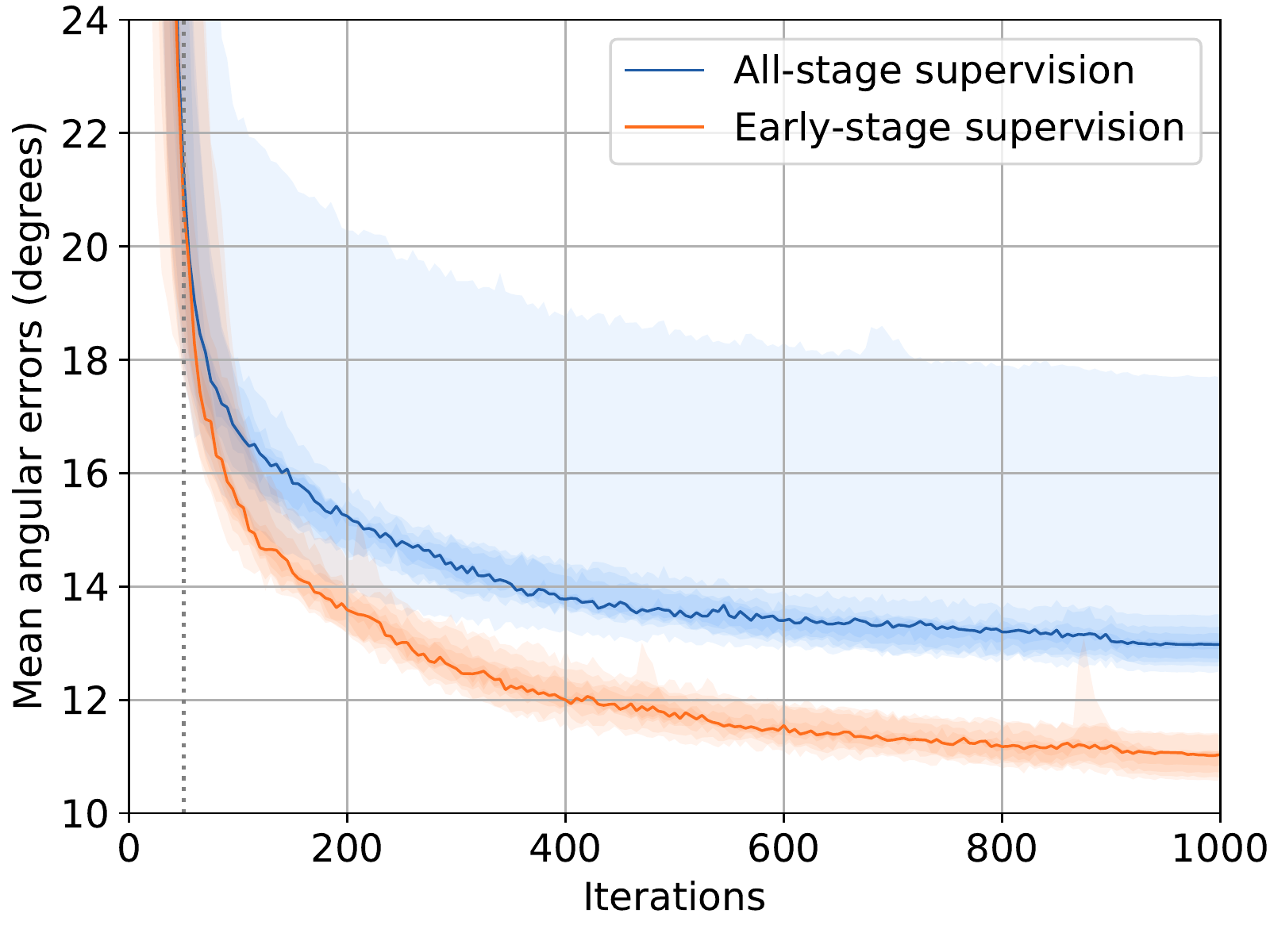}\hfil
\includegraphics[width=\figw]{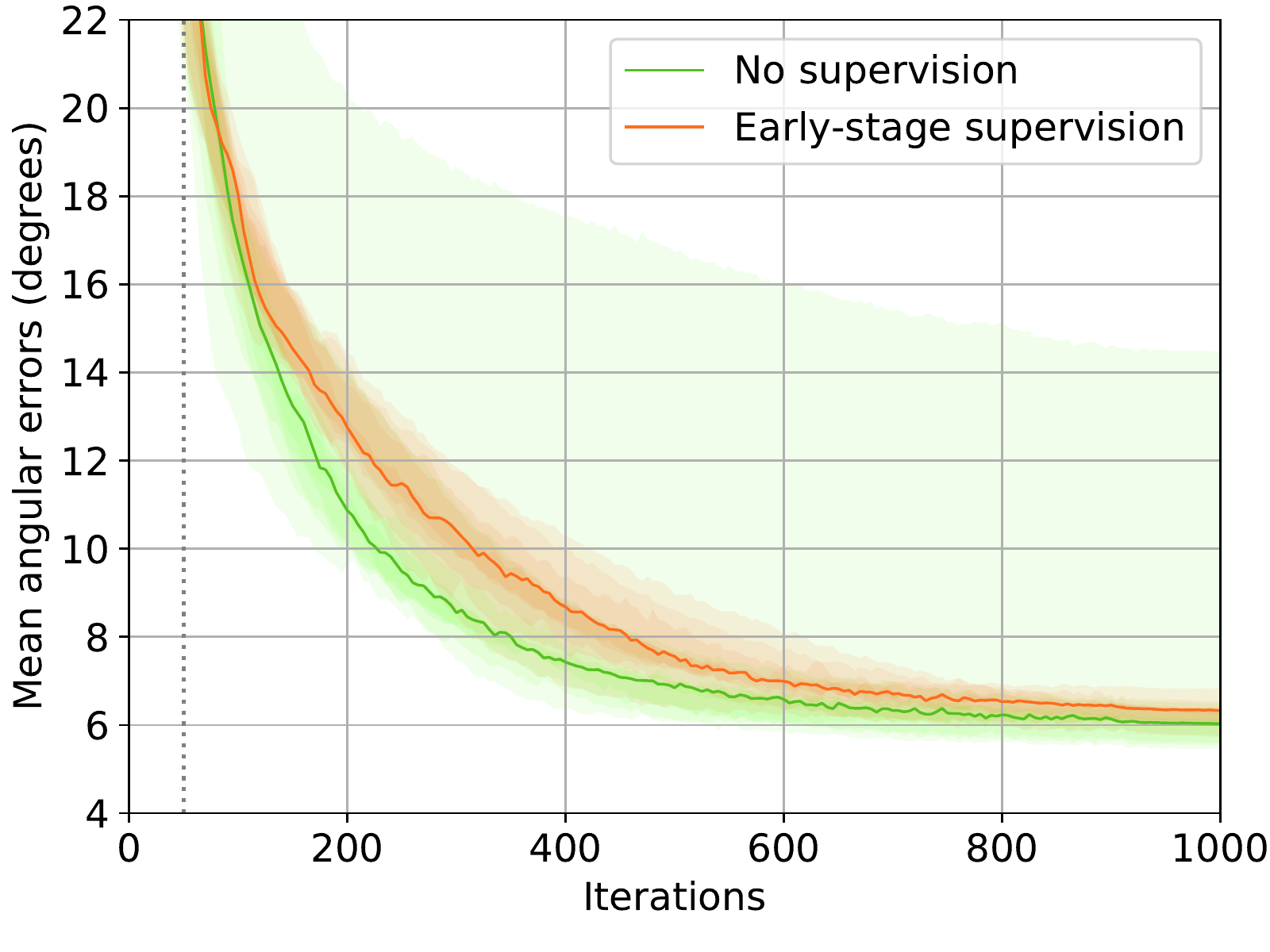}
\includegraphics[width=\figw]{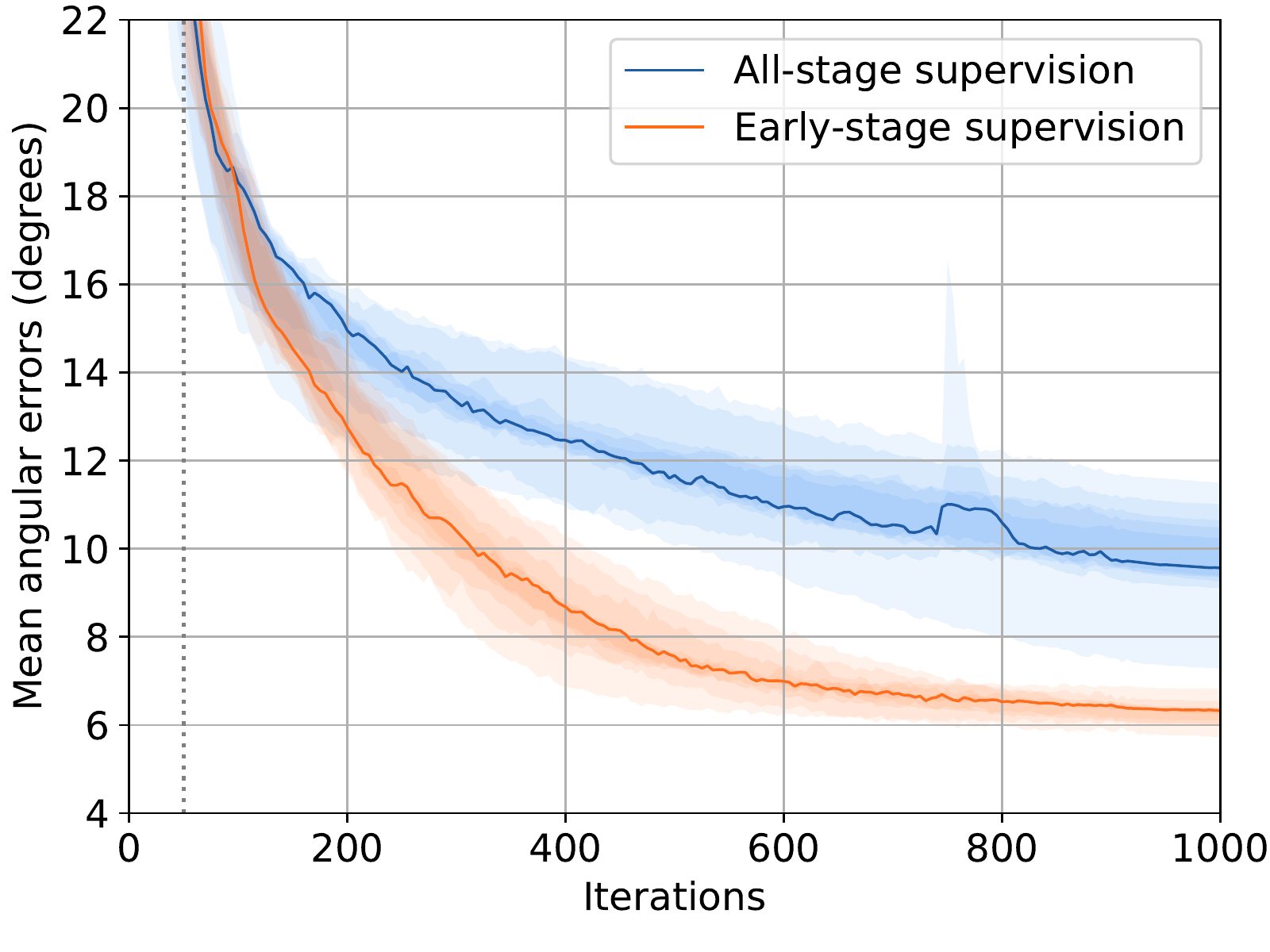}\\
\includegraphics[width=\figw]{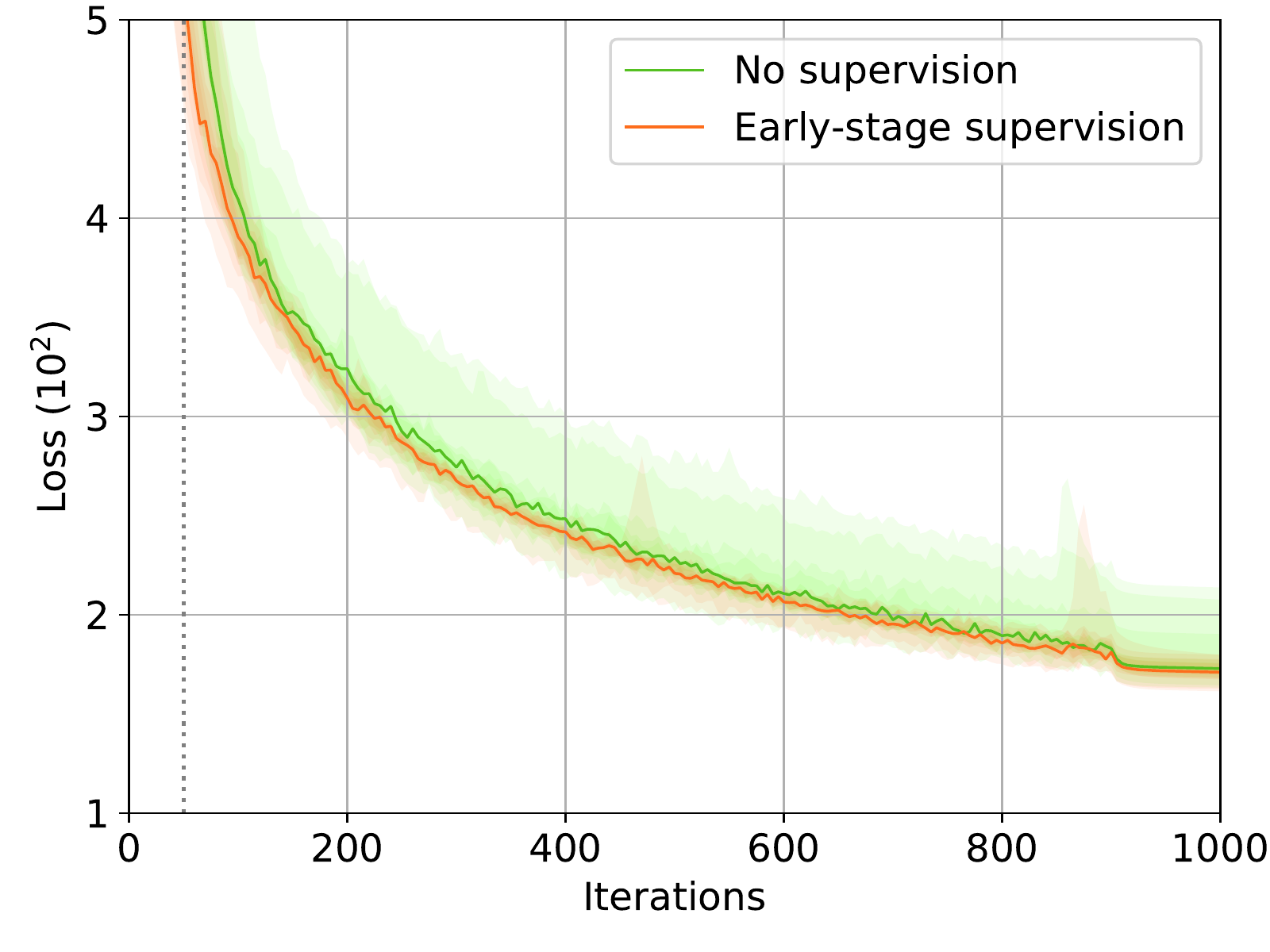}
\includegraphics[width=\figw]{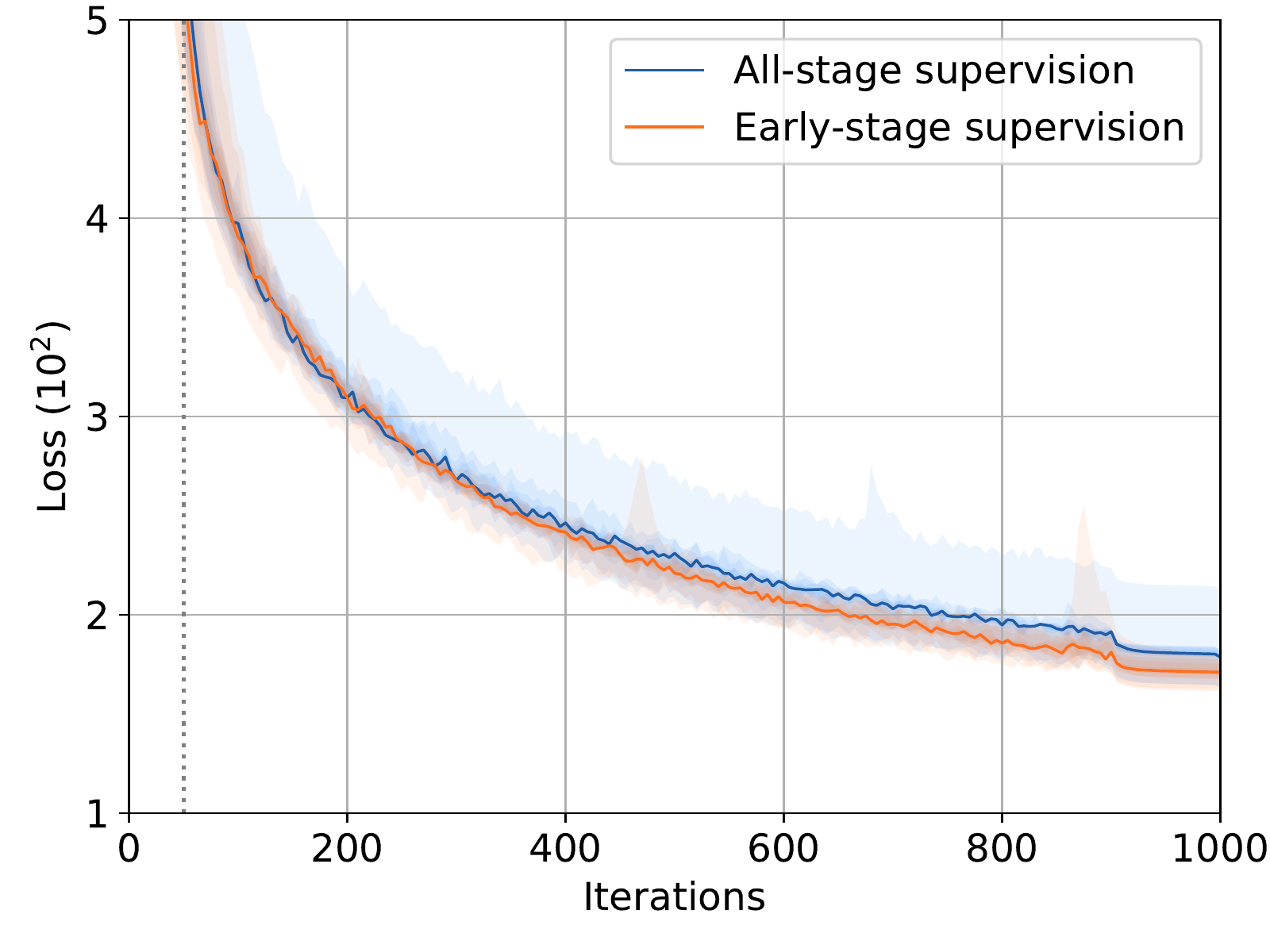}\hfil
\includegraphics[width=\figw]{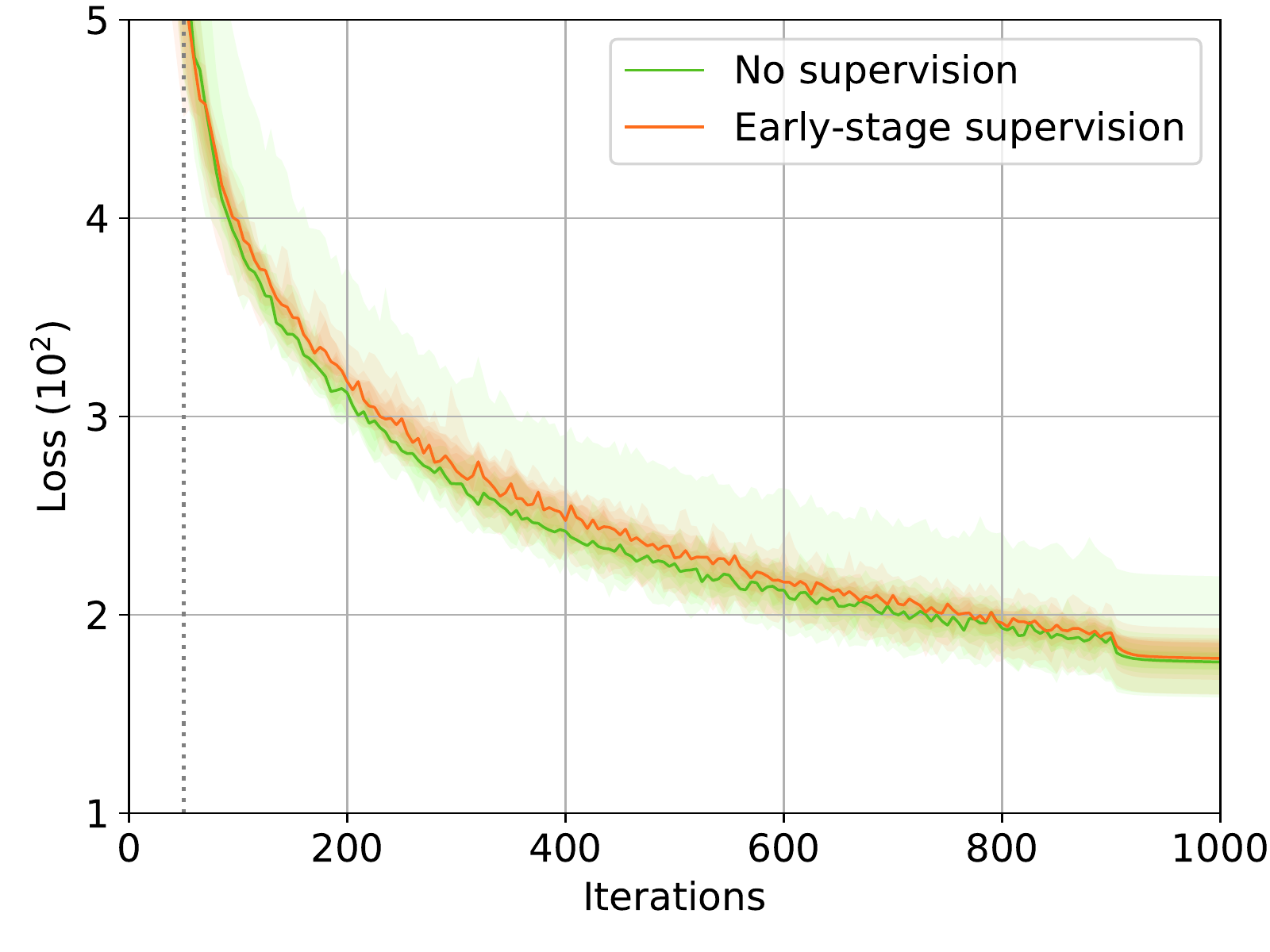}
\includegraphics[width=\figw]{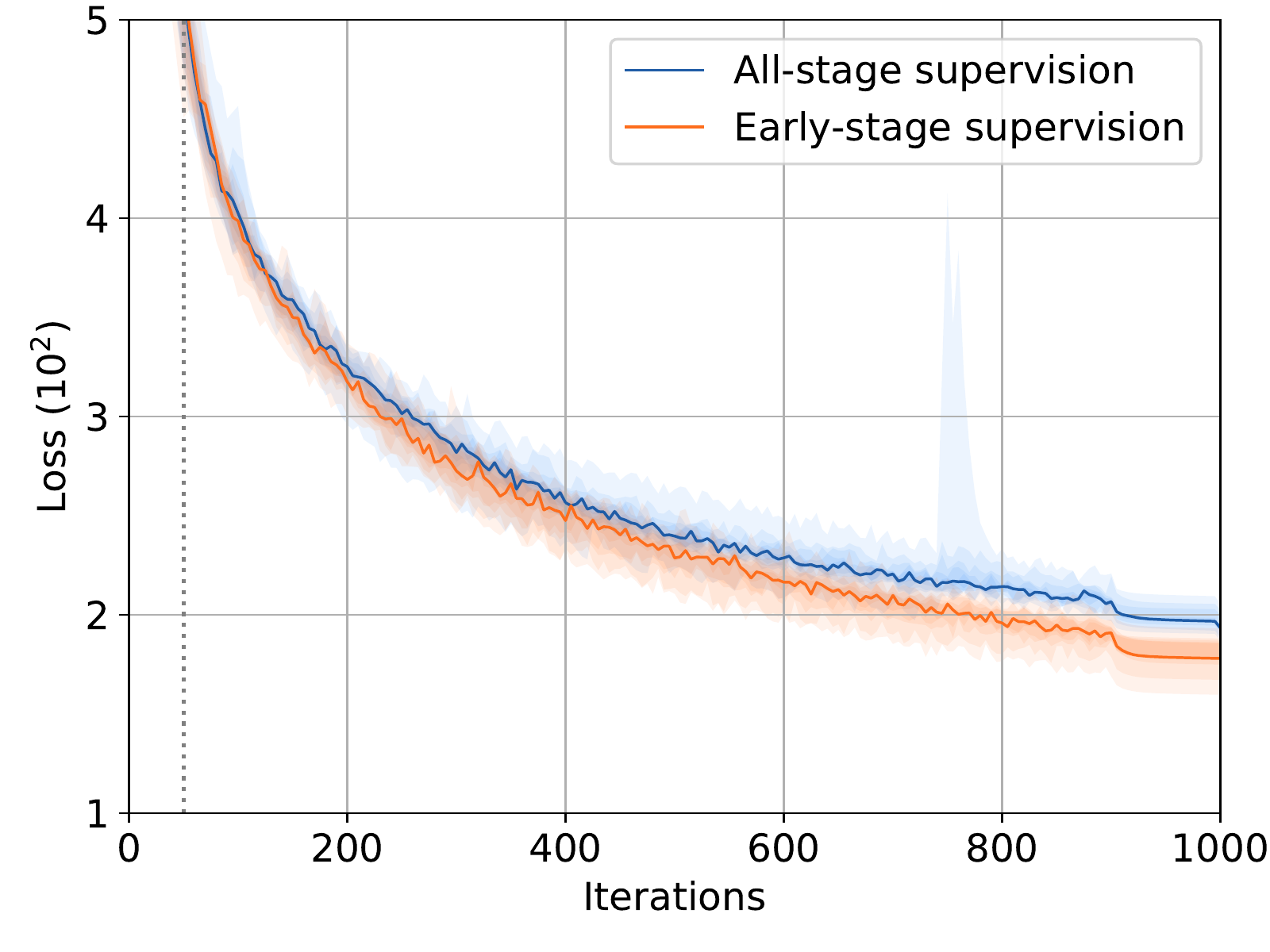}\\
%\begin{minipage}{0.33\linewidth}\centering \textsc{ball}\end{minipage}\hfill
\begin{minipage}{0.5\linewidth}\centering \textsc{reading} (early-stage vs. no/all-stage supervision)\end{minipage}\hfil
\begin{minipage}{0.5\linewidth}\centering \textsc{cow} (early-stage vs. no/all-stage supervision)\end{minipage}\\
\caption{\textbf{Convergence analysis with different types of weak supervision.}
We show learning curves of mean angular errors (top) and loss values (bottom) for \textsc{reading} and \textsc{cow}, profiled by distributions of 11 rounds run (colored region) and medians (solid line).
Compared to the proposed early-stage supervision (red), using no/all-stage supervision (green/blue) is often unstable or inaccurate.
%For each scene, the proposed early-stage supervision (red) is compared with  no (green) and all-stage (blue) supervision.
Vertical lines at $t=50$ indicate termination of early-stage supervision.
See the supplementary material for results of other scenes. Best viewed in color.
}
\label{fig:training}
\end{figure*}

\section{Experiments}
In this section we evaluate our method using a challenging real-world scene benchmark called DiLiGenT~\cite{Shi18}. 
In Sec.~\ref{sec:benchmark}, we show comparisons with state-of-the-art photometric stereo methods.
We then more analyze our network architecture in Sec.~\ref{sec:eval_components} and  weak supervision technique in Sec.~\ref{sec:eval_weaksup}. 
In the experiments, we use $M=96$ of observed images for each scene provided by the DiLiGenT dataset.
%DiLiGenT provides $M=96$ of observed images for each scene.
Our method is implemented in Chainer~\cite{chainer} and is run on a single nVidia Tesla V100 GPU with 16 GB memory and 32 bit floating-point precision.

\subsection{Real-world scene benchmark (DiLiGenT)}
\label{sec:benchmark}
We show our results on the DiLiGenT benchmark~\cite{Shi18} in Table~\ref{tab:benchmark}, where we compare our method with ten existing methods by mean angular errors. 
We also show visual comparisons of the top three and baseline methods for \textsc{reading} and \textsc{harvest} in Fig.~\ref{fig:results}.
Our method achieves the best average score and best individual scores for eight scenes (excepting only two scenes of \textsc{goblet} and \textsc{harvest}) that contain various materials and reflectance surfaces.
This is remarkable considering that another neural network method~\cite{Santo17} outperforms the other existing methods only for \textsc{harvest}, in spite of its supervised learning.
This \textsc{harvest} is the most difficult scene of all due to heavy interactions of cast shows and interreflections as well as spatially-varying materials and complex metallic BRDFs. For such complex scenes, supervised pre-training~\cite{Santo17} is effective.
The baseline method poorly performs especially for specular objects. Although we use its results as guidance priors, its low accuracy is not critical to our method thanks to the proposed early-stage supervision. We more analyze it in Sec.~\ref{sec:eval_weaksup}.

\subsection{Analysis of the network architecture}
\label{sec:eval_components}
In the middle part of Table~\ref{tab:analysis}, we show performance changes of our method by modifying its architecture. Specifically, we test two settings where we disable two connections from PSNet to IRNet, \ie, the specularity channel input and the global observation blending described in Sec.~\ref{sec:irnet}. 
As shown, the proposed full architecture performs best, while the removal of the specularity channel input has the most negative impact. 
As expected, directly inputting a specularity channel indeed eases learning of complex BRDFs (\eg, metallic surfaces in \textsc{cow}), demonstrating a strength of our physics-based network architecture that can exploit known pysical reflectance properties for BRDF learning.

\subsection{Effects of early-stage weak supervision}
\label{sec:eval_weaksup}
We here evaluate the effectiveness of our learning strategy using early-stage weak supervision, by comparing with two cases where we use no or all-stage  supervision (\ie, $\lambda_t$ is $0$ or constant). See the bottom part of Table~\ref{tab:analysis} for performance comparisons. Learning with no supervision produces comparable median scores but worse mean scores, compared to early-stage supervision. This indicates that learning with no supervision is very unstable and often gets stuck at bad local minimums, as shown in Fig.~\ref{fig:training} (green profiles). On the other hand, learning with all-stage supervision is relatively stable but is strongly biased by inaccurate least-squares priors, often producing worse solutions as shown  in Fig.~\ref{fig:training} (blue profiles).
In contrast, learning with the proposed early-stage supervision (red profiles) is more stable and persistently continues to improve accuracy even after terminating the supervision at $t=50$ (shown as vertical dashed lines).

\section{Discussions and related work}
Our method is inspired by recent work of \emph{deep image prior} by \citet{Ulyanov18}. They show that architectures of CNNs themselves behave as good regularizers for natural images, and show successful results for unsupervised tasks such as image super-resolution and inpainting by fitting a CNN for a single test image. However, their simple glass-hour network does not directly apply to photometric stereo, because we here need to 
simultaneously consider surface normal estimation that accounts for global statistics of observations, as well as reconstruction of individual observations for defining the loss. Our novel architecture addresses this problem by resorting to ideas of classical physics-based approaches to photometric stereo.

Our network architecture is also partly influenced by that of \cite{Santo17}, which regresses per-pixel observations $\bm{I}_p \in \mathbb{R}^M$ to a 3D normal vector using a simple feed-forward network of five fully-connected and ReLU layers plus an output layer. Our PSNet becomes similar to theirs, if we use 1x1 Conv with more layers and channels (\ie, they use channels of $4096$ and $2048$ for the five internal layers). Since our method only needs to learn reflectance properties of a single test scene, our PSNet requires fewer layers and channels. More importantly, we additionally introduce IRNet, which allows direct unsupervised learning on test data.

There are some other early studies on photometric stereo using (shallow) neural networks.
These methods work under more restricted conditions, \eg, assuming pre-training by a calibration sphere of the same material with target objects~\cite{Iwahori93,Iwahori95b}, special image capturing setups~\cite{Iwahori02,Ding09}, or the Lambertian surfaces~\cite{Cheng06,Elizondo08}, whereas none of them is required by our method.

Currently, our method has limitations of a slow running time (\eg, 1 hour to do 1000 SGD iterations for each scene) and limited performances to complex scenes (\eg, \textsc{harvest}). However, several studies~\cite{Akiba17,You17,GoyalDGNWKTJH17} show fast training of CNNs using extremely large minibatches and tuned scheduling of SGD step-sizes. Since our dense prediction method can use at most a large minibatch of $M\times H\times W$ pixel samples, the use of such acceleration schemes may improve the convergence speed.
Also, a pre-training approach similar to \cite{Santo17} is still feasible for our method, which will accelerate the convergence and will also increase accuracy to complex scenes (with the loss of permutation invariance). 
%Exploiting more sophisticated BRDF representations and properties~\cite{Ikehata14,Shi14} may further promote BRDF learning.
Thorough analyses in such directions are left as our future work. 

\section{Conclusions}
In this paper, we have presented a novel CNN architecture for photometric stereo.
The proposed unsupervised learning approach bridges a gap between existing supervised neural network methods and many other classical physics-based unsupervised methods.
Consequently, our method can learn complicated BRDFs by leveraging both powerful expressibility of deep neural networks and physical reflectance properties known by past studies, achieving the state-of-the-art performance in an unsupervised fashion just like classical methods.
We also hope that our idea of physics-based unsupervised learning stimulates further research on tasks 
%where preparing ground truth for training is difficult, 
%where desired data is not directly observable for training data,
that lack of ground truth data for training,
because even so the physics is everywhere in the real world, which will provide strong clues for the hidden data we desire.
%Since lack of annotated training data is a serious challenge in deep learning methodologies, \HL{we hope that our idea of the physics-based unsupervised learning stimulates further research }
%We hope that our work stimulates further research in this and other areas on reflectance analysis, where difficulties of making large training data have been largely confronted in application of deep learning methodologies.

\section*{Acknowledgements}
The authors would thank \citet{Shi18} for building a photometric stereo benchmark, \citet{Santo17} for providing us their results, and Profs.~Yoichi~Sato and ~Ryo~Yonetani and anonymous reviewers for their helpful feedback.
The authors would gratefully acknowledge the support of NVIDIA Corporation with the donation of a Titan Xp GPU.

\bibliography{reference}
\bibliographystyle{icml2018}

\clearpage
\onecolumn
\setcounter{page}{1}
\setcounter{section}{0}
\setcounter{figure}{0}
\setcounter{table}{0}
\setcounter{equation}{0}
\setcounter{footnote}{0}
\renewcommand{\thepage}{A\arabic{page}}
\renewcommand{\thesection}{\Alph{section}}
\renewcommand{\thefigure}{A\arabic{figure}}
\renewcommand{\thetable}{A\arabic{table}}
\renewcommand{\theequation}{A\arabic{equation}}

\icmltitlerunning{Supplementary Material of Neural Inverse Rendering for General Reflectance Photometric Stereo}

%\twocolumn[
\icmltitle{Supplementary Material of\\Neural Inverse Rendering for General Reflectance Photometric Stereo}



\icmlsetsymbol{equal}{*}

\begin{icmlauthorlist}
	\icmlauthor{Tatsunori Taniai}{to}
	\icmlauthor{Takanori Maehara}{to}
\end{icmlauthorlist}

%\icmlcorrespondingauthor{Takanori Maehara}{takanori.maehara@riken.jp}

\icmlkeywords{photometric stereo, physics-based vision, surface normal estimation, reflectance analysis, general BRDF}
\begin{center}
${}^1$~RIKEN AIP, Nihonbashi, Tokyo, Japan
\end{center}
\vskip 0.3in
%]
\setstretch{1.2}

\begin{center}
\begin{minipage}{0.8\textwidth}
In this supplementary material, we provide results for all the ten benchmark scenes that are omitted in the paper due to the page limitation. Specifically, we provide the following additional results.
\begin{itemize}
	\setlength{\itemsep}{-10pt}    % ブロック間の余白
	\setlength{\parskip}{10pt}    % 段落間余白．
	\setlength{\itemindent}{0pt}
	\item[\textbf{A.}] Visual comparisons of our method and existing methods.
	\item[\textbf{B.}] Our image reconstruction results.
	\item[\textbf{C.}] Convergence analyses of our method by different types of weak supervision.
\end{itemize}
\end{minipage}
\end{center}
\vskip 1cm

\section{Additional visual comparisons}
\label{asec:results}
\begin{figure}[p]
\centering
\begin{minipage}{0.16\linewidth}\centering\small Ground truth /\\ Observed image\end{minipage}\hfil
\begin{minipage}{0.16\linewidth}\centering\small Ours\end{minipage}\hfil
\begin{minipage}{0.16\linewidth}\centering\small \citet{Santo17}\end{minipage}\hfil
\begin{minipage}{0.16\linewidth}\centering\small \citet{Shi14}\end{minipage}\hfil
\begin{minipage}{0.16\linewidth}\centering\small \citet{Ikehata14}\end{minipage}\hfil
\begin{minipage}{0.16\linewidth}\centering\small Baseline\\(least squares)\end{minipage}\\
\vskip 2mm
\includegraphics[width=0.15\linewidth]{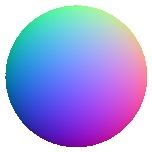}\hfil
\includegraphics[width=0.15\linewidth]{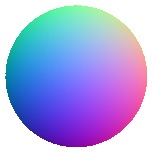}\hfil
\includegraphics[width=0.15\linewidth]{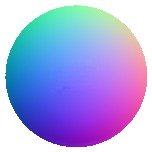}\hfil
\includegraphics[width=0.15\linewidth]{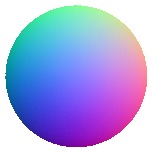}\hfil
\includegraphics[width=0.15\linewidth]{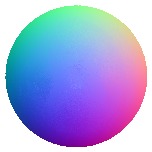}\hfil
\includegraphics[width=0.15\linewidth]{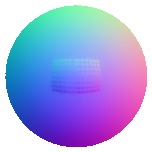}\\
\vskip 3mm
\includegraphics[width=0.15\linewidth]{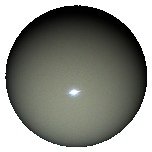}\hfil
\includegraphics[width=0.15\linewidth]{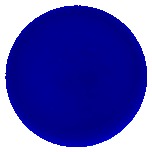}\hfil
\includegraphics[width=0.15\linewidth]{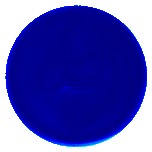}\hfil
\includegraphics[width=0.15\linewidth]{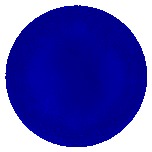}\hfil
\includegraphics[width=0.15\linewidth]{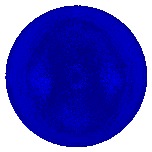}\hfil
\includegraphics[width=0.15\linewidth]{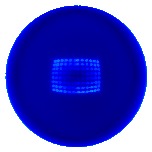}%
\begin{minipage}{0mm}{\vskip -24mm}\hspace*{1mm}\includegraphics[width=6.9mm]{figures/jet_legend}\end{minipage}\\
\begin{minipage}{0.15\linewidth}\centering \textsc{ball}\end{minipage}\hfil
\begin{minipage}{0.15\linewidth}\centering \textbf{1.47}\end{minipage}\hfil
\begin{minipage}{0.15\linewidth}\centering 2.02\end{minipage}\hfil
\begin{minipage}{0.15\linewidth}\centering 1.74\end{minipage}\hfil
\begin{minipage}{0.15\linewidth}\centering 3.34\end{minipage}\hfil
\begin{minipage}{0.15\linewidth}\centering 4.10\end{minipage}\\
\caption{\textbf{Visual comparisons for \textsc{ball} scene.} From left to right columns, we show 1) the ground truth normal map and an observed image,  2) our normal estimate and its error map, 3--6) four pairs of a normal estimate and its error map by three state-of-the-art methods~\cite{Santo17,Shi14,Ikehata14} and the baseline least squares method.}
\label{afig:results_ball}
\end{figure}

\begin{figure}
\centering
\begin{minipage}{0.16\linewidth}\centering\small Ground truth /\\ Observed image\end{minipage}\hfil
\begin{minipage}{0.16\linewidth}\centering\small Ours\end{minipage}\hfil
\begin{minipage}{0.16\linewidth}\centering\small \citet{Santo17}\end{minipage}\hfil
\begin{minipage}{0.16\linewidth}\centering\small \citet{Shi14}\end{minipage}\hfil
\begin{minipage}{0.16\linewidth}\centering\small \citet{Ikehata14}\end{minipage}\hfil
\begin{minipage}{0.16\linewidth}\centering\small Baseline\\(least squares)\end{minipage}\\
\vskip 2mm
\includegraphics[width=0.15\linewidth]{figures/results/gt/harvestPNG/normal}\hfil
\includegraphics[width=0.15\linewidth]{figures/results/ours/harvestPNG/normal}\hfil
\includegraphics[width=0.15\linewidth]{figures/results/ICCVW17Santo/harvestPNG/normal}\hfil
\includegraphics[width=0.15\linewidth]{figures/results/CVPR12Shi/harvestPNG/normal}\hfil
\includegraphics[width=0.15\linewidth]{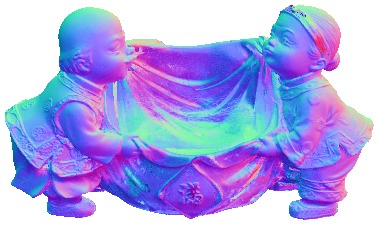}\hfil
\includegraphics[width=0.15\linewidth]{figures/results/l2/harvestPNG/normal}\\
\vskip 2mm
\includegraphics[width=0.15\linewidth]{figures/results/ours/harvestPNG/img/001}\hfil
\includegraphics[width=0.15\linewidth]{figures/results/ours/harvestPNG/error_jet}\hfil
\includegraphics[width=0.15\linewidth]{figures/results/ICCVW17Santo/harvestPNG/error_jet}\hfil
\includegraphics[width=0.15\linewidth]{figures/results/CVPR12Shi/harvestPNG/error_jet}\hfil
\includegraphics[width=0.15\linewidth]{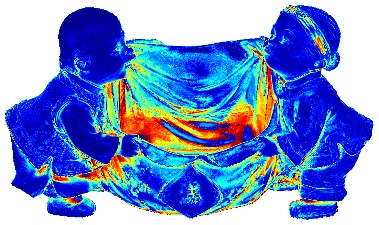}\hfil
\includegraphics[width=0.15\linewidth]{figures/results/l2/harvestPNG/error_jet}\\
\begin{minipage}{0.15\linewidth}\small\centering \textsc{harvest}\end{minipage}\hfil
\begin{minipage}{0.15\linewidth}\small\centering 22.59\end{minipage}\hfil
\begin{minipage}{0.15\linewidth}\small\centering \textbf{16.86}\end{minipage}\hfil
\begin{minipage}{0.15\linewidth}\small\centering 25.44\end{minipage}\hfil
\begin{minipage}{0.15\linewidth}\small\centering 25.95\end{minipage}\hfil
\begin{minipage}{0.15\linewidth}\small\centering 30.62\end{minipage}\\
\caption{\textbf{Visual comparisons for \textsc{harvest} scene.} See also explanations in Fig.~\ref{afig:results_ball}.}
\label{afig:results_harvest}
\end{figure}

In Figures~\ref{afig:results_ball}--\ref{afig:results_harvest}, we show surface normal estimates and their angular error maps for all ten scenes from the DiLiGenT dataset, comparing our method with three methods by \citet{Santo17}, \citet{Shi14}, and \citet{Ikehata14}, and also the baseline least squares method.
Here, the best result for each scene is obtained by either of the top four methods (where 8 of 10 best results are by our method), whose mean angular error is shown by a bold font number.

\section{Additional image reconstruction results}
\label{asec:reconst}
In Figures~\ref{afig:reconst_ball}--\ref{afig:reconst_harvest}, we show our image reconstruction results for all ten scenes from the DiLiGenT dataset. For each scene we show 6 images from 96 observed images, and for each observed image we show our synthesized image, intermediate reflectance image, and reconstruction error map. Here, the reflectance images represent multiplication of spatially-varying BRDFs and cast shadows under a particular illumination condition. We can clearly see cast shadows in reflectance images appearing in the results of \textsc{bear} and \textsc{buddha}.
Note that for better visualization, the image intensities are scaled by a factor of $255/2$ after the proposed global scaling normalization.

\section{Additional convergence analyses}
\label{asec:convergence}
\begin{figure*}[p]
	\centering
	\small
	\def\figw{0.30\linewidth}
	% 0:ball, 1:bear, 2:buddha, 3:cat, 4:cow, 5:goblet, 6:harbest, 7:pot1, 8:pot2, 9:reading
	\includegraphics[width=\figw]{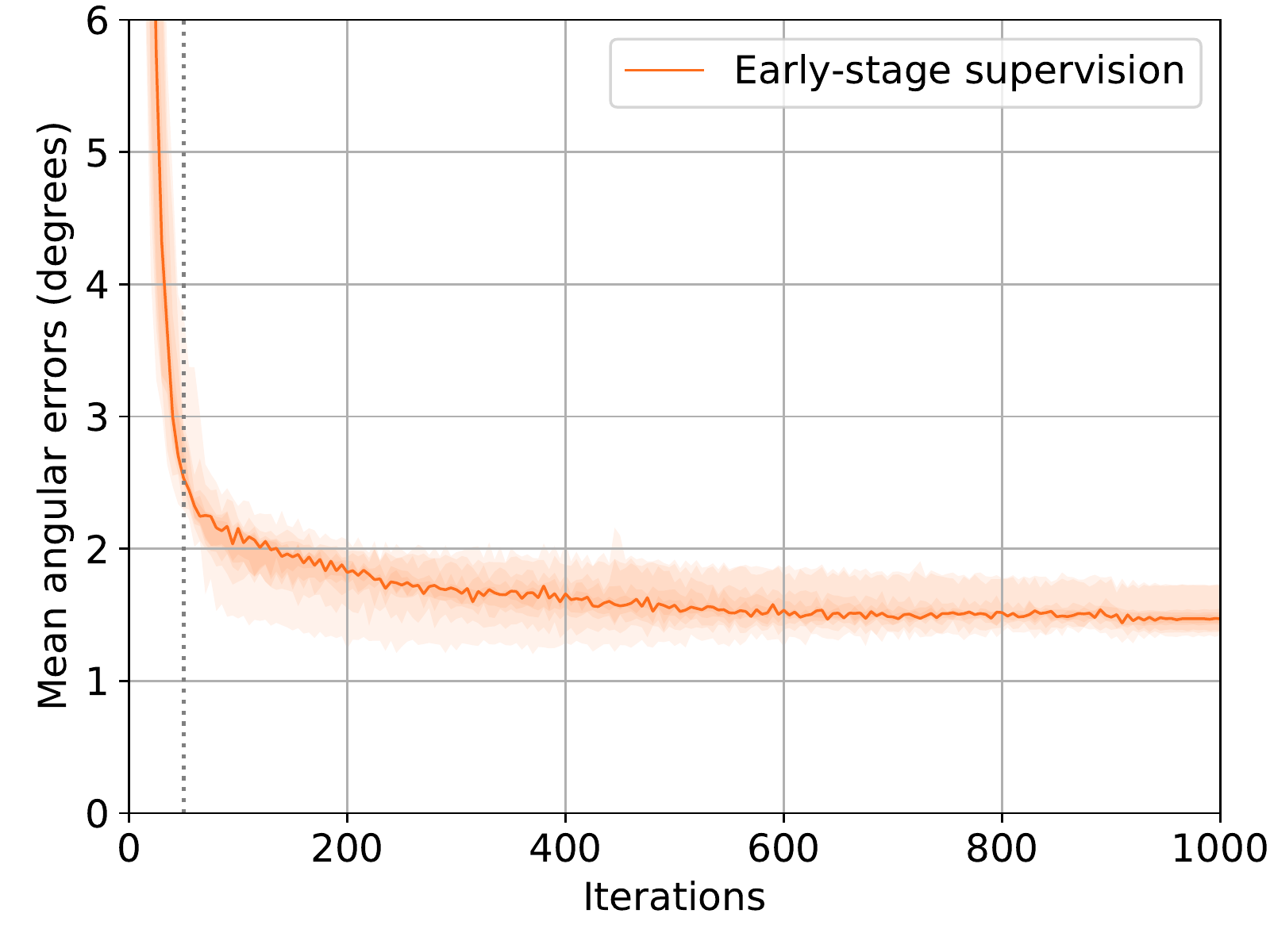}\hfil
	\includegraphics[width=\figw]{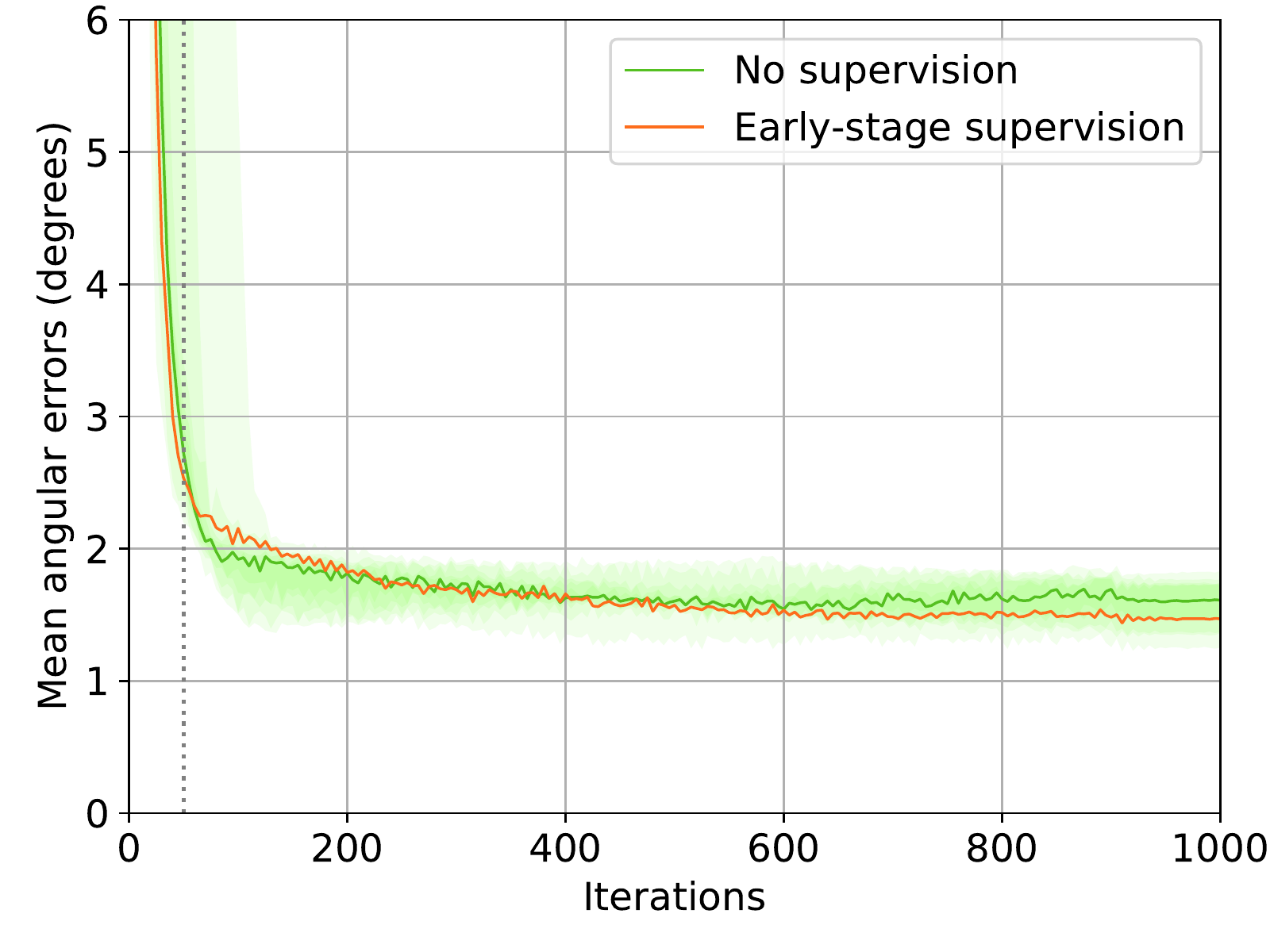}\hfil
	\includegraphics[width=\figw]{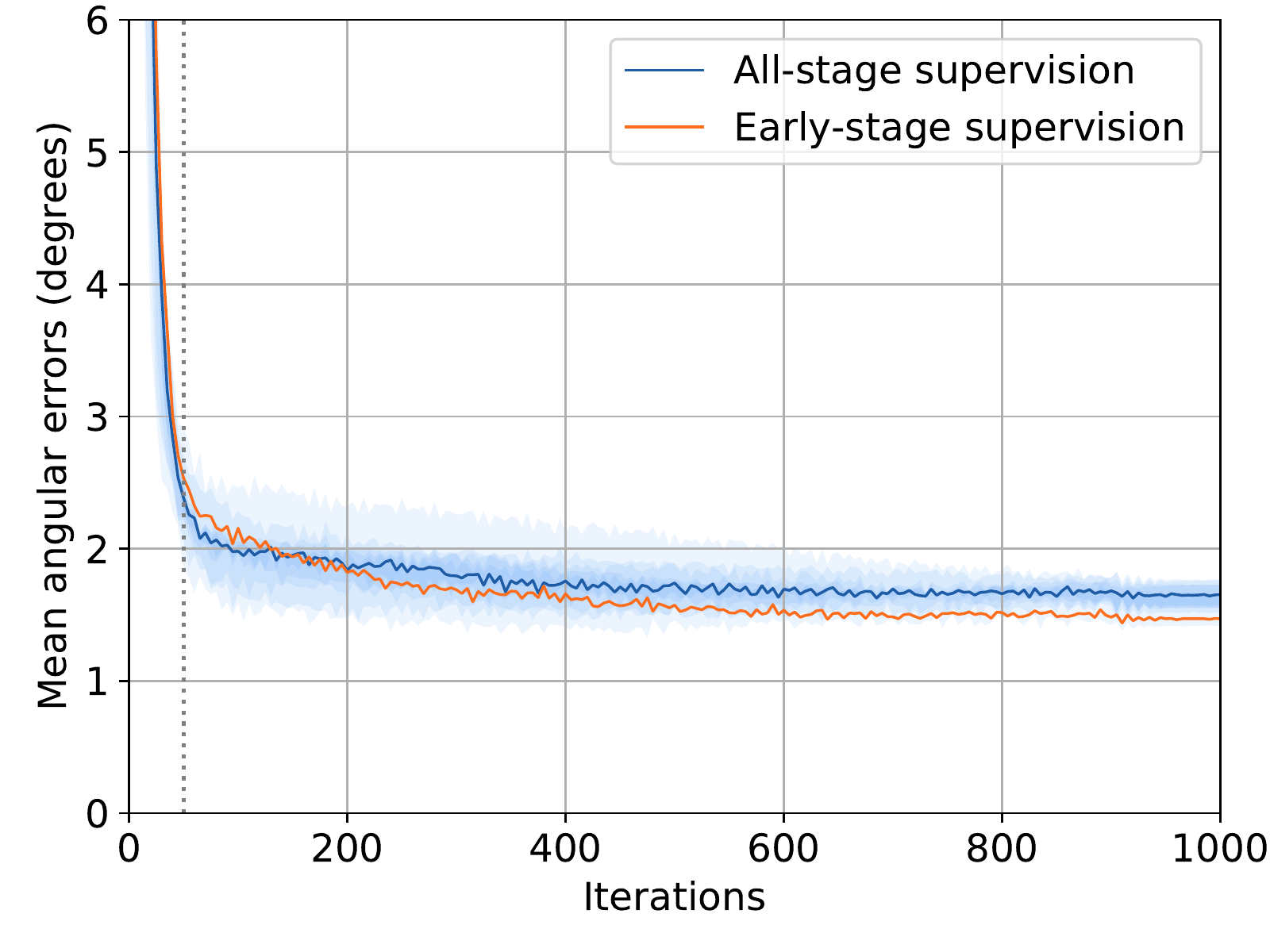}\\
	\includegraphics[width=\figw]{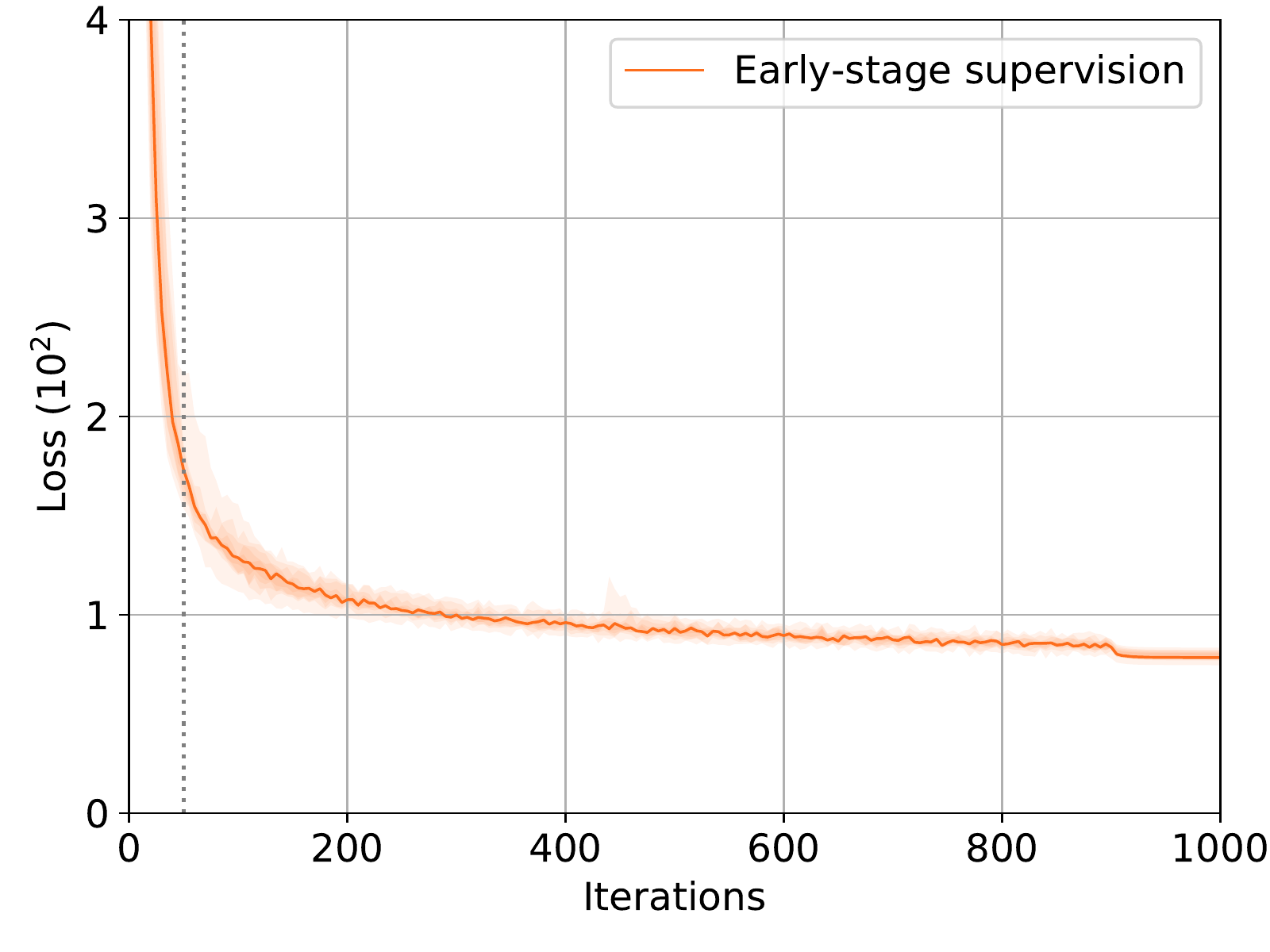}\hfil
	\includegraphics[width=\figw]{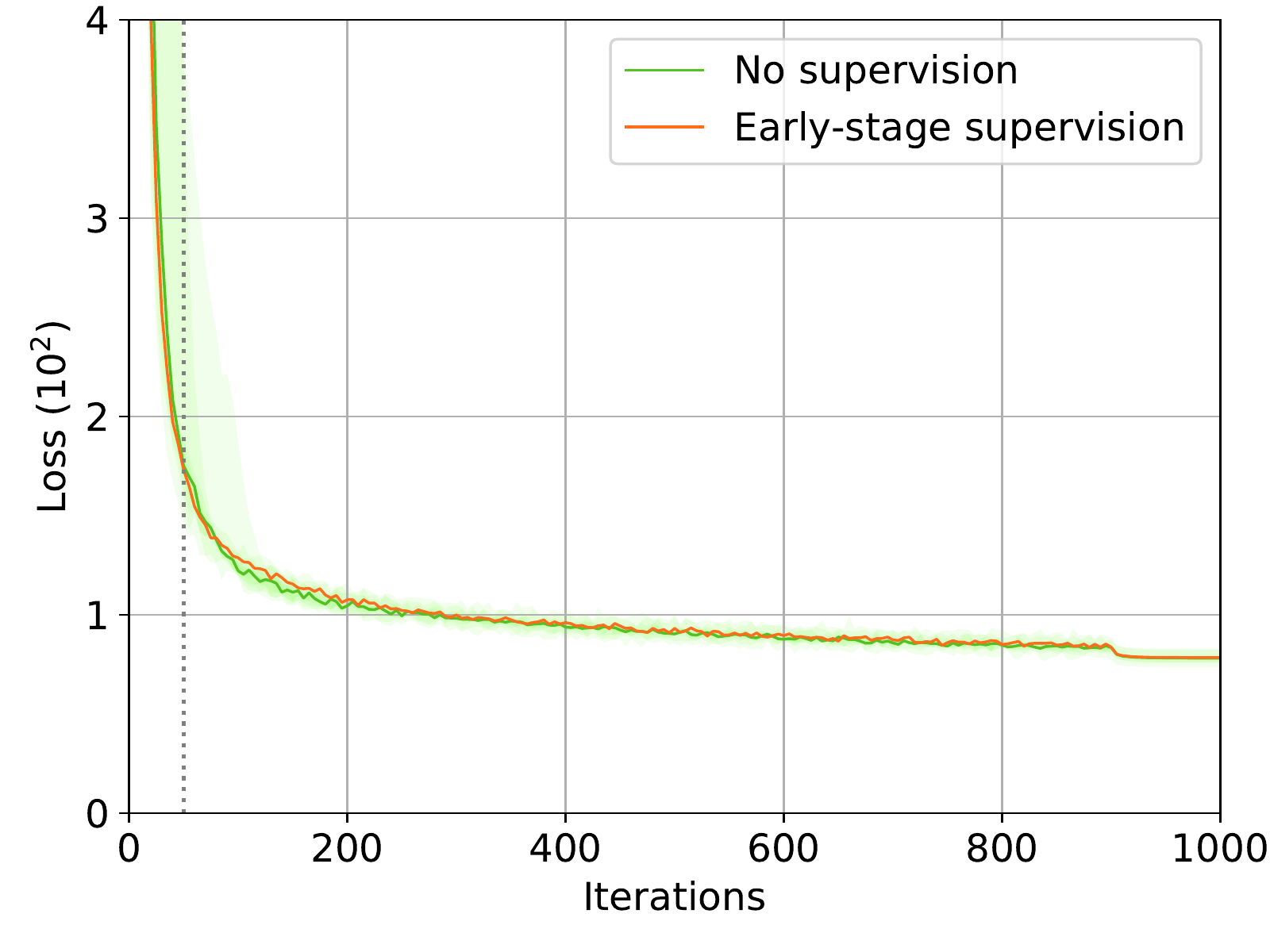}\hfil
	\includegraphics[width=\figw]{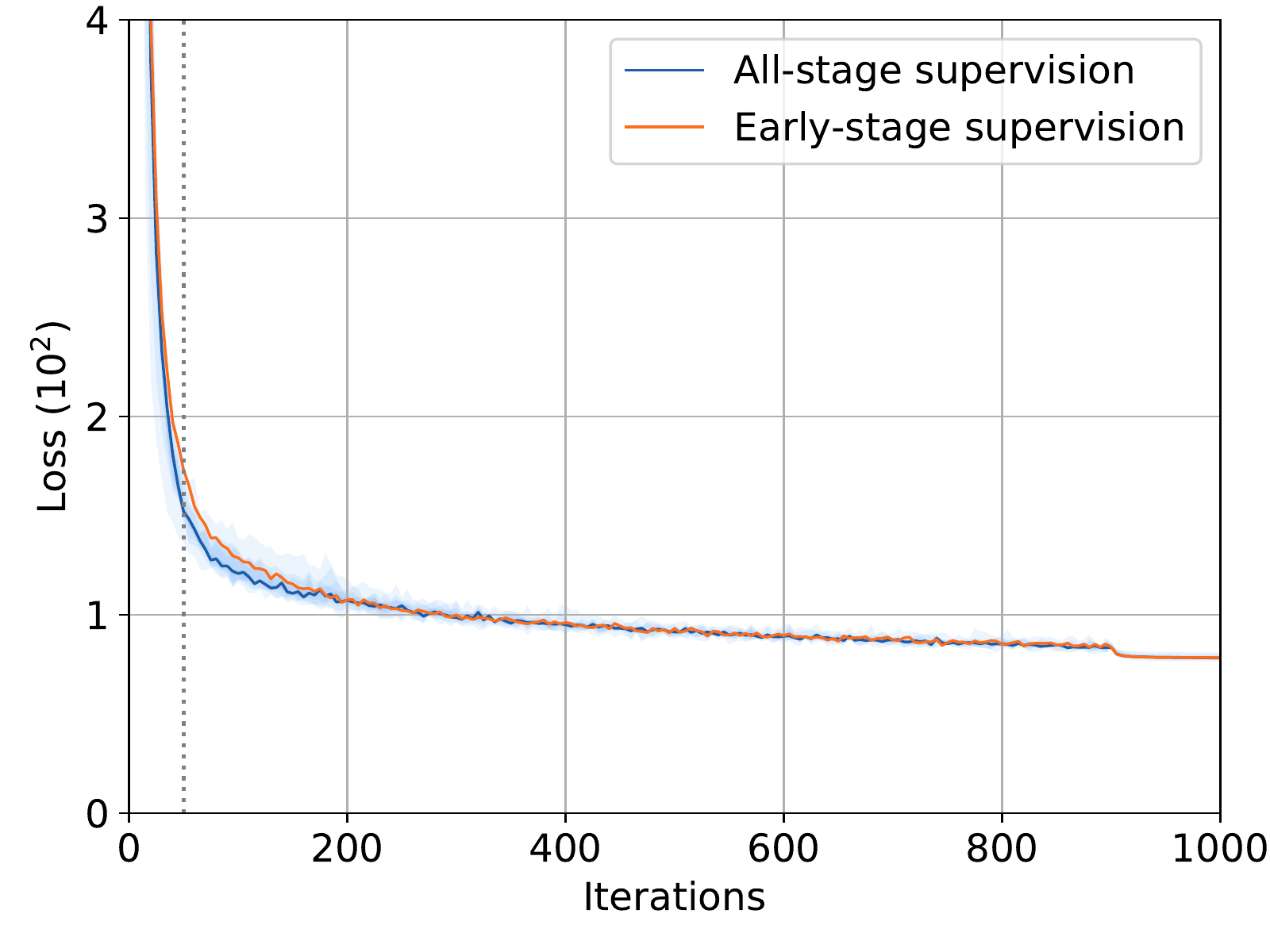}\\
	\begin{minipage}{\figw}\centering {Early-stage supervision}\end{minipage}\hfil
	\begin{minipage}{\figw}\centering {No supervision}\end{minipage}\hfil
	\begin{minipage}{\figw}\centering {All-stage supervision}\end{minipage}\\
	\caption{\textbf{Convergence analysis with different types of weak supervision for \textsc{ball} scene.}
		We compare training curves by using the proposed early-stage supervision (left) with no supervision (middle), and all-stage supervision (left). For each case, profiles of mean angular errors and loss values are shown in top and bottom, respectively, which are visualized by distributions of 11 rounds run (colored region) and medians (solid line).
		Vertical lines at $t=50$ show termination of early-stage supervision.
	}
	\label{afig:training_ball}
\end{figure*}

\begin{figure*}[p]
	\centering
	\small
	\def\figw{0.30\linewidth}
	% 0:ball, 1:bear, 2:buddha, 3:cat, 4:cow, 5:goblet, 6:harbest, 7:pot1, 8:pot2, 9:reading
	\includegraphics[width=\figw]{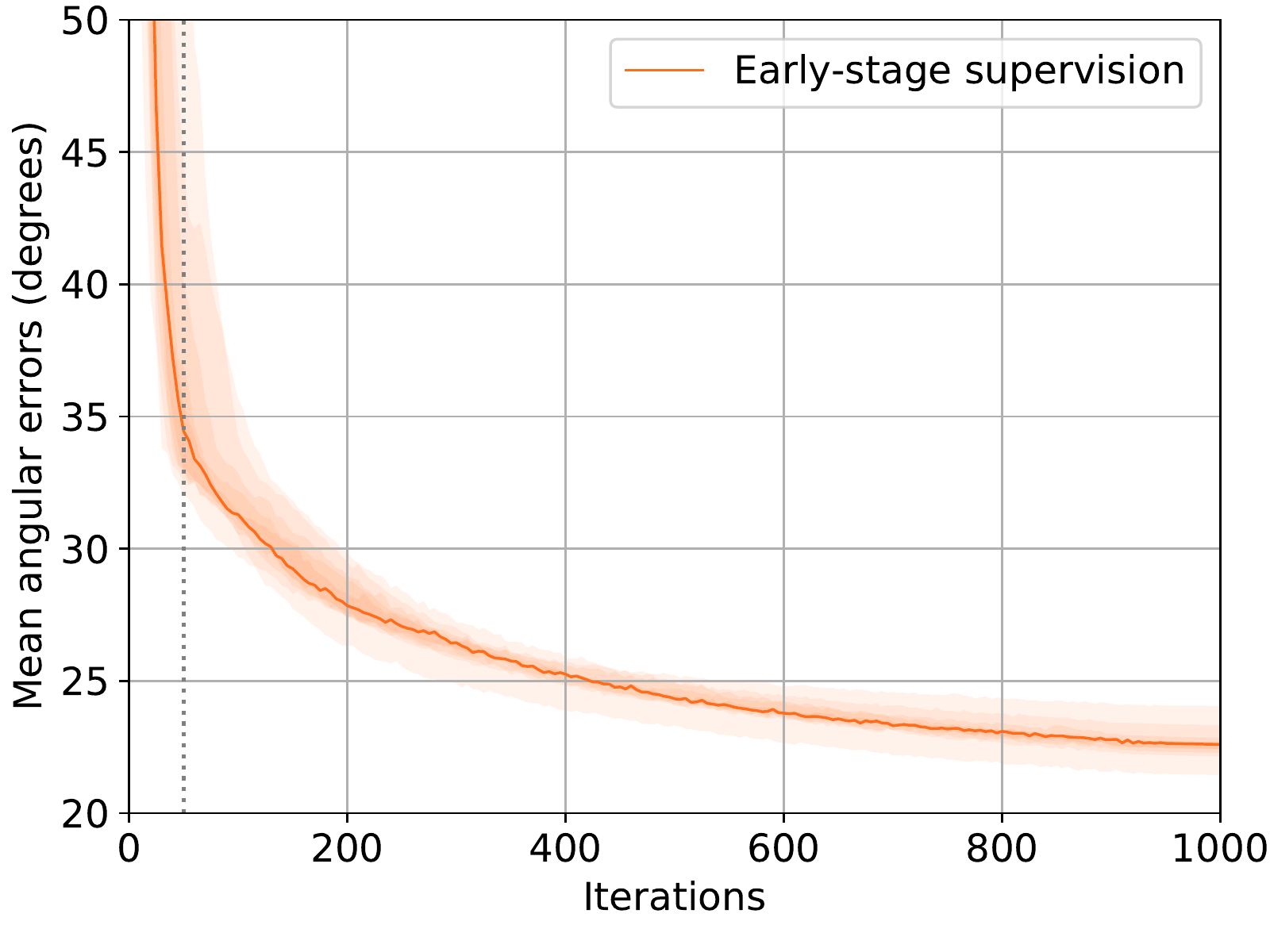}\hfil
	\includegraphics[width=\figw]{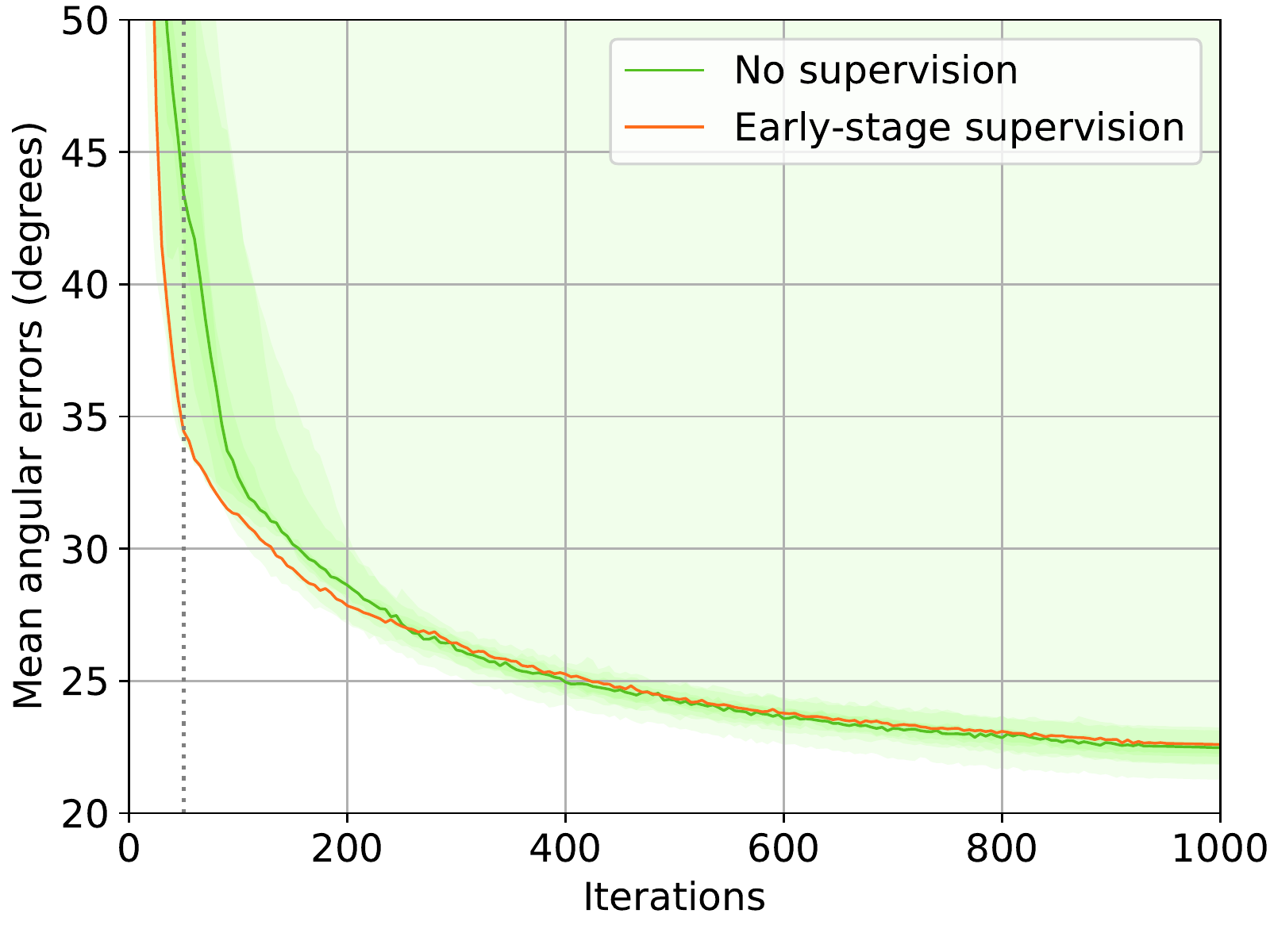}\hfil
	\includegraphics[width=\figw]{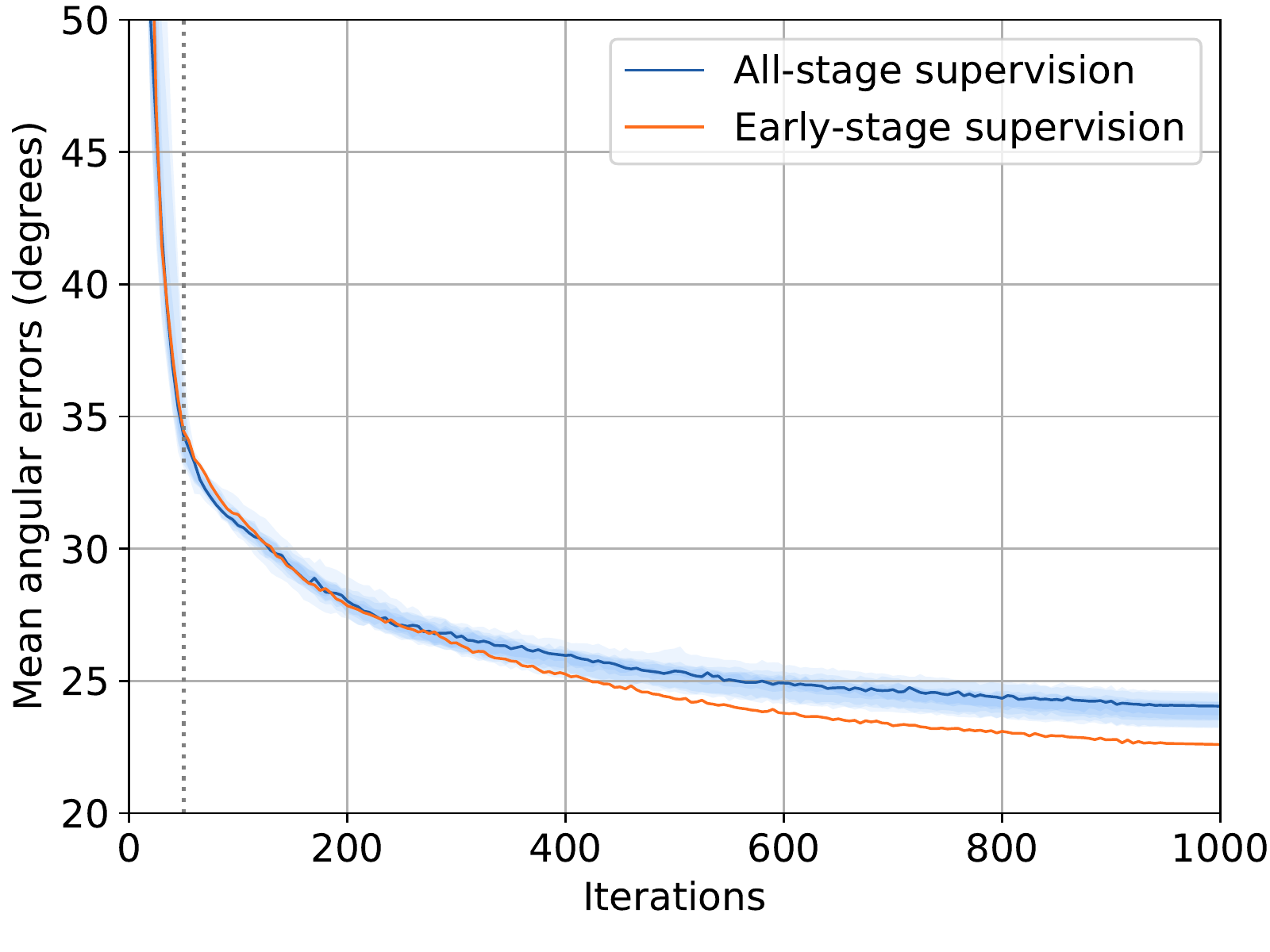}\\
	\includegraphics[width=\figw]{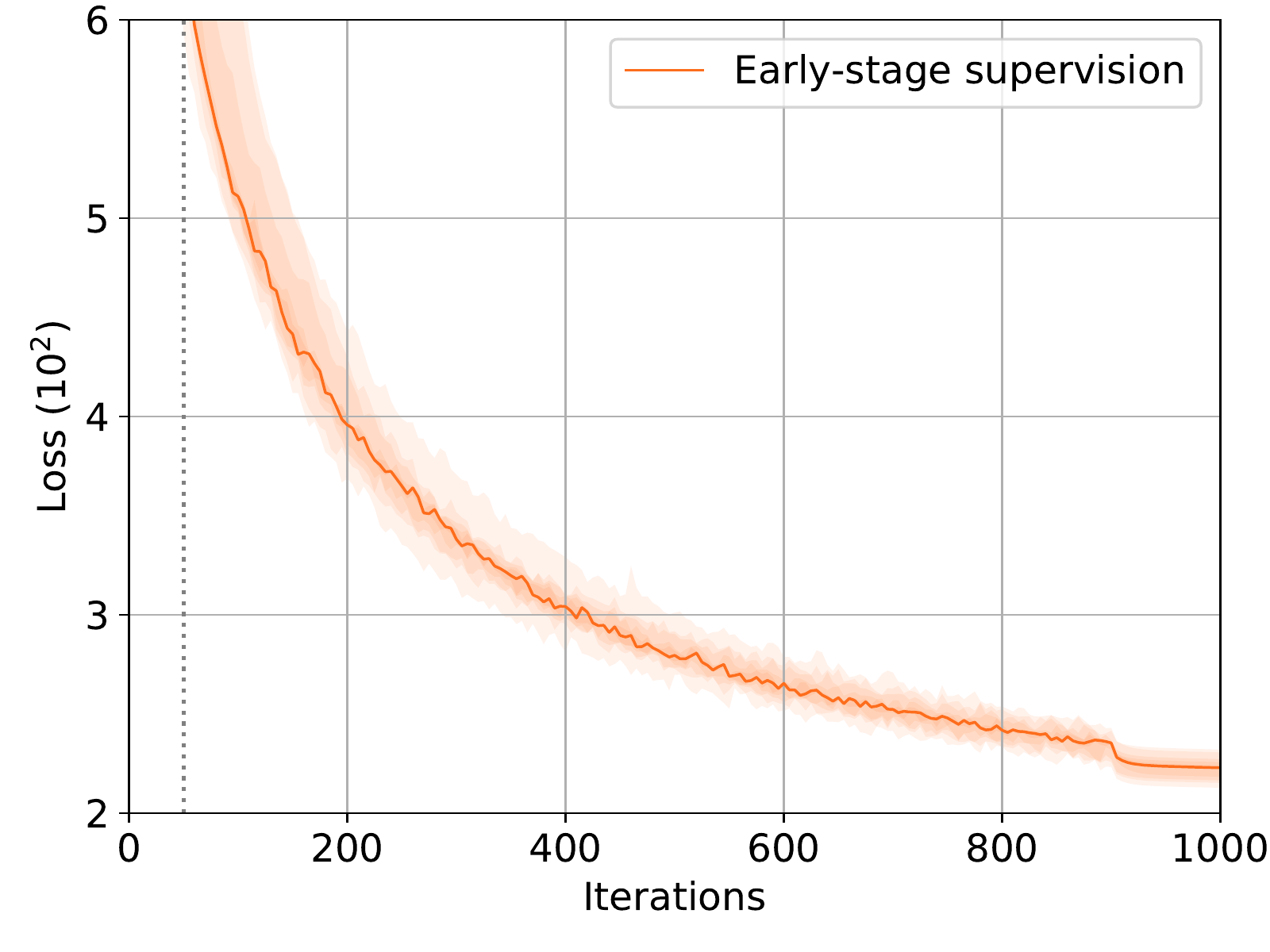}\hfil
	\includegraphics[width=\figw]{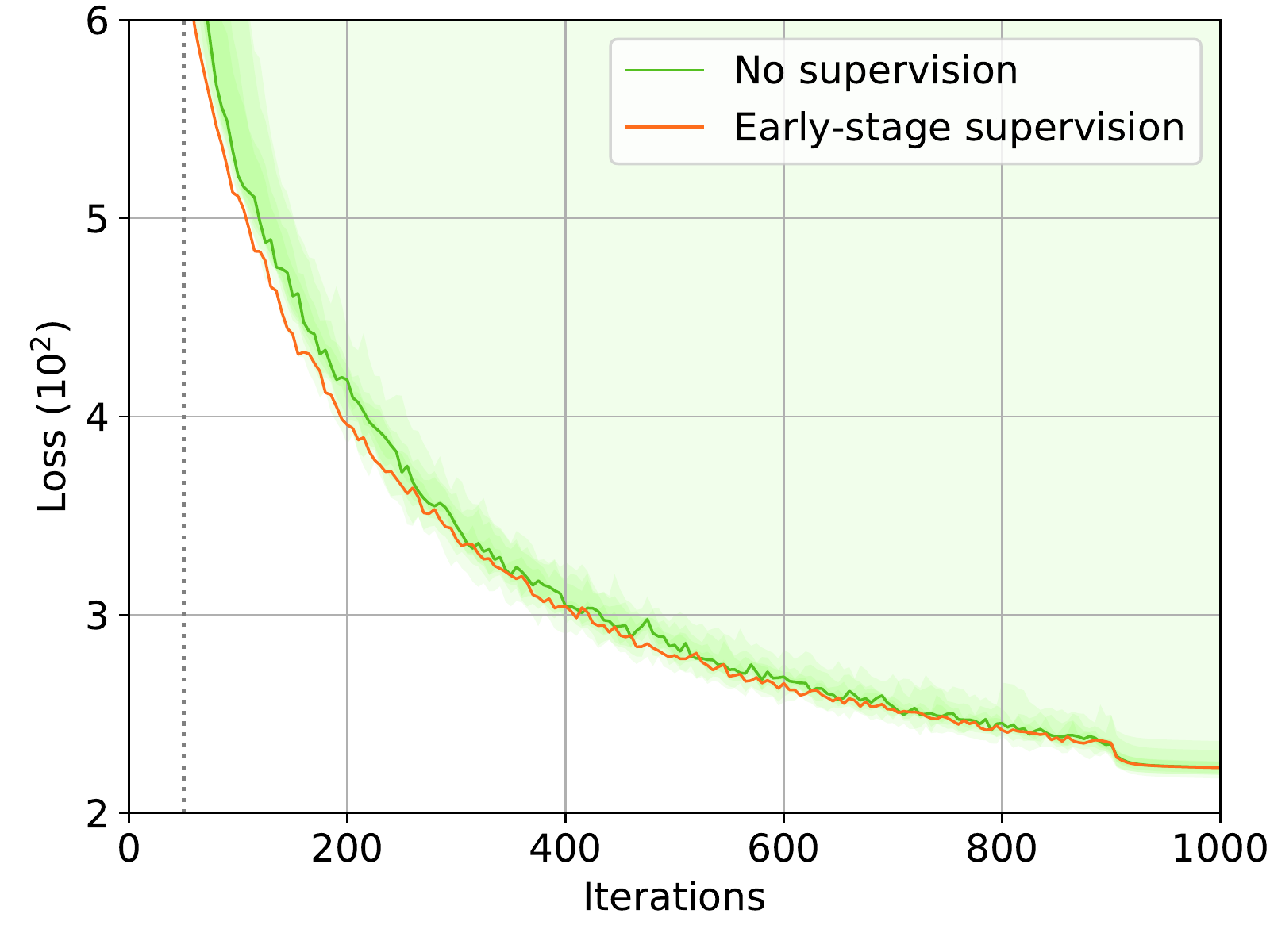}\hfil
	\includegraphics[width=\figw]{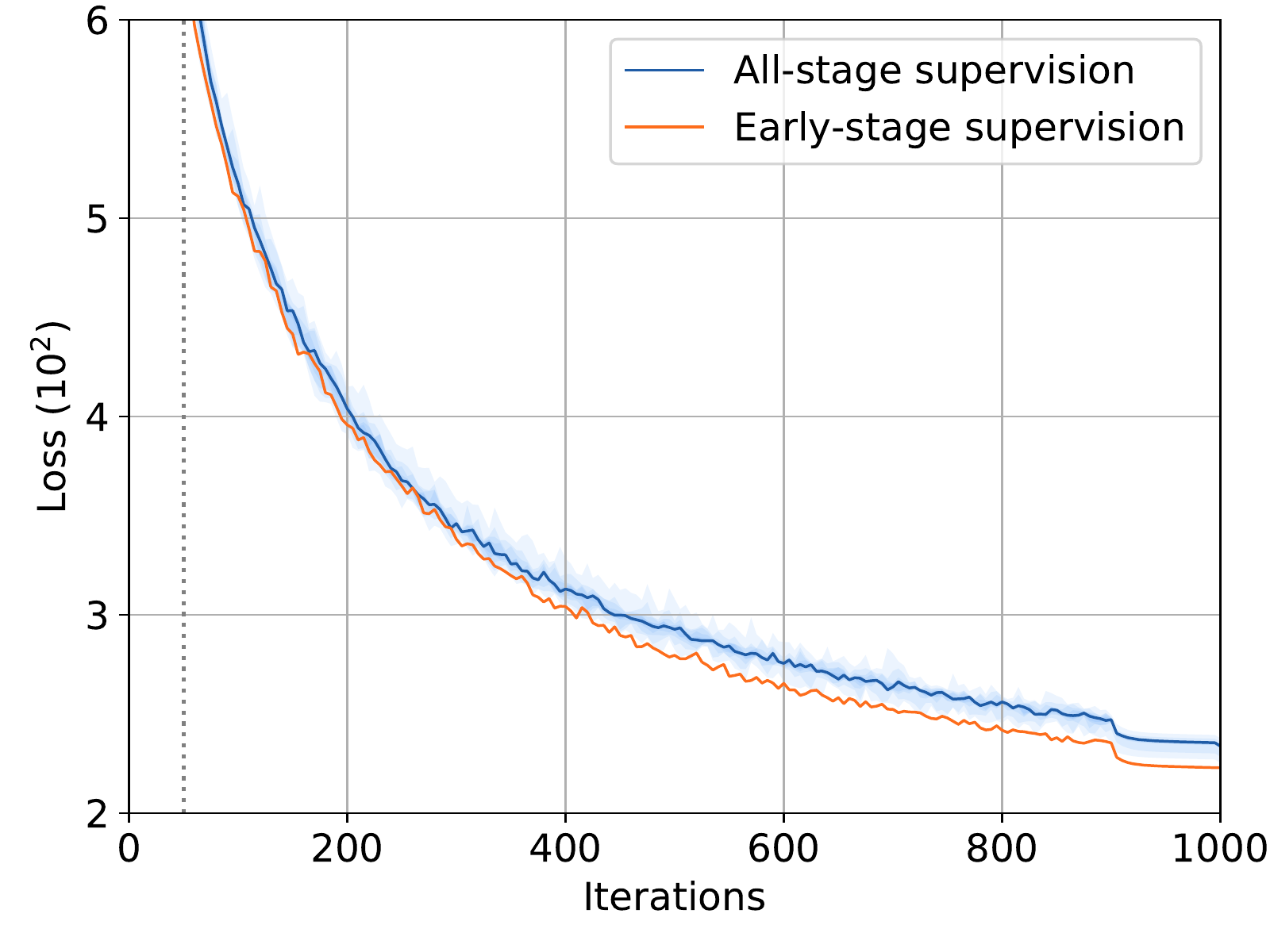}\\
	\begin{minipage}{\figw}\centering {Early-stage supervision}\end{minipage}\hfil
	\begin{minipage}{\figw}\centering {No supervision}\end{minipage}\hfil
	\begin{minipage}{\figw}\centering {All-stage supervision}\end{minipage}\\
	\caption{\textbf{Convergence analysis with different types of weak supervision for \textsc{harvest} scene.}
		See also explanations in Fig.~\ref{afig:training_ball}.
	}
	\label{afig:training_harvest}
\end{figure*}

In Figures~\ref{afig:training_ball}--\ref{afig:training_harvest}, we show convergence analyses for all ten scenes from the DiLiGenT dataset, where the proposed early-stage supervision is compared with no and all-stage supervision.
Training without supervision is very unstable, while training with all-stage supervision is strongly biased by inaccurate least square priors.
Note that for better comparison, the median profiles of the proposed early-stage weak supervision (red solid lines) are overlayed in the plots of no and all-stage supervision.

\clearpage

\begin{figure}[p]
	\centering
	\begin{minipage}{0.16\linewidth}\centering\small Ground truth /\\ Observed image\end{minipage}\hfil
	\begin{minipage}{0.16\linewidth}\centering\small Ours\end{minipage}\hfil
	\begin{minipage}{0.16\linewidth}\centering\small \citet{Santo17}\end{minipage}\hfil
	\begin{minipage}{0.16\linewidth}\centering\small \citet{Shi14}\end{minipage}\hfil
	\begin{minipage}{0.16\linewidth}\centering\small \citet{Ikehata14}\end{minipage}\hfil
	\begin{minipage}{0.16\linewidth}\centering\small Baseline\\(least squares)\end{minipage}\\
	\vskip 2mm
	\includegraphics[width=0.15\linewidth]{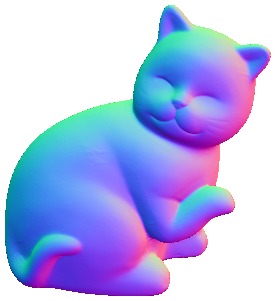}\hfil
	\includegraphics[width=0.15\linewidth]{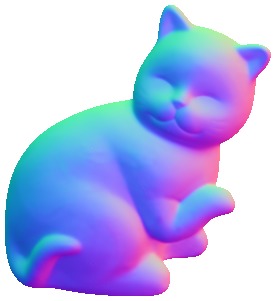}\hfil
	\includegraphics[width=0.15\linewidth]{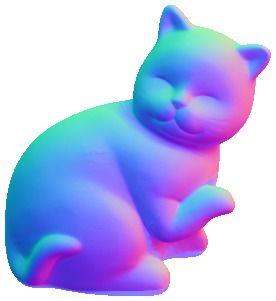}\hfil
	\includegraphics[width=0.15\linewidth]{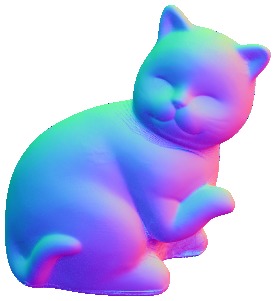}\hfil
	\includegraphics[width=0.15\linewidth]{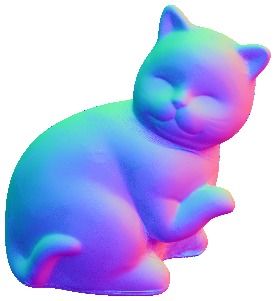}\hfil
	\includegraphics[width=0.15\linewidth]{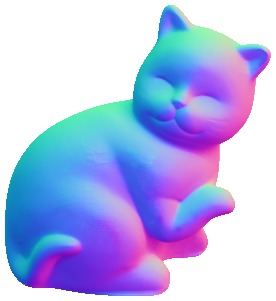}\\
	\vskip 2mm
	\includegraphics[width=0.15\linewidth]{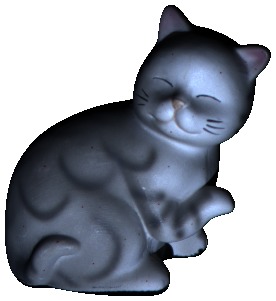}\hfil
	\includegraphics[width=0.15\linewidth]{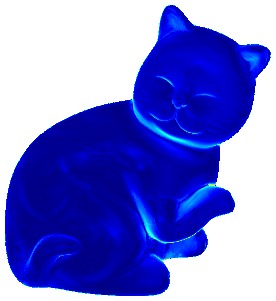}\hfil
	\includegraphics[width=0.15\linewidth]{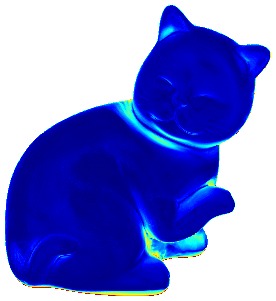}\hfil
	\includegraphics[width=0.15\linewidth]{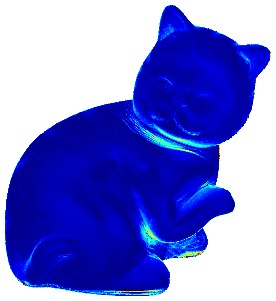}\hfil
	\includegraphics[width=0.15\linewidth]{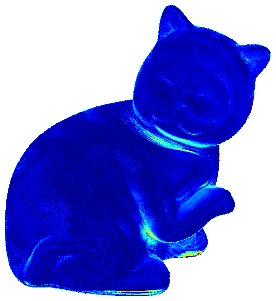}\hfil
	\includegraphics[width=0.15\linewidth]{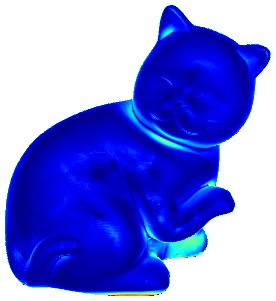}\\
	%\begin{minipage}{0mm}{\vskip -24mm}\hspace*{1mm}\includegraphics[width=6.9mm]{figures/jet_legend}\end{minipage}\\
	\begin{minipage}{0.15\linewidth}\centering \textsc{cat}\end{minipage}\hfil
	\begin{minipage}{0.15\linewidth}\centering \textbf{5.44}\end{minipage}\hfil
	\begin{minipage}{0.15\linewidth}\centering 6.54\end{minipage}\hfil
	\begin{minipage}{0.15\linewidth}\centering 6.12\end{minipage}\hfil
	\begin{minipage}{0.15\linewidth}\centering 6.74\end{minipage}\hfil
	\begin{minipage}{0.15\linewidth}\centering 8.41\end{minipage}\\
	\caption{\textbf{Visual comparisons for \textsc{cat} scene.} See also explanations in Fig.~\ref{afig:results_ball}.}
\end{figure}

\clearpage

\begin{figure}[p]
	\centering
	\begin{minipage}{0.16\linewidth}\centering\small Ground truth /\\ Observed image\end{minipage}\hfil
	\begin{minipage}{0.16\linewidth}\centering\small Ours\end{minipage}\hfil
	\begin{minipage}{0.16\linewidth}\centering\small \citet{Santo17}\end{minipage}\hfil
	\begin{minipage}{0.16\linewidth}\centering\small \citet{Shi14}\end{minipage}\hfil
	\begin{minipage}{0.16\linewidth}\centering\small \citet{Ikehata14}\end{minipage}\hfil
	\begin{minipage}{0.16\linewidth}\centering\small Baseline\\(least squares)\end{minipage}\\
	\vskip 2mm
	\includegraphics[width=0.15\linewidth]{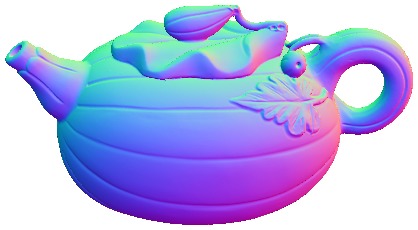}\hfil
	\includegraphics[width=0.15\linewidth]{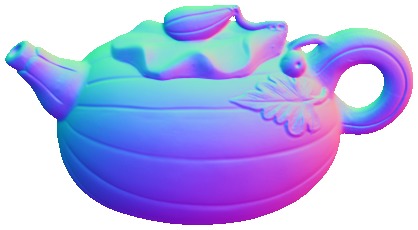}\hfil
	\includegraphics[width=0.15\linewidth]{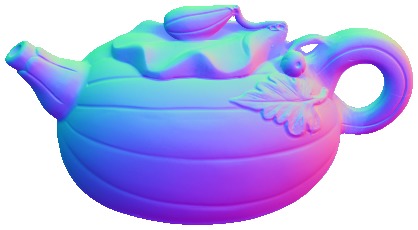}\hfil
	\includegraphics[width=0.15\linewidth]{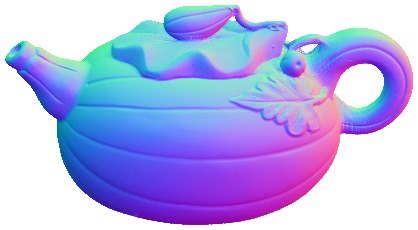}\hfil
	\includegraphics[width=0.15\linewidth]{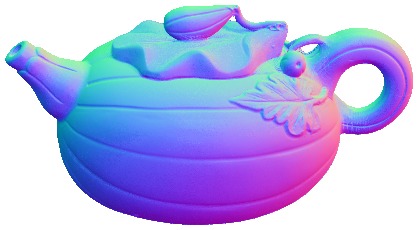}\hfil
	\includegraphics[width=0.15\linewidth]{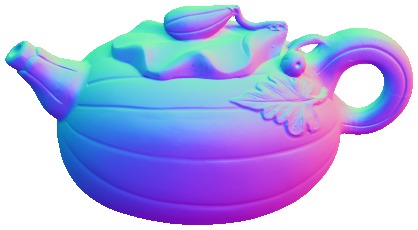}\\
	\vskip 2mm
	\includegraphics[width=0.15\linewidth]{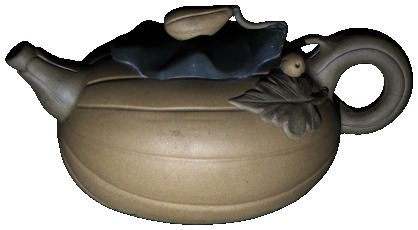}\hfil
	\includegraphics[width=0.15\linewidth]{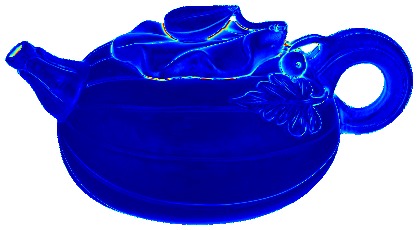}\hfil
	\includegraphics[width=0.15\linewidth]{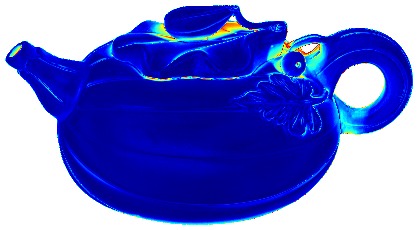}\hfil
	\includegraphics[width=0.15\linewidth]{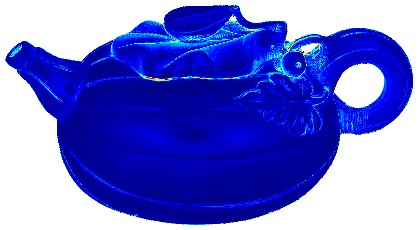}\hfil
	\includegraphics[width=0.15\linewidth]{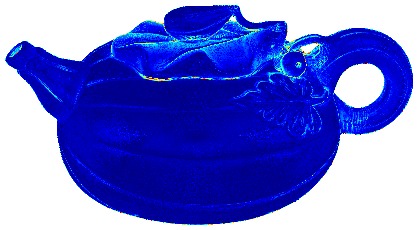}\hfil
	\includegraphics[width=0.15\linewidth]{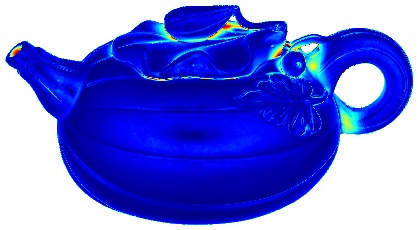}\\
	%\begin{minipage}{0mm}{\vskip -24mm}\hspace*{1mm}\includegraphics[width=6.9mm]{figures/jet_legend}\end{minipage}\\
	\begin{minipage}{0.15\linewidth}\centering \textsc{pot1}\end{minipage}\hfil
	\begin{minipage}{0.15\linewidth}\centering \textbf{6.09}\end{minipage}\hfil
	\begin{minipage}{0.15\linewidth}\centering 7.05\end{minipage}\hfil
	\begin{minipage}{0.15\linewidth}\centering 6.51\end{minipage}\hfil
	\begin{minipage}{0.15\linewidth}\centering 6.64\end{minipage}\hfil
	\begin{minipage}{0.15\linewidth}\centering 8.89\end{minipage}\\
	\caption{\textbf{Visual comparisons for \textsc{pot1} scene.} See also explanations in Fig.~\ref{afig:results_ball}.}
\end{figure}

%\vskip 2mm
%%
%%
%% Bear
\begin{figure}
\centering
\begin{minipage}{0.16\linewidth}\centering\small Ground truth /\\ Observed image\end{minipage}\hfil
\begin{minipage}{0.16\linewidth}\centering\small Ours\end{minipage}\hfil
\begin{minipage}{0.16\linewidth}\centering\small \citet{Santo17}\end{minipage}\hfil
\begin{minipage}{0.16\linewidth}\centering\small \citet{Shi14}\end{minipage}\hfil
\begin{minipage}{0.16\linewidth}\centering\small \citet{Ikehata14}\end{minipage}\hfil
\begin{minipage}{0.16\linewidth}\centering\small Baseline\\(least squares)\end{minipage}\\
\vskip 2mm
\includegraphics[width=0.15\linewidth]{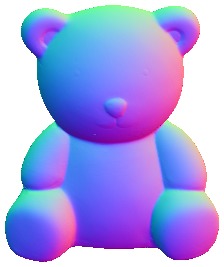}\hfil
\includegraphics[width=0.15\linewidth]{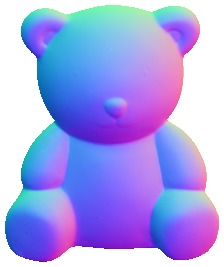}\hfil
\includegraphics[width=0.15\linewidth]{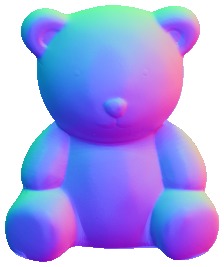}\hfil
\includegraphics[width=0.15\linewidth]{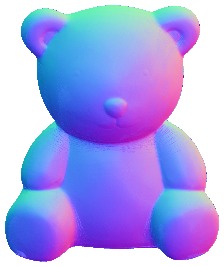}\hfil
\includegraphics[width=0.15\linewidth]{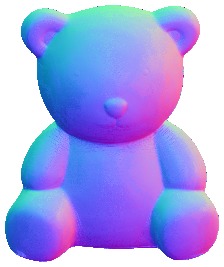}\hfil
\includegraphics[width=0.15\linewidth]{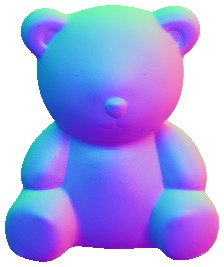}\\
\vskip 2mm
\includegraphics[width=0.15\linewidth]{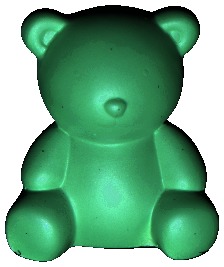}\hfil
\includegraphics[width=0.15\linewidth]{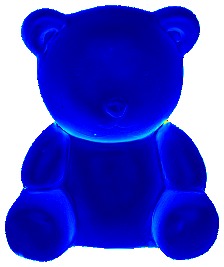}\hfil
\includegraphics[width=0.15\linewidth]{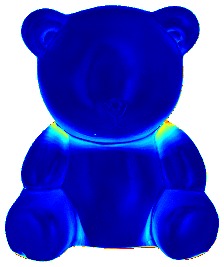}\hfil
\includegraphics[width=0.15\linewidth]{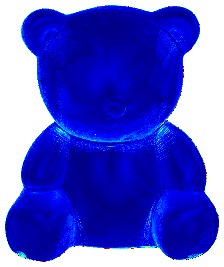}\hfil
\includegraphics[width=0.15\linewidth]{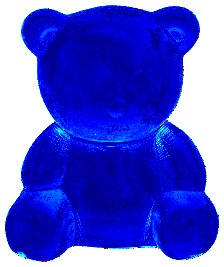}\hfil
\includegraphics[width=0.15\linewidth]{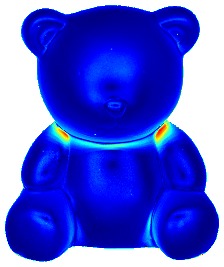}\\
\begin{minipage}{0.15\linewidth}\centering \textsc{bear}\end{minipage}\hfil
\begin{minipage}{0.15\linewidth}\centering \textbf{5.79}\end{minipage}\hfil
\begin{minipage}{0.15\linewidth}\centering 6.31\end{minipage}\hfil
\begin{minipage}{0.15\linewidth}\centering 6.12\end{minipage}\hfil
\begin{minipage}{0.15\linewidth}\centering 7.11\end{minipage}\hfil
\begin{minipage}{0.15\linewidth}\centering 8.39\end{minipage}
\caption{\textbf{Visual comparisons for \textsc{bear} scene.} See also explanations in Fig.~\ref{afig:results_ball}.}
\end{figure}
%\vskip 2mm
%%
%%
%% Pot2
\begin{figure}
\centering
\begin{minipage}{0.16\linewidth}\centering\small Ground truth /\\ Observed image\end{minipage}\hfil
\begin{minipage}{0.16\linewidth}\centering\small Ours\end{minipage}\hfil
\begin{minipage}{0.16\linewidth}\centering\small \citet{Santo17}\end{minipage}\hfil
\begin{minipage}{0.16\linewidth}\centering\small \citet{Shi14}\end{minipage}\hfil
\begin{minipage}{0.16\linewidth}\centering\small \citet{Ikehata14}\end{minipage}\hfil
\begin{minipage}{0.16\linewidth}\centering\small Baseline\\(least squares)\end{minipage}\\
\vskip 2mm
\includegraphics[width=0.15\linewidth]{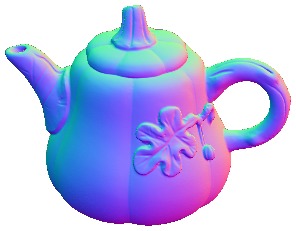}\hfil
\includegraphics[width=0.15\linewidth]{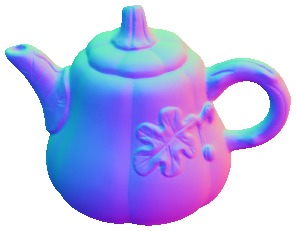}\hfil
\includegraphics[width=0.15\linewidth]{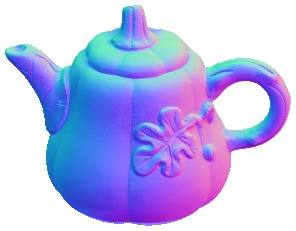}\hfil
\includegraphics[width=0.15\linewidth]{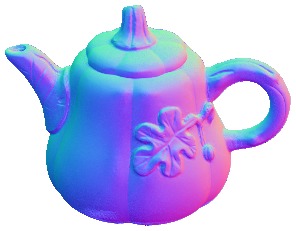}\hfil
\includegraphics[width=0.15\linewidth]{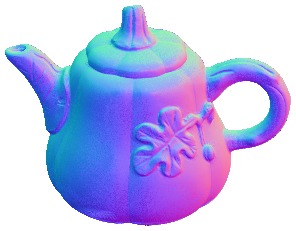}\hfil
\includegraphics[width=0.15\linewidth]{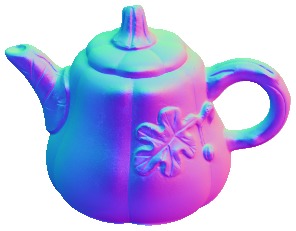}\\
\vskip 2mm
\includegraphics[width=0.15\linewidth]{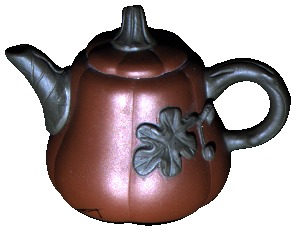}\hfil
\includegraphics[width=0.15\linewidth]{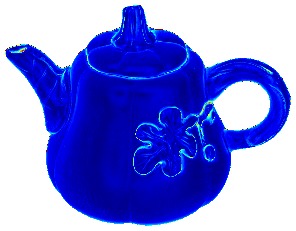}\hfil
\includegraphics[width=0.15\linewidth]{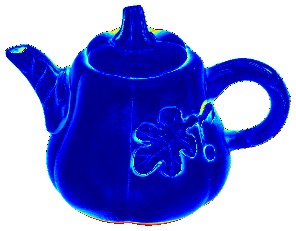}\hfil
\includegraphics[width=0.15\linewidth]{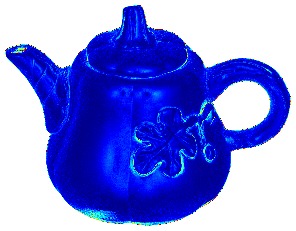}\hfil
\includegraphics[width=0.15\linewidth]{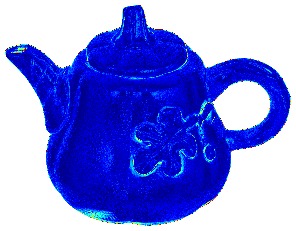}\hfil
\includegraphics[width=0.15\linewidth]{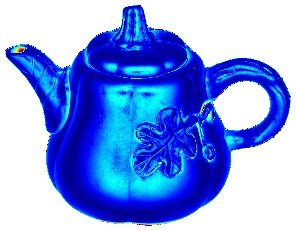}\\
\begin{minipage}{0.15\linewidth}\centering \textsc{pot2}\end{minipage}\hfil
\begin{minipage}{0.15\linewidth}\centering \textbf{7.76}\end{minipage}\hfil
\begin{minipage}{0.15\linewidth}\centering 7.86\end{minipage}\hfil
\begin{minipage}{0.15\linewidth}\centering 8.78\end{minipage}\hfil
\begin{minipage}{0.15\linewidth}\centering 8.77\end{minipage}\hfil
\begin{minipage}{0.15\linewidth}\centering 14.65\end{minipage}
\caption{\textbf{Visual comparisons for \textsc{pot2} scene.} See also explanations in Fig.~\ref{afig:results_ball}.}
\end{figure}
%\vskip 2mm
%%
%%
%% Buddha
\begin{figure}
\centering
\begin{minipage}{0.16\linewidth}\centering\small Ground truth /\\ Observed image\end{minipage}\hfil
\begin{minipage}{0.16\linewidth}\centering\small Ours\end{minipage}\hfil
\begin{minipage}{0.16\linewidth}\centering\small \citet{Santo17}\end{minipage}\hfil
\begin{minipage}{0.16\linewidth}\centering\small \citet{Shi14}\end{minipage}\hfil
\begin{minipage}{0.16\linewidth}\centering\small \citet{Ikehata14}\end{minipage}\hfil
\begin{minipage}{0.16\linewidth}\centering\small Baseline\\(least squares)\end{minipage}\\
\vskip 2mm
\includegraphics[width=0.15\linewidth]{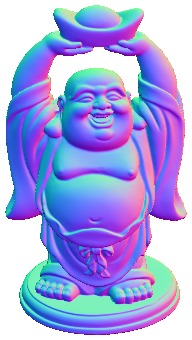}\hfil
\includegraphics[width=0.15\linewidth]{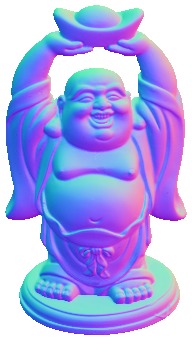}\hfil
\includegraphics[width=0.15\linewidth]{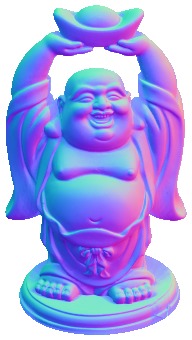}\hfil
\includegraphics[width=0.15\linewidth]{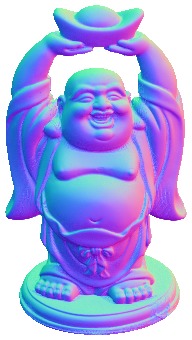}\hfil
\includegraphics[width=0.15\linewidth]{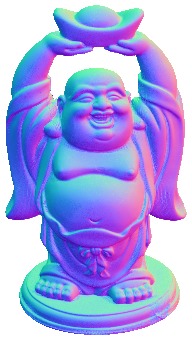}\hfil
\includegraphics[width=0.15\linewidth]{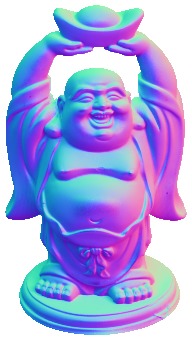}\\
\vskip 2mm
\includegraphics[width=0.15\linewidth]{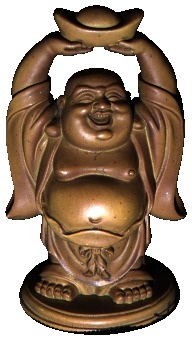}\hfil
\includegraphics[width=0.15\linewidth]{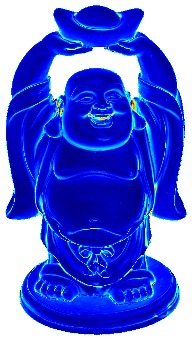}\hfil
\includegraphics[width=0.15\linewidth]{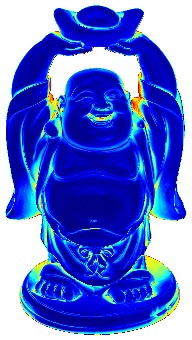}\hfil
\includegraphics[width=0.15\linewidth]{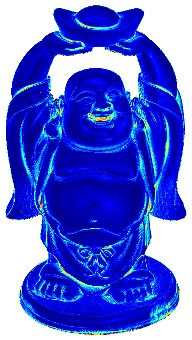}\hfil
\includegraphics[width=0.15\linewidth]{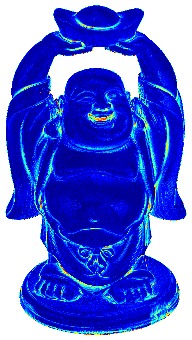}\hfil
\includegraphics[width=0.15\linewidth]{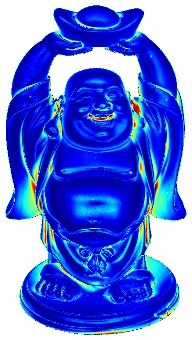}\\
\begin{minipage}{0.15\linewidth}\centering \textsc{buddha}\end{minipage}\hfil
\begin{minipage}{0.15\linewidth}\centering \textbf{10.36}\end{minipage}\hfil
\begin{minipage}{0.15\linewidth}\centering 12.68\end{minipage}\hfil
\begin{minipage}{0.15\linewidth}\centering 10.60\end{minipage}\hfil
\begin{minipage}{0.15\linewidth}\centering 10.47\end{minipage}\hfil
\begin{minipage}{0.15\linewidth}\centering 14.92\end{minipage}
\caption{\textbf{Visual comparisons for \textsc{buddha} scene.} See also explanations in Fig.~\ref{afig:results_ball}.}
\end{figure}
%\vskip 2mm
%
%
%% Goblet
\begin{figure}
\centering
\begin{minipage}{0.16\linewidth}\centering\small Ground truth /\\ Observed image\end{minipage}\hfil
\begin{minipage}{0.16\linewidth}\centering\small Ours\end{minipage}\hfil
\begin{minipage}{0.16\linewidth}\centering\small \citet{Santo17}\end{minipage}\hfil
\begin{minipage}{0.16\linewidth}\centering\small \citet{Shi14}\end{minipage}\hfil
\begin{minipage}{0.16\linewidth}\centering\small \citet{Ikehata14}\end{minipage}\hfil
\begin{minipage}{0.16\linewidth}\centering\small Baseline\\(least squares)\end{minipage}\\
\vskip 2mm
\includegraphics[width=0.15\linewidth]{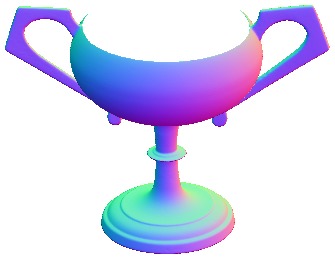}\hfil
\includegraphics[width=0.15\linewidth]{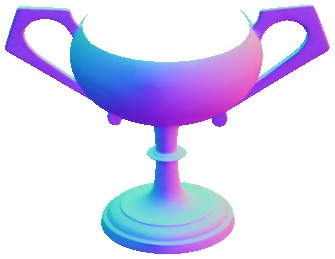}\hfil
\includegraphics[width=0.15\linewidth]{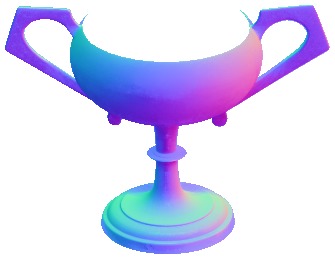}\hfil
\includegraphics[width=0.15\linewidth]{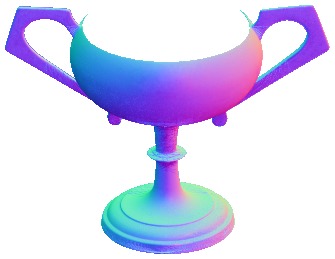}\hfil
\includegraphics[width=0.15\linewidth]{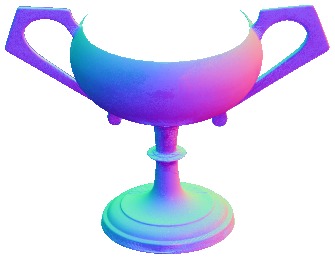}\hfil
\includegraphics[width=0.15\linewidth]{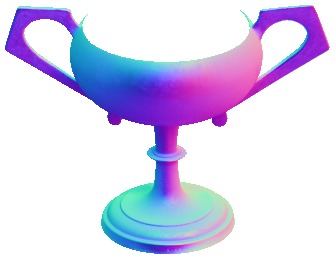}\\
\vskip 2mm
\includegraphics[width=0.15\linewidth]{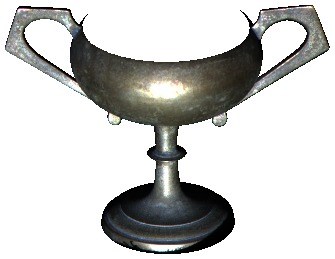}\hfil
\includegraphics[width=0.15\linewidth]{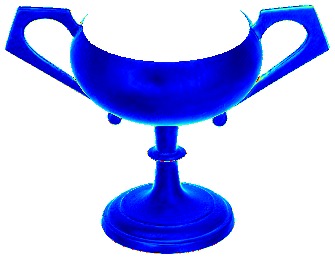}\hfil
\includegraphics[width=0.15\linewidth]{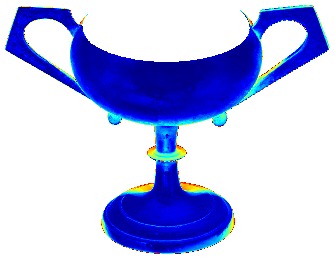}\hfil
\includegraphics[width=0.15\linewidth]{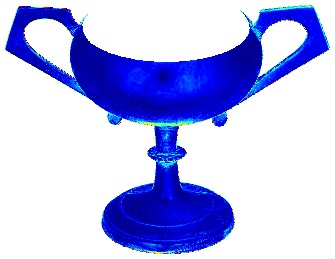}\hfil
\includegraphics[width=0.15\linewidth]{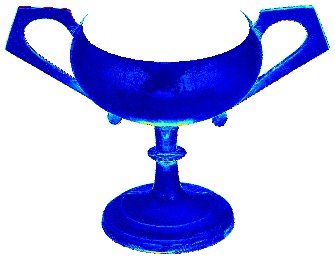}\hfil
\includegraphics[width=0.15\linewidth]{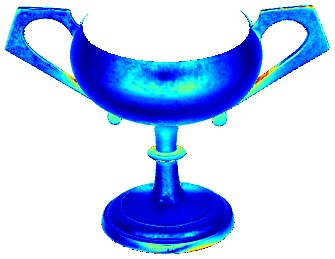}\\
\begin{minipage}{0.15\linewidth}\centering \textsc{goblet}\end{minipage}\hfil
\begin{minipage}{0.15\linewidth}\centering 11.47\end{minipage}\hfil
\begin{minipage}{0.15\linewidth}\centering 11.28\end{minipage}\hfil
\begin{minipage}{0.15\linewidth}\centering 10.09\end{minipage}\hfil
\begin{minipage}{0.15\linewidth}\centering \textbf{9.71}\end{minipage}\hfil
\begin{minipage}{0.15\linewidth}\centering 18.5\end{minipage}
\caption{\textbf{Visual comparisons for \textsc{goblet} scene.} See also explanations in Fig.~\ref{afig:results_ball}.}
\end{figure}
\begin{figure}
\centering
\begin{minipage}{0.16\linewidth}\centering\small Ground truth /\\ Observed image\end{minipage}\hfil
\begin{minipage}{0.16\linewidth}\centering\small Ours\end{minipage}\hfil
\begin{minipage}{0.16\linewidth}\centering\small \citet{Santo17}\end{minipage}\hfil
\begin{minipage}{0.16\linewidth}\centering\small \citet{Shi14}\end{minipage}\hfil
\begin{minipage}{0.16\linewidth}\centering\small \citet{Ikehata14}\end{minipage}\hfil
\begin{minipage}{0.16\linewidth}\centering\small Baseline\\(least squares)\end{minipage}\\
\vskip 2mm
\includegraphics[width=0.15\linewidth]{figures/results/gt/readingPNG/normal}\hfil
\includegraphics[width=0.15\linewidth]{figures/results/ours/readingPNG/normal}\hfil
\includegraphics[width=0.15\linewidth]{figures/results/ICCVW17Santo/readingPNG/normal}\hfil
\includegraphics[width=0.15\linewidth]{figures/results/CVPR12Shi/readingPNG/normal}\hfil
\includegraphics[width=0.15\linewidth]{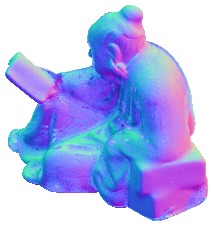}\hfil
\includegraphics[width=0.15\linewidth]{figures/results/l2/readingPNG/normal}\\
\vskip 2mm
\includegraphics[width=0.15\linewidth]{figures/results/ours/readingPNG/img/001}\hfil
\includegraphics[width=0.15\linewidth]{figures/results/ours/readingPNG/error_jet}\hfil
\includegraphics[width=0.15\linewidth]{figures/results/ICCVW17Santo/readingPNG/error_jet}\hfil
\includegraphics[width=0.15\linewidth]{figures/results/CVPR12Shi/readingPNG/error_jet}\hfil
\includegraphics[width=0.15\linewidth]{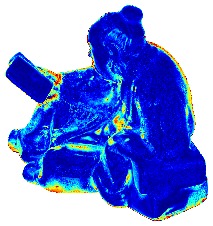}\hfil
\includegraphics[width=0.15\linewidth]{figures/results/l2/readingPNG/error_jet}\\
\begin{minipage}{0.15\linewidth}\centering \textsc{reading}\end{minipage}\hfil
\begin{minipage}{0.15\linewidth}\centering \textbf{11.03}\end{minipage}\hfil
\begin{minipage}{0.15\linewidth}\centering 15.51\end{minipage}\hfil
\begin{minipage}{0.15\linewidth}\centering 13.63\end{minipage}\hfil
\begin{minipage}{0.15\linewidth}\centering 14.19\end{minipage}\hfil
\begin{minipage}{0.15\linewidth}\centering 19.80\end{minipage}
\caption{\textbf{Visual comparisons for \textsc{reading} scene.} See also explanations in Fig.~\ref{afig:results_ball}.}
\end{figure}
%
%
% Cow
\begin{figure}
\centering
\begin{minipage}{0.16\linewidth}\centering\small Ground truth /\\ Observed image\end{minipage}\hfil
\begin{minipage}{0.16\linewidth}\centering\small Ours\end{minipage}\hfil
\begin{minipage}{0.16\linewidth}\centering\small \citet{Santo17}\end{minipage}\hfil
\begin{minipage}{0.16\linewidth}\centering\small \citet{Shi14}\end{minipage}\hfil
\begin{minipage}{0.16\linewidth}\centering\small \citet{Ikehata14}\end{minipage}\hfil
\begin{minipage}{0.16\linewidth}\centering\small Baseline\\(least squares)\end{minipage}\\
\vskip 2mm
\includegraphics[width=0.15\linewidth]{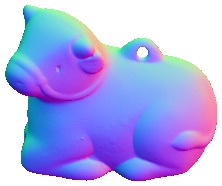}\hfil
\includegraphics[width=0.15\linewidth]{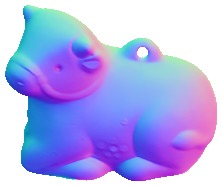}\hfil
\includegraphics[width=0.15\linewidth]{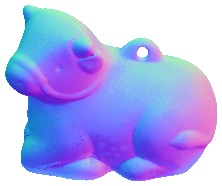}\hfil
\includegraphics[width=0.15\linewidth]{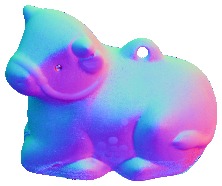}\hfil
\includegraphics[width=0.15\linewidth]{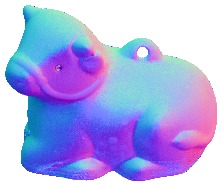}\hfil
\includegraphics[width=0.15\linewidth]{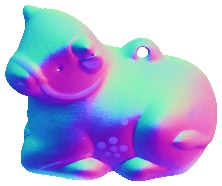}\\
\vskip 2mm
\includegraphics[width=0.15\linewidth]{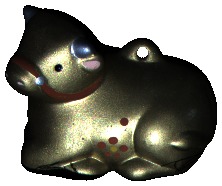}\hfil
\includegraphics[width=0.15\linewidth]{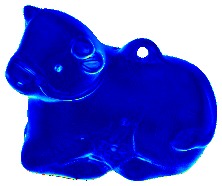}\hfil
\includegraphics[width=0.15\linewidth]{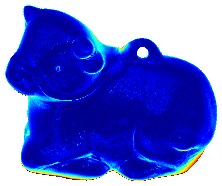}\hfil
\includegraphics[width=0.15\linewidth]{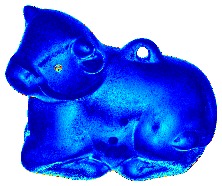}\hfil
\includegraphics[width=0.15\linewidth]{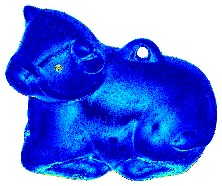}\hfil
\includegraphics[width=0.15\linewidth]{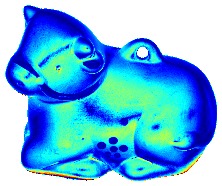}\\
\begin{minipage}{0.15\linewidth}\centering \textsc{cow}\end{minipage}\hfil
\begin{minipage}{0.15\linewidth}\centering \textbf{6.32}\end{minipage}\hfil
\begin{minipage}{0.15\linewidth}\centering 8.01\end{minipage}\hfil
\begin{minipage}{0.15\linewidth}\centering 13.93\end{minipage}\hfil
\begin{minipage}{0.15\linewidth}\centering 13.05\end{minipage}\hfil
\begin{minipage}{0.15\linewidth}\centering 25.60\end{minipage}
\caption{\textbf{Visual comparisons for \textsc{cow} scene.} See also explanations in Fig.~\ref{afig:results_ball}.}
\end{figure}
%
%
% Harvest

\def\ImgA{01}
\def\ImgB{20}
\def\ImgC{39}
\def\ImgD{58}
\def\ImgE{77}
\def\ImgF{96}

\begin{figure}[p]
	\centering
	\small
	\begin{minipage}{\linewidth}
		\flushright
		\begin{minipage}{0.15\linewidth}\centering\small Image \ImgA\end{minipage}\hfil
		\begin{minipage}{0.15\linewidth}\centering\small Image \ImgB\end{minipage}\hfil
		\begin{minipage}{0.15\linewidth}\centering\small Image \ImgC\end{minipage}\hfil
		\begin{minipage}{0.15\linewidth}\centering\small Image \ImgD\end{minipage}\hfil
		\begin{minipage}{0.15\linewidth}\centering\small Image \ImgE\end{minipage}\hfil
		\begin{minipage}{0.15\linewidth}\centering\small Image \ImgF\end{minipage}\\
	\end{minipage}
	\vskip 2mm
	\begin{minipage}{\linewidth}
		\flushright
		\begin{minipage}{0mm}{\vskip -25mm}\hspace{-1em}\rotatebox[origin=l]{90}{\emph{Observed}}\end{minipage}
		\includegraphics[width=0.15\linewidth]{figures/results/ours/ballPNG/img/0\ImgA}\hfil
		\includegraphics[width=0.15\linewidth]{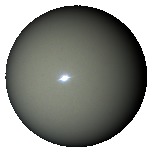}\hfil
		\includegraphics[width=0.15\linewidth]{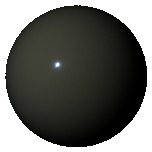}\hfil
		\includegraphics[width=0.15\linewidth]{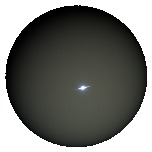}\hfil
		\includegraphics[width=0.15\linewidth]{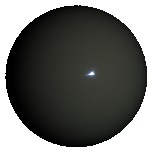}\hfil
		\includegraphics[width=0.15\linewidth]{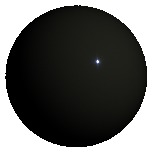}\\
	\end{minipage}\\
	\vskip 4mm
	\dotfill
	\vskip 4mm
	\begin{minipage}{\linewidth}
		\flushright
		\begin{minipage}{0mm}{\vskip -25mm}\hspace{-1em}\rotatebox[origin=l]{90}{\emph{Synthesized}}\end{minipage}
		\includegraphics[width=0.15\linewidth]{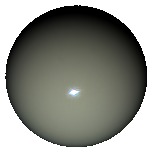}\hfil
		\includegraphics[width=0.15\linewidth]{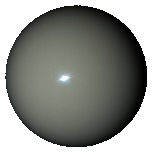}\hfil
		\includegraphics[width=0.15\linewidth]{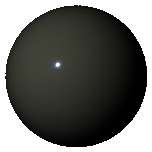}\hfil
		\includegraphics[width=0.15\linewidth]{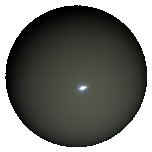}\hfil
		\includegraphics[width=0.15\linewidth]{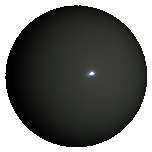}\hfil
		\includegraphics[width=0.15\linewidth]{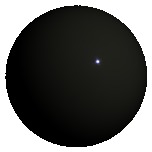}\\
	\end{minipage}
	\vskip 5mm
	\begin{minipage}{\linewidth}
		\flushright
		\begin{minipage}{0mm}{\vskip -25mm}\hspace{-1em}\rotatebox[origin=l]{90}{\emph{Reflectance}}\end{minipage}
		\includegraphics[width=0.15\linewidth]{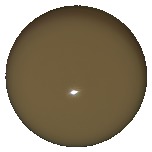}\hfil
		\includegraphics[width=0.15\linewidth]{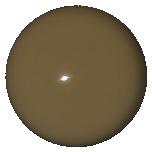}\hfil
		\includegraphics[width=0.15\linewidth]{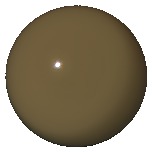}\hfil
		\includegraphics[width=0.15\linewidth]{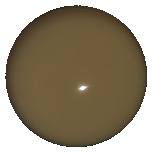}\hfil
		\includegraphics[width=0.15\linewidth]{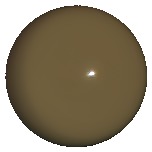}\hfil
		\includegraphics[width=0.15\linewidth]{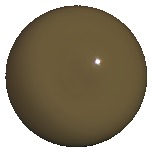}\\
	\end{minipage}
	\vskip 5mm
	\begin{minipage}{\linewidth}
		\flushright
		\begin{minipage}{0mm}{\vskip -25mm}\hspace{-1em}\rotatebox[origin=l]{90}{\emph{Errors}}\end{minipage}
		\colorbox{bgcolor}{\includegraphics[width=0.15\linewidth]{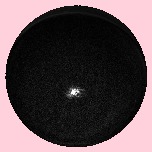}}\hfil
		\colorbox{bgcolor}{\includegraphics[width=0.15\linewidth]{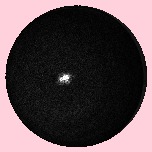}}\hfil
		\colorbox{bgcolor}{\includegraphics[width=0.15\linewidth]{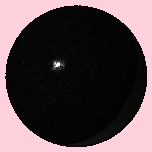}}\hfil
		\colorbox{bgcolor}{\includegraphics[width=0.15\linewidth]{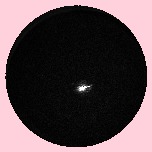}}\hfil
		\colorbox{bgcolor}{\includegraphics[width=0.15\linewidth]{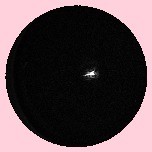}}\hfil
		\colorbox{bgcolor}{\includegraphics[width=0.15\linewidth]{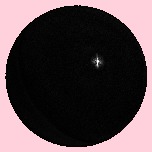}}\\
	\end{minipage}
	\caption{\textbf{Our image reconstruction results for \textsc{ball} scene.} We show our image reconstruction results for six selected images. From top to bottom, we show observed images, our synthesized images, our reflectance images, and reconstruction errors between observed and synthesized images. The error maps show absolute intensity differences scaled by a factor of 10.}
	\label{afig:reconst_ball}
\end{figure}

\begin{figure}[p]
	\centering
	\small
	\begin{minipage}{\linewidth}
		\flushright
		\begin{minipage}{0.15\linewidth}\centering\small Image \ImgA\end{minipage}\hfil
		\begin{minipage}{0.15\linewidth}\centering\small Image \ImgB\end{minipage}\hfil
		\begin{minipage}{0.15\linewidth}\centering\small Image \ImgC\end{minipage}\hfil
		\begin{minipage}{0.15\linewidth}\centering\small Image \ImgD\end{minipage}\hfil
		\begin{minipage}{0.15\linewidth}\centering\small Image \ImgE\end{minipage}\hfil
		\begin{minipage}{0.15\linewidth}\centering\small Image \ImgF\end{minipage}\\
	\end{minipage}
	\vskip 2mm
	\begin{minipage}{\linewidth}
		\flushright
		\begin{minipage}{0mm}{\vskip -25mm}\hspace{-1em}\rotatebox[origin=l]{90}{\emph{Observed}}\end{minipage}
		\includegraphics[width=0.15\linewidth]{figures/results/ours/catPNG/img/0\ImgA}\hfil
		\includegraphics[width=0.15\linewidth]{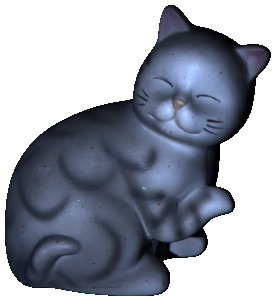}\hfil
		\includegraphics[width=0.15\linewidth]{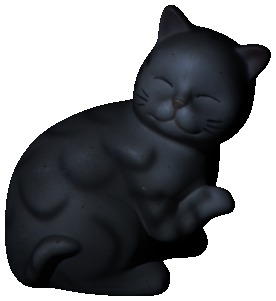}\hfil
		\includegraphics[width=0.15\linewidth]{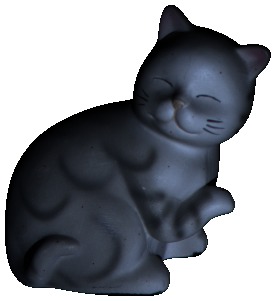}\hfil
		\includegraphics[width=0.15\linewidth]{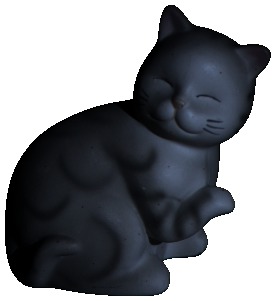}\hfil
		\includegraphics[width=0.15\linewidth]{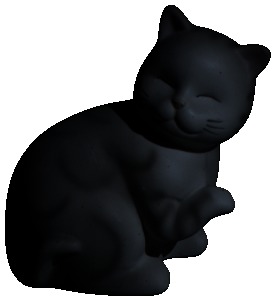}\\
	\end{minipage}\\
\vskip 4mm
\dotfill
\vskip 4mm
	\begin{minipage}{\linewidth}
		\flushright
		\begin{minipage}{0mm}{\vskip -26mm}\hspace{-1em}\rotatebox[origin=l]{90}{\emph{Synthesized}}\end{minipage}
		\includegraphics[width=0.15\linewidth]{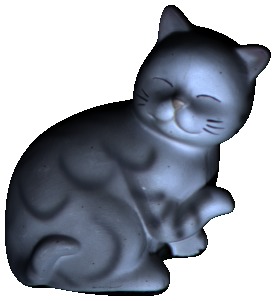}\hfil
		\includegraphics[width=0.15\linewidth]{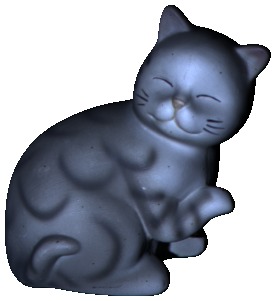}\hfil
		\includegraphics[width=0.15\linewidth]{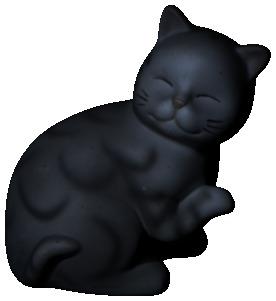}\hfil
		\includegraphics[width=0.15\linewidth]{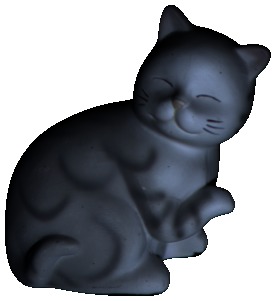}\hfil
		\includegraphics[width=0.15\linewidth]{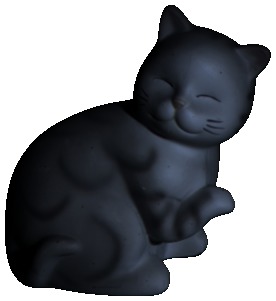}\hfil
		\includegraphics[width=0.15\linewidth]{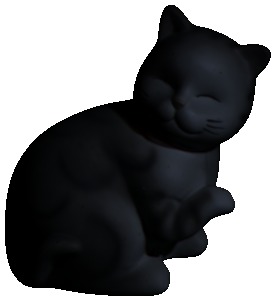}\\
	\end{minipage}
	\vskip 5mm
	\begin{minipage}{\linewidth}
		\flushright
		\begin{minipage}{0mm}{\vskip -26mm}\hspace{-1em}\rotatebox[origin=l]{90}{\emph{Reflectance}}\end{minipage}
		\includegraphics[width=0.15\linewidth]{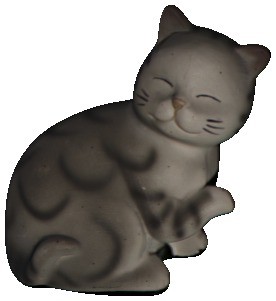}\hfil
		\includegraphics[width=0.15\linewidth]{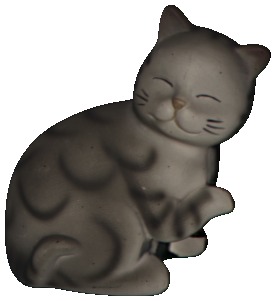}\hfil
		\includegraphics[width=0.15\linewidth]{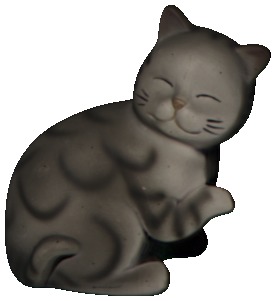}\hfil
		\includegraphics[width=0.15\linewidth]{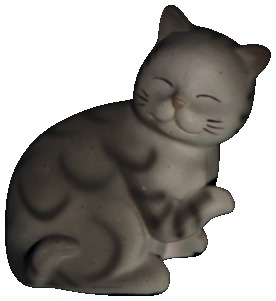}\hfil
		\includegraphics[width=0.15\linewidth]{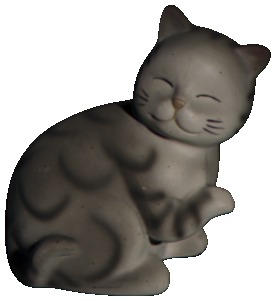}\hfil
		\includegraphics[width=0.15\linewidth]{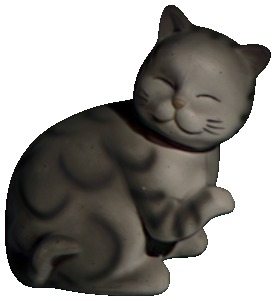}\\
	\end{minipage}
	\vskip 5mm
	\begin{minipage}{\linewidth}
		\flushright
		\begin{minipage}{0mm}{\vskip -26mm}\hspace{-1em}\rotatebox[origin=l]{90}{\emph{Errors}}\end{minipage}
		\colorbox{bgcolor}{\includegraphics[width=0.15\linewidth]{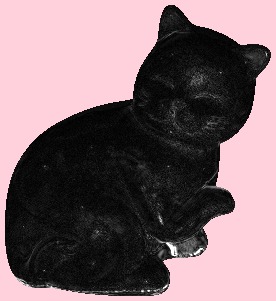}}\hfil
		\colorbox{bgcolor}{\includegraphics[width=0.15\linewidth]{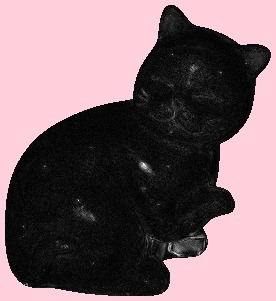}}\hfil
		\colorbox{bgcolor}{\includegraphics[width=0.15\linewidth]{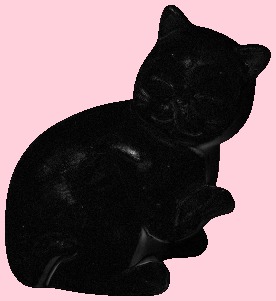}}\hfil
		\colorbox{bgcolor}{\includegraphics[width=0.15\linewidth]{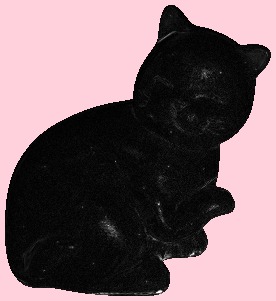}}\hfil
		\colorbox{bgcolor}{\includegraphics[width=0.15\linewidth]{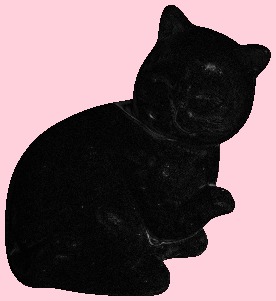}}\hfil
		\colorbox{bgcolor}{\includegraphics[width=0.15\linewidth]{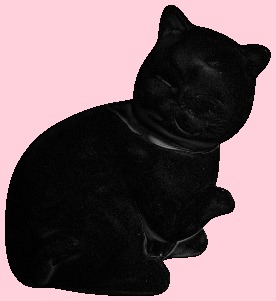}}\\
	\end{minipage}
	\caption{\textbf{Our image reconstruction results for \textsc{cat} scene.} See also explanations in Fig.~\ref{afig:reconst_ball}.}
\end{figure}

\begin{figure}[p]
	\centering
	\small
	\begin{minipage}{\linewidth}
		\flushright
		\begin{minipage}{0.15\linewidth}\centering\small Image \ImgA\end{minipage}\hfil
		\begin{minipage}{0.15\linewidth}\centering\small Image \ImgB\end{minipage}\hfil
		\begin{minipage}{0.15\linewidth}\centering\small Image \ImgC\end{minipage}\hfil
		\begin{minipage}{0.15\linewidth}\centering\small Image \ImgD\end{minipage}\hfil
		\begin{minipage}{0.15\linewidth}\centering\small Image \ImgE\end{minipage}\hfil
		\begin{minipage}{0.15\linewidth}\centering\small Image \ImgF\end{minipage}\\
	\end{minipage}
	\vskip 2mm
	\begin{minipage}{\linewidth}
		\flushright
		\begin{minipage}{0mm}{\vskip -12mm}\hspace{-1em}\rotatebox[origin=l]{90}{\emph{Observed}}\end{minipage}
		\includegraphics[width=0.15\linewidth]{figures/results/ours/pot1PNG/img/0\ImgA}\hfil
		\includegraphics[width=0.15\linewidth]{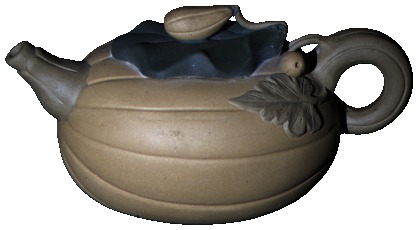}\hfil
		\includegraphics[width=0.15\linewidth]{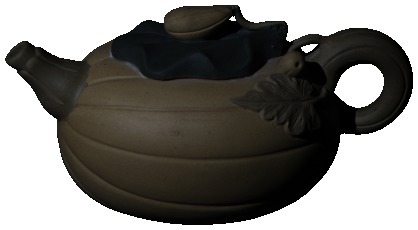}\hfil
		\includegraphics[width=0.15\linewidth]{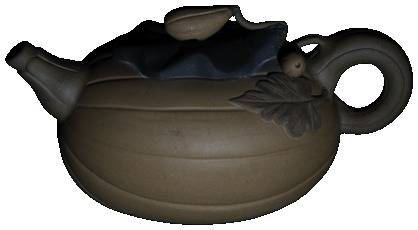}\hfil
		\includegraphics[width=0.15\linewidth]{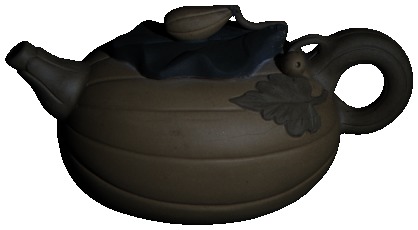}\hfil
		\includegraphics[width=0.15\linewidth]{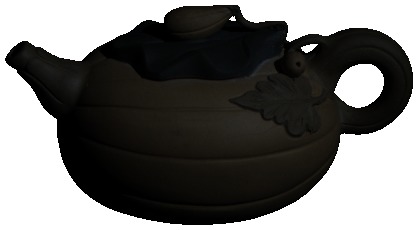}\\
	\end{minipage}\\
\vskip 4mm
\dotfill
\vskip 4mm
	\begin{minipage}{\linewidth}
		\flushright
		\begin{minipage}{0mm}{\vskip -12mm}\hspace{-1em}\rotatebox[origin=l]{90}{\emph{Synthesized}}\end{minipage}
		\includegraphics[width=0.15\linewidth]{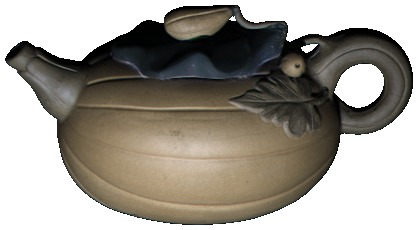}\hfil
		\includegraphics[width=0.15\linewidth]{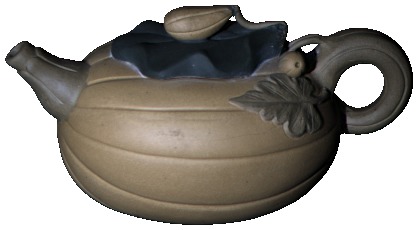}\hfil
		\includegraphics[width=0.15\linewidth]{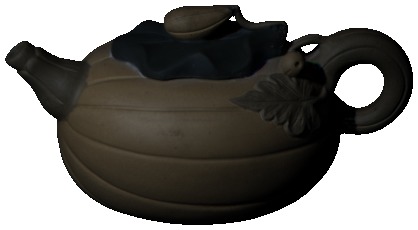}\hfil
		\includegraphics[width=0.15\linewidth]{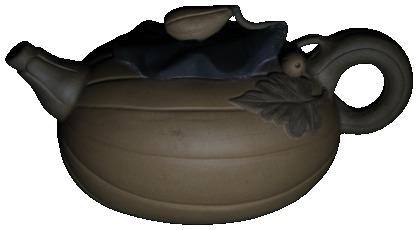}\hfil
		\includegraphics[width=0.15\linewidth]{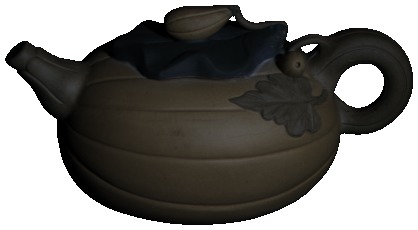}\hfil
		\includegraphics[width=0.15\linewidth]{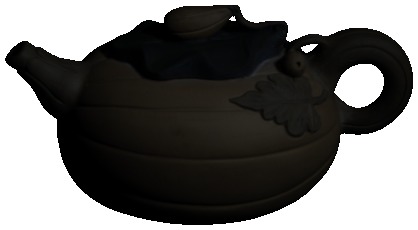}\\
	\end{minipage}
	\vskip 5mm
	\begin{minipage}{\linewidth}
		\flushright
		\begin{minipage}{0mm}{\vskip -12mm}\hspace{-1em}\rotatebox[origin=l]{90}{\emph{Reflectance}}\end{minipage}
		\includegraphics[width=0.15\linewidth]{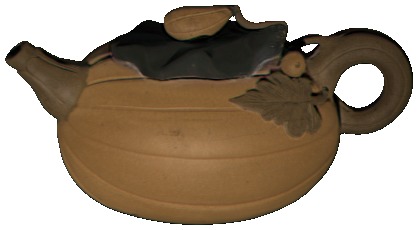}\hfil
		\includegraphics[width=0.15\linewidth]{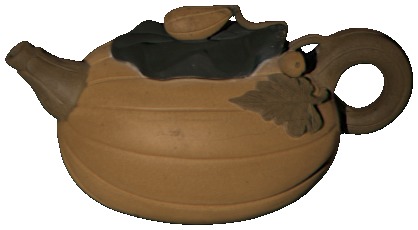}\hfil
		\includegraphics[width=0.15\linewidth]{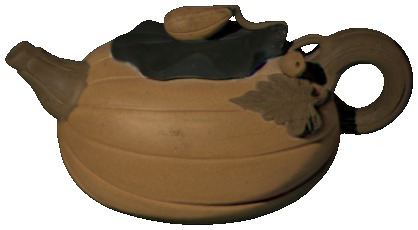}\hfil
		\includegraphics[width=0.15\linewidth]{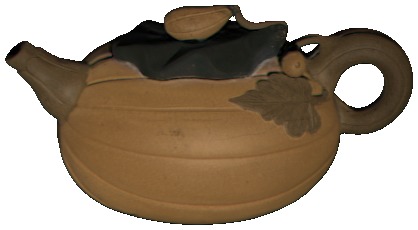}\hfil
		\includegraphics[width=0.15\linewidth]{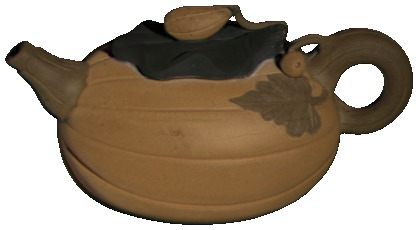}\hfil
		\includegraphics[width=0.15\linewidth]{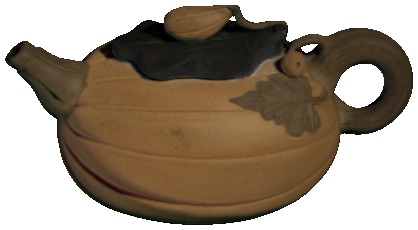}\\
	\end{minipage}
	\vskip 5mm
	\begin{minipage}{\linewidth}
		\flushright
		\begin{minipage}{0mm}{\vskip -12mm}\hspace{-1em}\rotatebox[origin=l]{90}{\emph{Errors}}\end{minipage}
		\colorbox{bgcolor}{\includegraphics[width=0.15\linewidth]{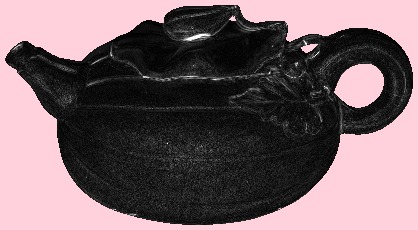}}\hfil
		\colorbox{bgcolor}{\includegraphics[width=0.15\linewidth]{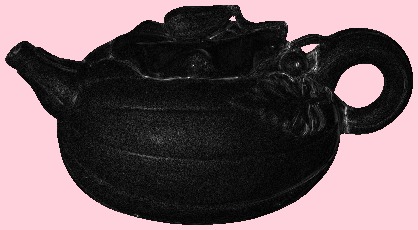}}\hfil
		\colorbox{bgcolor}{\includegraphics[width=0.15\linewidth]{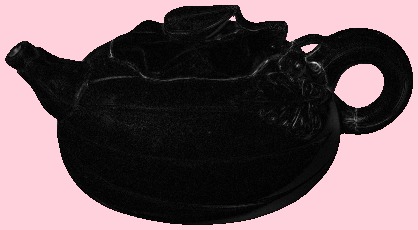}}\hfil
		\colorbox{bgcolor}{\includegraphics[width=0.15\linewidth]{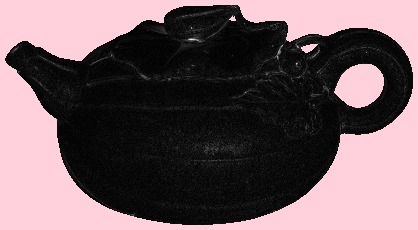}}\hfil
		\colorbox{bgcolor}{\includegraphics[width=0.15\linewidth]{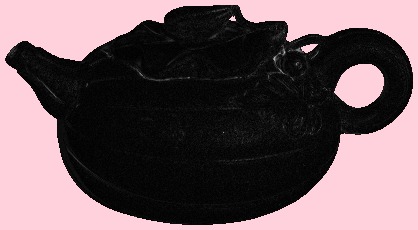}}\hfil
		\colorbox{bgcolor}{\includegraphics[width=0.15\linewidth]{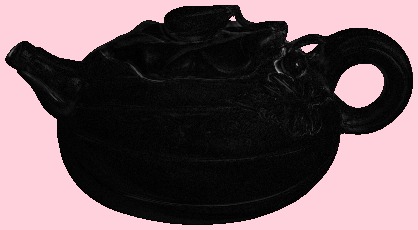}}\\
	\end{minipage}
	\caption{\textbf{Our image reconstruction results for \textsc{pot1} scene.} See also explanations in Fig.~\ref{afig:reconst_ball}.}
\end{figure}

\begin{figure}[p]
	\centering
	\small
	\begin{minipage}{\linewidth}
		\flushright
		\begin{minipage}{0.15\linewidth}\centering\small Image \ImgA\end{minipage}\hfil
		\begin{minipage}{0.15\linewidth}\centering\small Image \ImgB\end{minipage}\hfil
		\begin{minipage}{0.15\linewidth}\centering\small Image \ImgC\end{minipage}\hfil
		\begin{minipage}{0.15\linewidth}\centering\small Image \ImgD\end{minipage}\hfil
		\begin{minipage}{0.15\linewidth}\centering\small Image \ImgE\end{minipage}\hfil
		\begin{minipage}{0.15\linewidth}\centering\small Image \ImgF\end{minipage}\\
	\end{minipage}
	\vskip 2mm
	\begin{minipage}{\linewidth}
		\flushright
		\begin{minipage}{0mm}{\vskip -30mm}\hspace{-1em}\rotatebox[origin=l]{90}{\emph{Observed}}\end{minipage}
		\includegraphics[width=0.15\linewidth]{figures/results/ours/bearPNG/img/0\ImgA}\hfil
		\includegraphics[width=0.15\linewidth]{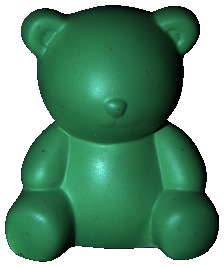}\hfil
		\includegraphics[width=0.15\linewidth]{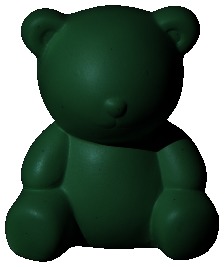}\hfil
		\includegraphics[width=0.15\linewidth]{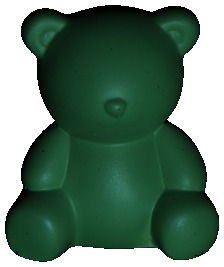}\hfil
		\includegraphics[width=0.15\linewidth]{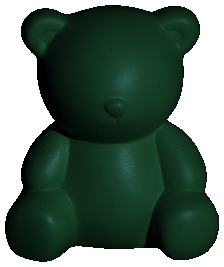}\hfil
		\includegraphics[width=0.15\linewidth]{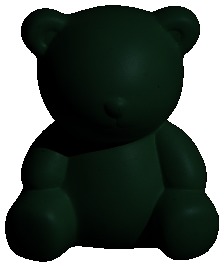}\\
	\end{minipage}\\
\vskip 4mm
\dotfill
\vskip 4mm
	\begin{minipage}{\linewidth}
		\flushright
		\begin{minipage}{0mm}{\vskip -30mm}\hspace{-1em}\rotatebox[origin=l]{90}{\emph{Synthesized}}\end{minipage}
		\includegraphics[width=0.15\linewidth]{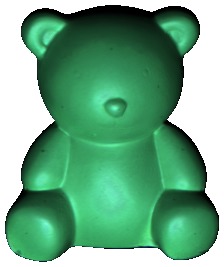}\hfil
		\includegraphics[width=0.15\linewidth]{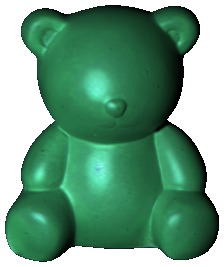}\hfil
		\includegraphics[width=0.15\linewidth]{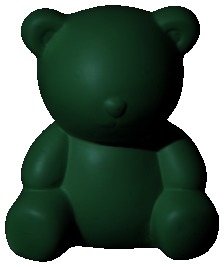}\hfil
		\includegraphics[width=0.15\linewidth]{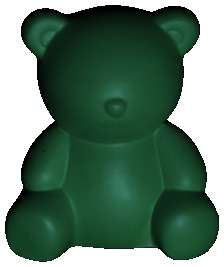}\hfil
		\includegraphics[width=0.15\linewidth]{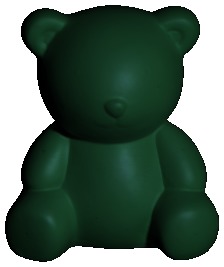}\hfil
		\includegraphics[width=0.15\linewidth]{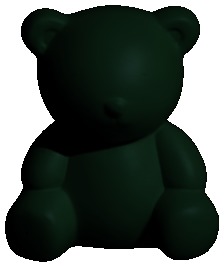}\\
	\end{minipage}
	\vskip 5mm
	\begin{minipage}{\linewidth}
		\flushright
		\begin{minipage}{0mm}{\vskip -30mm}\hspace{-1em}\rotatebox[origin=l]{90}{\emph{Reflectance}}\end{minipage}
		\includegraphics[width=0.15\linewidth]{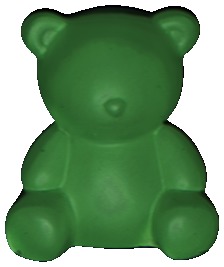}\hfil
		\includegraphics[width=0.15\linewidth]{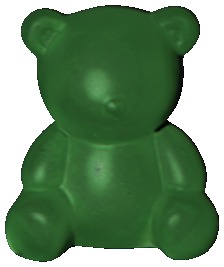}\hfil
		\includegraphics[width=0.15\linewidth]{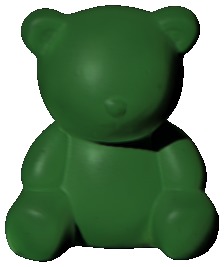}\hfil
		\includegraphics[width=0.15\linewidth]{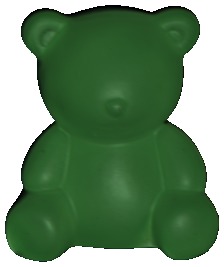}\hfil
		\includegraphics[width=0.15\linewidth]{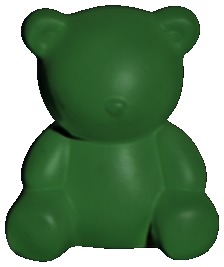}\hfil
		\includegraphics[width=0.15\linewidth]{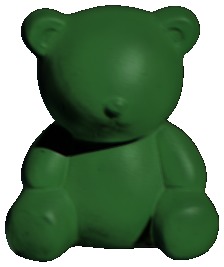}\\
	\end{minipage}
	\vskip 5mm
	\begin{minipage}{\linewidth}
		\flushright
		\begin{minipage}{0mm}{\vskip -30mm}\hspace{-1em}\rotatebox[origin=l]{90}{\emph{Errors}}\end{minipage}
		\colorbox{bgcolor}{\includegraphics[width=0.15\linewidth]{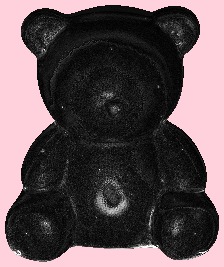}}\hfil
		\colorbox{bgcolor}{\includegraphics[width=0.15\linewidth]{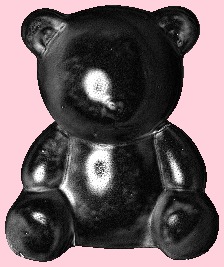}}\hfil
		\colorbox{bgcolor}{\includegraphics[width=0.15\linewidth]{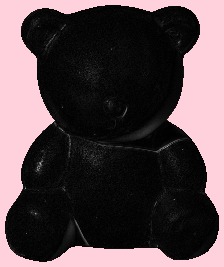}}\hfil
		\colorbox{bgcolor}{\includegraphics[width=0.15\linewidth]{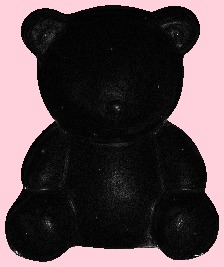}}\hfil
		\colorbox{bgcolor}{\includegraphics[width=0.15\linewidth]{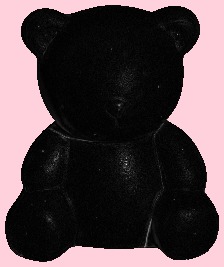}}\hfil
		\colorbox{bgcolor}{\includegraphics[width=0.15\linewidth]{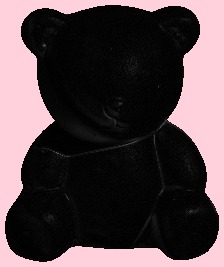}}\\
	\end{minipage}
	\caption{\textbf{Our image reconstruction results for \textsc{bear} scene.} See also explanations in Fig.~\ref{afig:reconst_ball}.}
\end{figure}

\begin{figure}[p]
	\centering
	\small
	\begin{minipage}{\linewidth}
		\flushright
		\begin{minipage}{0.15\linewidth}\centering\small Image \ImgA\end{minipage}\hfil
		\begin{minipage}{0.15\linewidth}\centering\small Image \ImgB\end{minipage}\hfil
		\begin{minipage}{0.15\linewidth}\centering\small Image \ImgC\end{minipage}\hfil
		\begin{minipage}{0.15\linewidth}\centering\small Image \ImgD\end{minipage}\hfil
		\begin{minipage}{0.15\linewidth}\centering\small Image \ImgE\end{minipage}\hfil
		\begin{minipage}{0.15\linewidth}\centering\small Image \ImgF\end{minipage}\\
	\end{minipage}
	\vskip 2mm
	\begin{minipage}{\linewidth}
		\flushright
		\begin{minipage}{0mm}{\vskip -17mm}\hspace{-1em}\rotatebox[origin=l]{90}{\emph{Observed}}\end{minipage}
		\includegraphics[width=0.15\linewidth]{figures/results/ours/pot2PNG/img/0\ImgA}\hfil
		\includegraphics[width=0.15\linewidth]{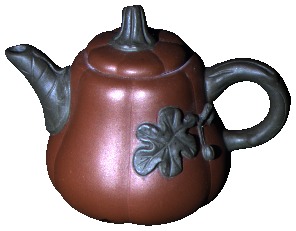}\hfil
		\includegraphics[width=0.15\linewidth]{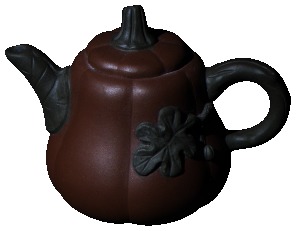}\hfil
		\includegraphics[width=0.15\linewidth]{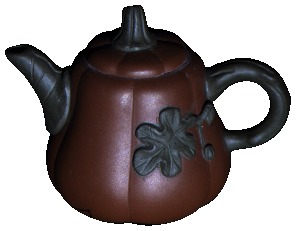}\hfil
		\includegraphics[width=0.15\linewidth]{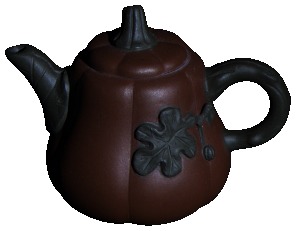}\hfil
		\includegraphics[width=0.15\linewidth]{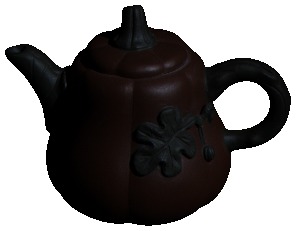}\\
	\end{minipage}\\
\vskip 4mm
\dotfill
\vskip 4mm
	\begin{minipage}{\linewidth}
		\flushright
		\begin{minipage}{0mm}{\vskip -17mm}\hspace{-1em}\rotatebox[origin=l]{90}{\emph{Synthesized}}\end{minipage}
		\includegraphics[width=0.15\linewidth]{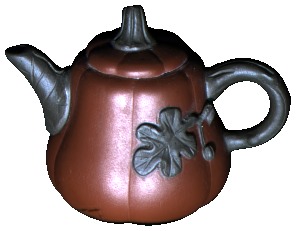}\hfil
		\includegraphics[width=0.15\linewidth]{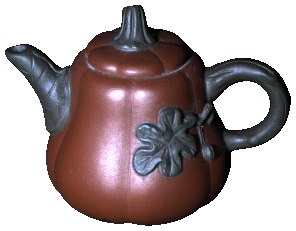}\hfil
		\includegraphics[width=0.15\linewidth]{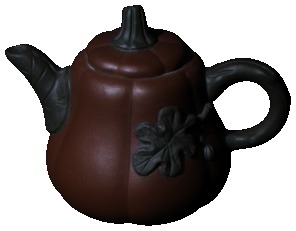}\hfil
		\includegraphics[width=0.15\linewidth]{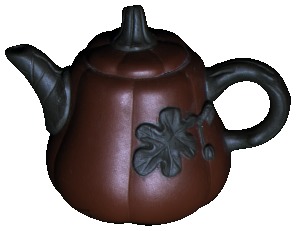}\hfil
		\includegraphics[width=0.15\linewidth]{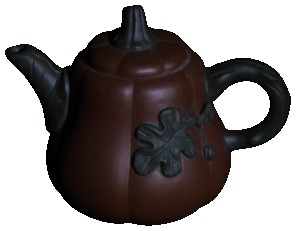}\hfil
		\includegraphics[width=0.15\linewidth]{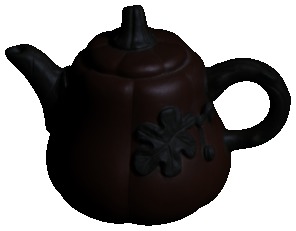}\\
	\end{minipage}
	\vskip 5mm
	\begin{minipage}{\linewidth}
		\flushright
		\begin{minipage}{0mm}{\vskip -17mm}\hspace{-1em}\rotatebox[origin=l]{90}{\emph{Reflectance}}\end{minipage}
		\includegraphics[width=0.15\linewidth]{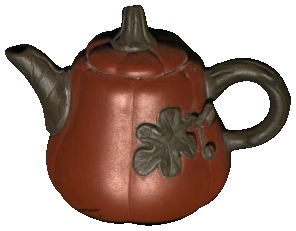}\hfil
		\includegraphics[width=0.15\linewidth]{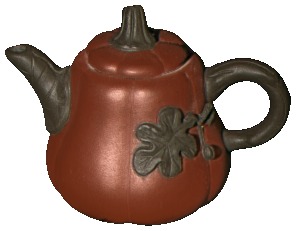}\hfil
		\includegraphics[width=0.15\linewidth]{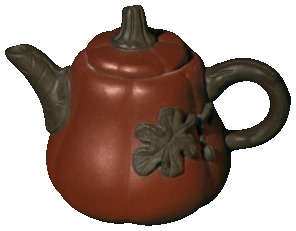}\hfil
		\includegraphics[width=0.15\linewidth]{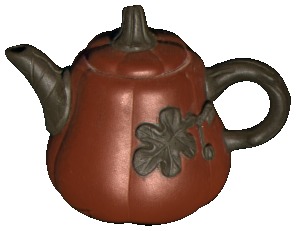}\hfil
		\includegraphics[width=0.15\linewidth]{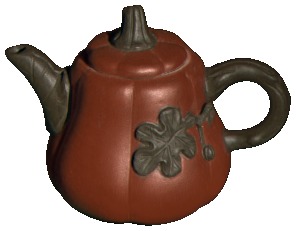}\hfil
		\includegraphics[width=0.15\linewidth]{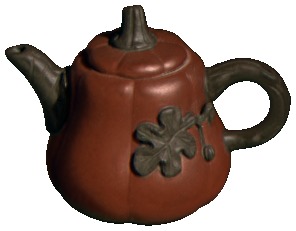}\\
	\end{minipage}
	\vskip 5mm
	\begin{minipage}{\linewidth}
		\flushright
		\begin{minipage}{0mm}{\vskip -17mm}\hspace{-1em}\rotatebox[origin=l]{90}{\emph{Errors}}\end{minipage}
		\colorbox{bgcolor}{\includegraphics[width=0.15\linewidth]{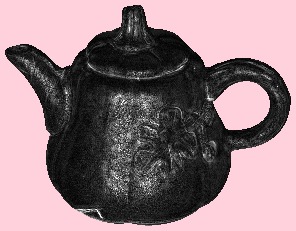}}\hfil
		\colorbox{bgcolor}{\includegraphics[width=0.15\linewidth]{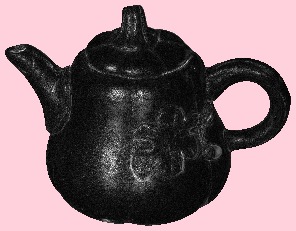}}\hfil
		\colorbox{bgcolor}{\includegraphics[width=0.15\linewidth]{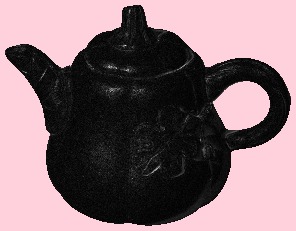}}\hfil
		\colorbox{bgcolor}{\includegraphics[width=0.15\linewidth]{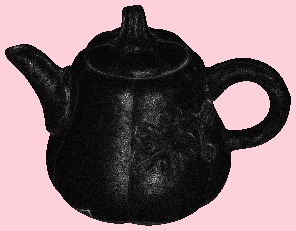}}\hfil
		\colorbox{bgcolor}{\includegraphics[width=0.15\linewidth]{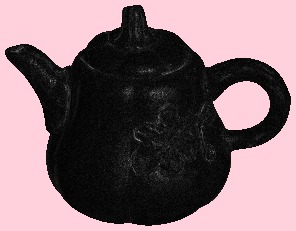}}\hfil
		\colorbox{bgcolor}{\includegraphics[width=0.15\linewidth]{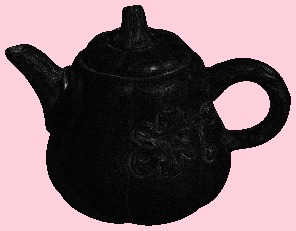}}\\
	\end{minipage}
	\caption{\textbf{Our image reconstruction results for \textsc{pot2} scene.} See also explanations in Fig.~\ref{afig:reconst_ball}.}
\end{figure}

\begin{figure}[p]
	\centering
	\small
	\begin{minipage}{\linewidth}
		\flushright
		\begin{minipage}{0.15\linewidth}\centering\small Image \ImgA\end{minipage}\hfil
		\begin{minipage}{0.15\linewidth}\centering\small Image \ImgB\end{minipage}\hfil
		\begin{minipage}{0.15\linewidth}\centering\small Image \ImgC\end{minipage}\hfil
		\begin{minipage}{0.15\linewidth}\centering\small Image \ImgD\end{minipage}\hfil
		\begin{minipage}{0.15\linewidth}\centering\small Image \ImgE\end{minipage}\hfil
		\begin{minipage}{0.15\linewidth}\centering\small Image \ImgF\end{minipage}\\
	\end{minipage}
	\vskip 2mm
	\begin{minipage}{\linewidth}
		\flushright
		\begin{minipage}{0mm}{\vskip -44mm}\hspace{-1em}\rotatebox[origin=l]{90}{\emph{Observed}}\end{minipage}%
		\includegraphics[width=0.15\linewidth]{figures/results/ours/buddhaPNG/img/0\ImgA}\hfil
		\includegraphics[width=0.15\linewidth]{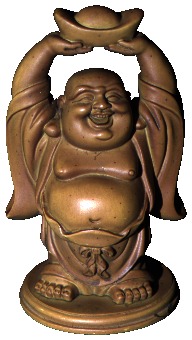}\hfil
		\includegraphics[width=0.15\linewidth]{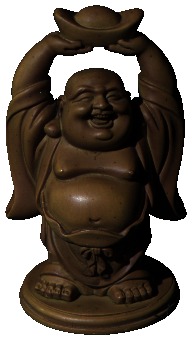}\hfil
		\includegraphics[width=0.15\linewidth]{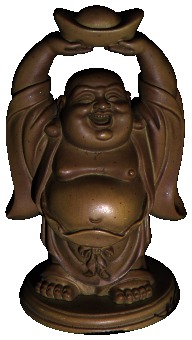}\hfil
		\includegraphics[width=0.15\linewidth]{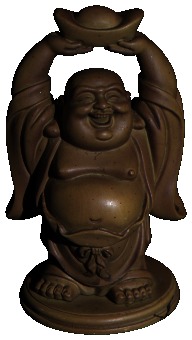}\hfil
		\includegraphics[width=0.15\linewidth]{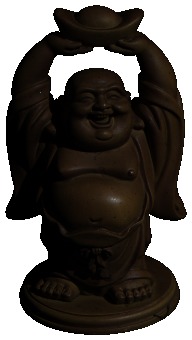}\\
	\end{minipage}\\
\vskip 4mm
\dotfill
\vskip 4mm
	\begin{minipage}{\linewidth}
		\flushright
		\begin{minipage}{0mm}{\vskip -44mm}\hspace{-1em}\rotatebox[origin=l]{90}{\emph{Synthesized}}\end{minipage}
		\includegraphics[width=0.15\linewidth]{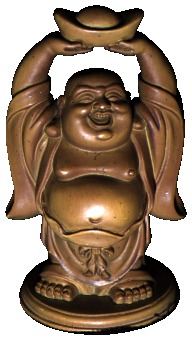}\hfil
		\includegraphics[width=0.15\linewidth]{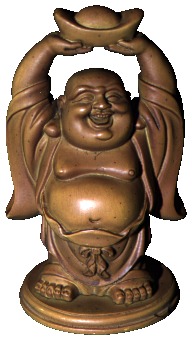}\hfil
		\includegraphics[width=0.15\linewidth]{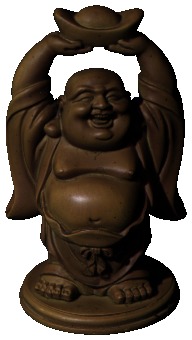}\hfil
		\includegraphics[width=0.15\linewidth]{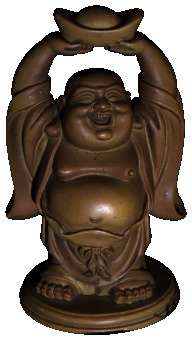}\hfil
		\includegraphics[width=0.15\linewidth]{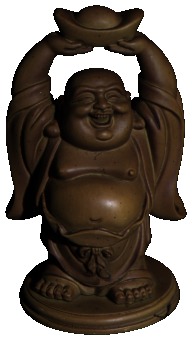}\hfil
		\includegraphics[width=0.15\linewidth]{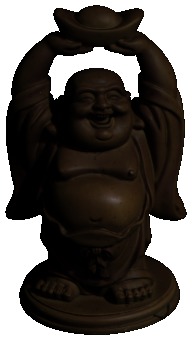}\\
	\end{minipage}
	\vskip 5mm
	\begin{minipage}{\linewidth}
		\flushright
		\begin{minipage}{0mm}{\vskip -44mm}\hspace{-1em}\rotatebox[origin=l]{90}{\emph{Reflectance}}\end{minipage}
		\includegraphics[width=0.15\linewidth]{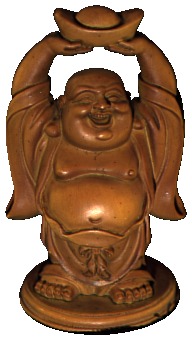}\hfil
		\includegraphics[width=0.15\linewidth]{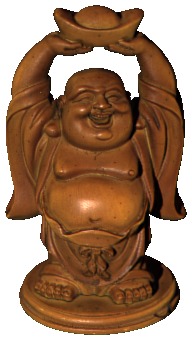}\hfil
		\includegraphics[width=0.15\linewidth]{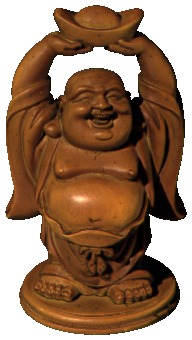}\hfil
		\includegraphics[width=0.15\linewidth]{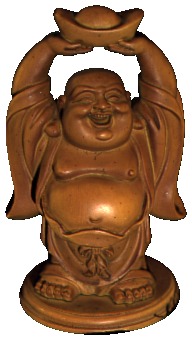}\hfil
		\includegraphics[width=0.15\linewidth]{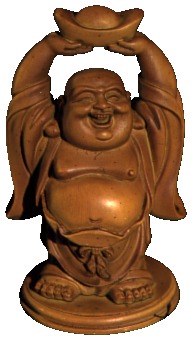}\hfil
		\includegraphics[width=0.15\linewidth]{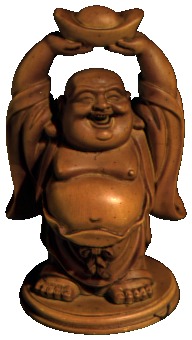}\\
	\end{minipage}
	\vskip 5mm
	\begin{minipage}{\linewidth}
		\flushright
		\begin{minipage}{0mm}{\vskip -44mm}\hspace{-1em}\rotatebox[origin=l]{90}{\emph{Errors}}\end{minipage}
		\colorbox{bgcolor}{\includegraphics[width=0.15\linewidth]{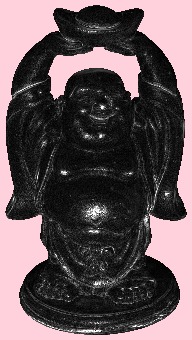}}\hfil
		\colorbox{bgcolor}{\includegraphics[width=0.15\linewidth]{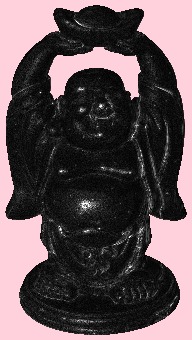}}\hfil
		\colorbox{bgcolor}{\includegraphics[width=0.15\linewidth]{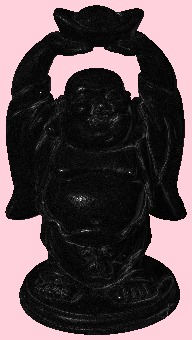}}\hfil
		\colorbox{bgcolor}{\includegraphics[width=0.15\linewidth]{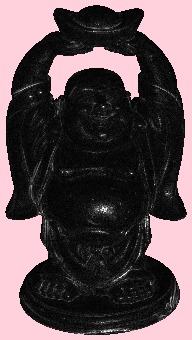}}\hfil
		\colorbox{bgcolor}{\includegraphics[width=0.15\linewidth]{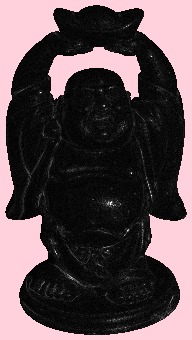}}\hfil
		\colorbox{bgcolor}{\includegraphics[width=0.15\linewidth]{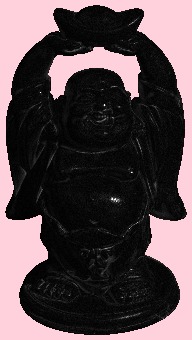}}\\
	\end{minipage}
	\caption{\textbf{Our image reconstruction results for \textsc{buddha} scene.} See also explanations in Fig.~\ref{afig:reconst_ball}.}
\end{figure}

\begin{figure}[p]
	\centering
	\small
	\begin{minipage}{\linewidth}
		\flushright
		\begin{minipage}{0.15\linewidth}\centering\small Image \ImgA\end{minipage}\hfil
		\begin{minipage}{0.15\linewidth}\centering\small Image \ImgB\end{minipage}\hfil
		\begin{minipage}{0.15\linewidth}\centering\small Image \ImgC\end{minipage}\hfil
		\begin{minipage}{0.15\linewidth}\centering\small Image \ImgD\end{minipage}\hfil
		\begin{minipage}{0.15\linewidth}\centering\small Image \ImgE\end{minipage}\hfil
		\begin{minipage}{0.15\linewidth}\centering\small Image \ImgF\end{minipage}\\
	\end{minipage}
	\vskip 2mm
	\begin{minipage}{\linewidth}
		\flushright
		\begin{minipage}{0mm}{\vskip -18mm}\hspace{-1em}\rotatebox[origin=l]{90}{\emph{Observed}}\end{minipage}
		\includegraphics[width=0.15\linewidth]{figures/results/ours/gobletPNG/img/0\ImgA}\hfil
		\includegraphics[width=0.15\linewidth]{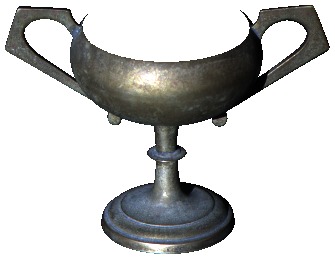}\hfil
		\includegraphics[width=0.15\linewidth]{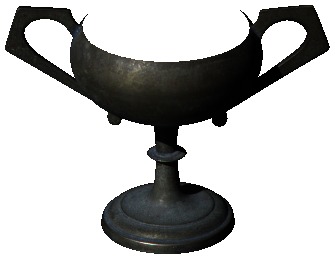}\hfil
		\includegraphics[width=0.15\linewidth]{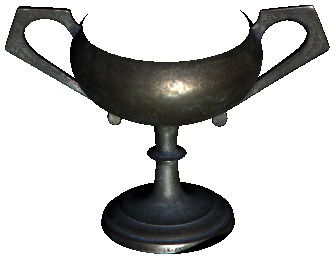}\hfil
		\includegraphics[width=0.15\linewidth]{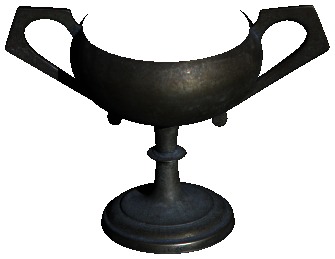}\hfil
		\includegraphics[width=0.15\linewidth]{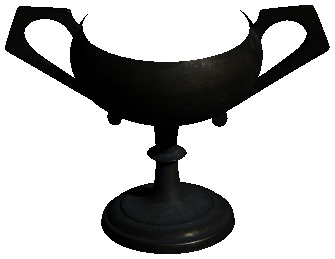}\\
	\end{minipage}\\
\vskip 4mm
\dotfill
\vskip 4mm
	\begin{minipage}{\linewidth}
		\flushright
		\begin{minipage}{0mm}{\vskip -18mm}\hspace{-1em}\rotatebox[origin=l]{90}{\emph{Synthesized}}\end{minipage}
		\includegraphics[width=0.15\linewidth]{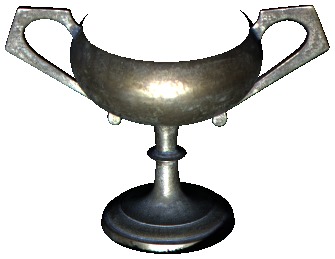}\hfil
		\includegraphics[width=0.15\linewidth]{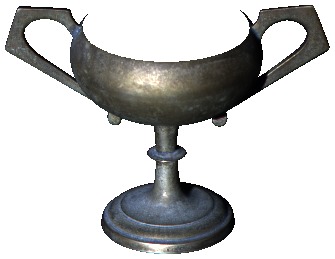}\hfil
		\includegraphics[width=0.15\linewidth]{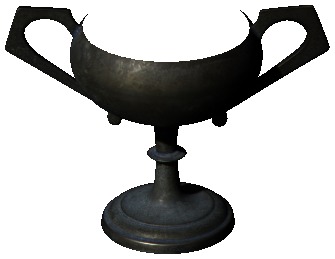}\hfil
		\includegraphics[width=0.15\linewidth]{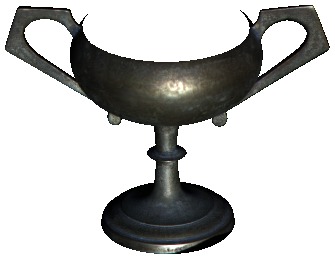}\hfil
		\includegraphics[width=0.15\linewidth]{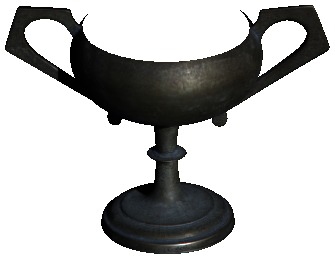}\hfil
		\includegraphics[width=0.15\linewidth]{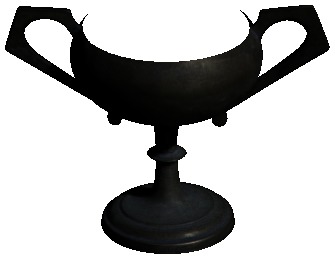}\\
	\end{minipage}
	\vskip 5mm
	\begin{minipage}{\linewidth}
		\flushright
		\begin{minipage}{0mm}{\vskip -18mm}\hspace{-1em}\rotatebox[origin=l]{90}{\emph{Reflectance}}\end{minipage}
		\includegraphics[width=0.15\linewidth]{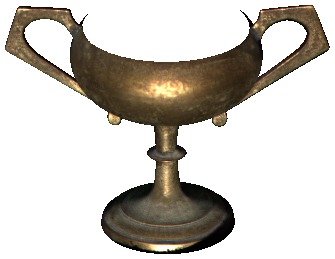}\hfil
		\includegraphics[width=0.15\linewidth]{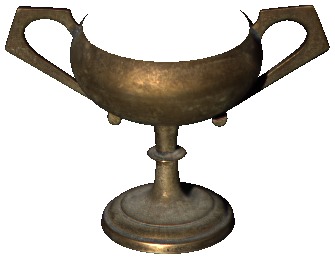}\hfil
		\includegraphics[width=0.15\linewidth]{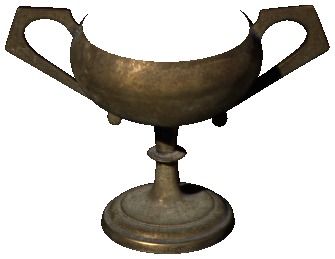}\hfil
		\includegraphics[width=0.15\linewidth]{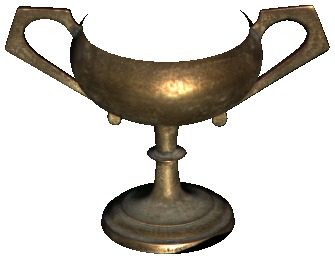}\hfil
		\includegraphics[width=0.15\linewidth]{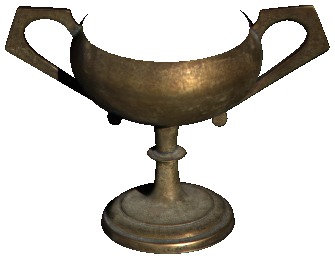}\hfil
		\includegraphics[width=0.15\linewidth]{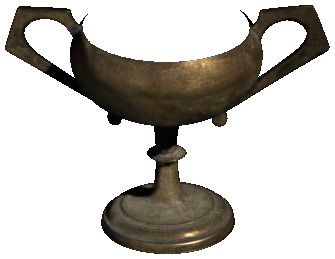}\\
	\end{minipage}
	\vskip 5mm
	\begin{minipage}{\linewidth}
		\flushright
		\begin{minipage}{0mm}{\vskip -18mm}\hspace{-1em}\rotatebox[origin=l]{90}{\emph{Errors}}\end{minipage}
		\colorbox{bgcolor}{\includegraphics[width=0.15\linewidth]{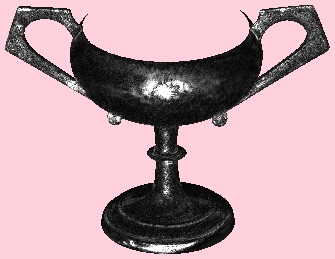}}\hfil
		\colorbox{bgcolor}{\includegraphics[width=0.15\linewidth]{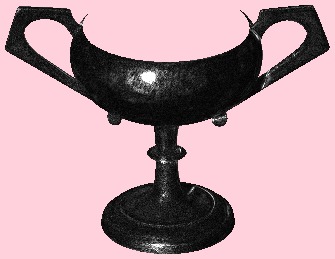}}\hfil
		\colorbox{bgcolor}{\includegraphics[width=0.15\linewidth]{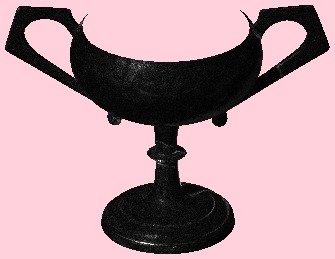}}\hfil
		\colorbox{bgcolor}{\includegraphics[width=0.15\linewidth]{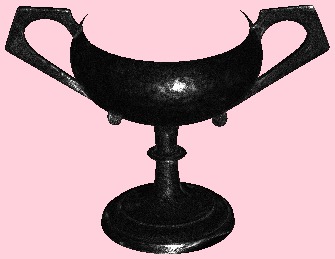}}\hfil
		\colorbox{bgcolor}{\includegraphics[width=0.15\linewidth]{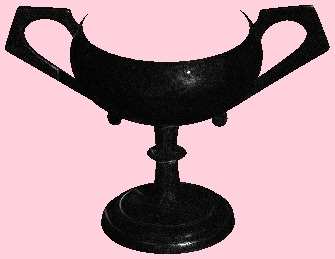}}\hfil
		\colorbox{bgcolor}{\includegraphics[width=0.15\linewidth]{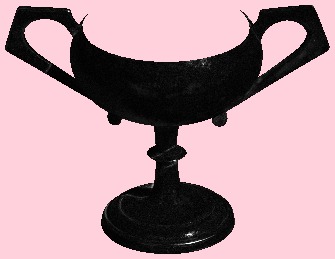}}\\
	\end{minipage}
	\caption{\textbf{Our image reconstruction results for \textsc{goblet} scene.} See also explanations in Fig.~\ref{afig:reconst_ball}.}
\end{figure}

\begin{figure}[p]
	\centering
	\small
	\begin{minipage}{\linewidth}
		\flushright
		\begin{minipage}{0.15\linewidth}\centering\small Image \ImgA\end{minipage}\hfil
		\begin{minipage}{0.15\linewidth}\centering\small Image \ImgB\end{minipage}\hfil
		\begin{minipage}{0.15\linewidth}\centering\small Image \ImgC\end{minipage}\hfil
		\begin{minipage}{0.15\linewidth}\centering\small Image \ImgD\end{minipage}\hfil
		\begin{minipage}{0.15\linewidth}\centering\small Image \ImgE\end{minipage}\hfil
		\begin{minipage}{0.15\linewidth}\centering\small Image \ImgF\end{minipage}\\
	\end{minipage}
	\vskip 2mm
	\begin{minipage}{\linewidth}
		\flushright
		\begin{minipage}{0mm}{\vskip -25mm}\hspace{-1em}\rotatebox[origin=l]{90}{\emph{Observed}}\end{minipage}
		\includegraphics[width=0.15\linewidth]{figures/results/ours/readingPNG/img/0\ImgA}\hfil
		\includegraphics[width=0.15\linewidth]{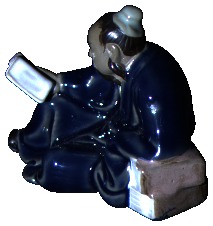}\hfil
		\includegraphics[width=0.15\linewidth]{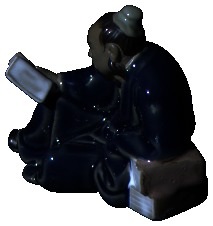}\hfil
		\includegraphics[width=0.15\linewidth]{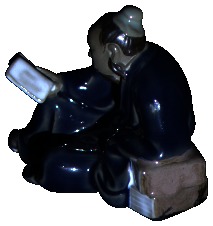}\hfil
		\includegraphics[width=0.15\linewidth]{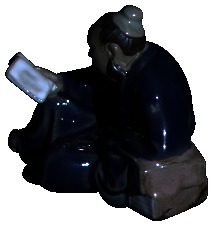}\hfil
		\includegraphics[width=0.15\linewidth]{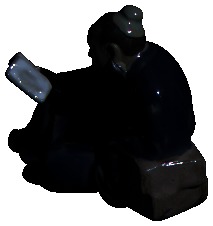}\\
	\end{minipage}\\
\vskip 4mm
\dotfill
\vskip 4mm
	\begin{minipage}{\linewidth}
		\flushright
		\begin{minipage}{0mm}{\vskip -25mm}\hspace{-1em}\rotatebox[origin=l]{90}{\emph{Synthesized}}\end{minipage}
		\includegraphics[width=0.15\linewidth]{figures/results/ours/readingPNG/syn/0\ImgA}\hfil
		\includegraphics[width=0.15\linewidth]{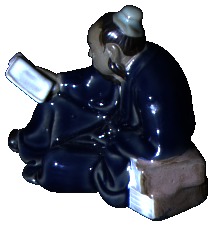}\hfil
		\includegraphics[width=0.15\linewidth]{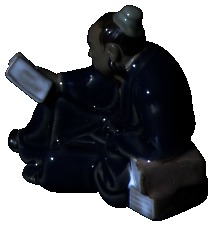}\hfil
		\includegraphics[width=0.15\linewidth]{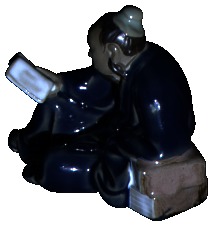}\hfil
		\includegraphics[width=0.15\linewidth]{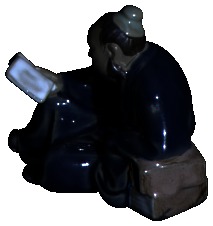}\hfil
		\includegraphics[width=0.15\linewidth]{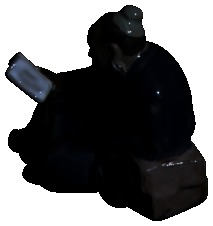}\\
	\end{minipage}
	\vskip 5mm
	\begin{minipage}{\linewidth}
		\flushright
		\begin{minipage}{0mm}{\vskip -25mm}\hspace{-1em}\rotatebox[origin=l]{90}{\emph{Reflectance}}\end{minipage}
		\includegraphics[width=0.15\linewidth]{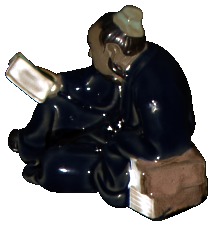}\hfil
		\includegraphics[width=0.15\linewidth]{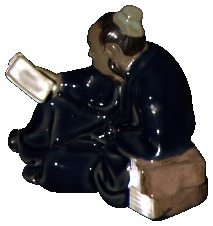}\hfil
		\includegraphics[width=0.15\linewidth]{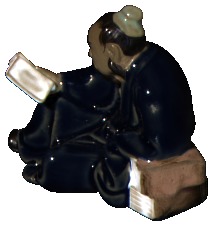}\hfil
		\includegraphics[width=0.15\linewidth]{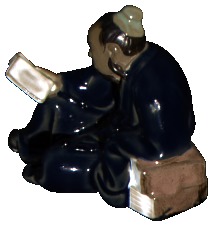}\hfil
		\includegraphics[width=0.15\linewidth]{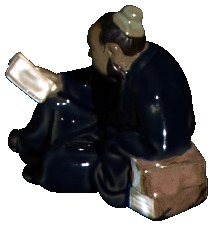}\hfil
		\includegraphics[width=0.15\linewidth]{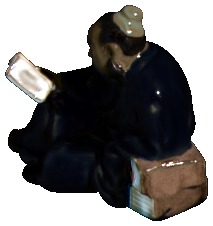}\\
	\end{minipage}
	\vskip 5mm
	\begin{minipage}{\linewidth}
		\flushright
		\begin{minipage}{0mm}{\vskip -25mm}\hspace{-1em}\rotatebox[origin=l]{90}{\emph{Errors}}\end{minipage}
		\colorbox{bgcolor}{\includegraphics[width=0.15\linewidth]{figures/results/ours/readingPNG/err/0\ImgA}}\hfil
		\colorbox{bgcolor}{\includegraphics[width=0.15\linewidth]{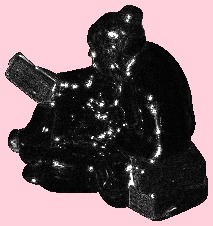}}\hfil
		\colorbox{bgcolor}{\includegraphics[width=0.15\linewidth]{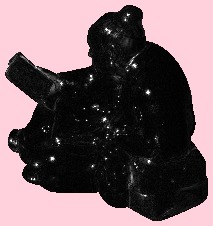}}\hfil
		\colorbox{bgcolor}{\includegraphics[width=0.15\linewidth]{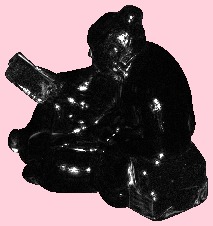}}\hfil
		\colorbox{bgcolor}{\includegraphics[width=0.15\linewidth]{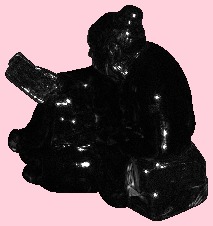}}\hfil
		\colorbox{bgcolor}{\includegraphics[width=0.15\linewidth]{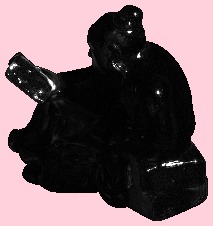}}\\
	\end{minipage}
	\caption{\textbf{Our image reconstruction results for \textsc{reading} scene.} See also explanations in Fig.~\ref{afig:reconst_ball}.}
\end{figure}

\begin{figure}[p]
	\centering
	\small
	\begin{minipage}{\linewidth}
		\flushright
		\begin{minipage}{0.15\linewidth}\centering\small Image \ImgA\end{minipage}\hfil
		\begin{minipage}{0.15\linewidth}\centering\small Image \ImgB\end{minipage}\hfil
		\begin{minipage}{0.15\linewidth}\centering\small Image \ImgC\end{minipage}\hfil
		\begin{minipage}{0.15\linewidth}\centering\small Image \ImgD\end{minipage}\hfil
		\begin{minipage}{0.15\linewidth}\centering\small Image \ImgE\end{minipage}\hfil
		\begin{minipage}{0.15\linewidth}\centering\small Image \ImgF\end{minipage}\\
	\end{minipage}
	\vskip 2mm
	\begin{minipage}{\linewidth}
		\flushright
		\begin{minipage}{0mm}{\vskip -20mm}\hspace{-1em}\rotatebox[origin=l]{90}{\emph{Observed}}\end{minipage}
		\includegraphics[width=0.15\linewidth]{figures/results/ours/cowPNG/img/0\ImgA}\hfil
		\includegraphics[width=0.15\linewidth]{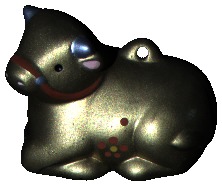}\hfil
		\includegraphics[width=0.15\linewidth]{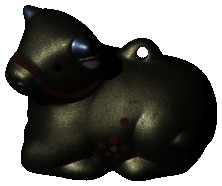}\hfil
		\includegraphics[width=0.15\linewidth]{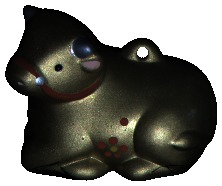}\hfil
		\includegraphics[width=0.15\linewidth]{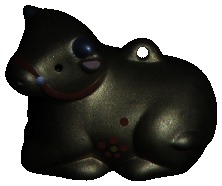}\hfil
		\includegraphics[width=0.15\linewidth]{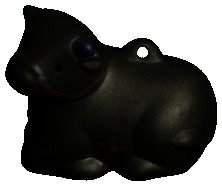}\\
	\end{minipage}\\
\vskip 4mm
\dotfill
\vskip 4mm
	\begin{minipage}{\linewidth}
		\flushright
		\begin{minipage}{0mm}{\vskip -20mm}\hspace{-1em}\rotatebox[origin=l]{90}{\emph{Synthesized}}\end{minipage}
		\includegraphics[width=0.15\linewidth]{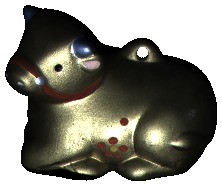}\hfil
		\includegraphics[width=0.15\linewidth]{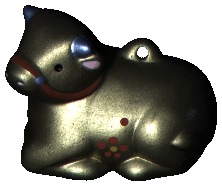}\hfil
		\includegraphics[width=0.15\linewidth]{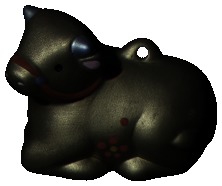}\hfil
		\includegraphics[width=0.15\linewidth]{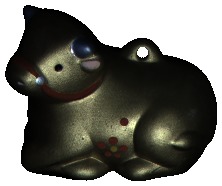}\hfil
		\includegraphics[width=0.15\linewidth]{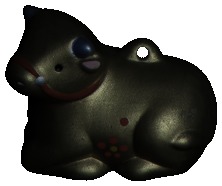}\hfil
		\includegraphics[width=0.15\linewidth]{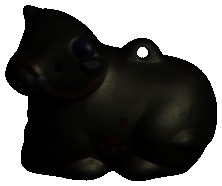}\\
	\end{minipage}
	\vskip 5mm
	\begin{minipage}{\linewidth}
		\flushright
		\begin{minipage}{0mm}{\vskip -20mm}\hspace{-1em}\rotatebox[origin=l]{90}{\emph{Reflectance}}\end{minipage}
		\includegraphics[width=0.15\linewidth]{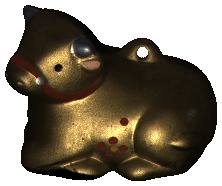}\hfil
		\includegraphics[width=0.15\linewidth]{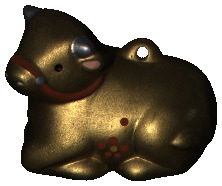}\hfil
		\includegraphics[width=0.15\linewidth]{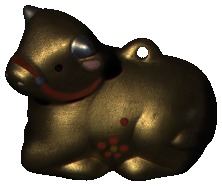}\hfil
		\includegraphics[width=0.15\linewidth]{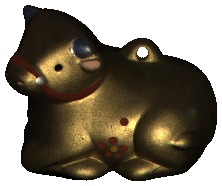}\hfil
		\includegraphics[width=0.15\linewidth]{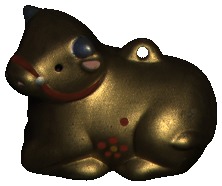}\hfil
		\includegraphics[width=0.15\linewidth]{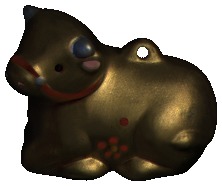}\\
	\end{minipage}
	\vskip 5mm
	\begin{minipage}{\linewidth}
		\flushright
		\begin{minipage}{0mm}{\vskip -20mm}\hspace{-1em}\rotatebox[origin=l]{90}{\emph{Errors}}\end{minipage}
		\colorbox{bgcolor}{\includegraphics[width=0.15\linewidth]{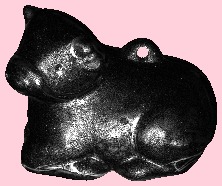}}\hfil
		\colorbox{bgcolor}{\includegraphics[width=0.15\linewidth]{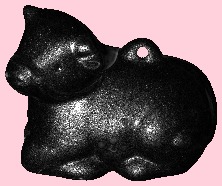}}\hfil
		\colorbox{bgcolor}{\includegraphics[width=0.15\linewidth]{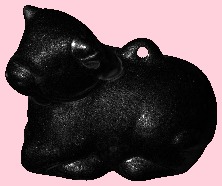}}\hfil
		\colorbox{bgcolor}{\includegraphics[width=0.15\linewidth]{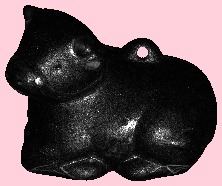}}\hfil
		\colorbox{bgcolor}{\includegraphics[width=0.15\linewidth]{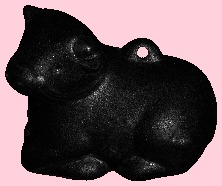}}\hfil
		\colorbox{bgcolor}{\includegraphics[width=0.15\linewidth]{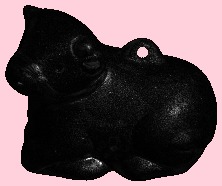}}\\
	\end{minipage}
	\caption{\textbf{Our image reconstruction results for \textsc{cow} scene.} See also explanations in Fig.~\ref{afig:reconst_ball}.}
\end{figure}
\clearpage
\begin{figure}[p]
	\centering
	\small
	\begin{minipage}{\linewidth}
		\flushright
		\begin{minipage}{0.15\linewidth}\centering\small Image \ImgA\end{minipage}\hfil
		\begin{minipage}{0.15\linewidth}\centering\small Image \ImgB\end{minipage}\hfil
		\begin{minipage}{0.15\linewidth}\centering\small Image \ImgC\end{minipage}\hfil
		\begin{minipage}{0.15\linewidth}\centering\small Image \ImgD\end{minipage}\hfil
		\begin{minipage}{0.15\linewidth}\centering\small Image \ImgE\end{minipage}\hfil
		\begin{minipage}{0.15\linewidth}\centering\small Image \ImgF\end{minipage}\\
	\end{minipage}
	\vskip 2mm
	\begin{minipage}{\linewidth}
		\flushright
		\begin{minipage}{0mm}{\vskip -13mm}\hspace{-1em}\rotatebox[origin=l]{90}{\emph{Observed}}\end{minipage}
		\includegraphics[width=0.15\linewidth]{figures/results/ours/harvestPNG/img/0\ImgA}\hfil
		\includegraphics[width=0.15\linewidth]{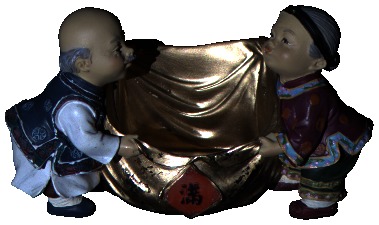}\hfil
		\includegraphics[width=0.15\linewidth]{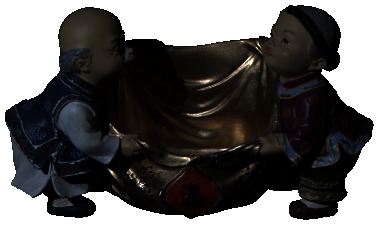}\hfil
		\includegraphics[width=0.15\linewidth]{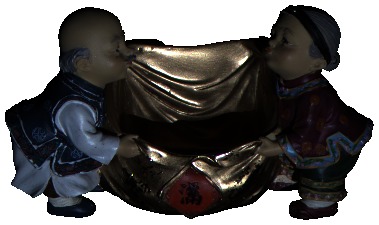}\hfil
		\includegraphics[width=0.15\linewidth]{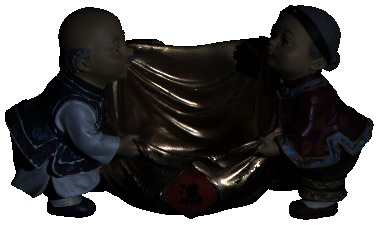}\hfil
		\includegraphics[width=0.15\linewidth]{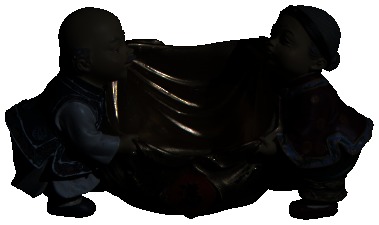}\\
	\end{minipage}
\vskip 4mm
\dotfill
\vskip 4mm
	\begin{minipage}{\linewidth}
		\flushright
		\begin{minipage}{0mm}{\vskip -13mm}\hspace{-1em}\rotatebox[origin=l]{90}{\emph{Synthesized}}\end{minipage}
		\includegraphics[width=0.15\linewidth]{figures/results/ours/harvestPNG/syn/0\ImgA}\hfil
		\includegraphics[width=0.15\linewidth]{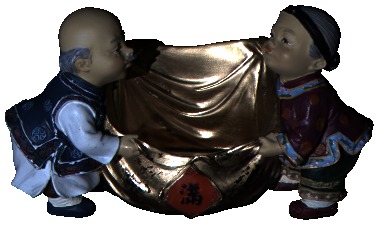}\hfil
		\includegraphics[width=0.15\linewidth]{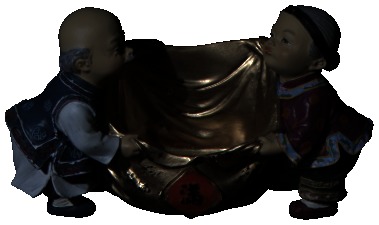}\hfil
		\includegraphics[width=0.15\linewidth]{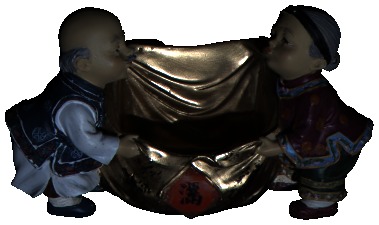}\hfil
		\includegraphics[width=0.15\linewidth]{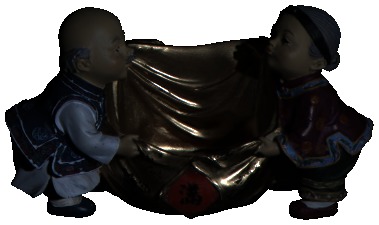}\hfil
		\includegraphics[width=0.15\linewidth]{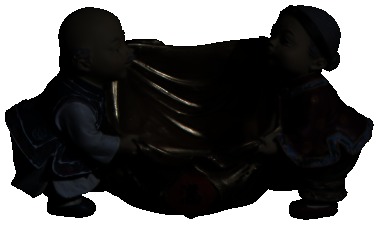}\\
	\end{minipage}
\vskip 5mm
	\begin{minipage}{\linewidth}
		\flushright
		\begin{minipage}{0mm}{\vskip -13mm}\hspace{-1em}\rotatebox[origin=l]{90}{\emph{Reflectance}}\end{minipage}
		\includegraphics[width=0.15\linewidth]{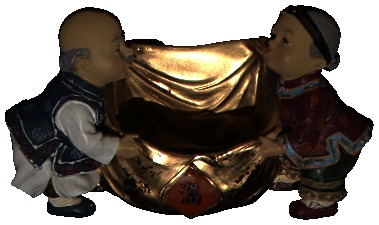}\hfil
		\includegraphics[width=0.15\linewidth]{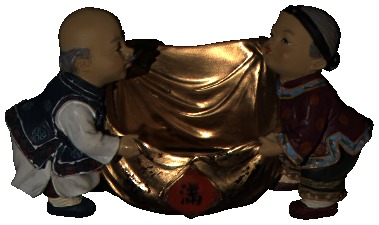}\hfil
		\includegraphics[width=0.15\linewidth]{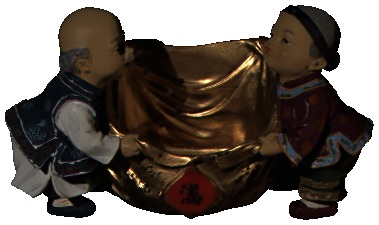}\hfil
		\includegraphics[width=0.15\linewidth]{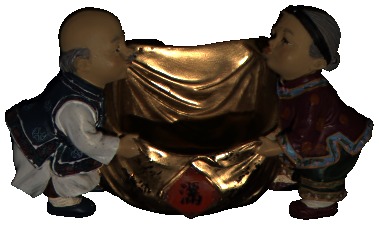}\hfil
		\includegraphics[width=0.15\linewidth]{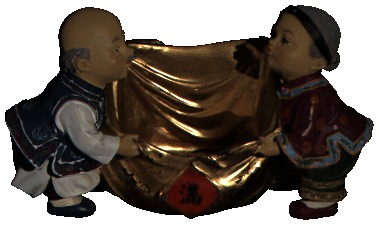}\hfil
		\includegraphics[width=0.15\linewidth]{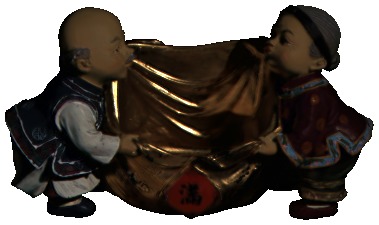}\\
	\end{minipage}
	\vskip 5mm
	\begin{minipage}{\linewidth}
		\flushright
		\begin{minipage}{0mm}{\vskip -13mm}\hspace{-1em}\rotatebox[origin=l]{90}{\emph{Errors}}\end{minipage}
		\colorbox{bgcolor}{\includegraphics[width=0.15\linewidth]{figures/results/ours/harvestPNG/err/0\ImgA}}\hfil
		\colorbox{bgcolor}{\includegraphics[width=0.15\linewidth]{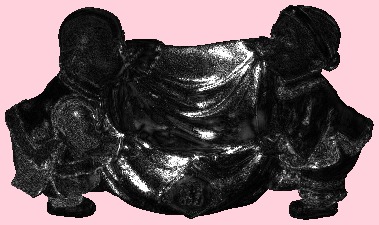}}\hfil
		\colorbox{bgcolor}{\includegraphics[width=0.15\linewidth]{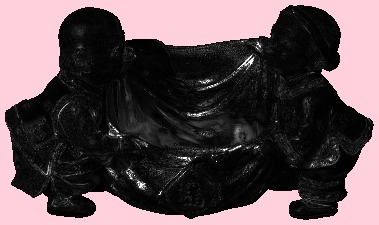}}\hfil
		\colorbox{bgcolor}{\includegraphics[width=0.15\linewidth]{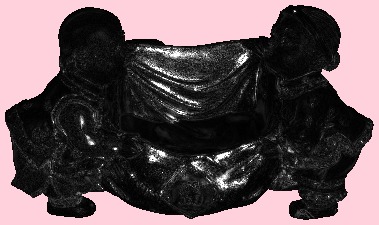}}\hfil
		\colorbox{bgcolor}{\includegraphics[width=0.15\linewidth]{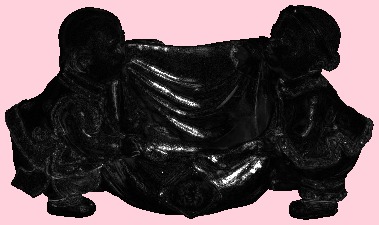}}\hfil
		\colorbox{bgcolor}{\includegraphics[width=0.15\linewidth]{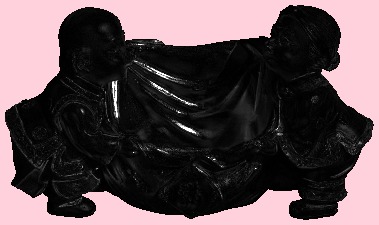}}\\
	\end{minipage}
	\caption{\textbf{Our image reconstruction results for \textsc{harvest} scene.} See also explanations in Fig.~\ref{afig:reconst_ball}.}
	\label{afig:reconst_harvest}
\end{figure}

\clearpage

\begin{figure*}[p]
	\centering
	\small
	\def\figw{0.30\linewidth}
	% 0:ball, 1:bear, 2:buddha, 3:cat, 4:cow, 5:goblet, 6:harbest, 7:pot1, 8:pot2, 9:reading
	\includegraphics[width=\figw]{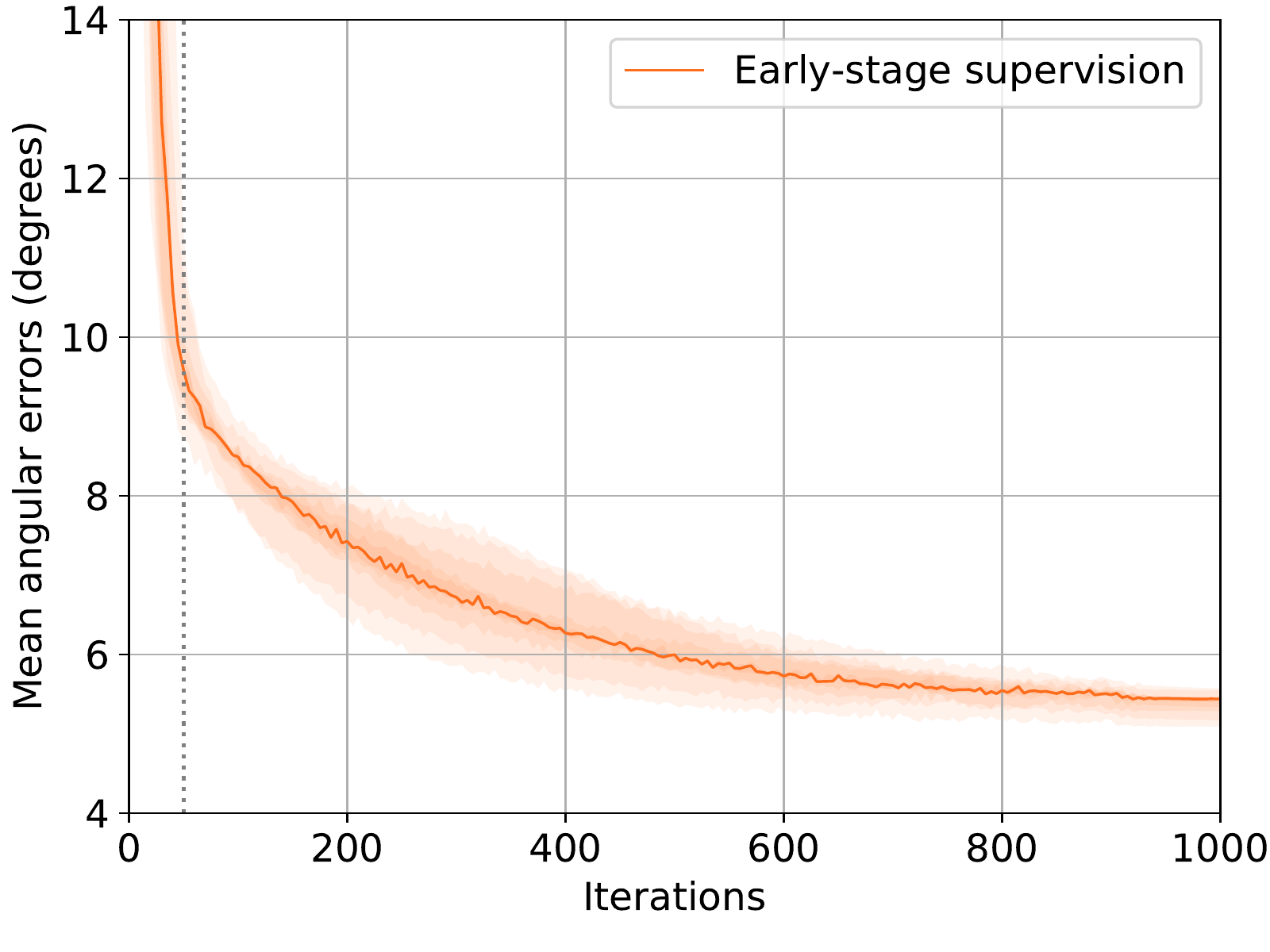}\hfil
	\includegraphics[width=\figw]{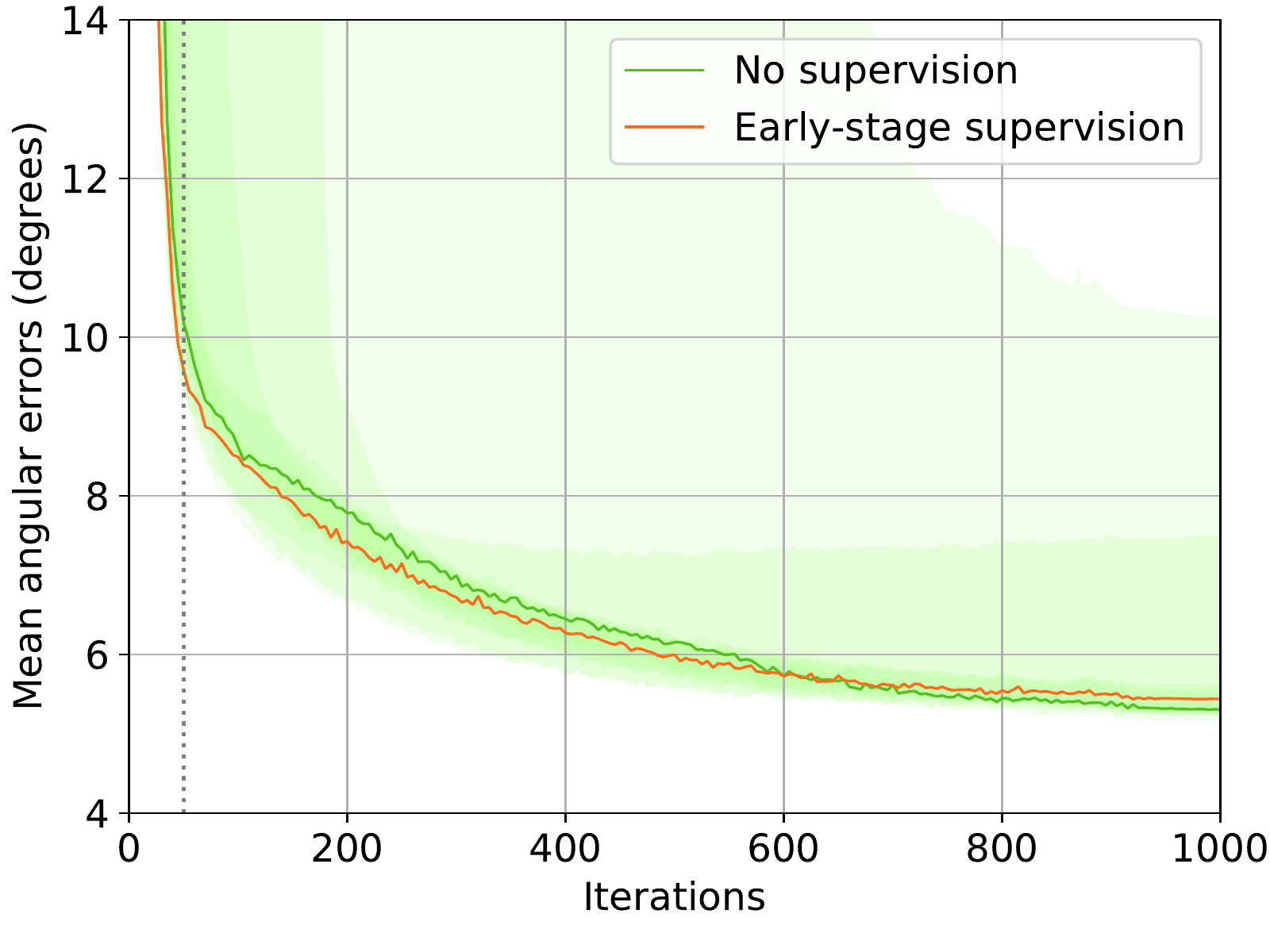}\hfil
	\includegraphics[width=\figw]{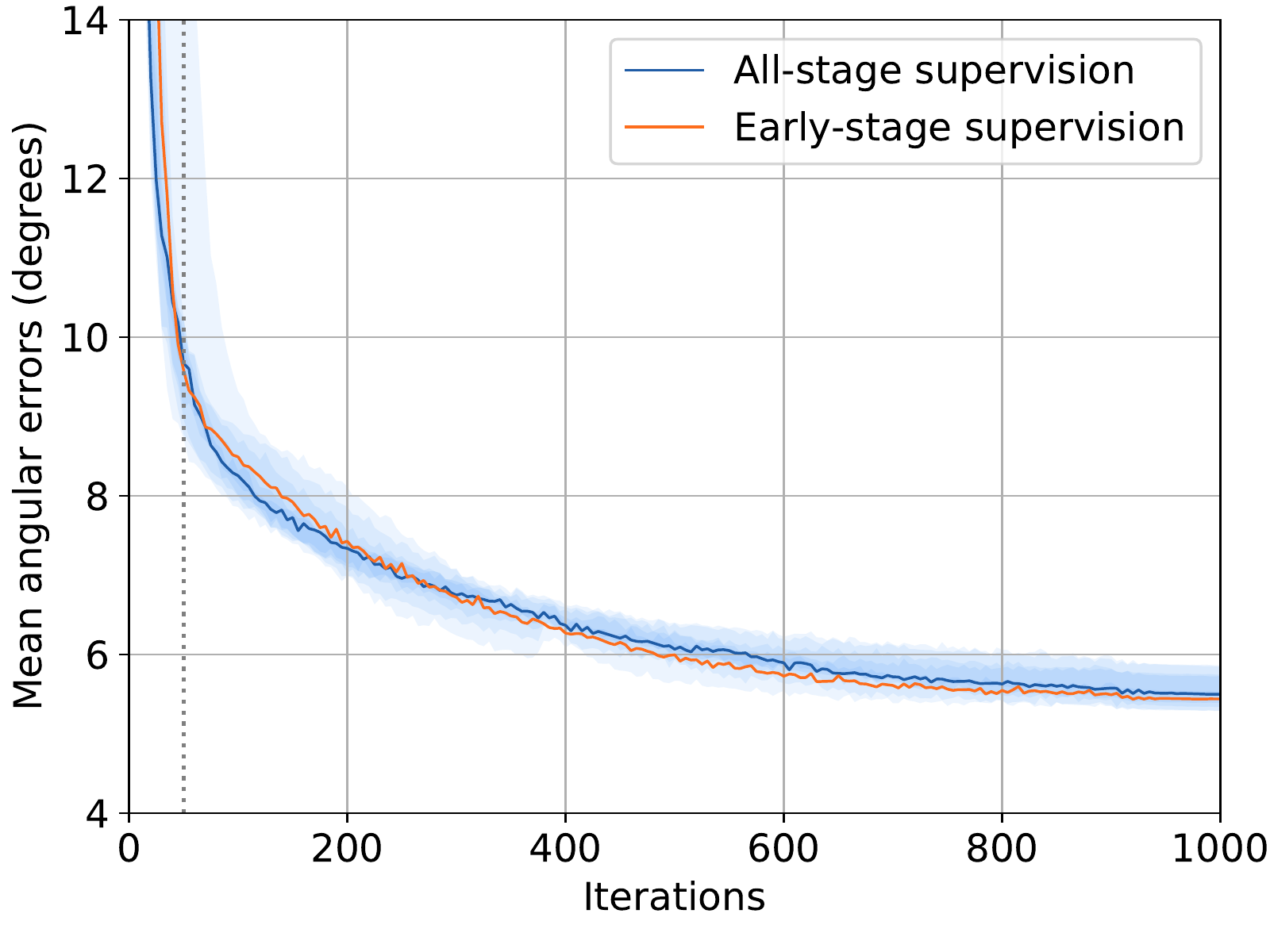}\\
	\includegraphics[width=\figw]{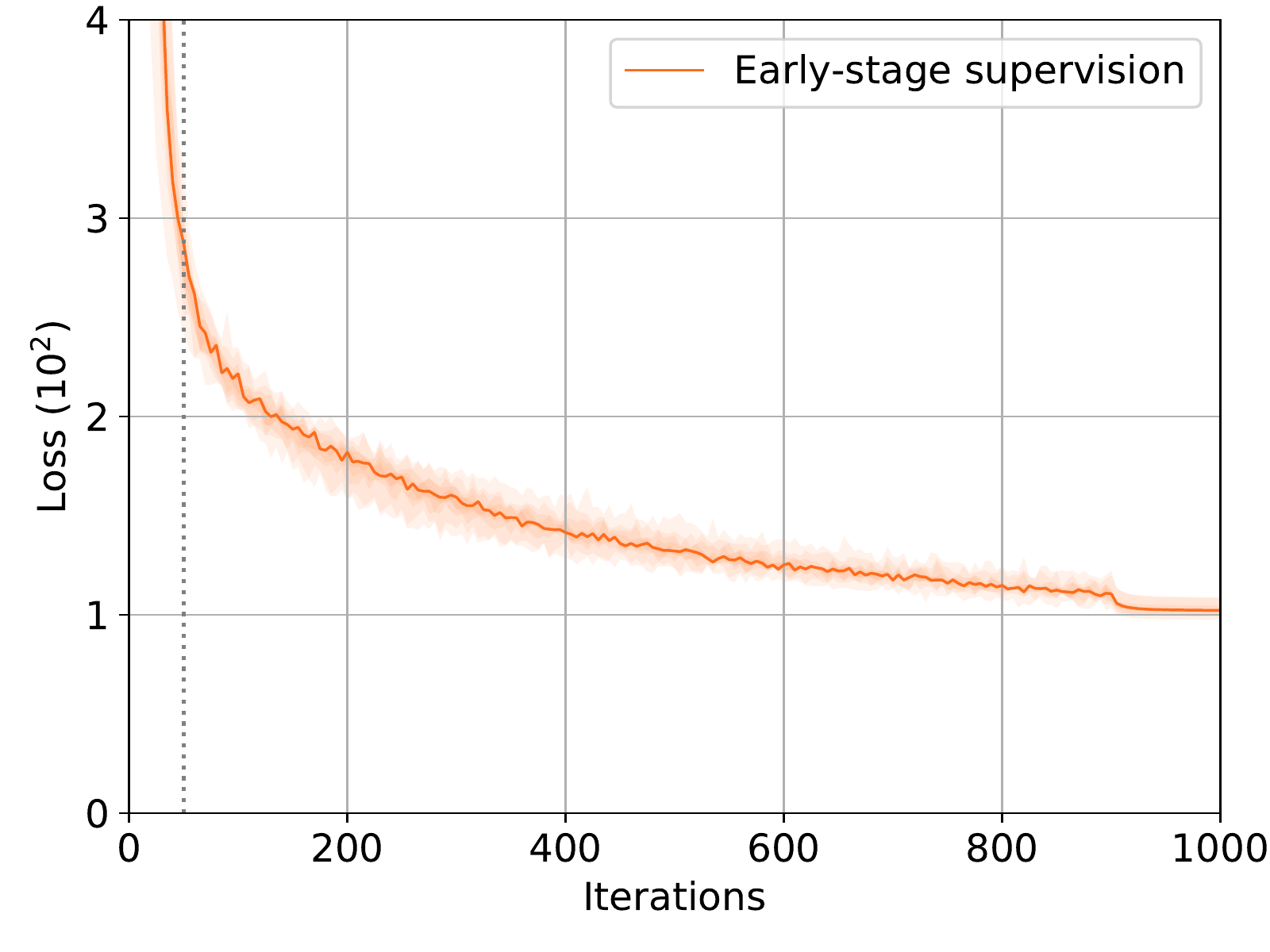}\hfil
	\includegraphics[width=\figw]{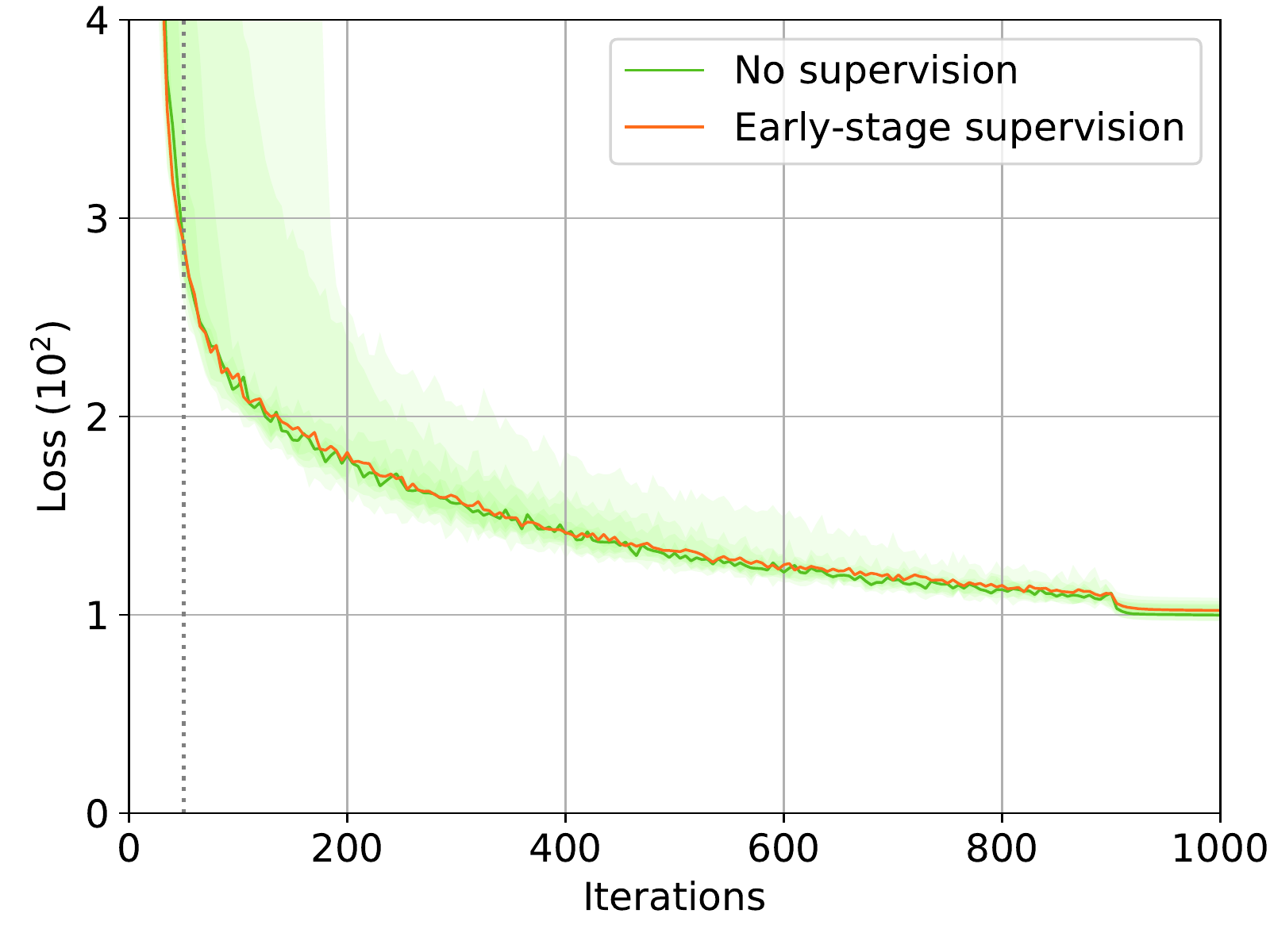}\hfil
	\includegraphics[width=\figw]{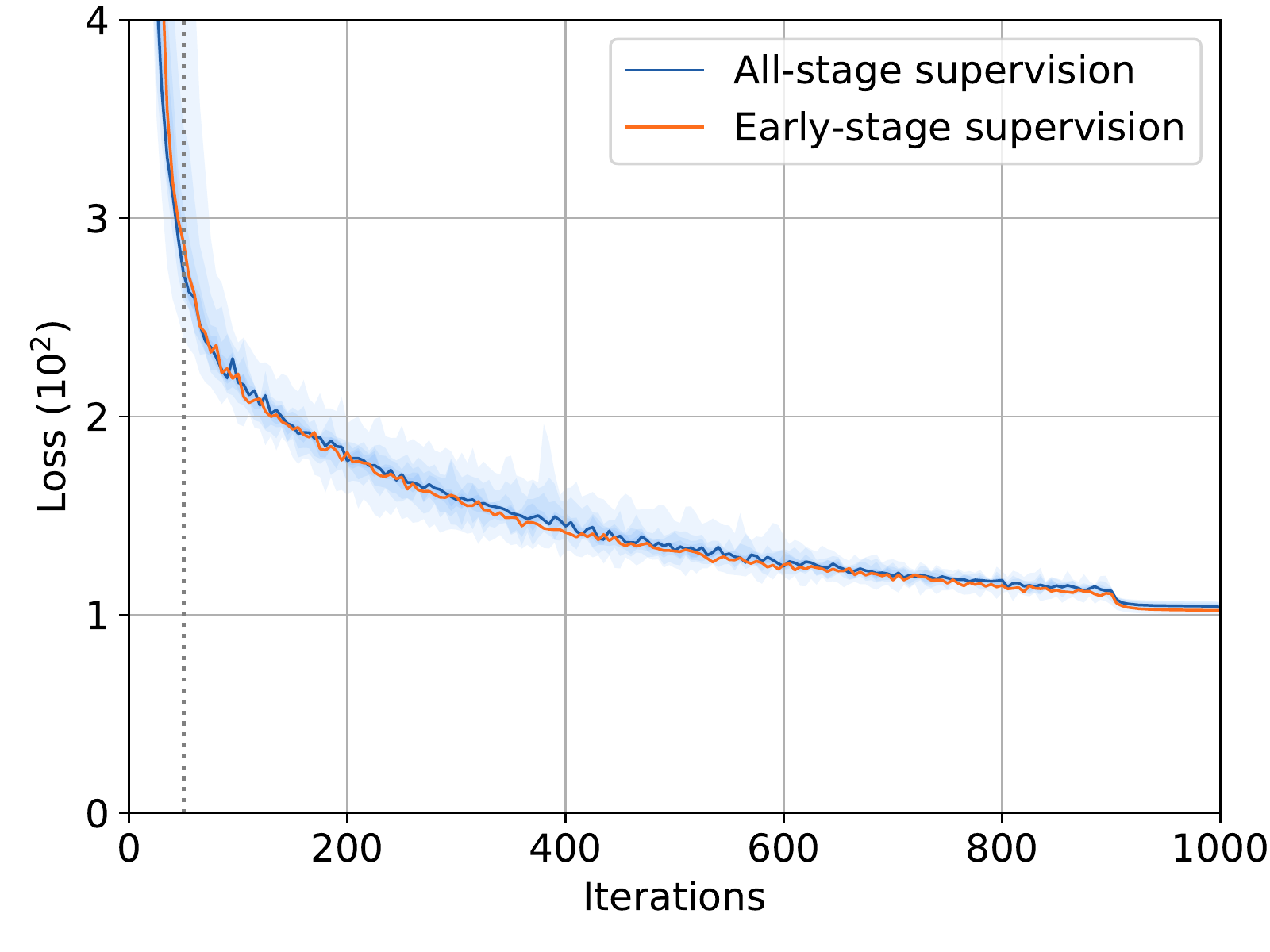}\\
	\begin{minipage}{\figw}\centering {Early-stage supervision}\end{minipage}\hfil
	\begin{minipage}{\figw}\centering {No supervision}\end{minipage}\hfil
	\begin{minipage}{\figw}\centering {All-stage supervision}\end{minipage}\\
	\caption{\textbf{Convergence analysis with different types of weak supervision for \textsc{cat} scene.}
		See also explanations in Fig.~\ref{afig:training_ball}.
	}
\end{figure*}

\begin{figure*}[p]
	\centering
	\small
	\def\figw{0.30\linewidth}
	% 0:ball, 1:bear, 2:buddha, 3:cat, 4:cow, 5:goblet, 6:harbest, 7:pot1, 8:pot2, 9:reading
	\includegraphics[width=\figw]{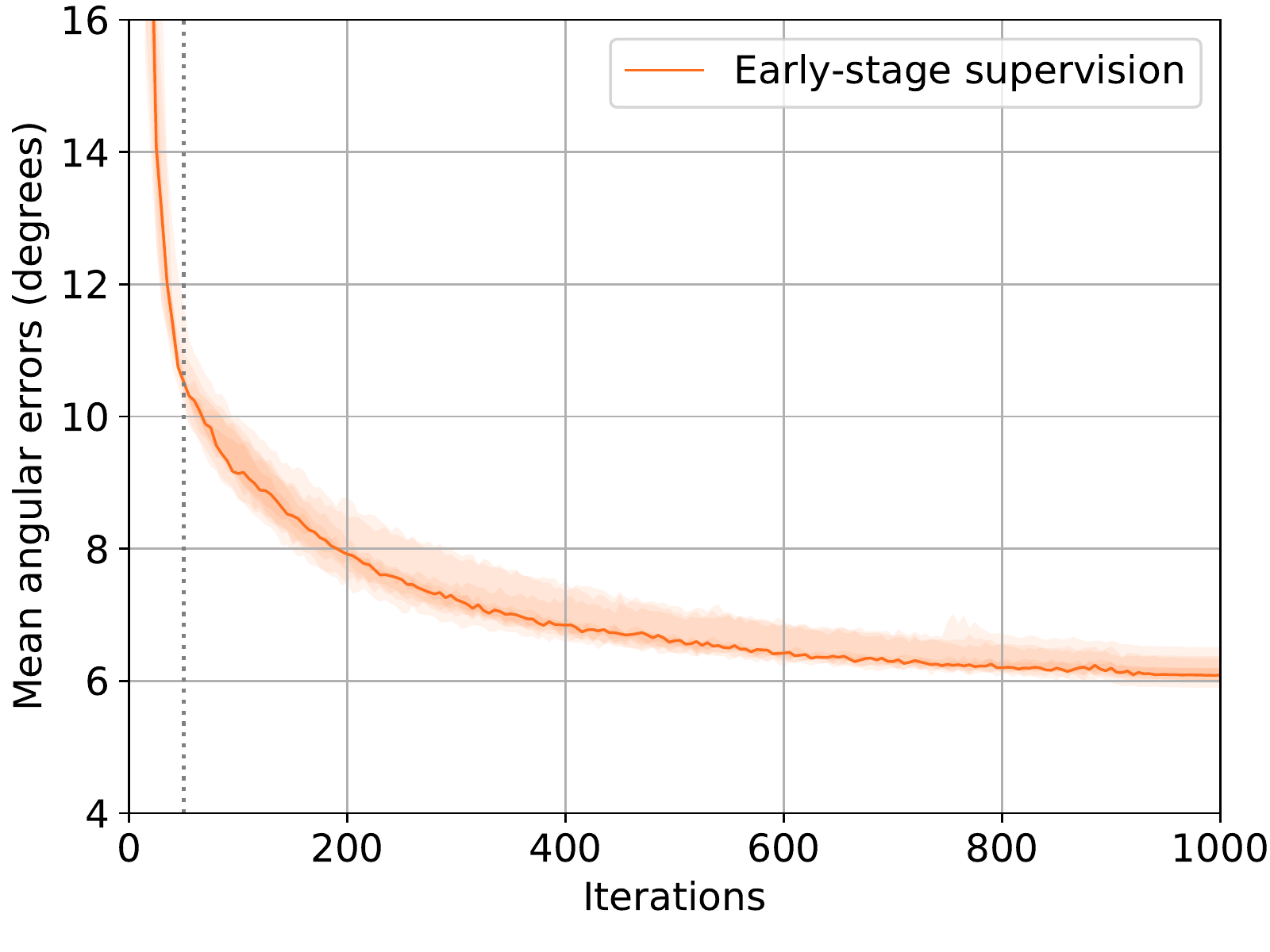}\hfil
	\includegraphics[width=\figw]{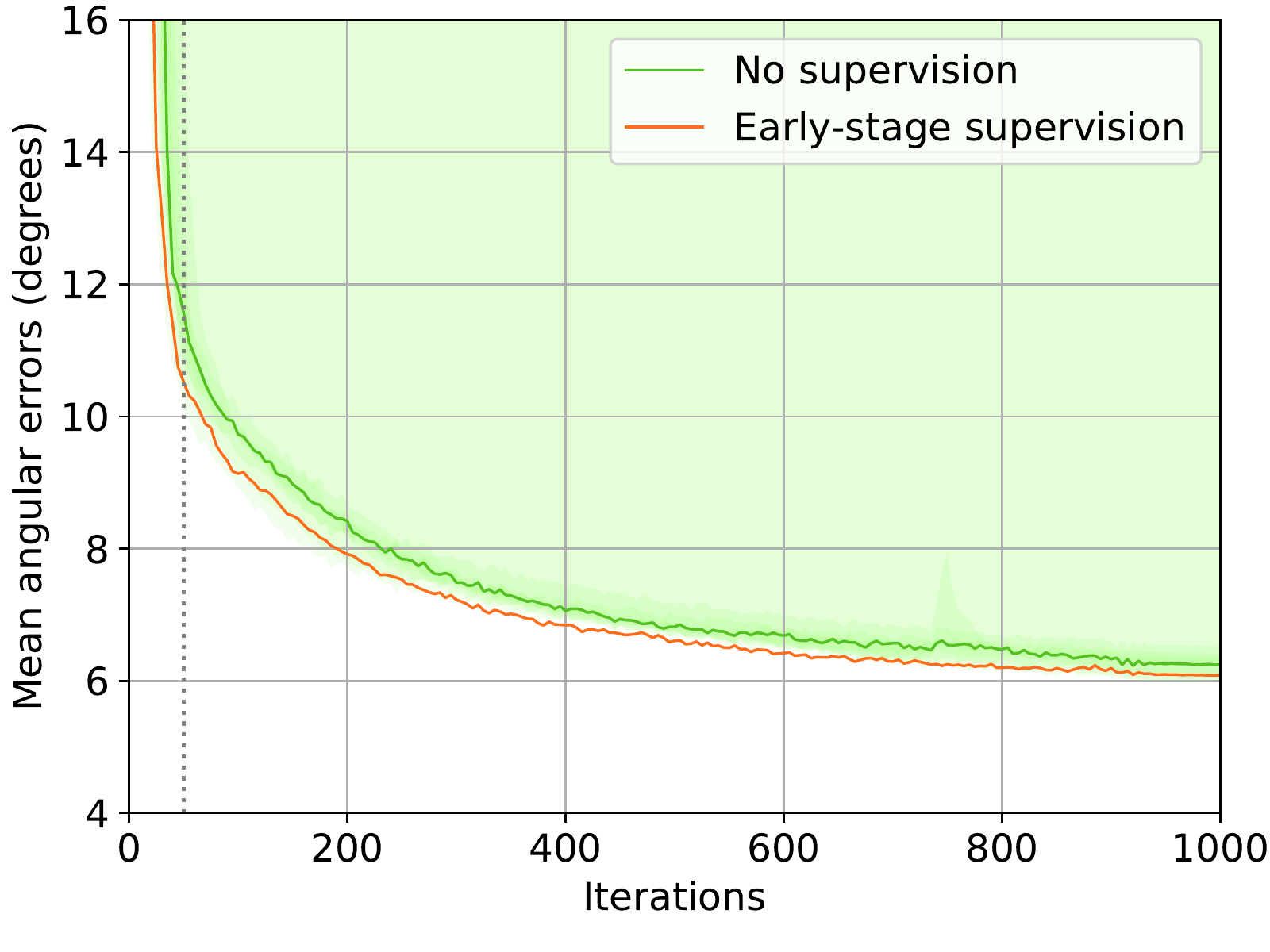}\hfil
	\includegraphics[width=\figw]{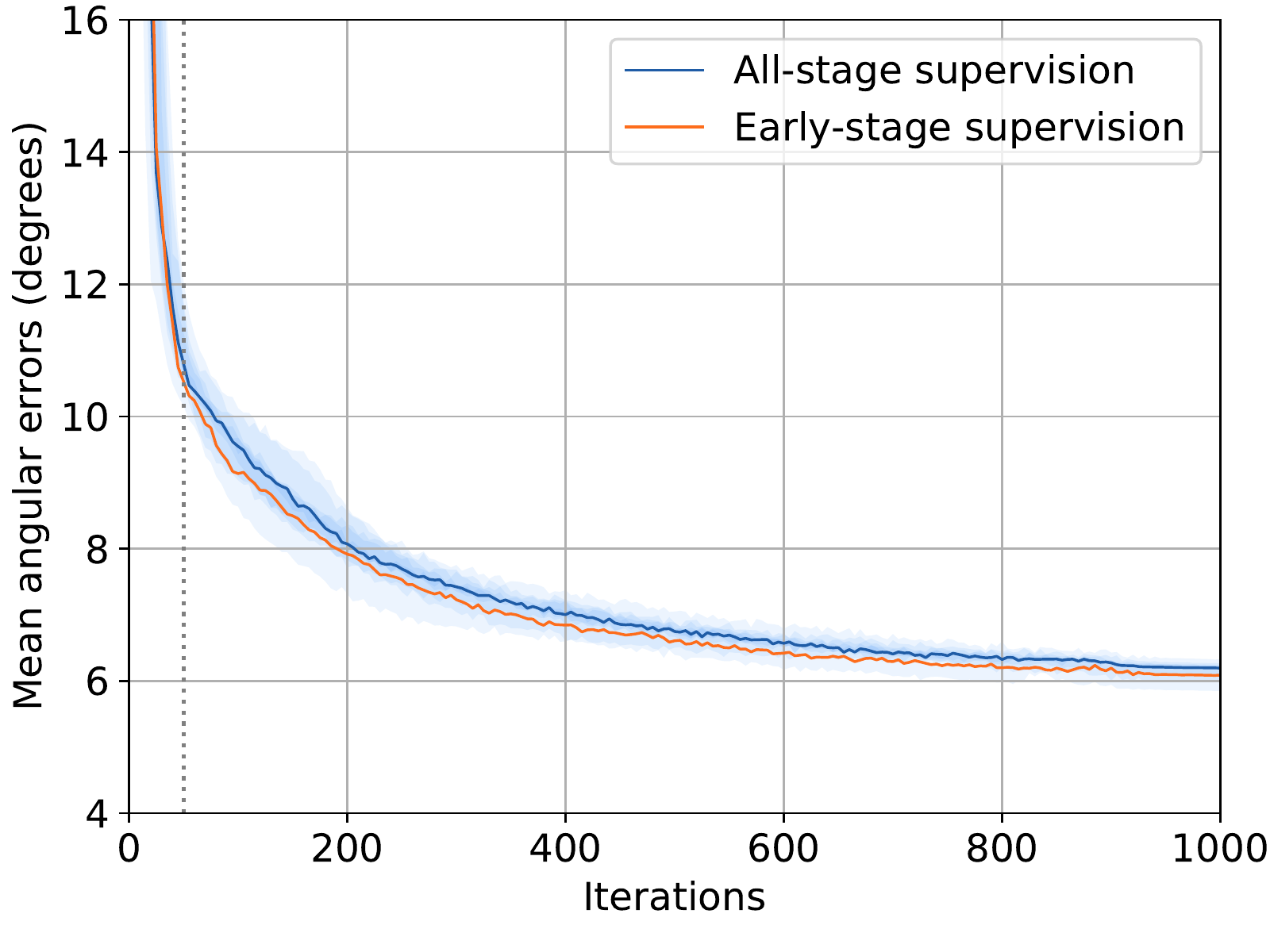}\\
	\includegraphics[width=\figw]{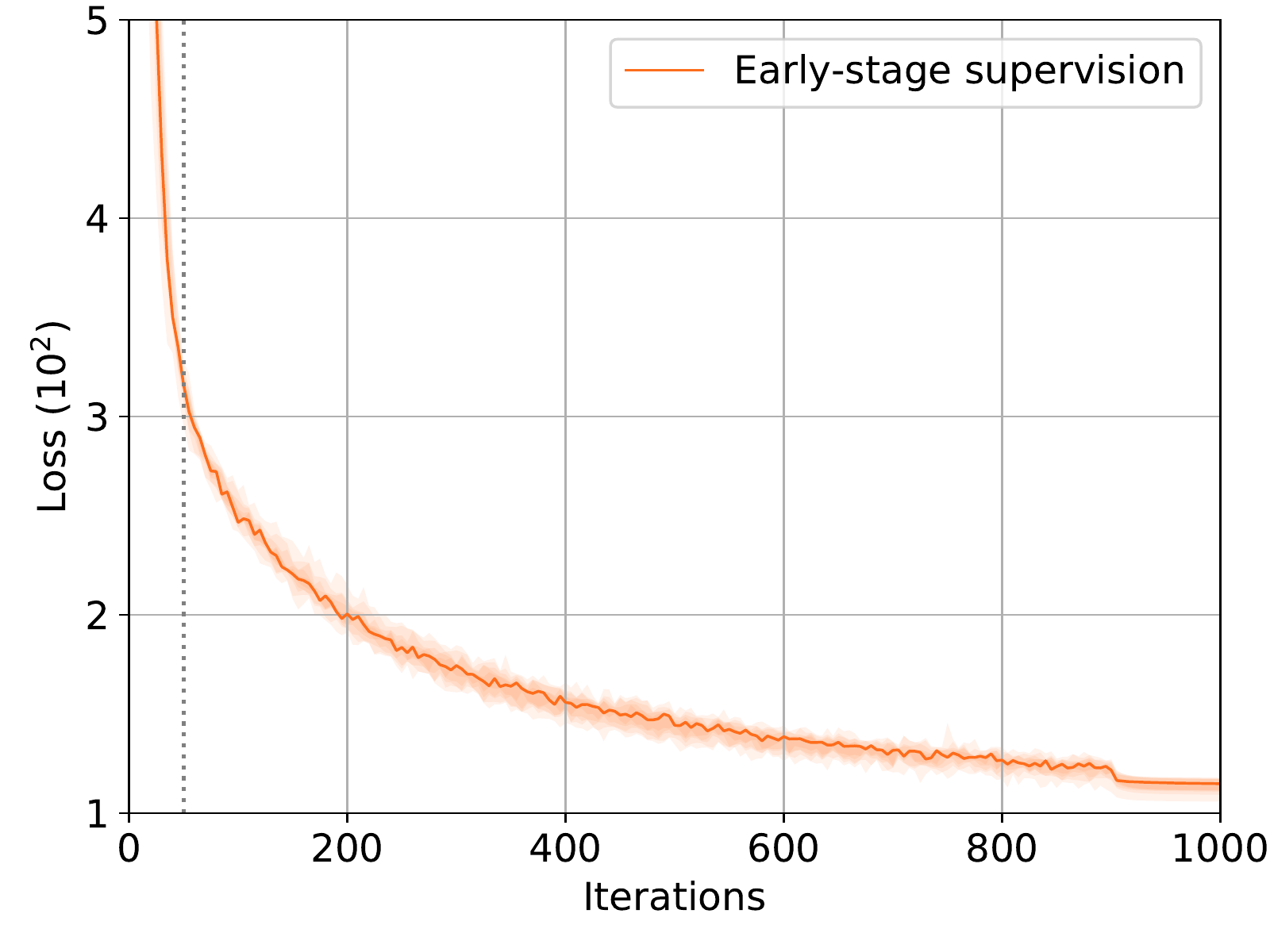}\hfil
	\includegraphics[width=\figw]{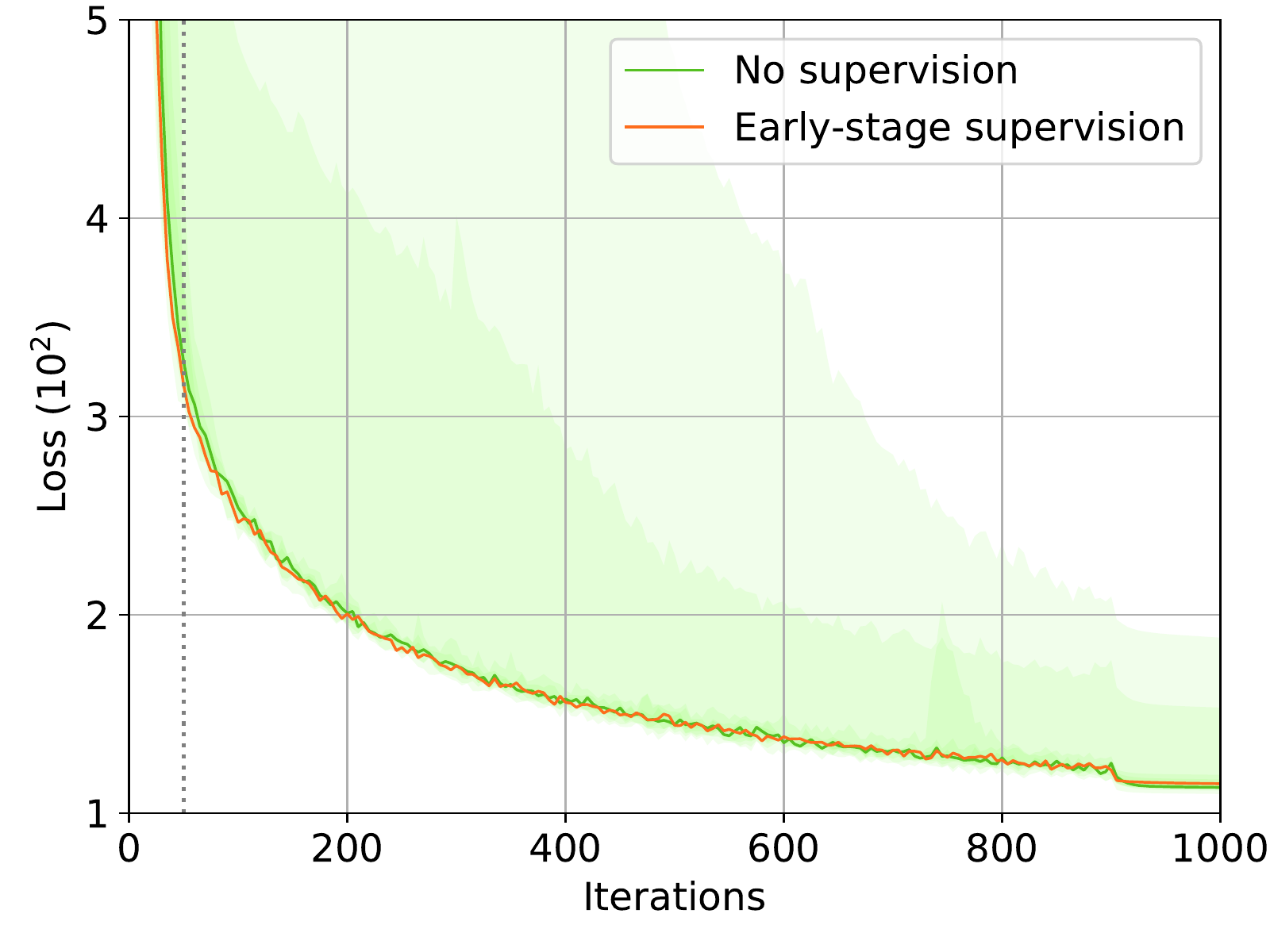}\hfil
	\includegraphics[width=\figw]{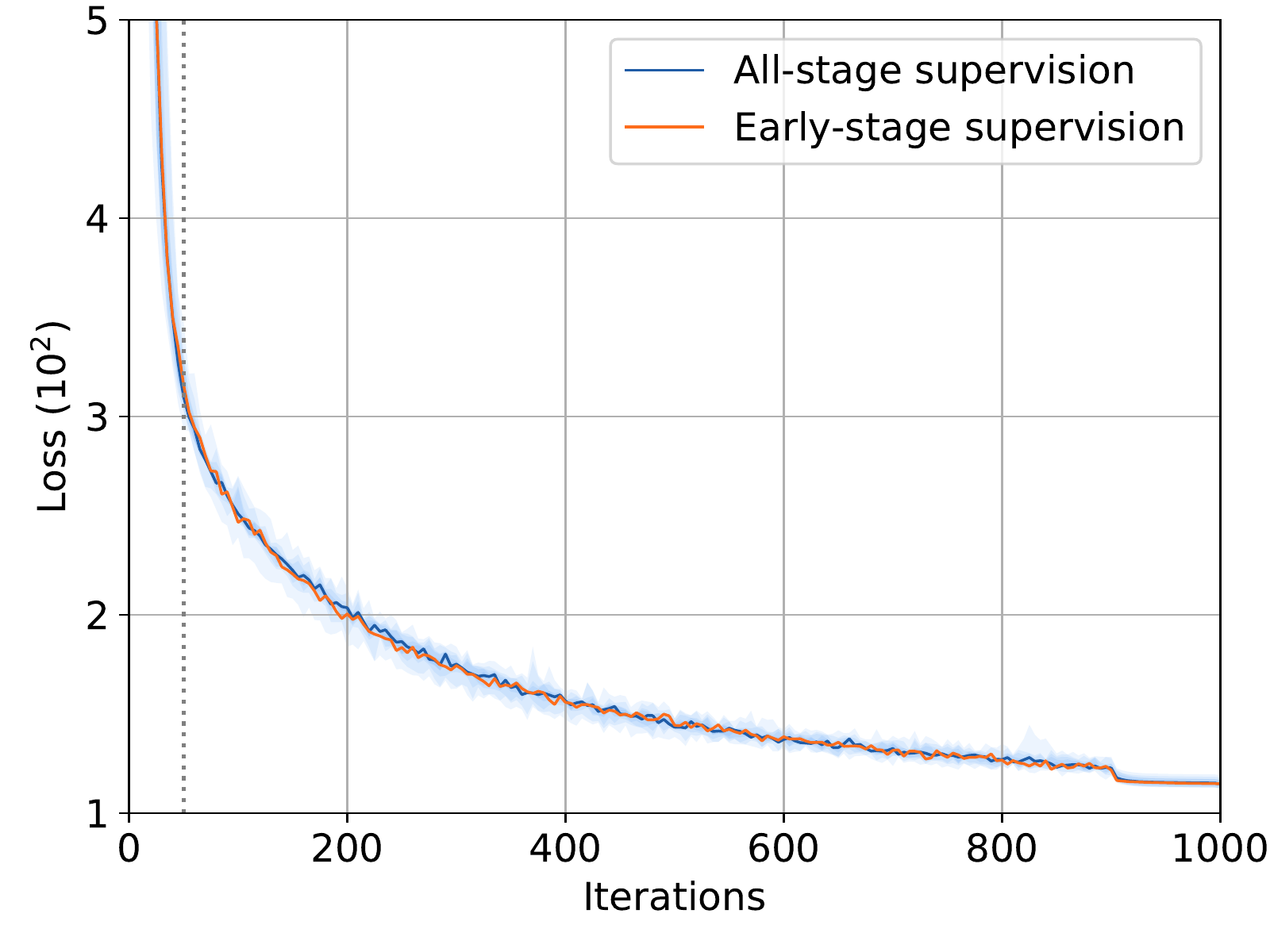}\\
	\begin{minipage}{\figw}\centering {Early-stage supervision}\end{minipage}\hfil
	\begin{minipage}{\figw}\centering {No supervision}\end{minipage}\hfil
	\begin{minipage}{\figw}\centering {All-stage supervision}\end{minipage}\\
	\caption{\textbf{Convergence analysis with different types of weak supervision for \textsc{pot1} scene.}
		See also explanations in Fig.~\ref{afig:training_ball}.
	}
\end{figure*}

\begin{figure*}[p]
	\centering
	\small
	\def\figw{0.30\linewidth}
	% 0:ball, 1:bear, 2:buddha, 3:cat, 4:cow, 5:goblet, 6:harbest, 7:pot1, 8:pot2, 9:reading
	\includegraphics[width=\figw]{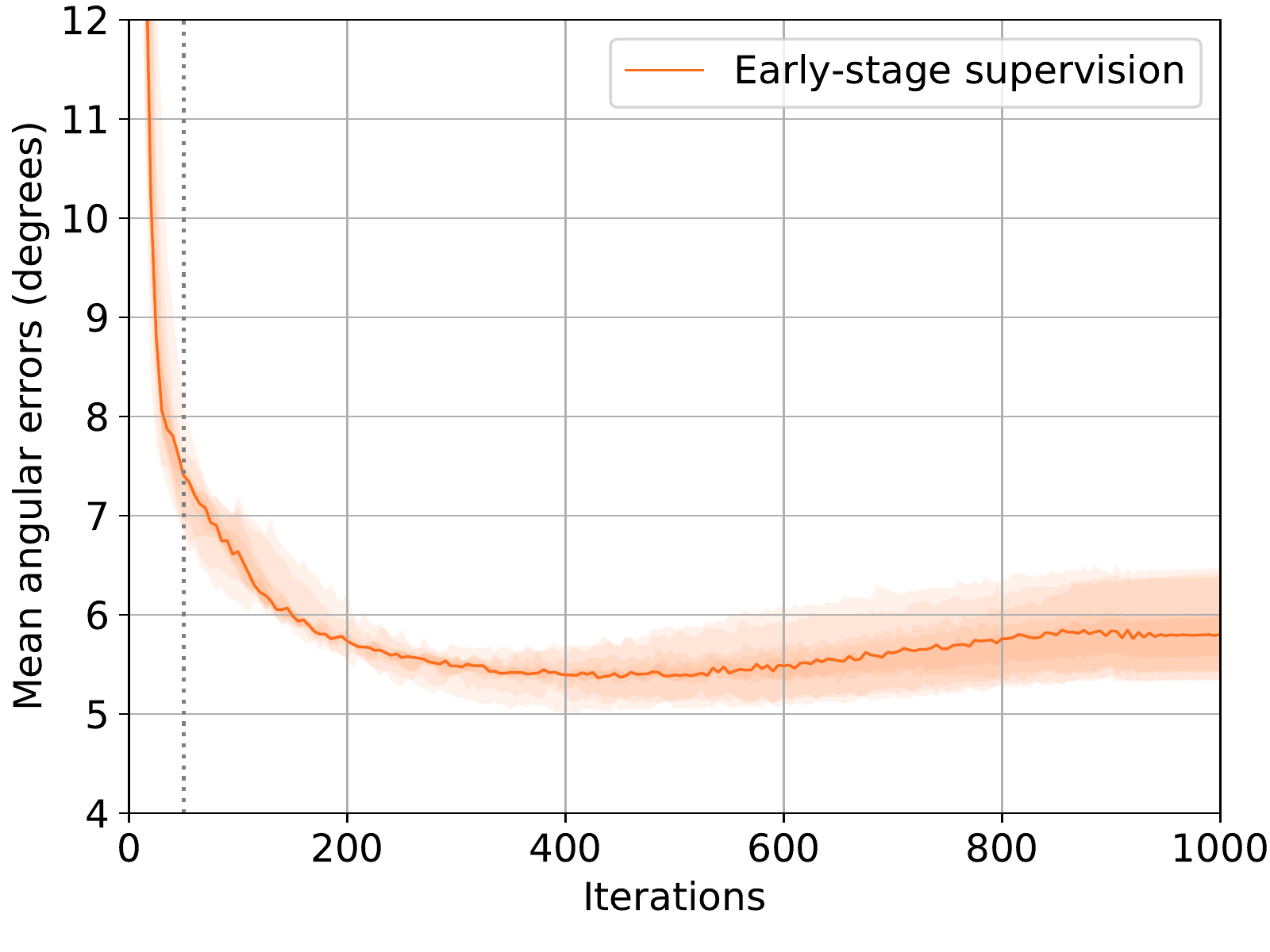}\hfil
	\includegraphics[width=\figw]{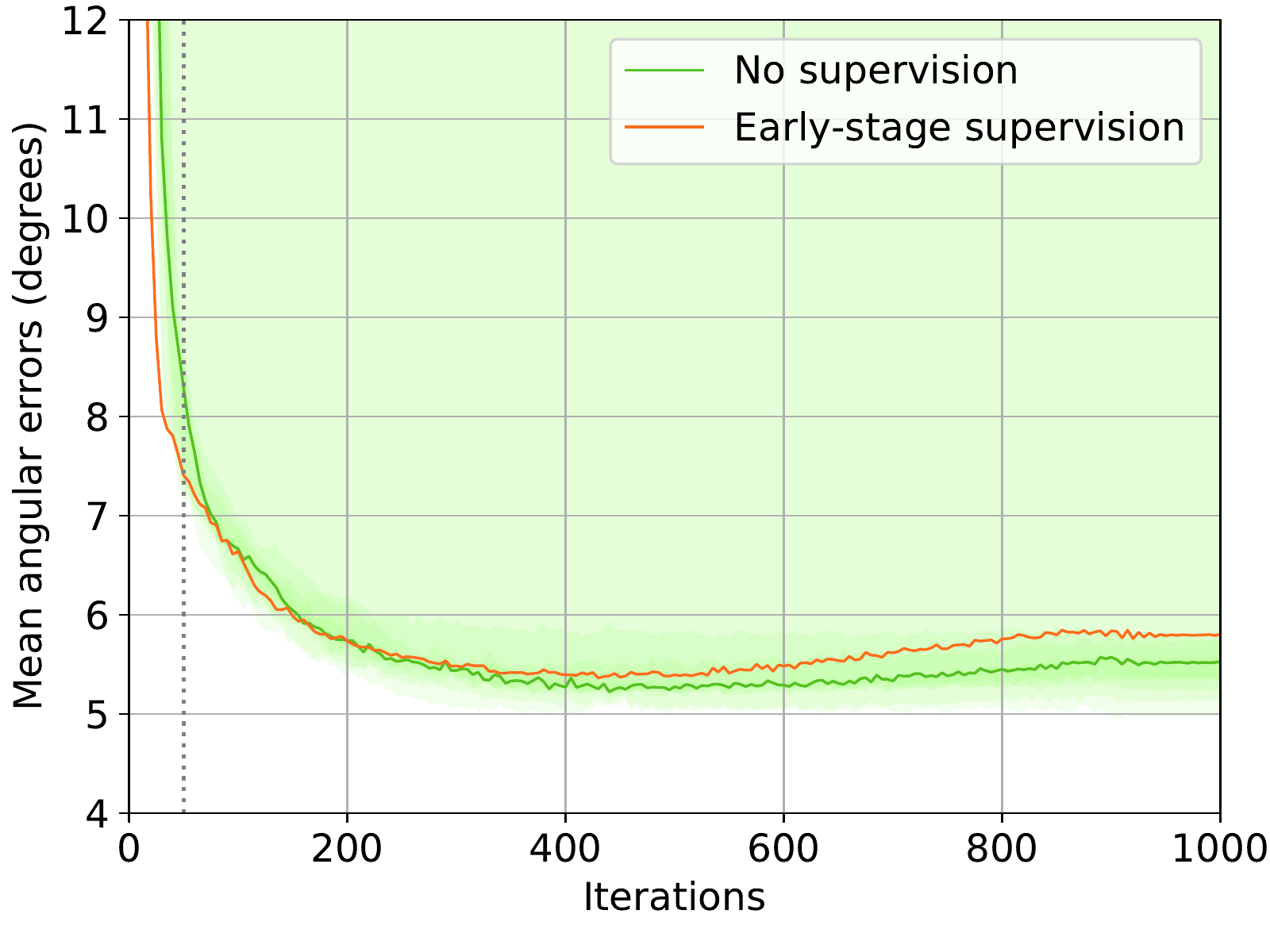}\hfil
	\includegraphics[width=\figw]{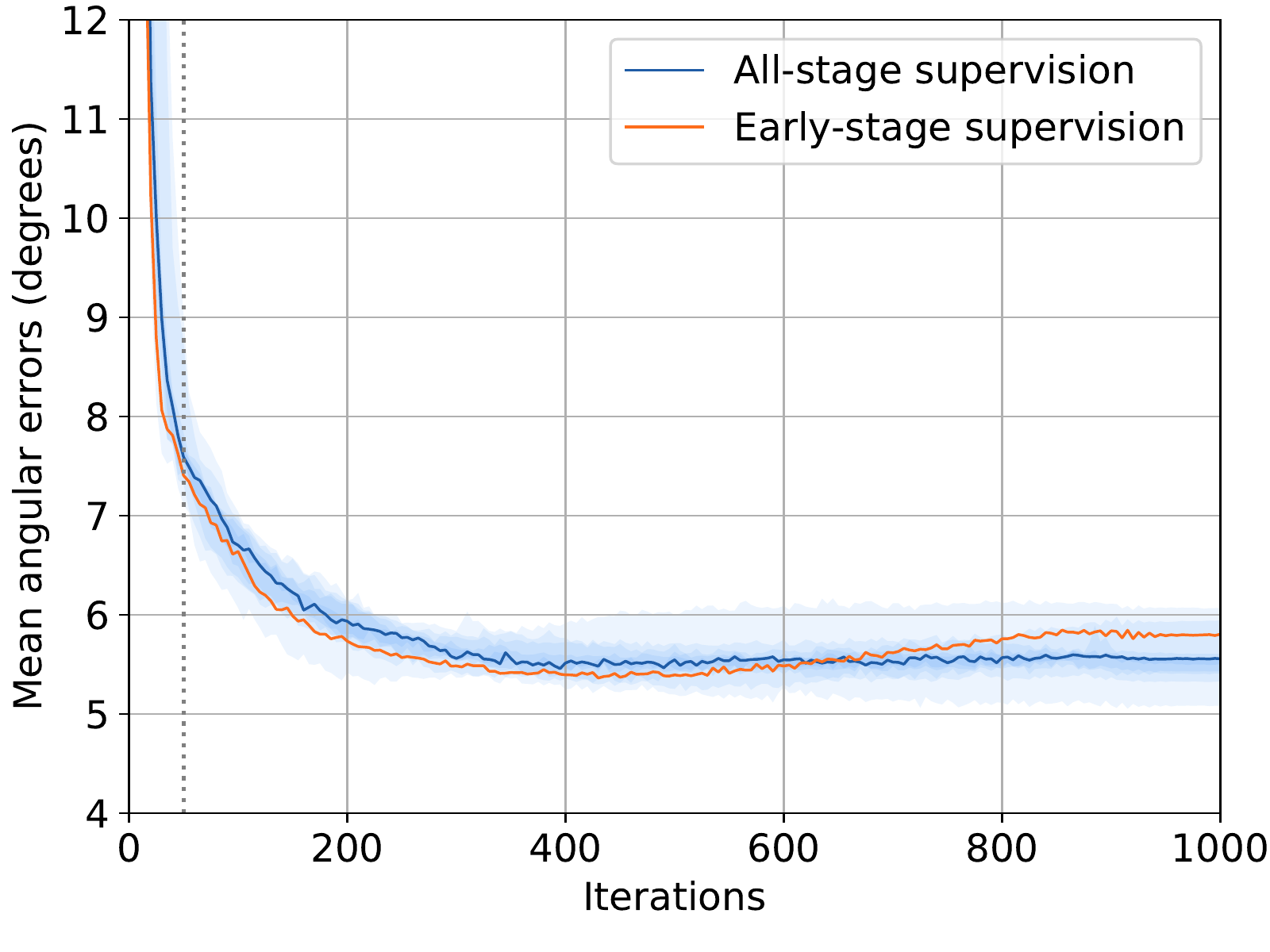}\\
	\includegraphics[width=\figw]{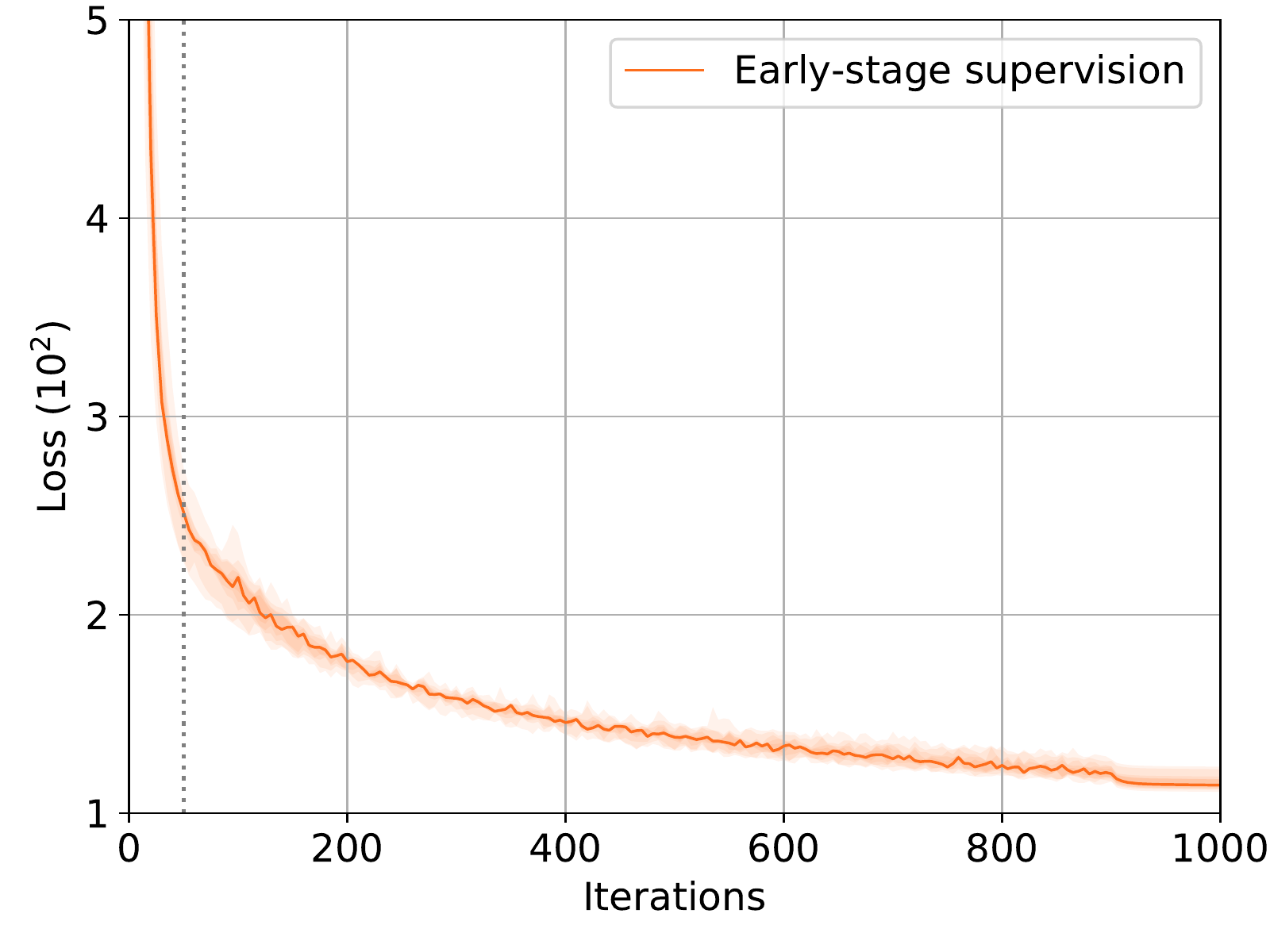}\hfil
	\includegraphics[width=\figw]{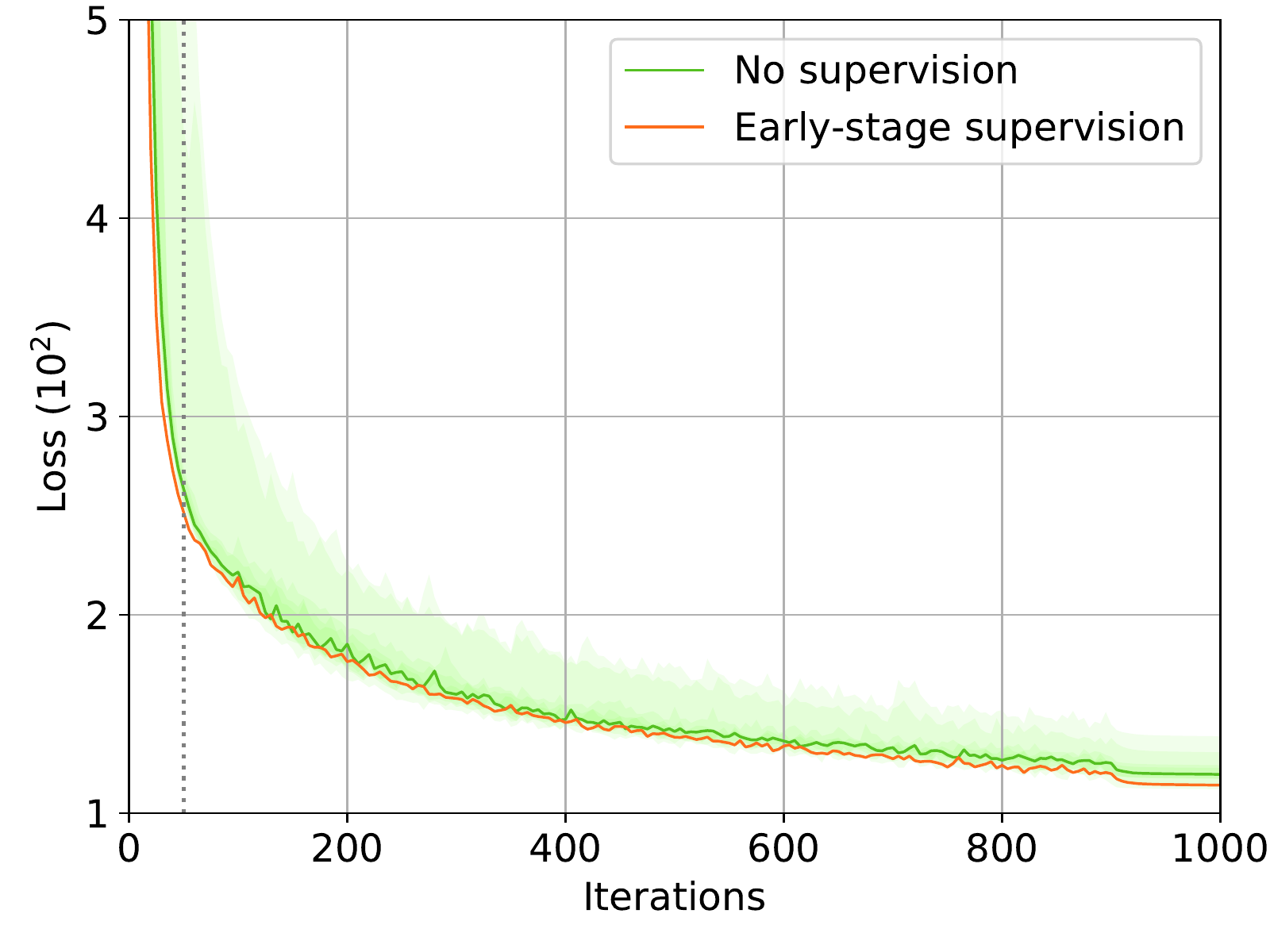}\hfil
	\includegraphics[width=\figw]{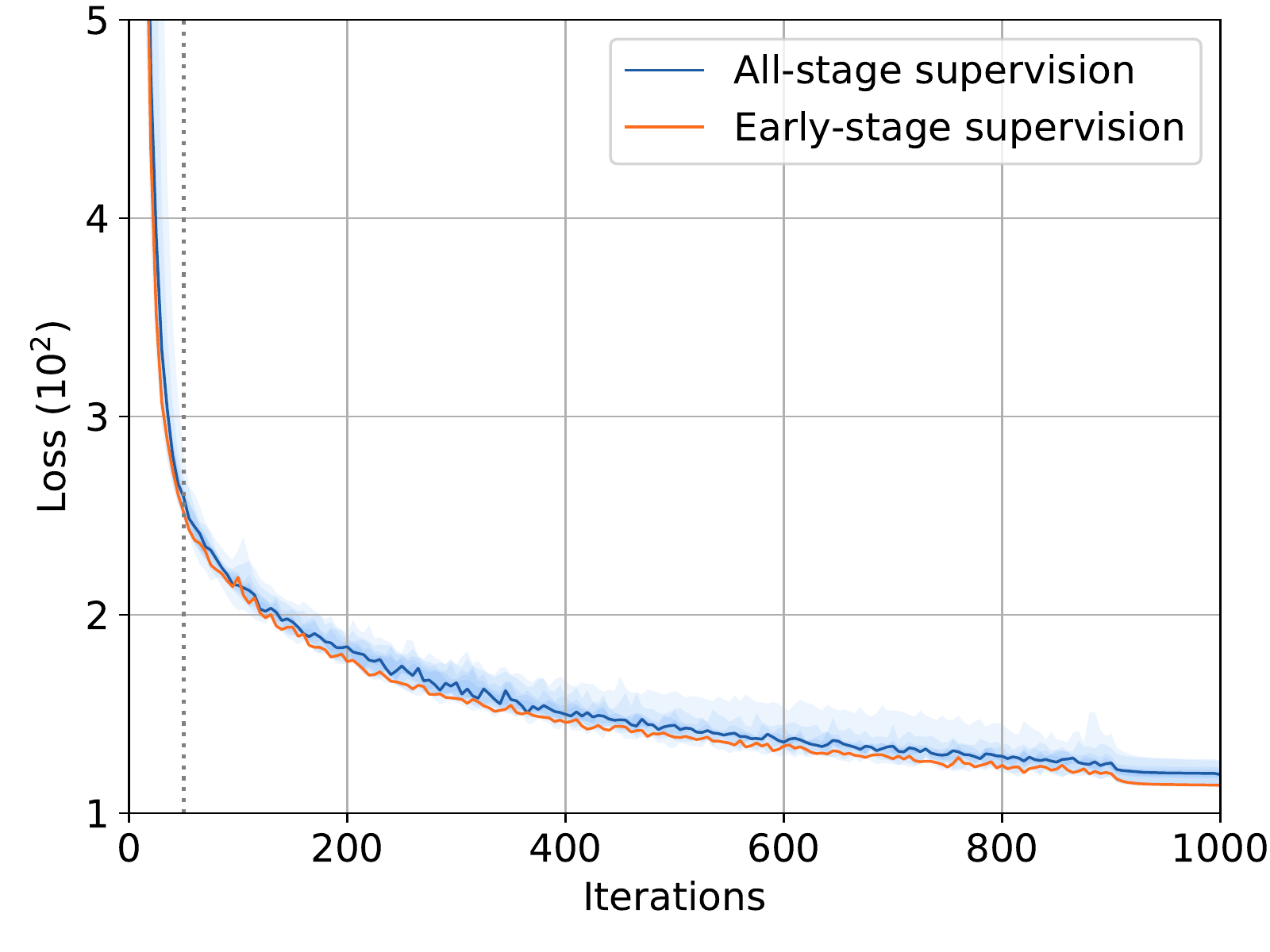}\\
	\begin{minipage}{\figw}\centering {Early-stage supervision}\end{minipage}\hfil
	\begin{minipage}{\figw}\centering {No supervision}\end{minipage}\hfil
	\begin{minipage}{\figw}\centering {All-stage supervision}\end{minipage}\\
	\caption{\textbf{Convergence analysis with different types of weak supervision for \textsc{bear} scene.}
		See also explanations in Fig.~\ref{afig:training_ball}.
	}
\end{figure*}

\begin{figure*}[p]
	\centering
	\small
	\def\figw{0.30\linewidth}
	% 0:ball, 1:bear, 2:buddha, 3:cat, 4:cow, 5:goblet, 6:harbest, 7:pot1, 8:pot2, 9:reading
	\includegraphics[width=\figw]{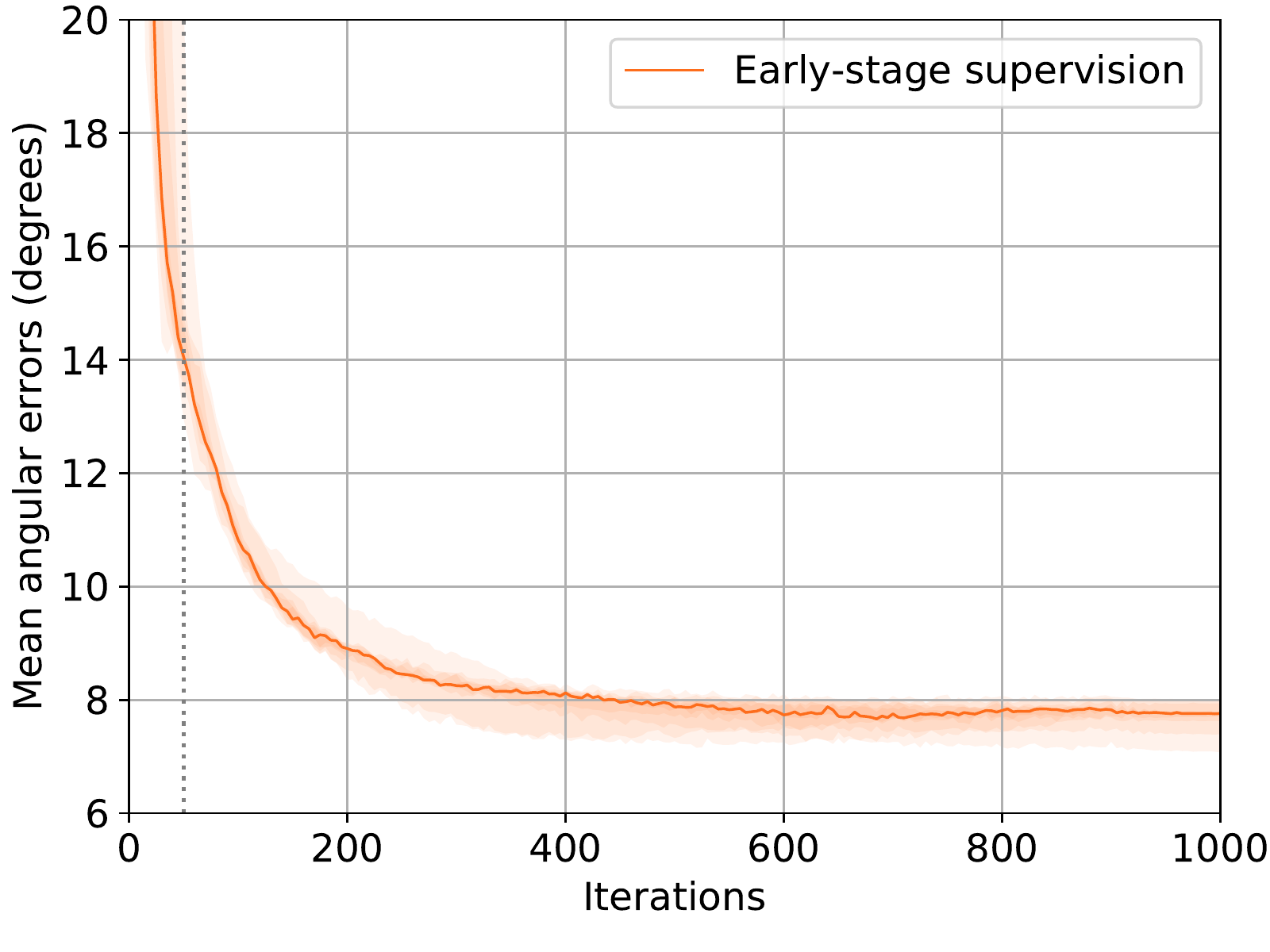}\hfil
	\includegraphics[width=\figw]{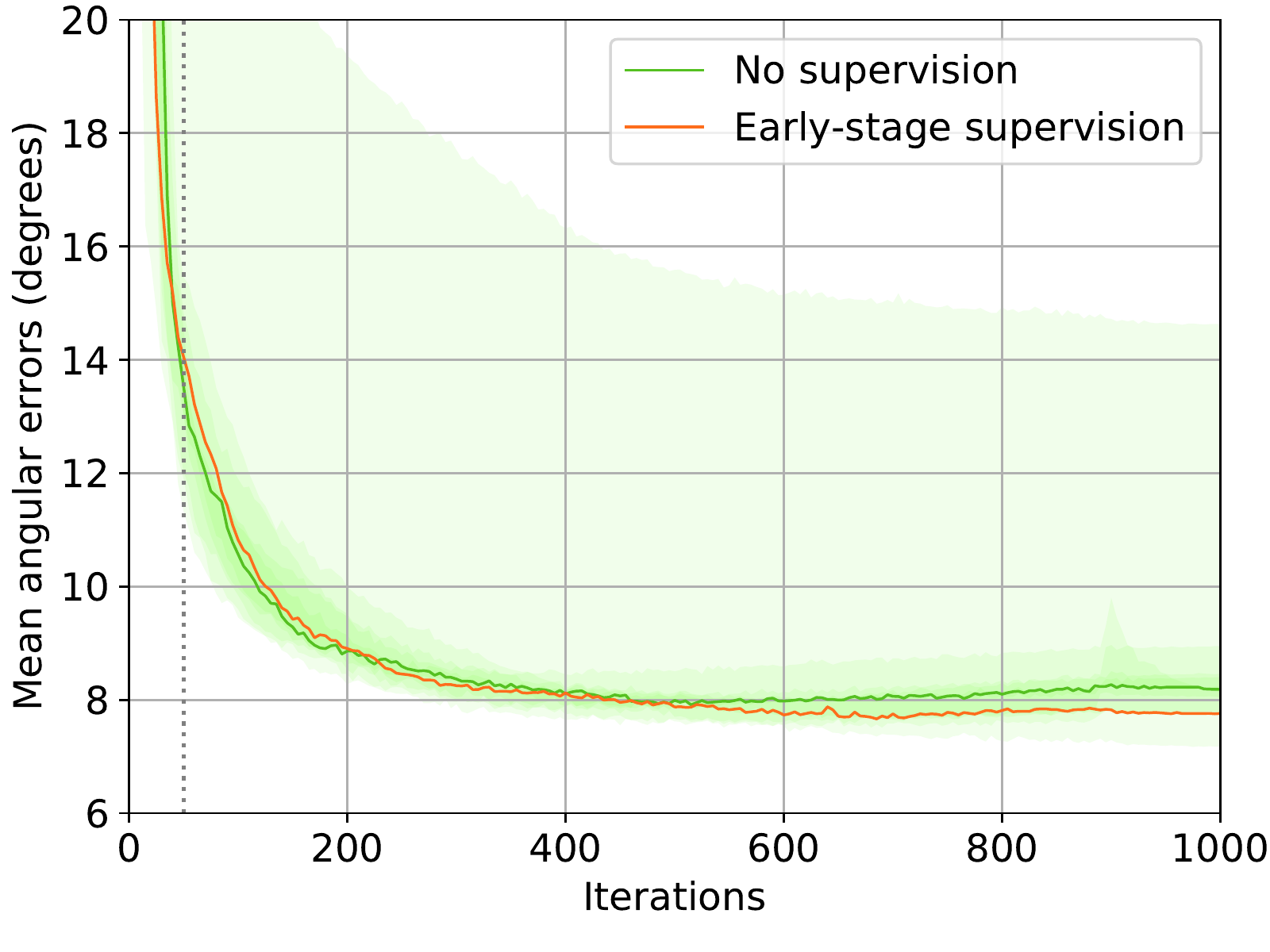}\hfil
	\includegraphics[width=\figw]{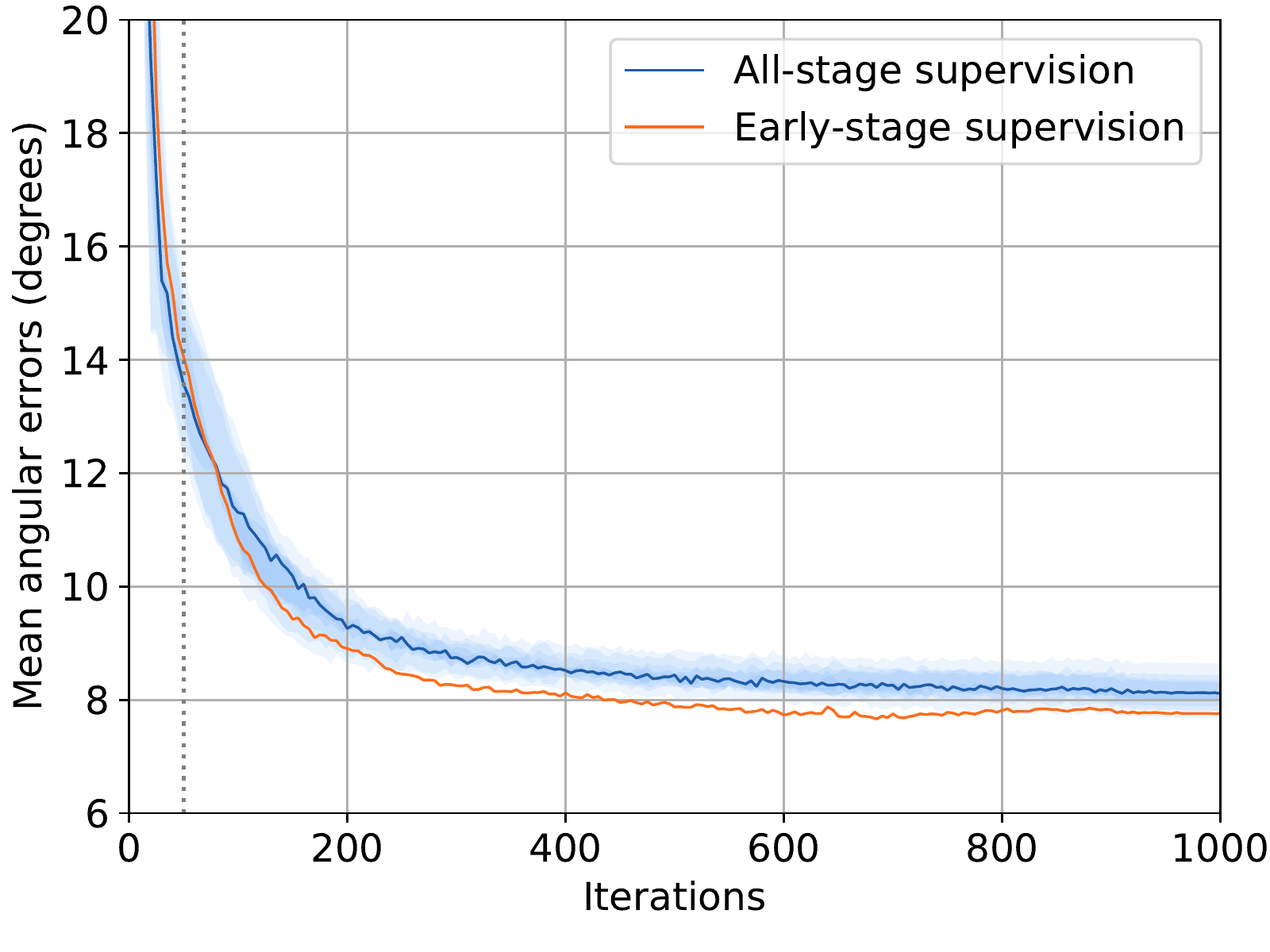}\\
	\includegraphics[width=\figw]{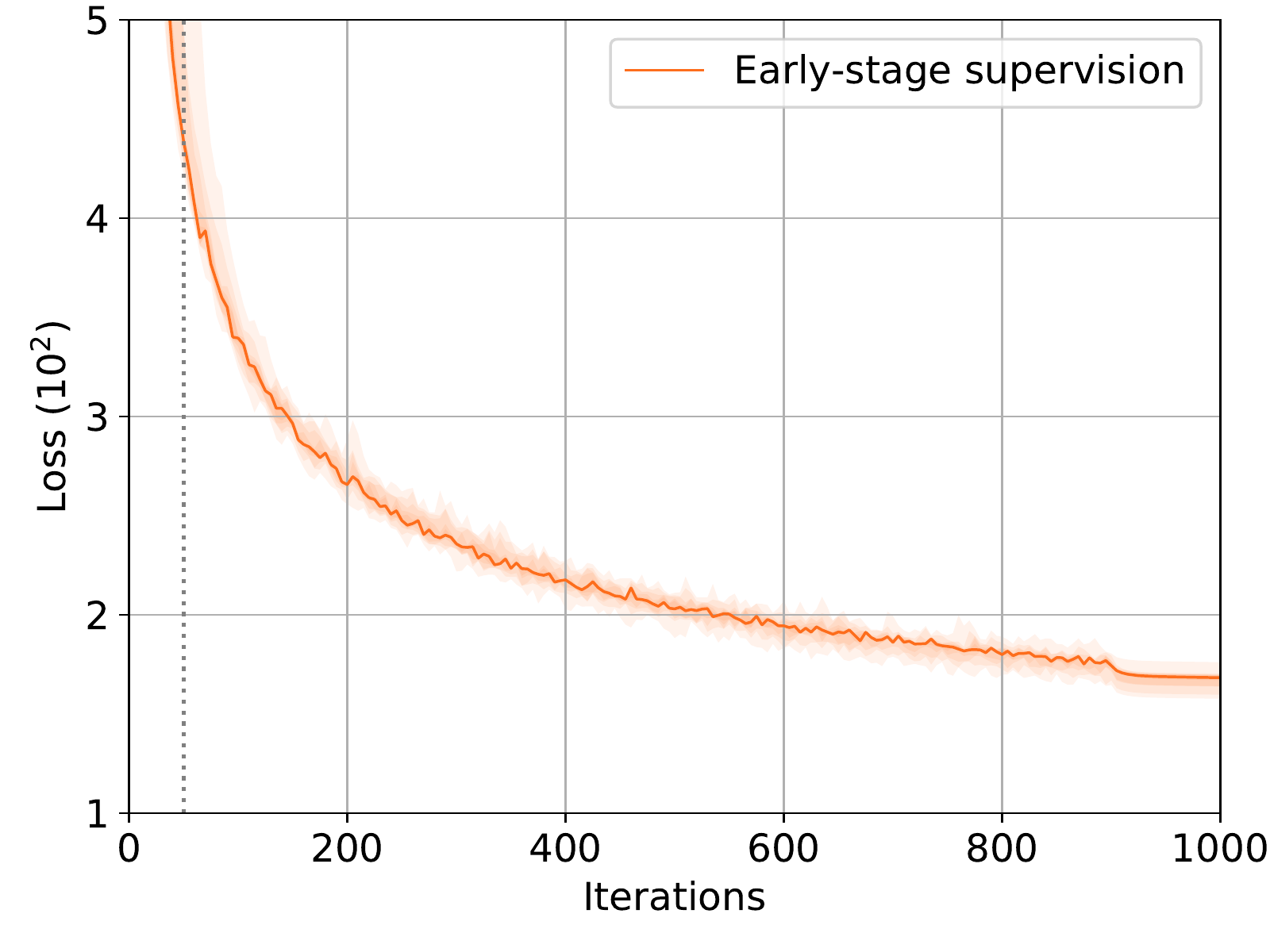}\hfil
	\includegraphics[width=\figw]{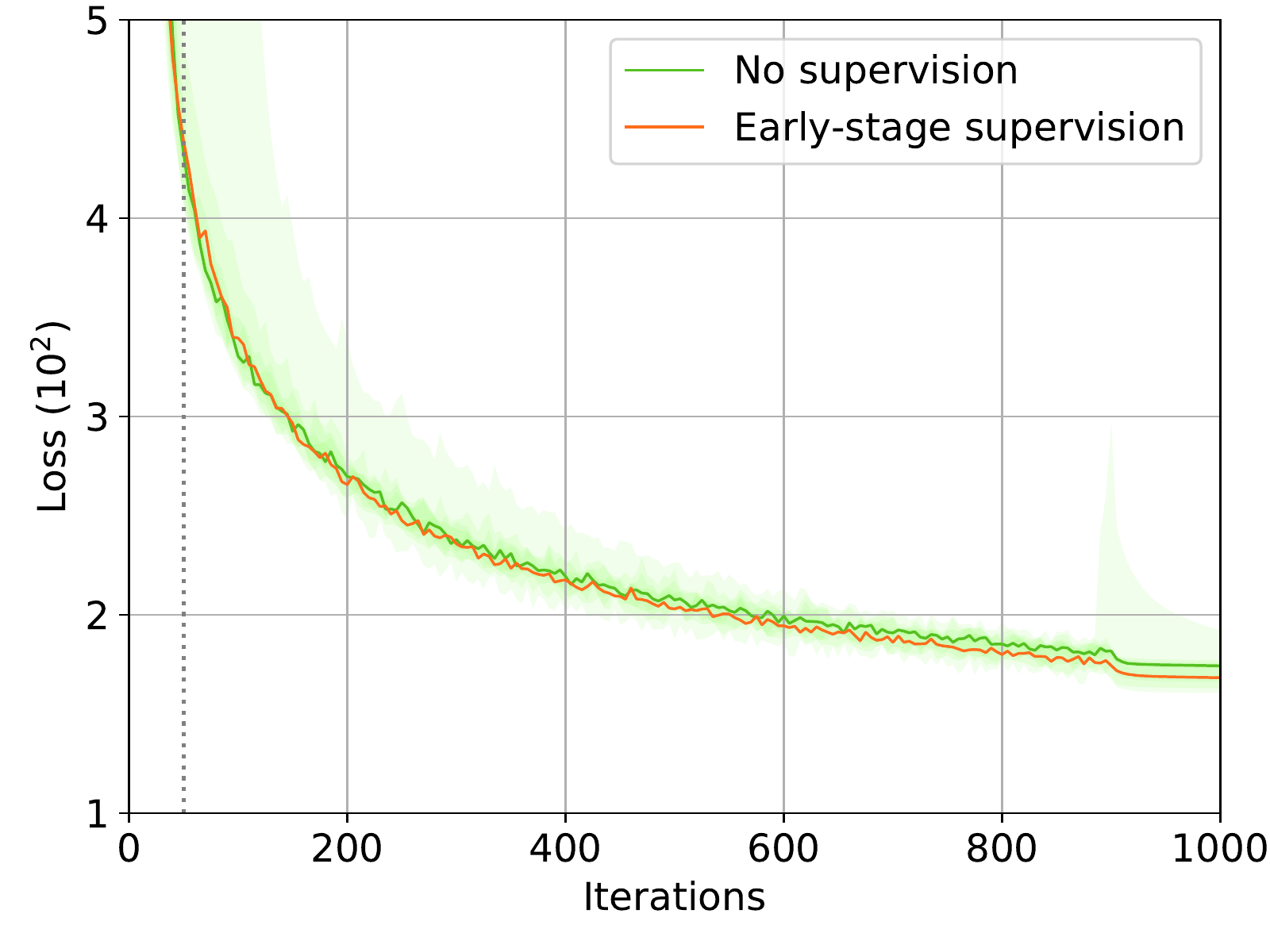}\hfil
	\includegraphics[width=\figw]{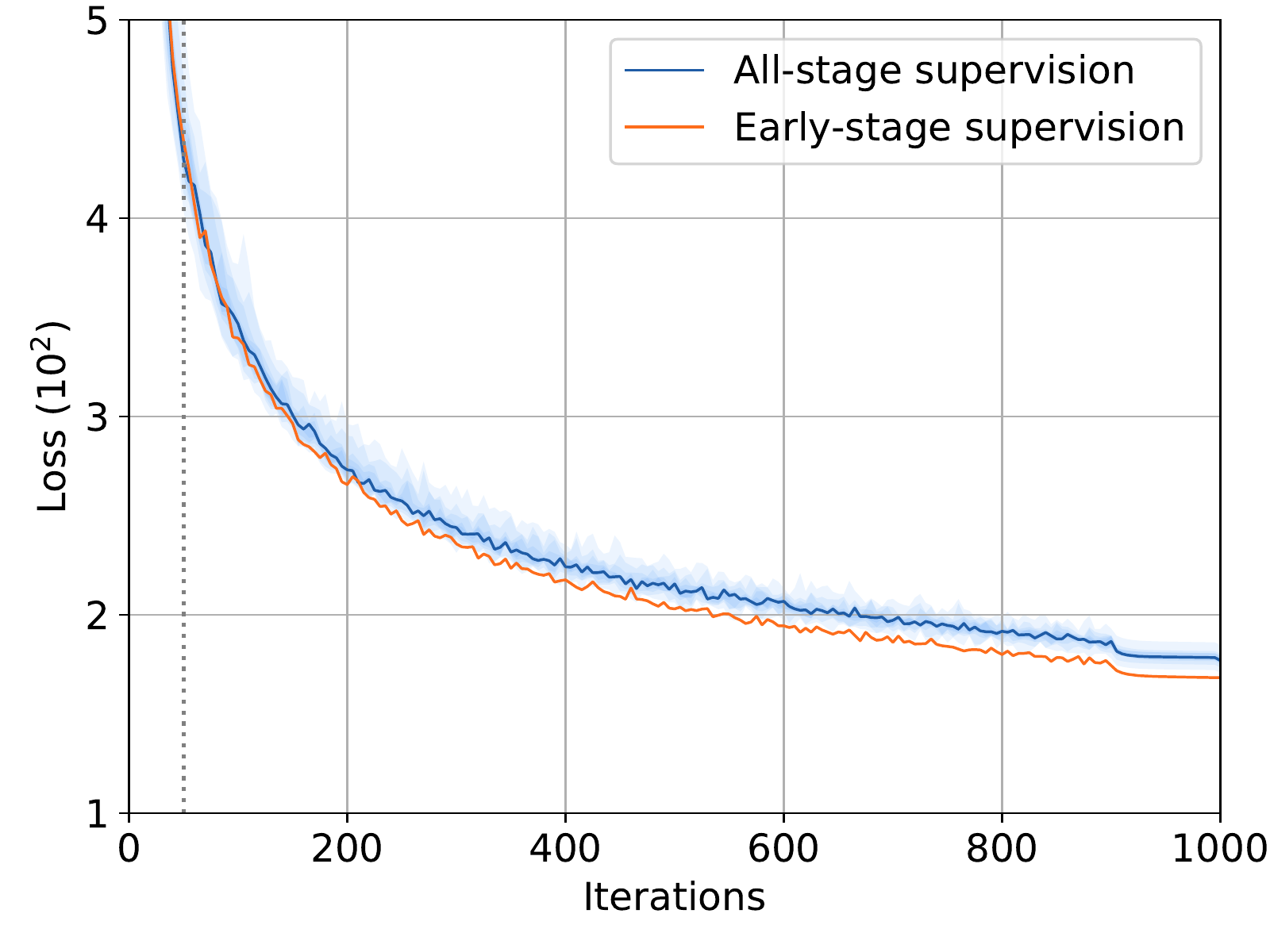}\\
	\begin{minipage}{\figw}\centering {Early-stage supervision}\end{minipage}\hfil
	\begin{minipage}{\figw}\centering {No supervision}\end{minipage}\hfil
	\begin{minipage}{\figw}\centering {All-stage supervision}\end{minipage}\\
	\caption{\textbf{Convergence analysis with different types of weak supervision for \textsc{pot2} scene.}
		See also explanations in Fig.~\ref{afig:training_ball}.
	}
\end{figure*}

\begin{figure*}[p]
	\centering
	\small
	\def\figw{0.30\linewidth}
	% 0:ball, 1:bear, 2:buddha, 3:cat, 4:cow, 5:goblet, 6:harbest, 7:pot1, 8:pot2, 9:reading
	\includegraphics[width=\figw]{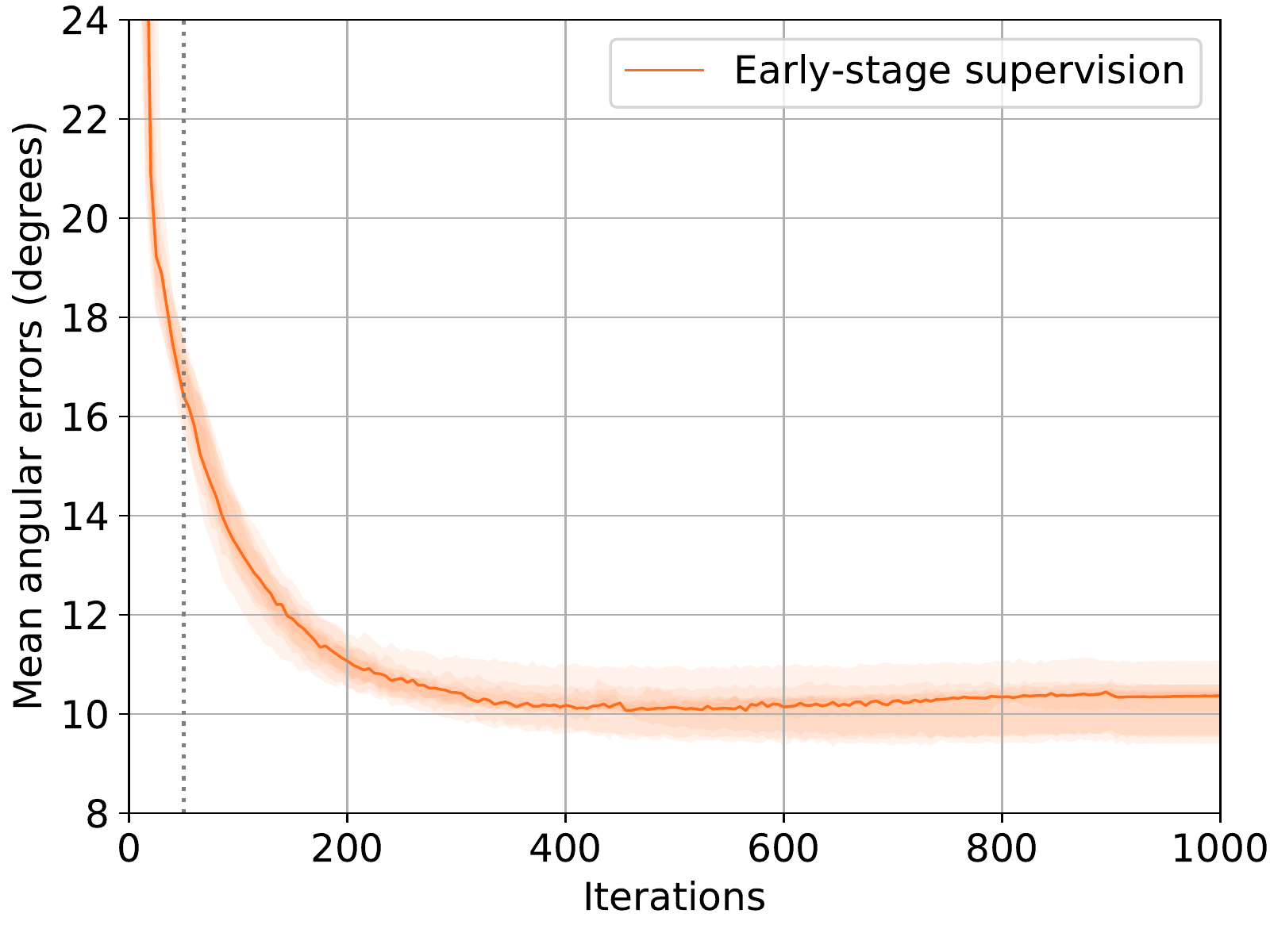}\hfil
	\includegraphics[width=\figw]{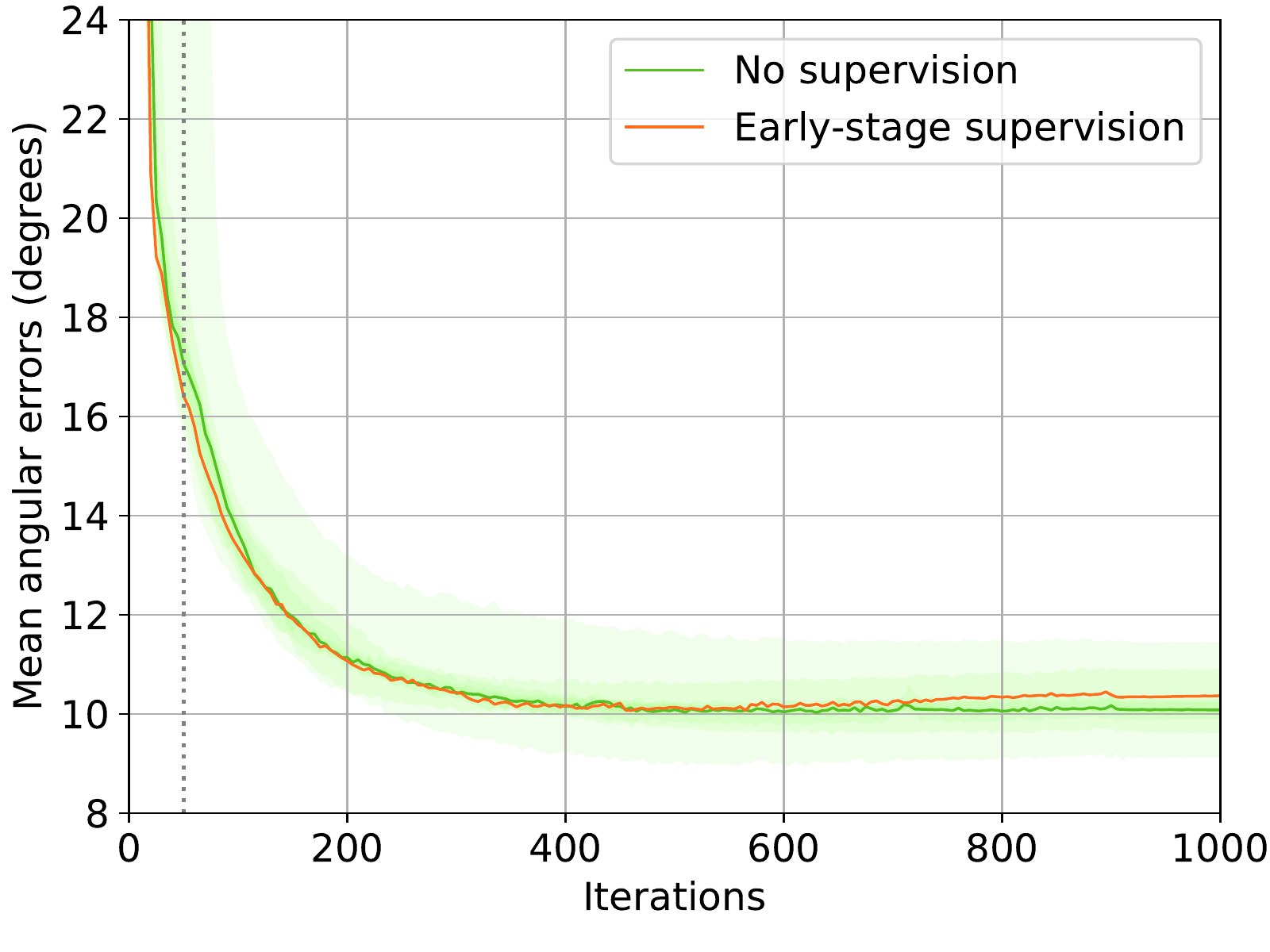}\hfil
	\includegraphics[width=\figw]{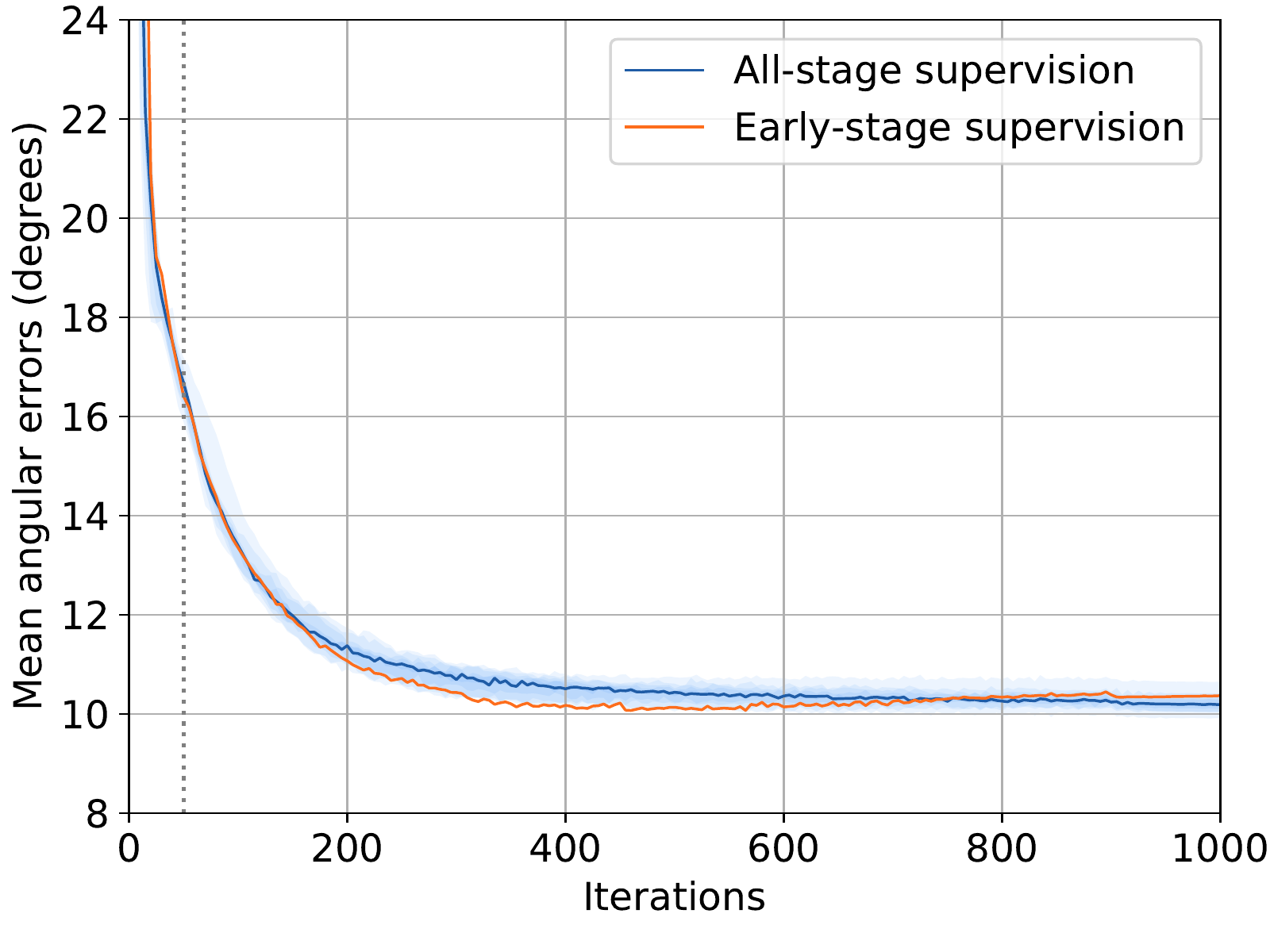}\\
	\includegraphics[width=\figw]{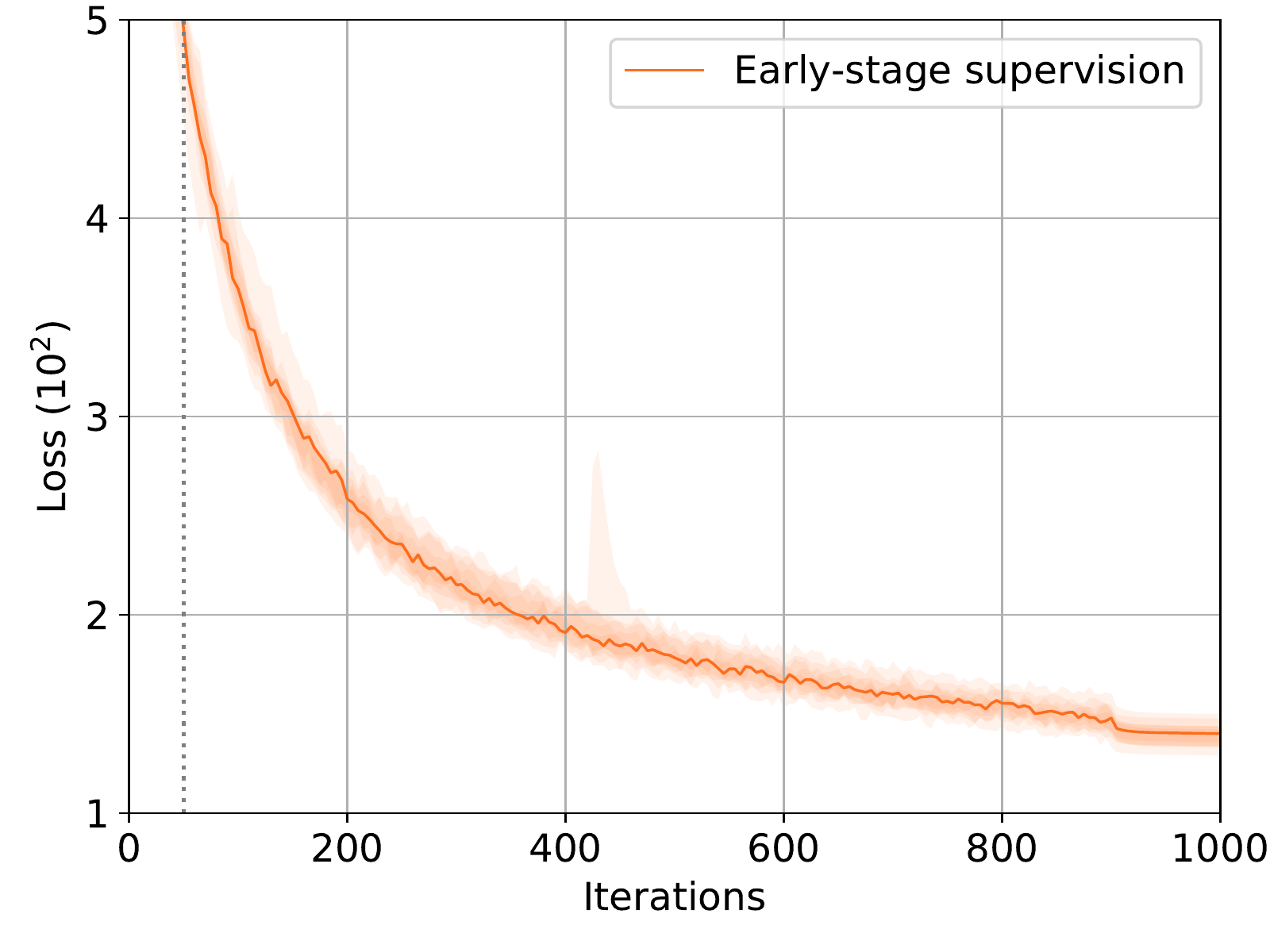}\hfil
	\includegraphics[width=\figw]{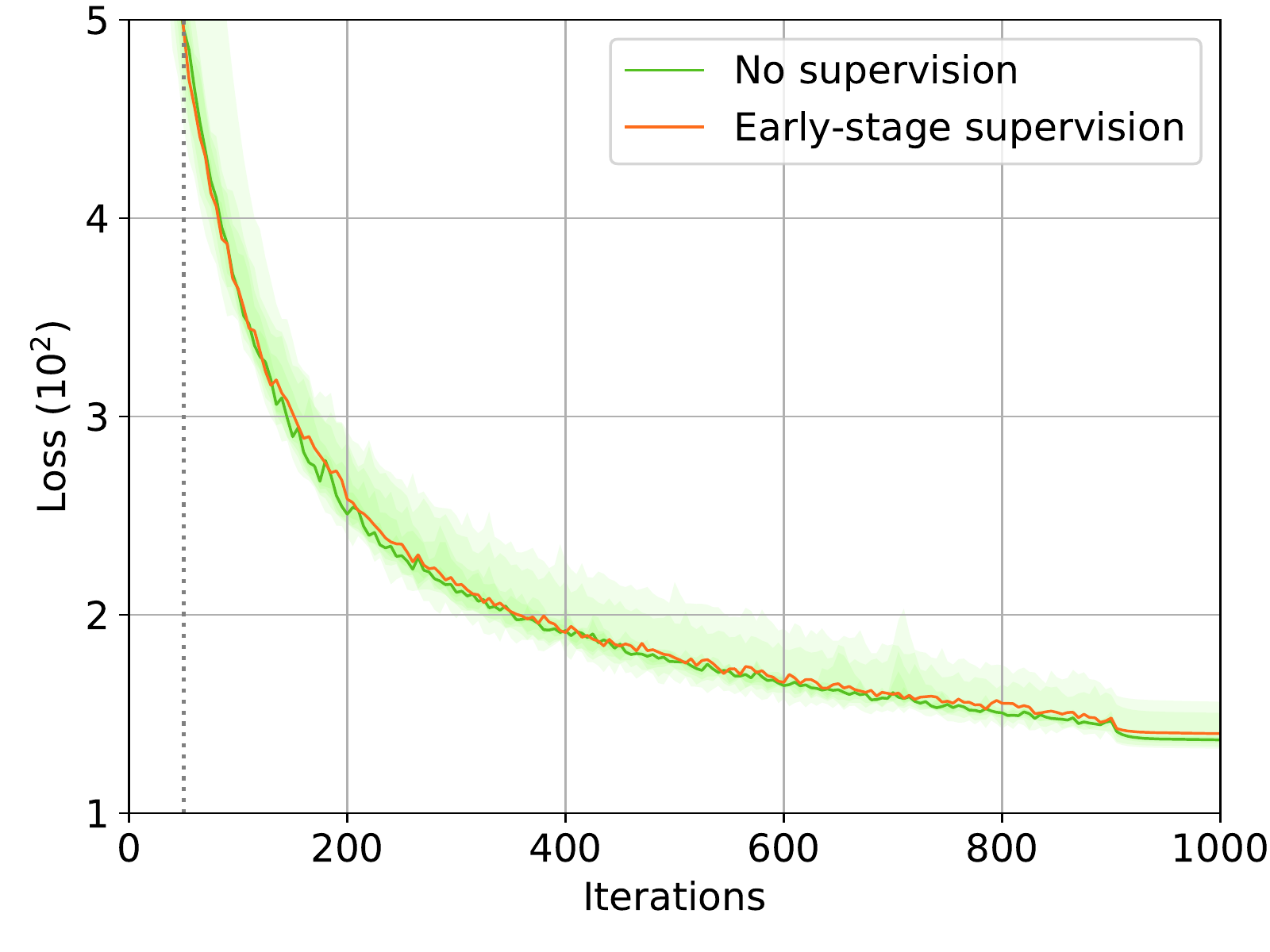}\hfil
	\includegraphics[width=\figw]{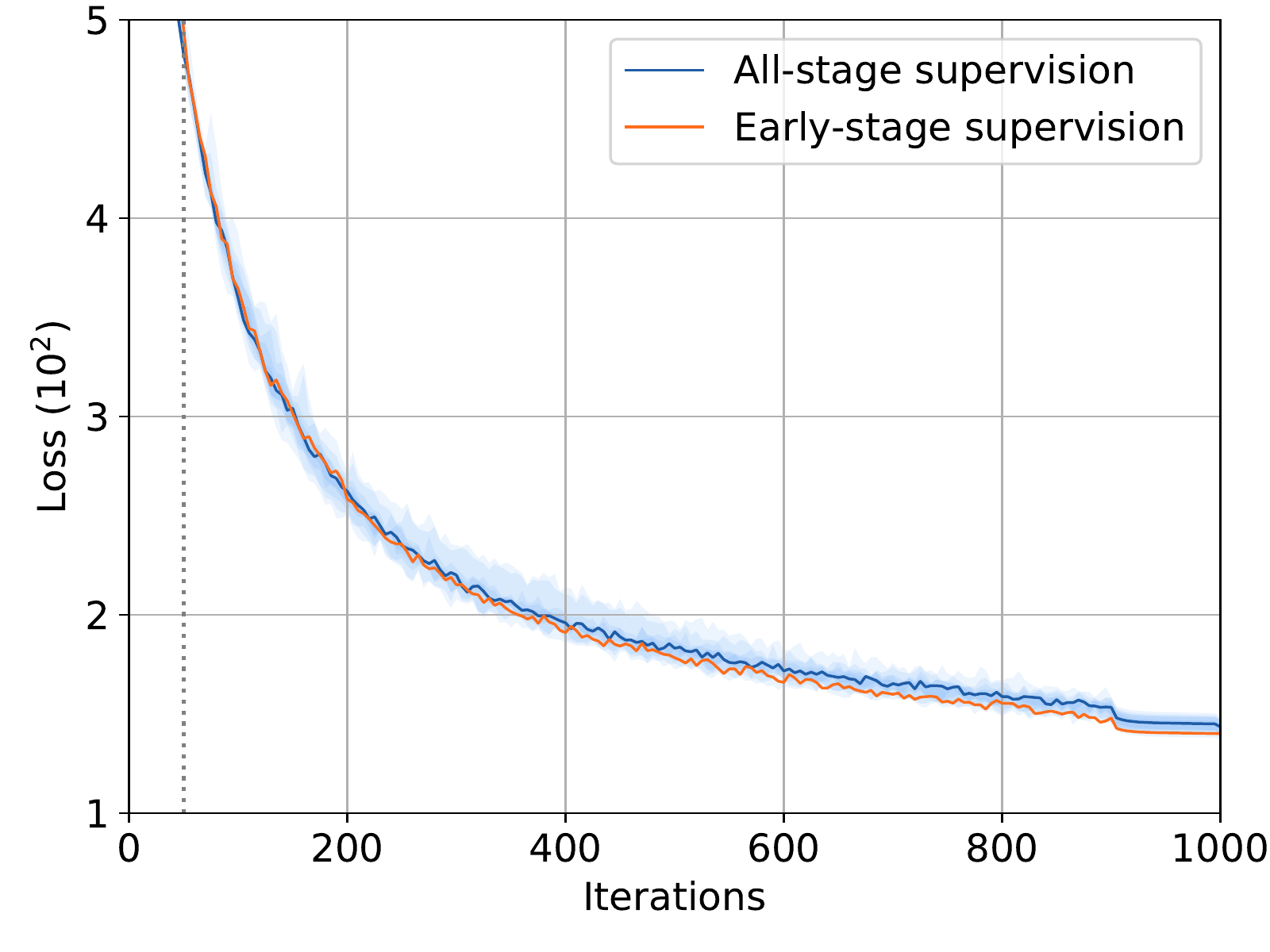}\\
	\begin{minipage}{\figw}\centering {Early-stage supervision}\end{minipage}\hfil
	\begin{minipage}{\figw}\centering {No supervision}\end{minipage}\hfil
	\begin{minipage}{\figw}\centering {All-stage supervision}\end{minipage}\\
	\caption{\textbf{Convergence analysis with different types of weak supervision for \textsc{buddha} scene.}
		See also explanations in Fig.~\ref{afig:training_ball}.
	}
\end{figure*}

\begin{figure*}[p]
	\centering
	\small
	\def\figw{0.30\linewidth}
	% 0:ball, 1:bear, 2:buddha, 3:cat, 4:cow, 5:goblet, 6:harbest, 7:pot1, 8:pot2, 9:reading
	\includegraphics[width=\figw]{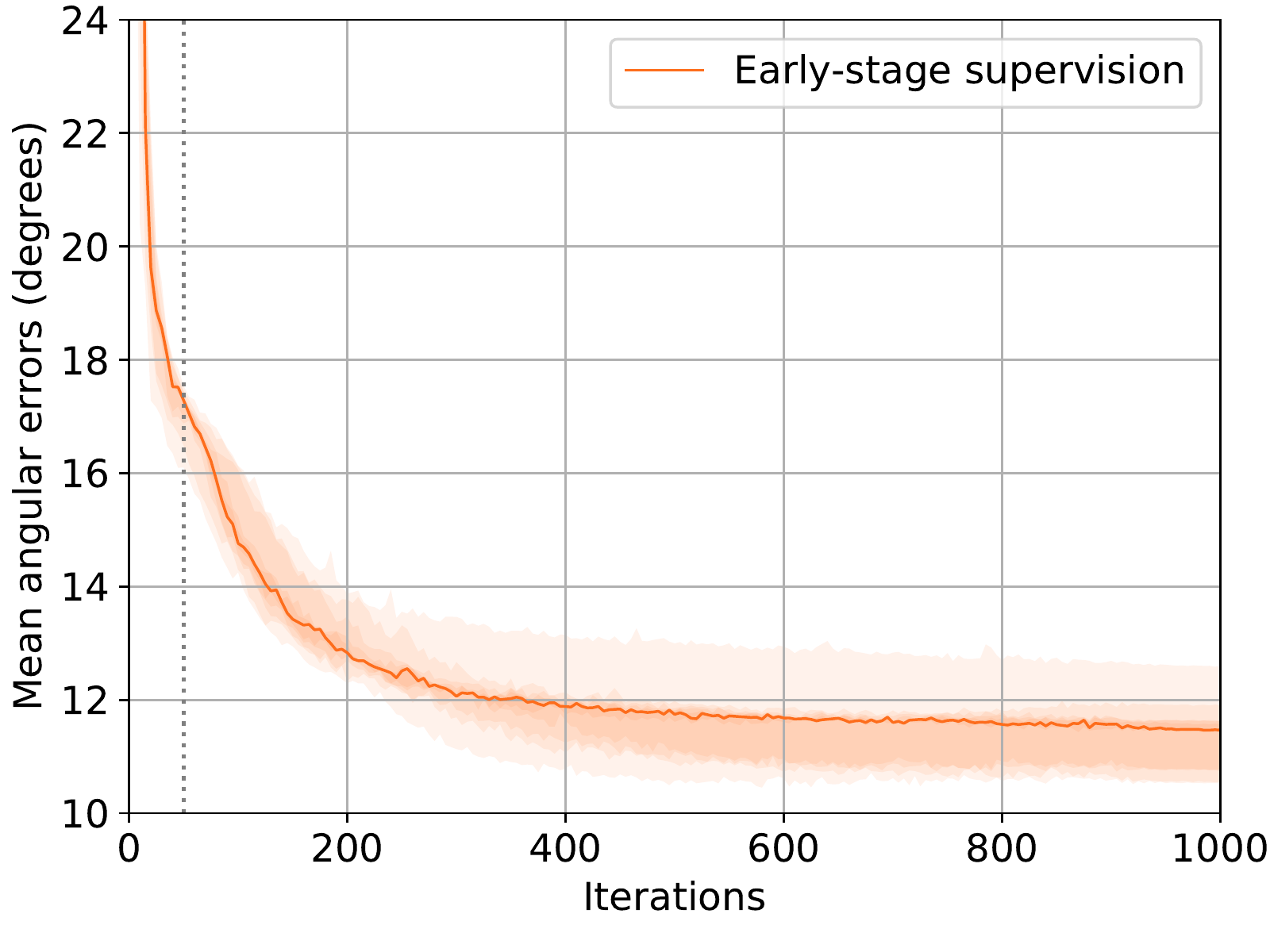}\hfil
	\includegraphics[width=\figw]{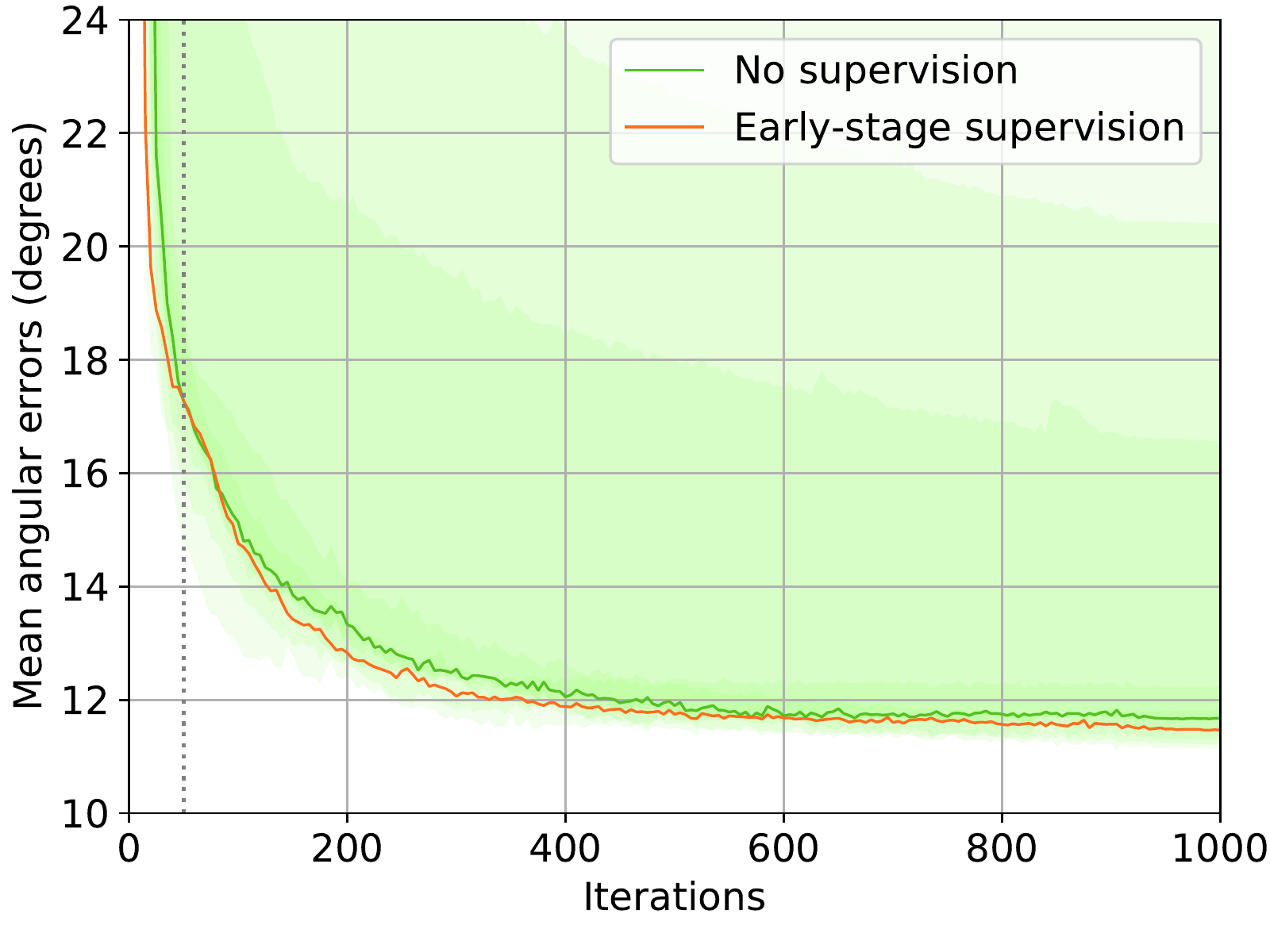}\hfil
	\includegraphics[width=\figw]{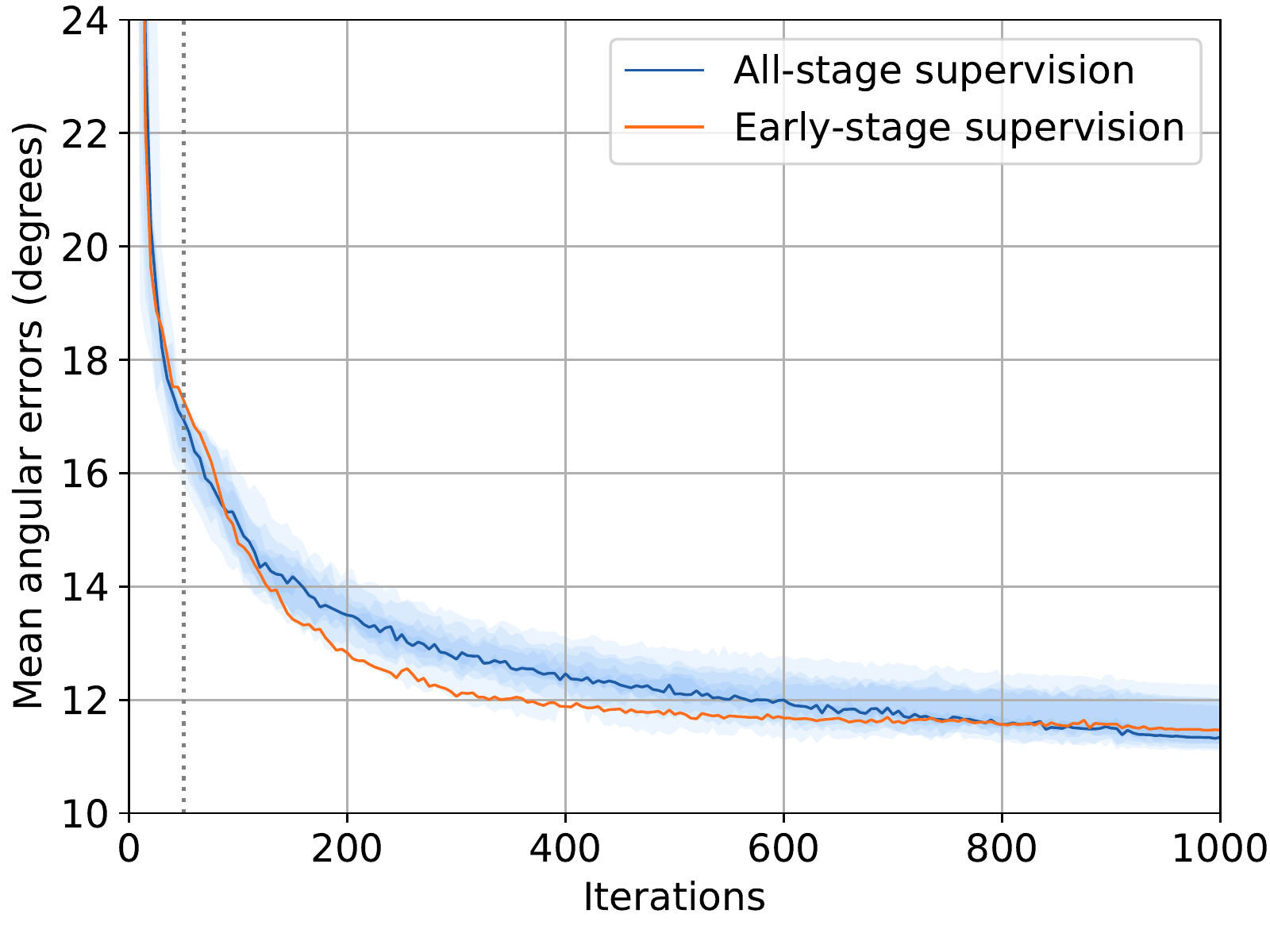}\\
	\includegraphics[width=\figw]{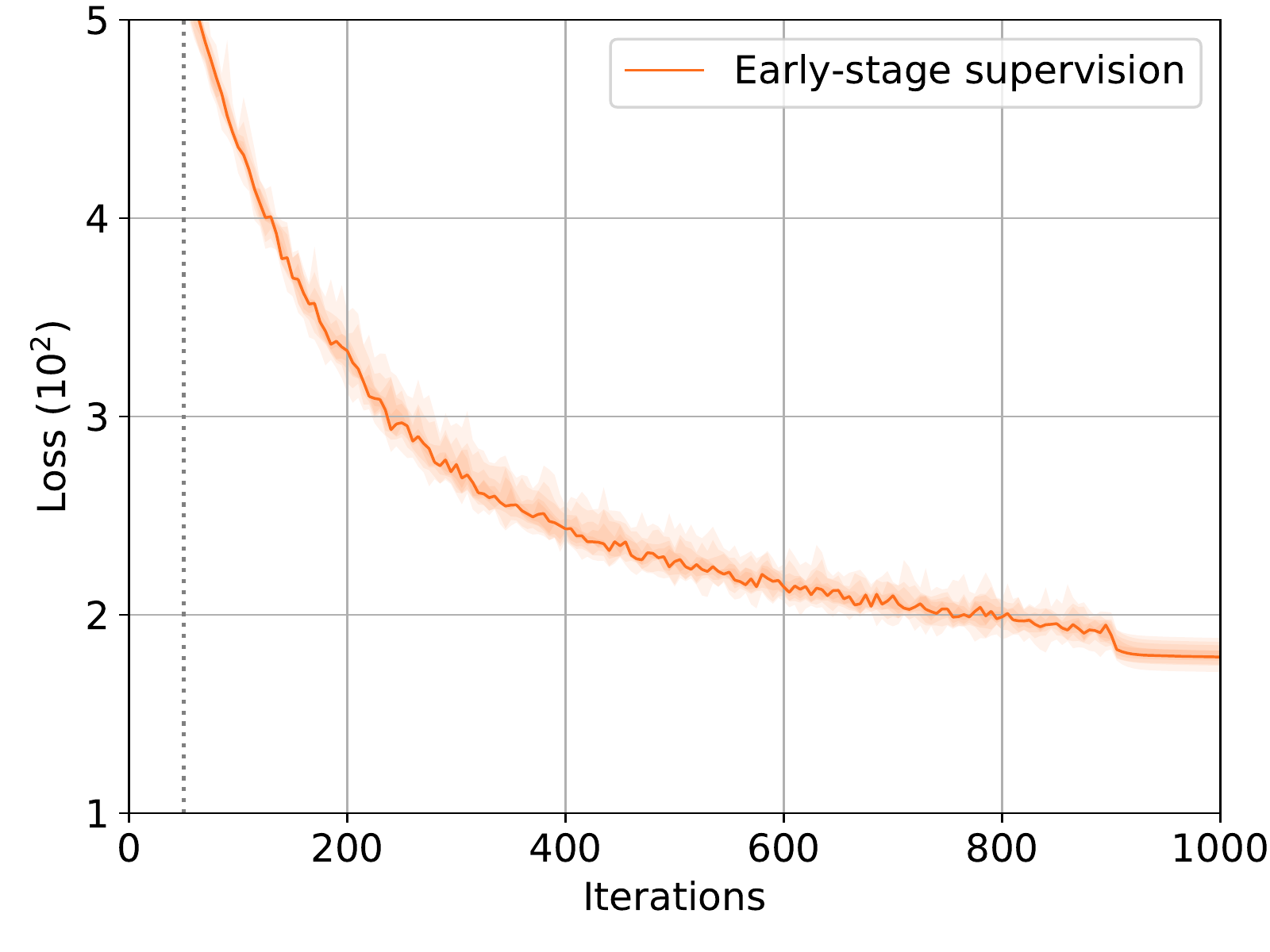}\hfil
	\includegraphics[width=\figw]{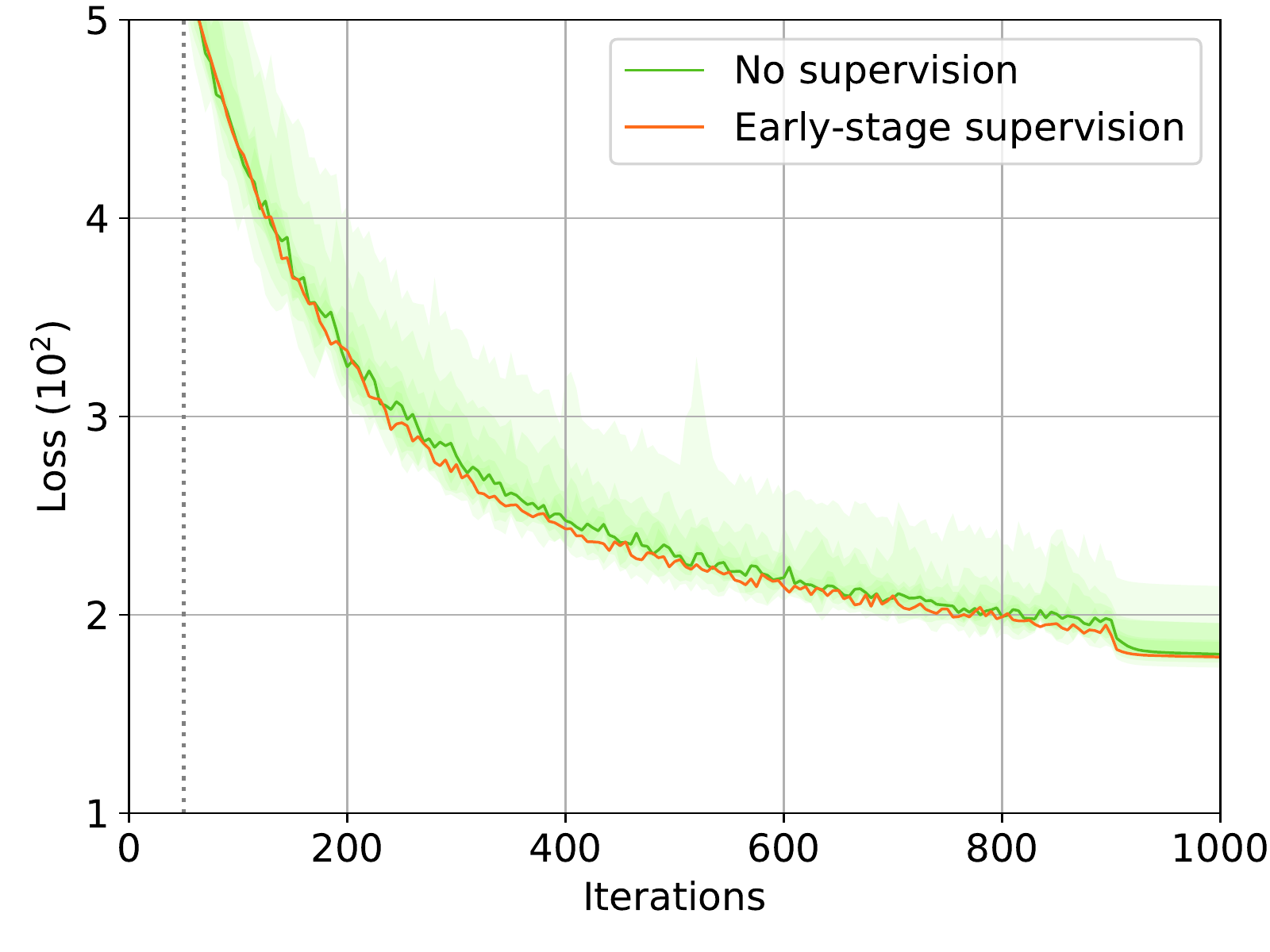}\hfil
	\includegraphics[width=\figw]{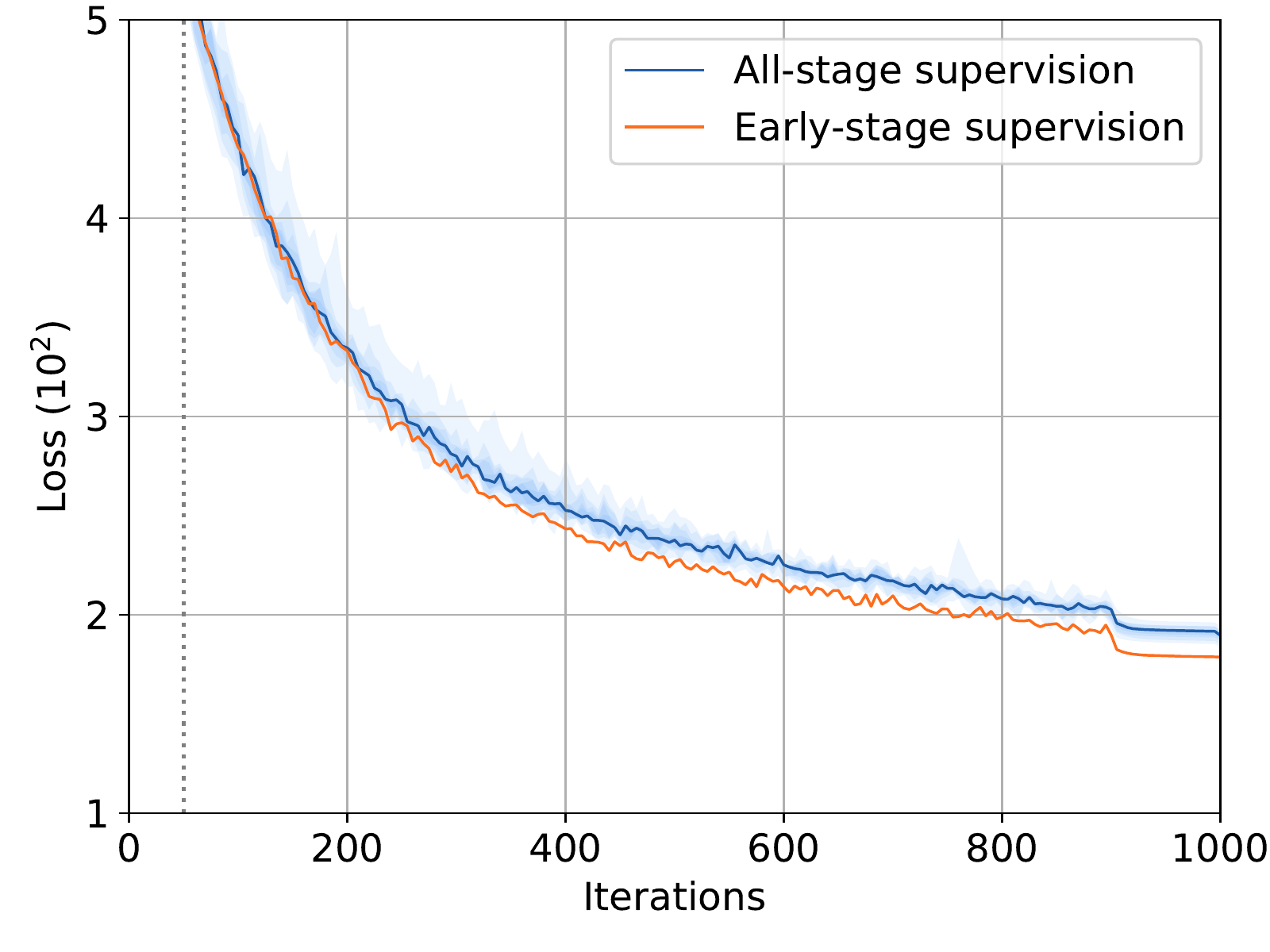}\\
	\begin{minipage}{\figw}\centering {Early-stage supervision}\end{minipage}\hfil
	\begin{minipage}{\figw}\centering {No supervision}\end{minipage}\hfil
	\begin{minipage}{\figw}\centering {All-stage supervision}\end{minipage}\\
	\caption{\textbf{Convergence analysis with different types of weak supervision for \textsc{goblet} scene.}
		See also explanations in Fig.~\ref{afig:training_ball}.
	}
\end{figure*}

\begin{figure*}[p]
	\centering
	\small
	\def\figw{0.30\linewidth}
	% 0:ball, 1:bear, 2:buddha, 3:cat, 4:cow, 5:goblet, 6:harbest, 7:pot1, 8:pot2, 9:reading
	\includegraphics[width=\figw]{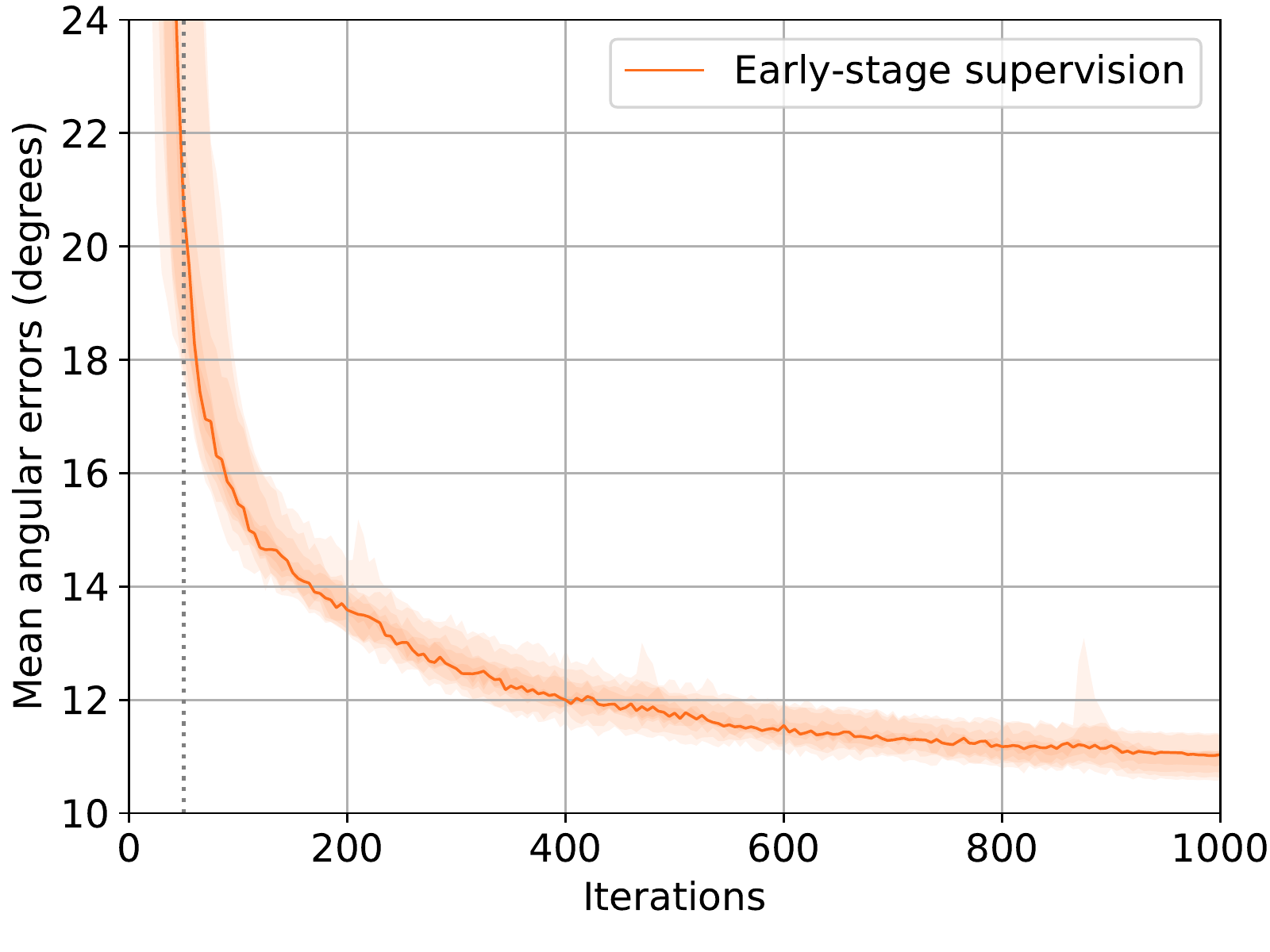}\hfil
	\includegraphics[width=\figw]{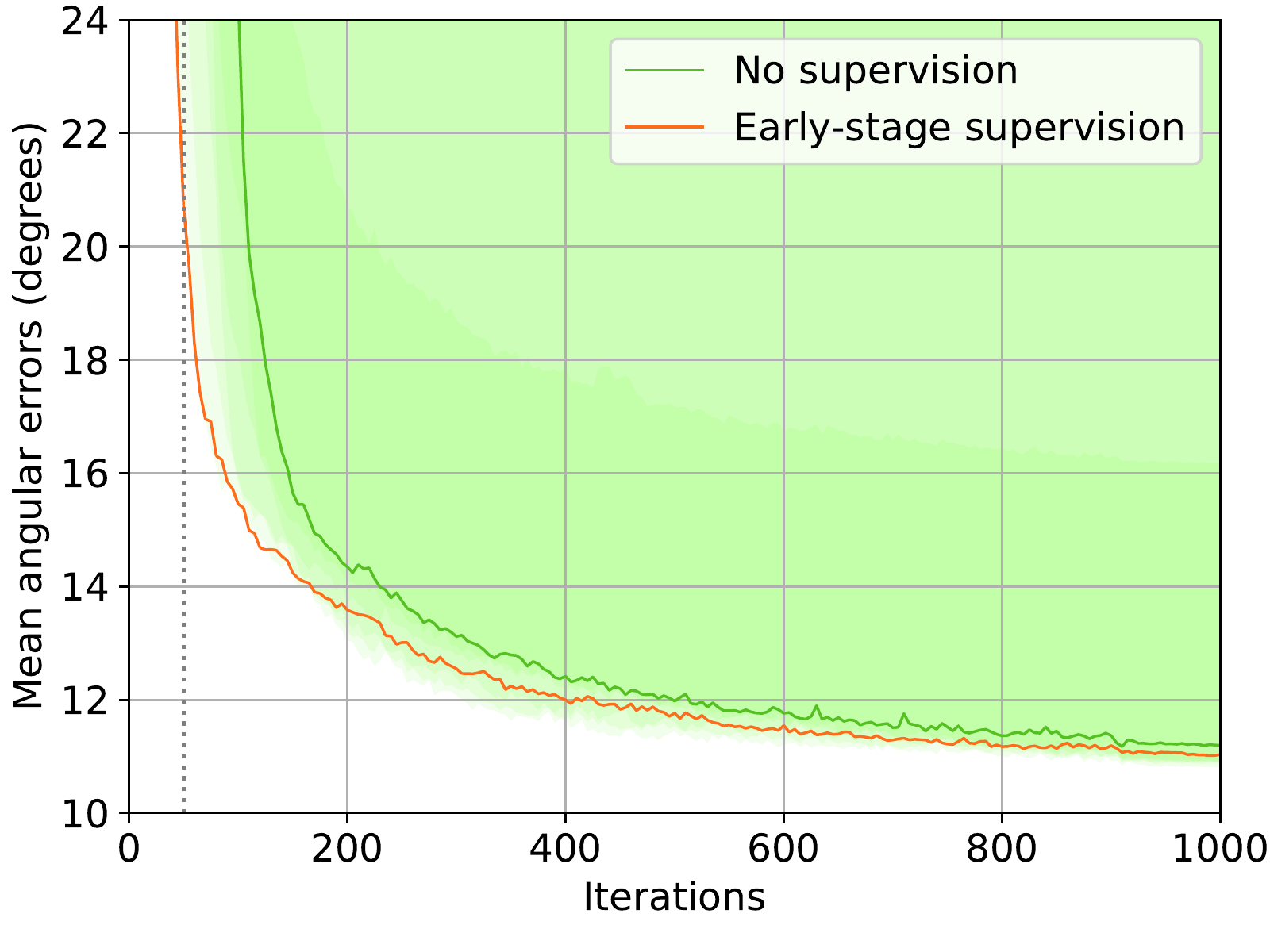}\hfil
	\includegraphics[width=\figw]{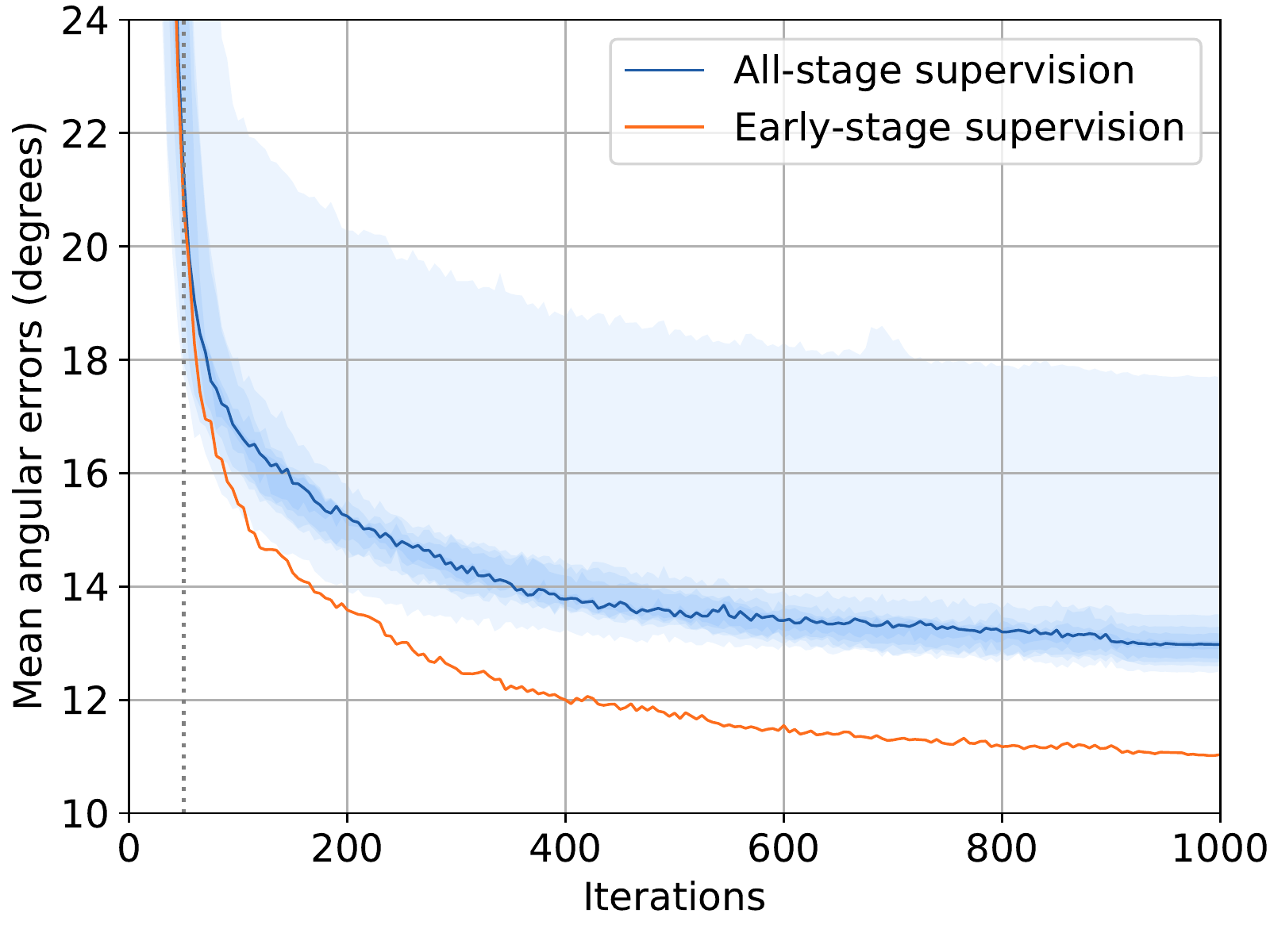}\\
	\includegraphics[width=\figw]{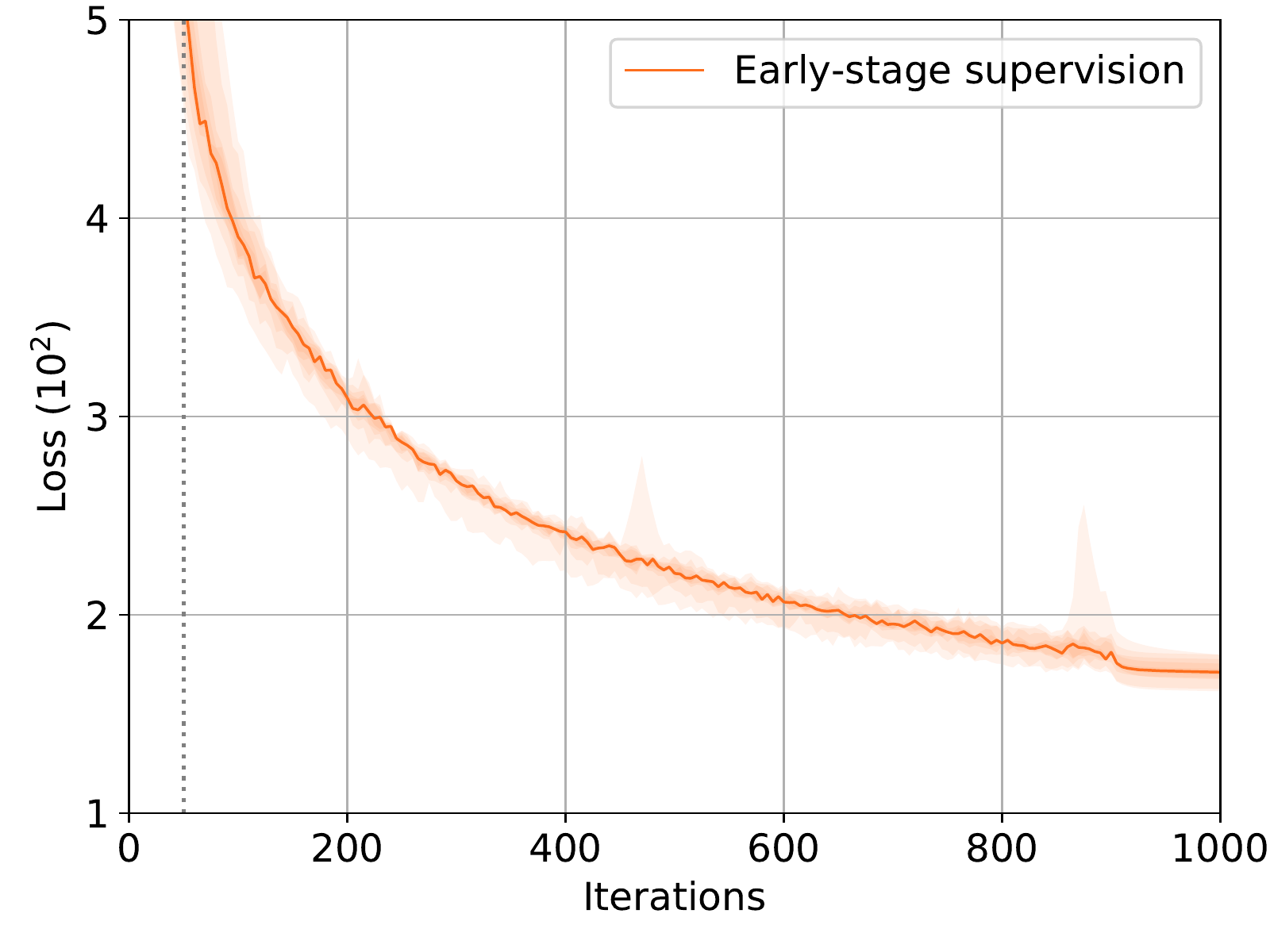}\hfil
	\includegraphics[width=\figw]{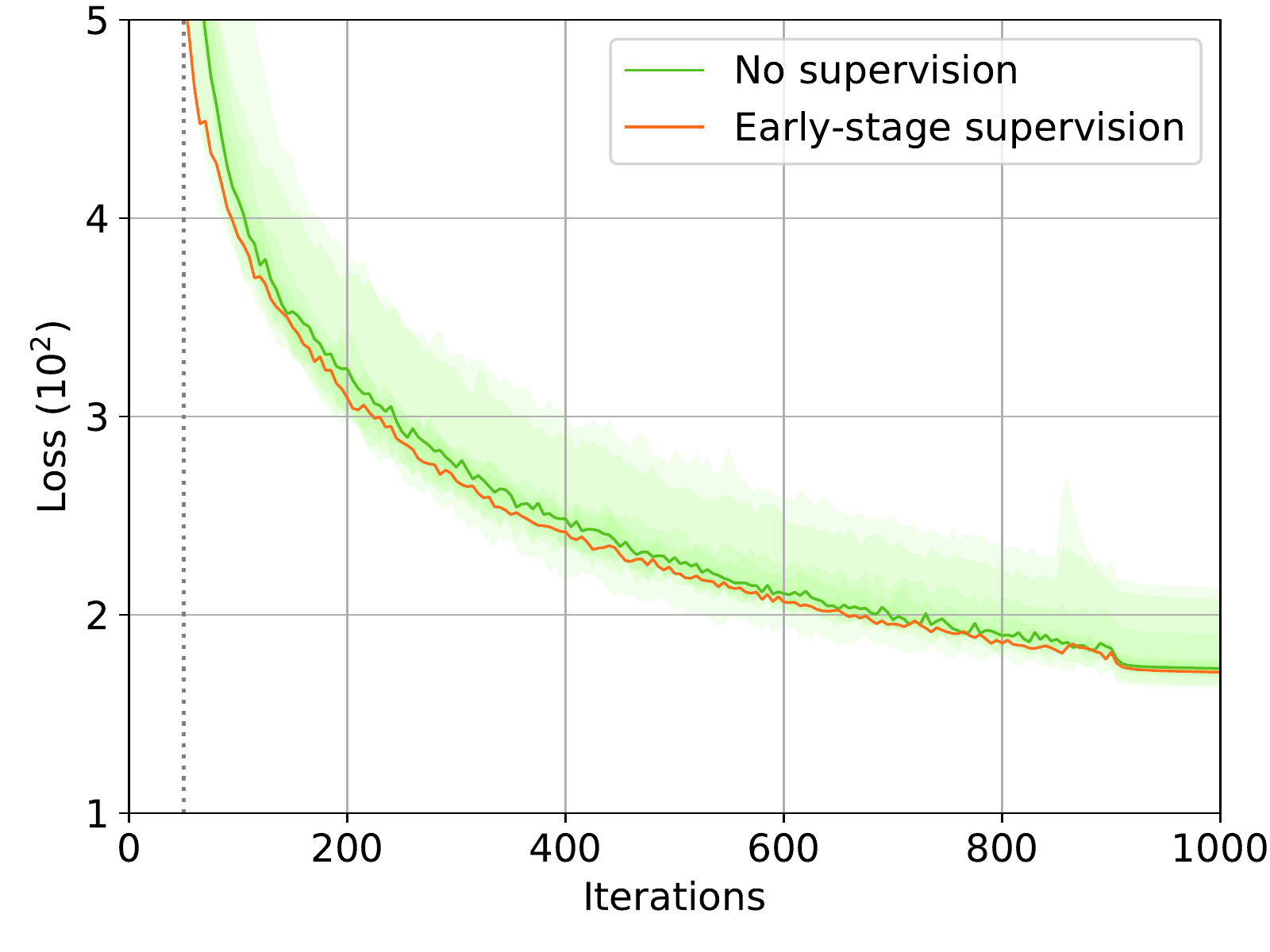}\hfil
	\includegraphics[width=\figw]{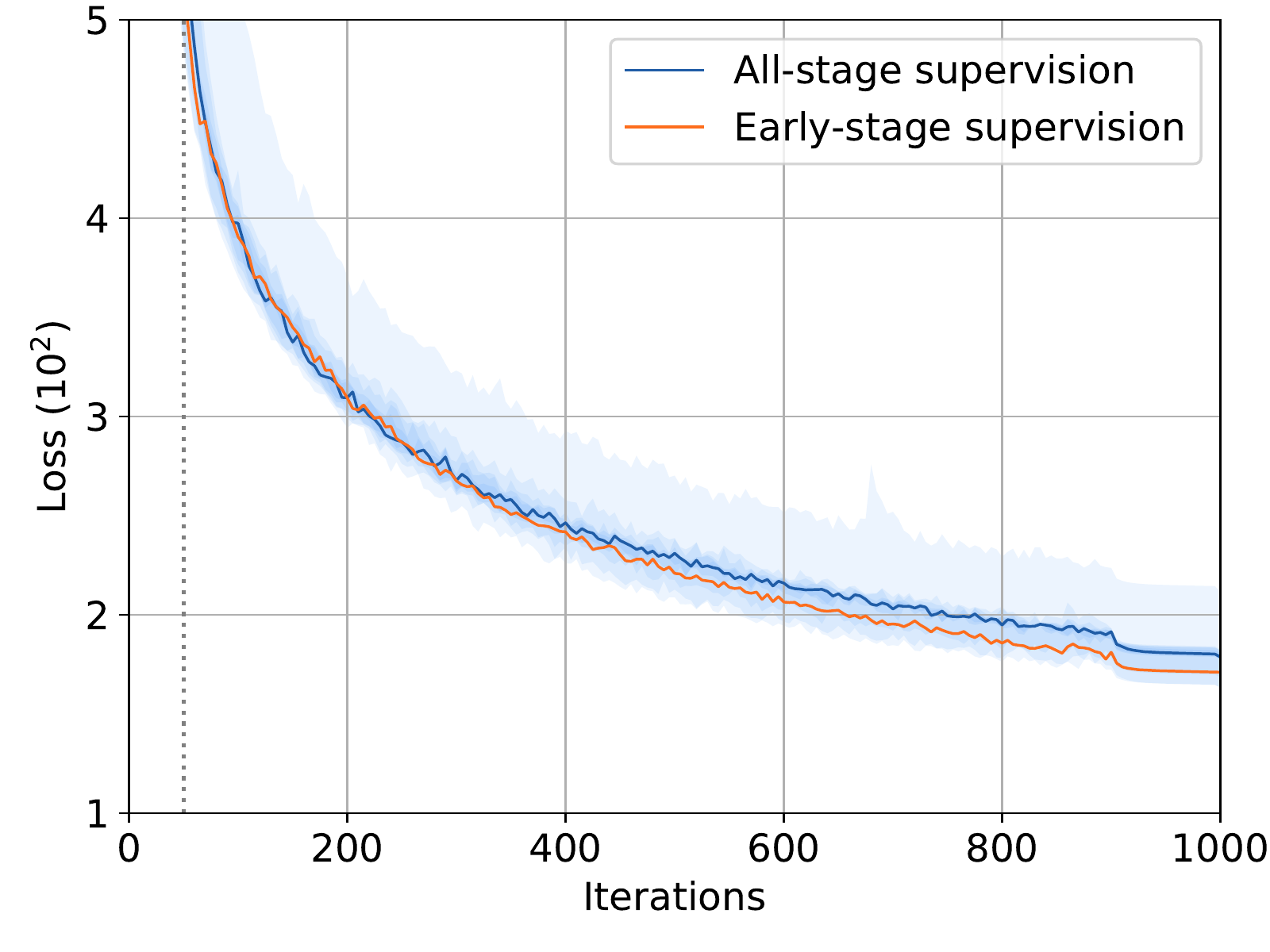}\\
	\begin{minipage}{\figw}\centering {Early-stage supervision}\end{minipage}\hfil
	\begin{minipage}{\figw}\centering {No supervision}\end{minipage}\hfil
	\begin{minipage}{\figw}\centering {All-stage supervision}\end{minipage}\\
	\caption{\textbf{Convergence analysis with different types of weak supervision for \textsc{reading} scene.}
		See also explanations in Fig.~\ref{afig:training_ball}.
	}
\end{figure*}

\begin{figure*}[p]
	\centering
	\small
	\def\figw{0.30\linewidth}
	% 0:ball, 1:bear, 2:buddha, 3:cat, 4:cow, 5:goblet, 6:harbest, 7:pot1, 8:pot2, 9:reading
	\includegraphics[width=\figw]{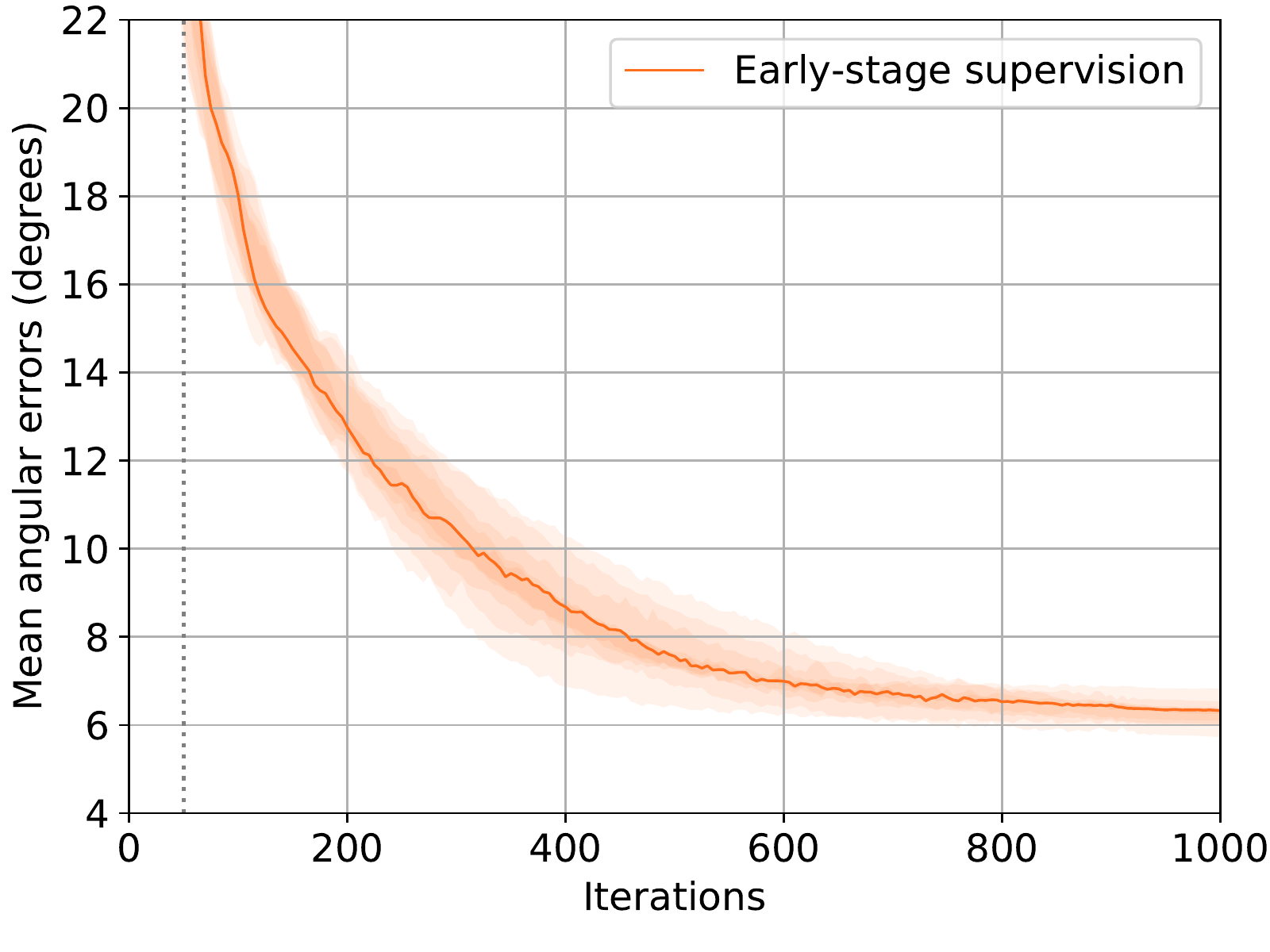}\hfil
	\includegraphics[width=\figw]{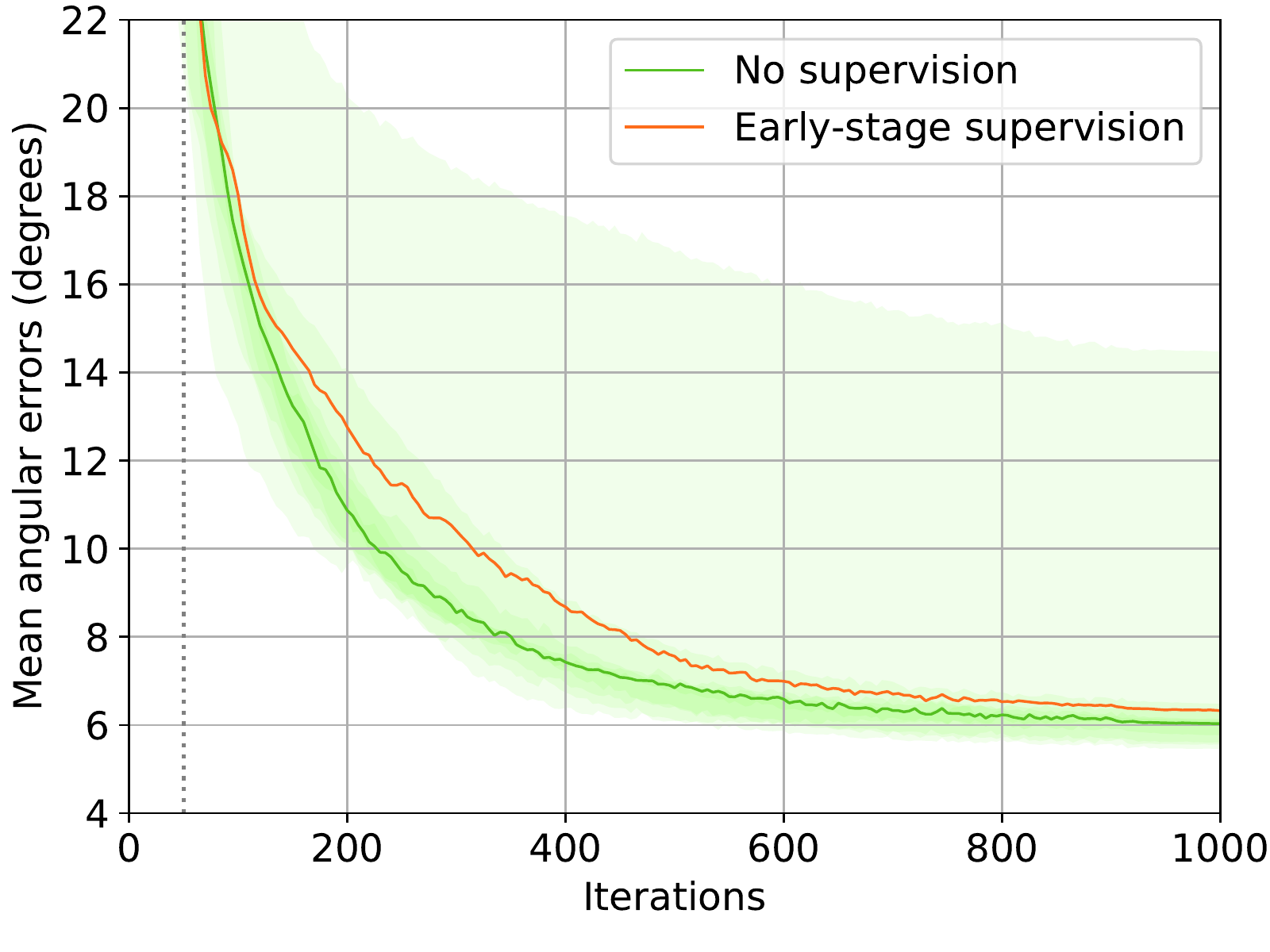}\hfil
	\includegraphics[width=\figw]{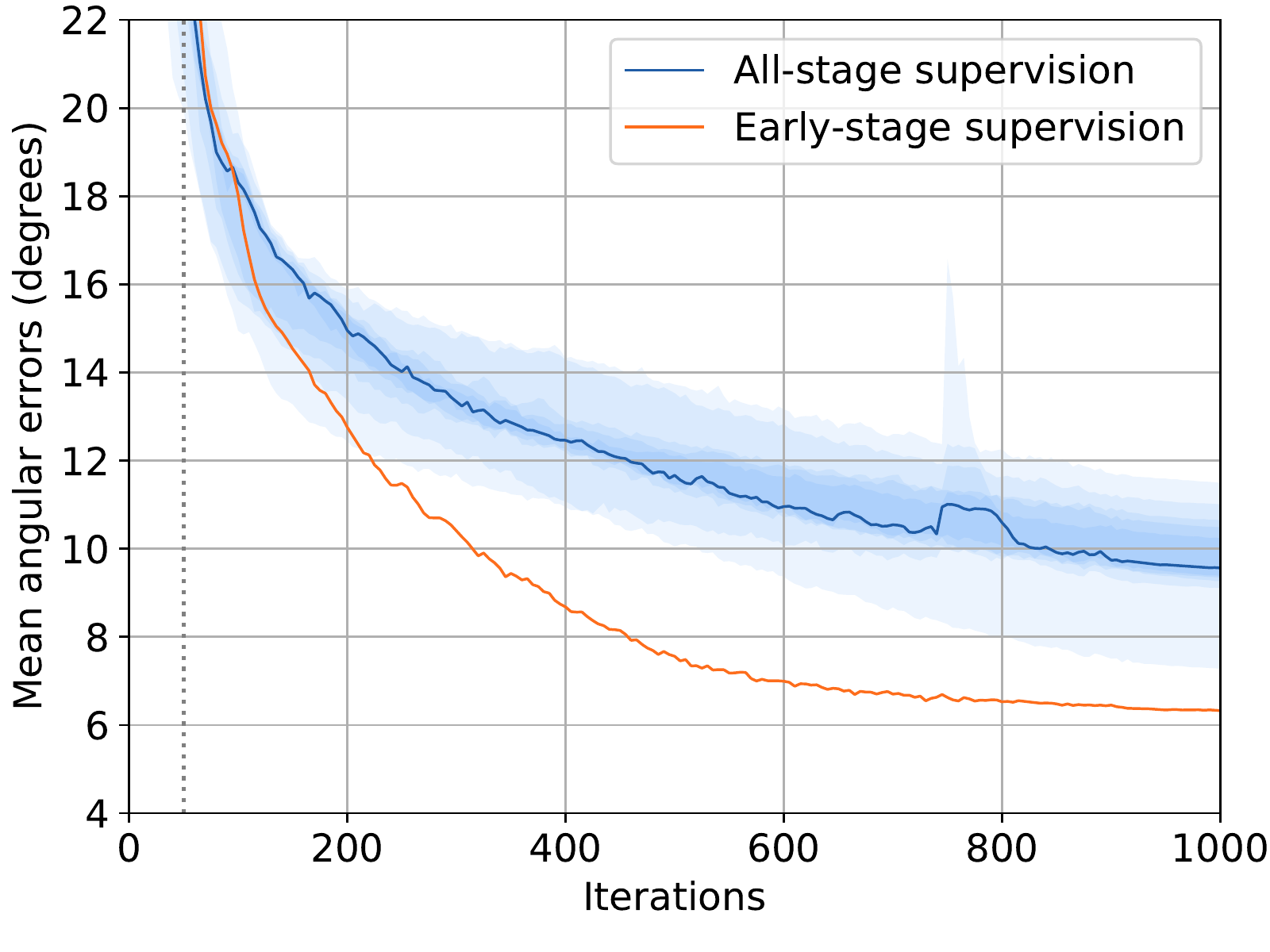}\\
	\includegraphics[width=\figw]{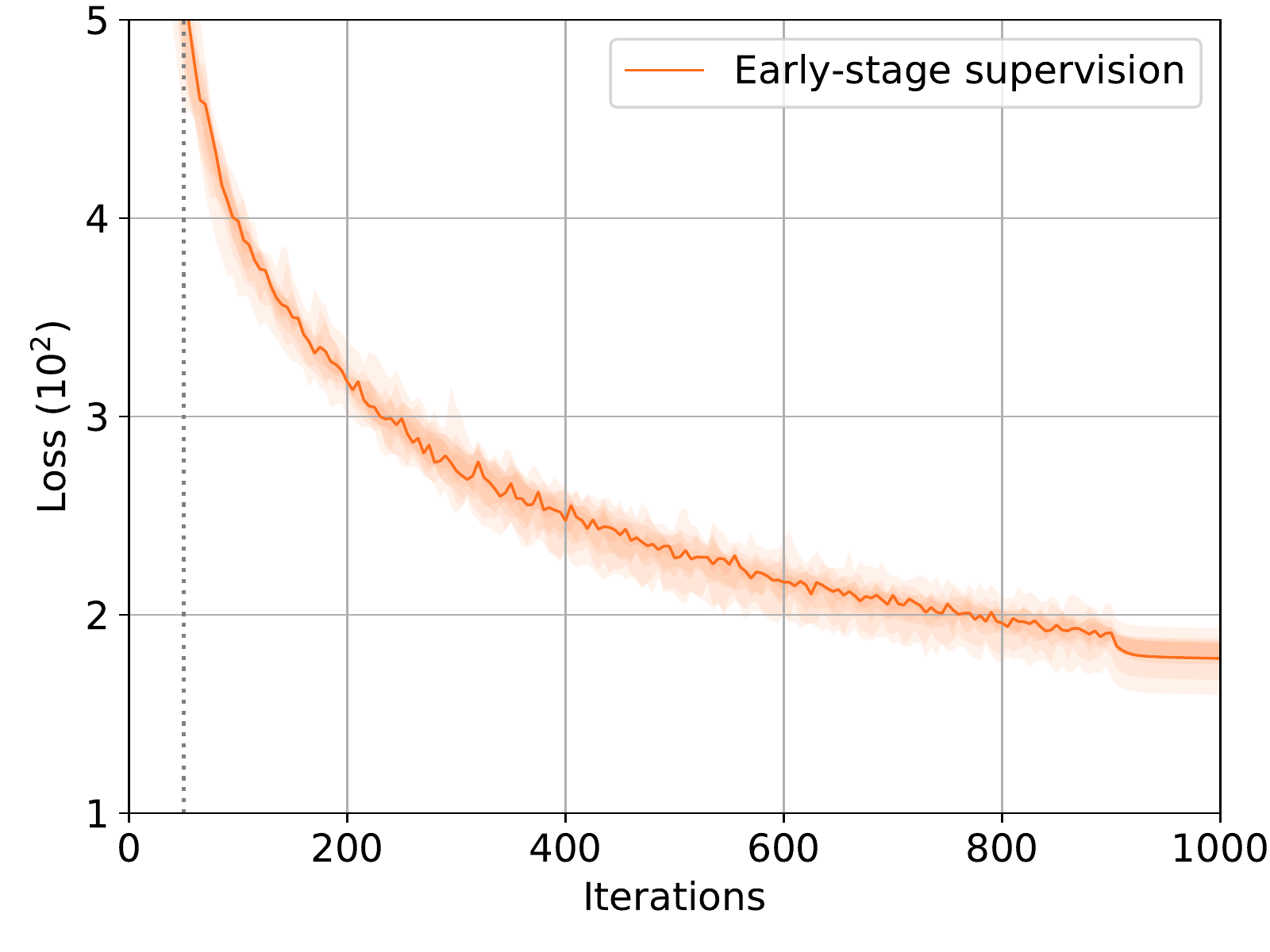}\hfil
	\includegraphics[width=\figw]{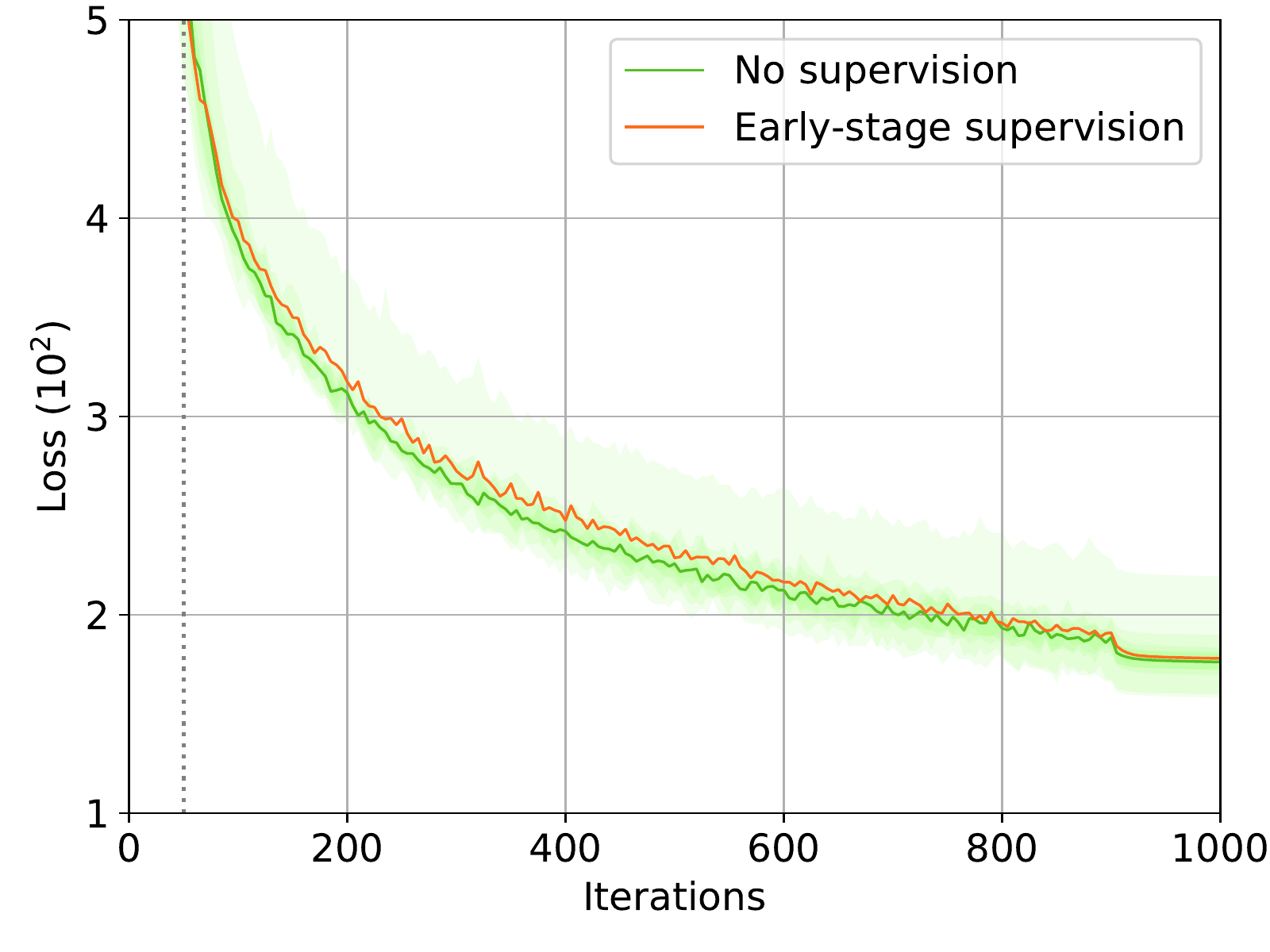}\hfil
	\includegraphics[width=\figw]{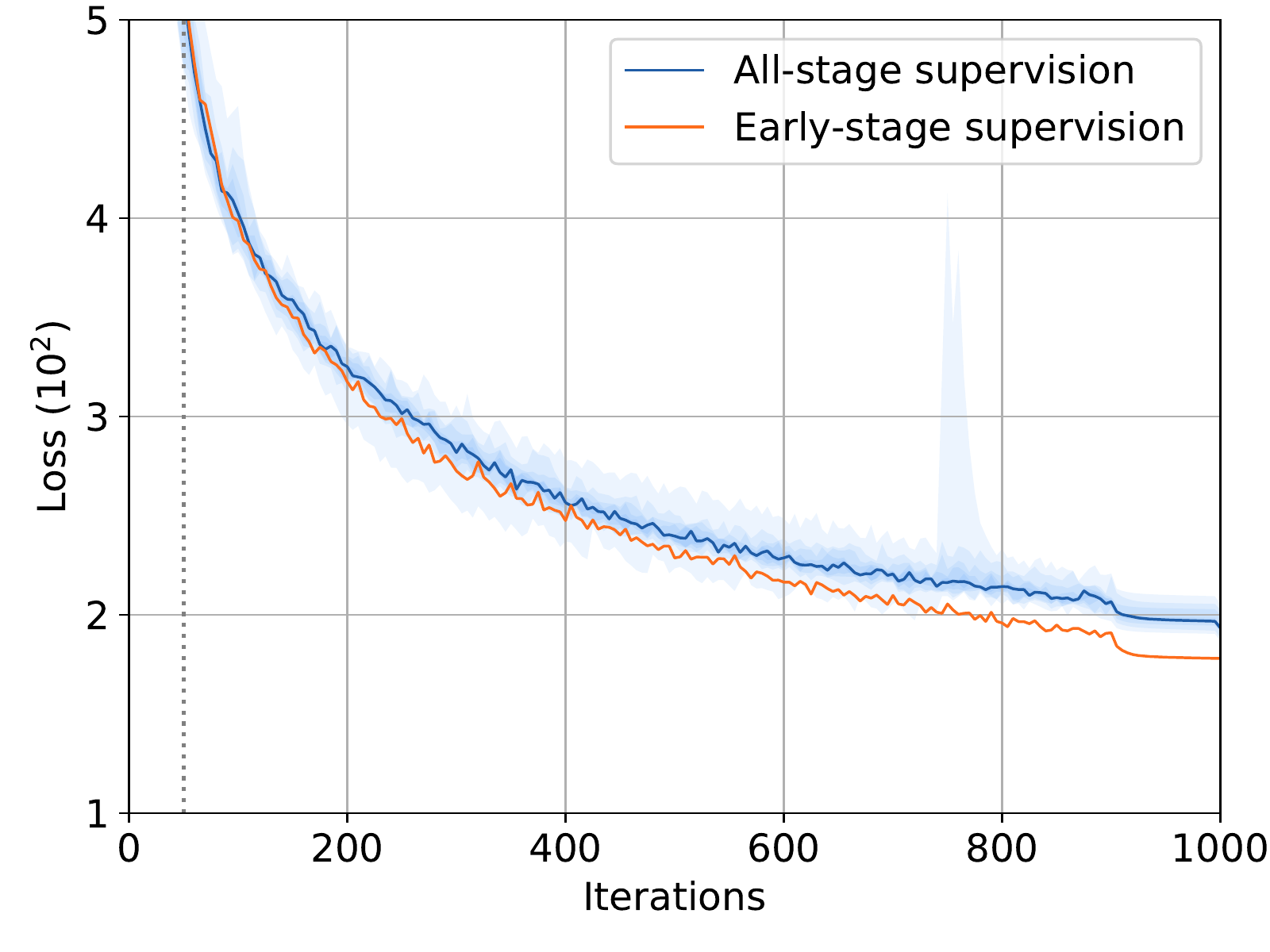}\\
	\begin{minipage}{\figw}\centering {Early-stage supervision}\end{minipage}\hfil
	\begin{minipage}{\figw}\centering {No supervision}\end{minipage}\hfil
	\begin{minipage}{\figw}\centering {All-stage supervision}\end{minipage}\\
	\caption{\textbf{Convergence analysis with different types of weak supervision for \textsc{cow} scene.}
		See also explanations in Fig.~\ref{afig:training_ball}.
	}
\end{figure*}

\end{document}